\documentclass[11pt]{article}

\usepackage[letterpaper, margin=0.98in]{geometry}
\usepackage[utf8]{inputenc}
\usepackage[T1]{fontenc}
\usepackage{times}
\usepackage{microtype}
\usepackage[scaled=.92]{inconsolata}

\usepackage{amsmath}
\usepackage{amssymb}
\usepackage{amsfonts} % For \mathbb, etc.
\usepackage{nicefrac} % For 1/2 symbols
\usepackage{bbm}      % For \mathbbm

\usepackage{graphicx}
\usepackage{xcolor} % xcolor is more powerful than color
\usepackage[labelfont={bf},font=small]{caption}
\usepackage{subcaption}
\usepackage{wrapfig}
\usepackage[breakable, skins]{tcolorbox} % Combined and added 'skins' for tikz
\usepackage{tikz}
\usetikzlibrary{positioning}

\usepackage{booktabs}

\usepackage{natbib}
\usepackage{url}
\usepackage{pifont} % For symbols like \ding
\usepackage{twemojis}
\usepackage{placeins}
\usepackage{lineno}
\usepackage{titletoc}
\usepackage{amsthm} % For theorem environments

\usepackage{hyperref}

\usepackage{enumitem}

\usepackage{array}
\newcolumntype{R}[1]{>{\raggedright\arraybackslash}p{#1}}

\newcommand{\rownote}[1]{\parbox[t]{\linewidth}{\footnotesize #1}}
\newcommand{\rowstrut}{\rule{0pt}{0pt}} % adjust 58pt as needed

\tikzset{every picture/.style={baseline=(current bounding box.north)}}

\newcommand{\noteoffset}{4.4ex}
\renewcommand{\rownote}[1]{\raisebox{-\noteoffset}{\parbox[t]{\linewidth}{\footnotesize #1}}}

\usepackage{mathtools}

\newcommand{\thetabtrain}{{\thetab}_\text{ID}}
\newcommand{\x}{\mathbf{x}}

\usepackage{dsfont}

\usepackage{nicefrac} 
\definecolor{darkred}{RGB}{200,0,0}
\definecolor{darkgreen}{RGB}{0,150,0}
\definecolor{darkblue}{RGB}{0,0,200}
\definecolor{darkcyan}{rgb}{0.0, 0.55, 0.55}
\definecolor{cyan}{rgb}{0.0, 1.0, 1.0}
\definecolor{lightorange}{rgb}{1.0, 0.55, 0.0}
\definecolor{darkpurple}{rgb}{0.7, 0., 0.7}
\definecolor{ballblue}{rgb}{0.13, 0.67, 0.8}
\definecolor{bluencs}{rgb}{0.0, 0.53, 0.74}
\definecolor{darkorange}{rgb}{0.9, 0.29, 0.0}

\newcommand{\darkred}{\color{darkred}}

\newcommand{\darkorange}{\color{darkorange}}

\theoremstyle{definition}

\renewcommand{\paragraph}[1]{\textbf{#1}.~}
\newif\ifshowedits
\showeditsfalse % uncomment to turn off
\newcommand{\editstyle}[1]{\textcolor{blue}{#1}}
\newcommand{\new}[1]{%
  \ifshowedits
    \editstyle{#1}%
  \else
    #1%
  \fi
}

\newif\ifshowcomments
\showcommentsfalse % hide all comments

\newcommand{\cmt}[2]{%
  \ifshowcomments
    \textcolor{#1}{#2}%
  \fi
}

\newcommand{\tina}[1]{\cmt{darkred}{\textbf{[TB]} #1}}
\newenvironment{fminipage}%
  {\begin{Sbox}\begin{minipage}}%
  {\end{minipage}\end{Sbox}\fbox{\TheSbox}}

\newcommand{\R}{\mathds{R}}

\newcommand{\sft}[1]{\mathbb{S}(#1)}

 \newcommand{\wt}{\widetilde}

\newcommand{\ab}{\mathbf{a}}
\newcommand{\Ab}{\mathbf{A}}

\newcommand{\Hb}{\mathbf{H}}
\newcommand{\Hbtilde}{\wt\Hb}
\newcommand{\hb}{\mathbf{h}}
\newcommand{\hbtilde}{\wt\hb}

\newcommand{\Mb}{\mathbf{M}}
\newcommand{\Pb}{\mathbf{P}}

\newcommand{\pb}{\mathbf{p}}

\newcommand{\xb}{\mathbf{x}}

\newcommand{\zb}{\mathbf{z}}
\newcommand{\ellb}{\boldsymbol{\ell}}

\newcommand{\thetab}{{\boldsymbol{\theta}}}

\newcommand{\Dc}{\mathcal{D}}

\newcommand{\Lc}{\mathcal{L}}

\newcommand{\Ebb}{\mathbb{E}}

\newcommand{\seqlen}{T}
\newcommand{\vocabsize}{V}
\newcommand{\vset}{\mathcal{\vocabsize}}
\newcommand{\vk}{\vset_\factset}

\newcommand{\vmsize}{\vocabsize_\Dc}
\newcommand{\vmc}{\vset_\Dc}

\newcommand{\factsize}{K}
\newcommand{\factset}{\mathcal{\factsize}}
\newcommand{\source}{a}
\newcommand{\sourceset}{\ensuremath{\mathcal{\expandafter\MakeUppercase\expandafter{\source}}}}
\newcommand{\target}{b}
\newcommand{\targetset}{\ensuremath{\mathcal{\expandafter\MakeUppercase\expandafter{\target}}}}

\newcommand{\dist}{\Dc}
\newcommand{\tmplnum}{N}

\newcommand{\pos}{I}
\newcommand{\pospair}{\mathbf{\pos}}
\newcommand{\posSource}{i}
\newcommand{\posTarget}{j}

\newcommand{\transletter}{P}
\newcommand{\transitionmat}{\mathbf{\transletter}}

\newcommand{\positionset}{\ensuremath{\mathcal{\expandafter\MakeUppercase\expandafter{\pos}}}}

\newcommand{\exposmat}{\new{\Mb}}
\newcommand{\exposmatin}{{\exposmat_{\rm{in}}}}

\newcommand{\Structure}{Position }
\newcommand{\structure}{position }

\newcommand{\frz}{{\text{\ding{100}}}}

\newcommand{\posexp}[1]{\textsc{MC1Pos#1}}
\newcommand{\mcexp}[1]{\textsc{MC#1Pos1}}
\newcommand{\mixexp}[1]{\textsc{MC#1Pos#1}}

\newcommand{\highmodel}{{\thetab_{\rm{hi}}}}
\newcommand{\lowmodel}{{\thetab_{\rm{low}}}}

\newcommand{\factacc}{{\textbf{\texttt{Acc}}_{\rm{fact}}}}
\newcommand{\factloss}{{\textbf{\texttt{Loss}}_{\rm{fact}}}}
\newcommand{\posacc}{{\textbf{\texttt{Acc}}_{\rm{pos}}}}
\newcommand{\posacck}{{\textbf{\texttt{Acc}}_{\rm{pos},\mathcal{\factset}}}}

\newcommand{\posloss}{{\textbf{\texttt{Loss}}_{\rm{pos}}}}
\newcommand{\KL}{{\textbf{\texttt{Loss}}_{\rm{stat}}}}

    \newcommand{\dvr}{\texttt{DIV}}

\newcommand{\colTwoThreeShift}{\raisebox{6pt}[0pt][0pt]}

\usepackage[capitalize]{cleveref}
\definecolor{darkblue}{rgb}{0, 0, 0.5}
\hypersetup{colorlinks=true, citecolor=darkblue, linkcolor=darkblue, urlcolor=darkblue}

\title{
Facts in Stats: Impacts of Pretraining Diversity\\ on Language Model Generalization
}

\author{Tina Behnia \quad Puneesh Deora \quad Christos Thrampoulidis\\
\vspace{-2.5mm}\\
University of British Columbia, Canada%\thanks{Corresponding author: \texttt{tina.behnia@ece.ubc.ca}}\\
}

\date{}

\newcommand\blfootnote[1]{%
  \begingroup
  \renewcommand\thefootnote{}\footnote{#1}%
  \addtocounter{footnote}{-1}%
  \endgroup
}

\begin{document}

\maketitle

\blfootnote{%
  \parbox{\textwidth}{%
    Code available at \url{https://github.com/TinaBehnia/FactStat}\\
    Corresponding author: \texttt{tina.behnia@ece.ubc.ca}%
  }%
}
 \begin{abstract}
 Language models are pretrained on sequences that blend statistical regularities (structures making text fluent) with factual associations between specific tokens (corresponding to knowledge of facts). While recent work suggests that the variability of their interaction, such as paraphrases of factual associations, critically determines generalization ability, we lack a systematic analysis of these impacts. This paper introduces a flexible synthetic testbed that combines a statistical stream of generic tokens with an abstract factual stream of source-target token pairs, enabling fine-grained control over their interaction. The design enables the independent control of diversity nature by manipulating stream composition (contextual structure) and the level of diversity by varying which statistical streams each fact appears in. Through controlled experiments, we find that while higher contextual diversity delays in-distribution (ID) factual accuracy, its effect on out-of-distribution (OOD) {factual} generalization depends critically on contextual structure. In some cases, OOD performance follows the same trend as ID, but in others, diversity becomes essential for non-trivial factual learning. Even when low diversity prohibits factual recall, optimal diversity levels depend on training duration. Beyond factual recall failures, we identify structures where statistical generalization fails independently, and others where both capabilities degrade simultaneously. This demonstrates how the interplay between contextual design and diversity level impacts different aspects of generalization. 
 \new{Furthermore, through a series of controlled interventions on the model components, we trace the {OOD} generalization failures to distinct optimization bottlenecks, highlighting the importance of the learned embedding and unembedding layers.}
 % Through detailed mechanistic analysis of transformer components, we find that learned embeddings are key to successful generalization under high-diversity data. 
 Overall, our synthetic framework allows us to isolate effects that would be confounded in large-scale studies, thus offering a controlled testbed for future investigations. \looseness=-1 %\footnote{Code available at \url{https://github.com/TinaBehnia/FactStat}.} 

   \end{abstract}

\section{Introduction}\label{sec:intro}

Modern transformer-based {language models} (LMs), trained on massive corpora of  natural-text sequences, simultaneously learn to generate contextually plausible sequences that follow statistical and linguistic patterns, while also learning significant amount of real-world knowledge. Their parameters effectively become an implicit knowledge base that can rival structured knowledge bases in question-answering and recall tasks \citep{petroni2019language, roberts2020much,dai2021knowledge,allen2023physics,akyurek2022towards,yang2024synthetic,mallen2023not,meng2023locating}.
%\citep{roberts2020much,petroni2019language,mallen2023not, meng2023locating,dai2021knowledge}
%\tina{} %\puneesh{maybe you kept the comment for this only, but 1-2 more refs would be nice} \tina{}. 
%
This dual capability is remarkable precisely because the LM receives no explicit fact supervision during training: it never encounters knowledge graphs, labeled facts, or any explicit differentiation between linguistic and factual content. 
%Yet somehow, it develops the capability to distinguish world knowledge from linguistic patterns, while simultaneously learning both. 
Yet somehow, it develops the ability to both follow linguistic patterns and encode world knowledge. 
\emph{How does the conceptually-simple next-token prediction objective enable a language model to implicitly {learn both linguistic structure and factual knowledge?} Are there inherent trade-offs in acquiring these two types of information?
% distinguish and simultaneously learn linguistic structure versus factual knowledge? Are there inherent trade-offs in acquiring these two types of information?
}

Intuitively, the repetition of factual information across varying \new{and diverse} linguistic contexts during training, essentially appearing as \emph{paraphrases} of the same underlying facts, likely plays a key role in facilitating factual recall from pure next-token prediction supervision. But, \new{ this raises several questions:}

\emph{How frequently must a model encounter a fact during training to reliably learn it, and how does this depend on the \emph{diversity} of linguistic contexts (paraphrases) in which the fact appears?}  \emph{Under what training conditions is it possible to isolate factual associations from linguistic structures?} {Moreover, \emph{is diversity only possibly impacting factual recall, or can it also impact the model's ability to learn statistical patterns?}} \looseness=-1
% \puneesh{why is diversity and moreover non-italicized; i dont think second question is clear "under what conditions is it possible to isolate factual associations from linguistic structures}

\begin{figure}
    \centering
    \vspace{-20pt}
    \includegraphics[width=0.95\linewidth]{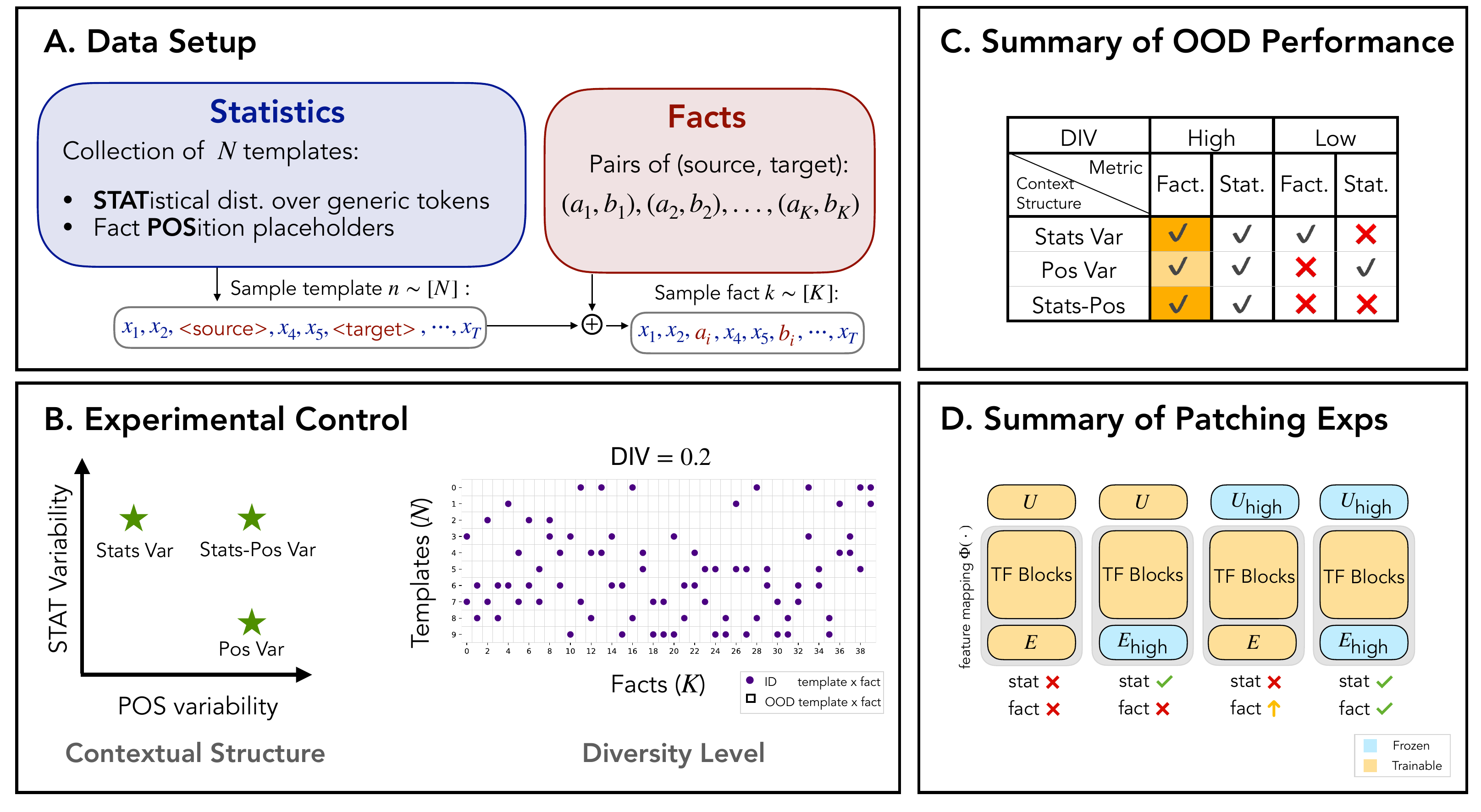}
    \vspace{-0.05in}
    \caption{
    {\textbf{Proposed testbed and summary of key findings.}
    \textbf{(A)} Sequences are generated by combining a \textbf{factual stream}, a set of atomic pairs, and a \textbf{statistical stream} of templates, each a statistical distribution over sequences with two fact placeholder positions. Final sequences insert a fact into a sequence sampled from a template \new{(Sec.~\ref{sec:setup})}.
    \textbf{(B)} We control two key data properties. \textbf{(Left) Contextual structure:} The degree of statistical and (fact) positional variation across templates. We explore three structures: \textbf{Stats Var} (varied statistics, fixed positions), \textbf{Pos Var} (fixed statistics, varied positions), and \textbf{Stats-Pos Var} (varied on both axes). \textbf{(Right) Diversity level:} The number of unique templates each fact is paired with in the training corpus. The table shows an \emph{exposure matrix}, where dots indicate template-fact pairs used for training \new{(Sec.~\ref{sec:div}-\ref{sec:templates})}.
    \textbf{(C)} Low diversity impairs {out-of-distribution} generalization, but the specific aspects affected (factual recall vs. statistical pattern generation) depend on contextual structure.  At high diversity, all structures enable both types of generalization, though with varying efficiency (colored boxes: darker \new{means more training iterations needed} with higher $\dvr$) \new{(Sec.~\ref{sec:results})}. \textbf{(D)} 
    {While training on low-diversity data impairs statistical and factual generalization, when the embeddings or unembeddings from a high-diversity model are ``patched'' in, the remaining modules can still learn a generalizing solution even when trained on that same low-diversity data \new{(Sec.~\ref{sec:bottleneck})}.}
    \looseness=-1 
    % \tina{fig!+add ref to sec in fig + add interventions?} 
    % \puneesh{just adding an incomplete version to see how it fits}
    }
    }
    \vspace{-0.15in}
    \label{fig:setup}
\end{figure}

% . Instead, it acquires factual knowledge purely as an emergent by-product of next-token prediction on raw text.

% Crucially, next-token prediction does not explicitly differentiate between tokens that form statistical linguistic structures and those that convey \new{deterministic} factual content. 
% Yet, LLMs somehow develop the capability to distinguish world knowledge from linguistic patterns, while simultaneously learning both.  The same factual information often appears multiple times during training within varying linguistic contexts, essentially as {paraphrases} of the original factual information. Despite being trained with only the next-token prediction objective 
% %
% This raises fundamental questions: How can a single objective guide models to learn these distinct types of information? Are there inherent trade-offs between acquiring linguistic structure versus factual knowledge? Specifically, focusing on factual recall, what conditions enable successful learning: how frequently must a model encounter a fact during training to reliably learn it, and how does this depend on the \emph{diversity} of linguistic contexts in which the fact appears? If learning truly means isolating factual associations from linguistic structures, under what conditions is this possible? 

\vspace{3pt}
The intriguing factual-recall capability of LLMs has motivated several recent studies probing how models acquire, represent, and recall facts using controlled setups. \citet{allen2023physics} and very recently  \citet{zucchet2025language} study synthetic biography corpora with fixed templates 
% \puneesh{think the word template is not clear, it's just introduced here all of a sudden; saying it bc it is an important point raised a few lines later} 
and varying factual associations (e.g., name, age, occupation) to examine how paraphrastic diversity affects factual recall. While their data setup provides more control than raw natural-text corpora, where linguistic patterns and facts are intricately interleaved, they still operate at considerable scale, requiring transformers with {8-12 layers/8-12 heads/512-768 dimension} for empirical analysis. 
This limits accessibility and makes comprehensive parameter sweeps over controllable variables computationally challenging. Additionally, by employing fixed templates, these setups isolate focus on factual learning while ignoring the model's ability to simultaneously acquire linguistic structure and leaving fine-grained impacts of diversity on learning of both components underexplored.

Motivated by these, we develop a small-scale synthetic testbed for studying at a fine-grained level the interactions between statistical and factual learning streams, specifically as captured in terms of \new{context} \emph{diversity}. 
% \puneesh{I think should mention here again what diversity means or even say pre-training data diversity} 
%Recognizing that language consists of more than isolated facts in fixed sequences, our data abstraction embeds factual mappings within sequences generated by statistical distributions that models must also learn. This cleanly disentangles statistical and factual generalization.
%
Our framework offers three key advantages. \emph{Tractability:} Its minimal scale enables comprehensive parameter sweeps and in-depth analysis that would be computationally challenging in large-scale setups. All our experiments use {4-layer/4-head/32-dim} transformers, and  phenomena can be reproduced even with 1-layer transformers. All experiments are performed on {on single Tesla V100-SXM2-GPU(16GB) and can be replicated on the free-version of Google-Colab}. 
% \puneesh{do you wanna mention gpu hours and compare with Zuccet and Allen-Zhu?} 
\emph{Reproducibility:} Despite its minimalism, our framework reproduces  empirical phenomena from large-scale studies, including stage-wise factual learning 
% \puneesh{do you want to say stagewise learning now?}
\citep{zucchet2025language} and improved factual recall with larger diversity \citep{allen2023physics}. \emph{Experimental control:} The explicit statistical stream enables exhaustive exploration of how pretraining data diversity, both in terms of exposure level  and structural variety, impacts transformer out-of-distribution generalization under next-token prediction training.

In broader perspective, our methodology mirrors research turning to deliberately small-scale tasks and transformers to study other next-token prediction phenomena, such as in-context learning \citep{garg2022can,statistical_induction_heads}, grokking \citep{power2022grokking,nanda2023progress}, and arithmetic reasoning \citep{lee2023teaching}. Within factual recall research, our data setup bears some abstract similarity to \citet{nichani2024understanding}, who use synthetic tasks to analyze scaling laws and dynamics of factual recall. Yet, their setup lacks statistical learning, any account of diversity effects, and operates in non-autoregressive setting. See App. \ref{app:related_works} for detailed comparison to related works. 

% In terms of abstraction and simplicity, our data setup is closest in spirit to \citet{nichani2024understanding}, who used a synthetic task to theoretically analyze scaling laws and dynamics of factual recall. However, our formulation is richer: it introduces auto-regressive statistical context patterns, enables precise tuning of Diversity level, and measures both statistical and factual accuracy in both ID and OOD settings.

% Overall, our testbed, enables exhaustive exploration of how diversity of pretraining data, both in terms of level (how many facts appear in how many stats) but also in terms of structure (how facts embed into stats), impacts the ability of transformers trained with the language modeling next-token prediction objective to generalize on out-of-distribution data. 

% the diversity-structure interaction space that remains computationally intractable in realistic settings.

% 
% \ct{Instead of this say recent works try to address this via controlled experiments with natural language. Challenges: .... you do pure synthetic instead}The primary challenge in addressing these questions stems from the inherently complex nature of natural language data that intricately interleaves linguistic patterns and factual content.
%  To address this, inspired by \

\begin{figure}[t]
  \centering
  \begin{tabular}{@{\hspace{-0pt}}c@{\hspace{15pt}}c@{\hspace{15pt}}|@{\hspace{-3pt}}c@{\hspace{15pt}}c@{\hspace{15pt}}|@{\hspace{-3pt}}c@{\hspace{15pt}}c@{}}
  % ─── header row: one cell for col 1, one spanning cols 2–3 ─────────
  \multicolumn{2}{c@{\hspace{0pt}}}{\fontsize{9pt}{8pt}\selectfont\textbf{(a) Statistical loss}}
  & \multicolumn{2}{c}{\fontsize{9pt}{8pt}\selectfont\textbf{(b) Position accuracy}}
  & \multicolumn{2}{c@{\hspace{5pt}}}{\fontsize{9pt}{8pt}\selectfont\textbf{(c) Factual accuracy}}  \\[8pt]
  % ─── row 1 ───────────────────────────────
    % \hspace{-25pt}
    \vspace{-0.2in}
    \begin{subfigure}{0.14\textwidth}
        \centering
        \begin{tikzpicture}[remember picture]
            \node at (0,0) {\includegraphics[scale=0.14]{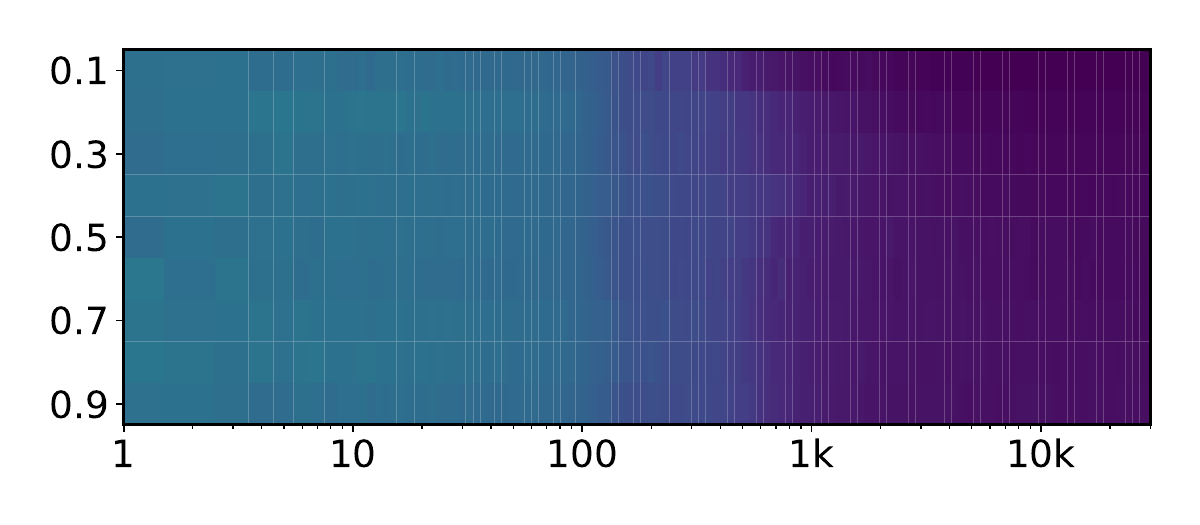}};
            % \node at (0,0) {\includegraphics[scale=0.14]{figures/kl_masked_completion_GT_in_dist_MC10Pos1_order1_L4H4d32_T50_heatmap_avg.pdf}};
            \node[scale=0.8] at (0.0, 0.8) { \textbf{ ID}};
            \node[overlay,scale=0.7,rotate=90] at (-2.4, 0.0) { \new{\textbf{Stats Var}}};
            \node[overlay, scale=0.5,rotate=90] at (-2.0, 0.0) { \textbf{(\mcexp{10}) }};
            \node[scale=0.8,rotate=90] at (-1.5, 0.05) { \textbf{{$\dvr$}}};            
            
        \end{tikzpicture}
    \end{subfigure} &
    \begin{subfigure}{0.14\textwidth}
        \centering
        \begin{tikzpicture}[remember picture]
            \node at (0,0) {\includegraphics[scale=0.14]{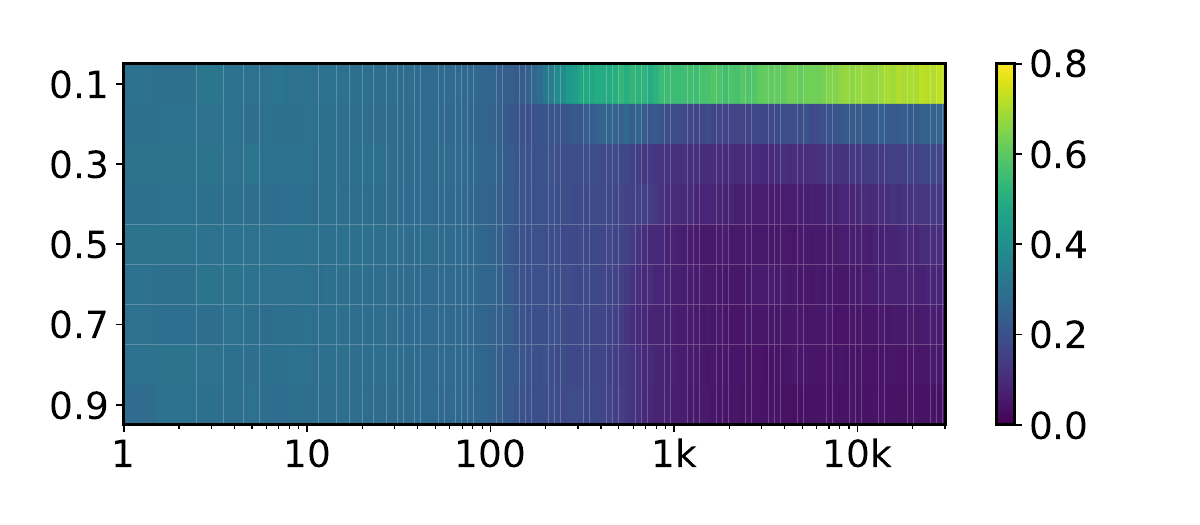}};
            % \node at (0,0) {\includegraphics[scale=0.14]{figures/kl_masked_completion_GT_out_dist_MC10Pos1_order1_L4H4d32_T50_heatmap_avg.pdf}};
            \node[scale=0.8] at (-0.3, 0.8) { \textbf{ OOD}};
            % \node[scale=0.8,rotate=90] at (-2.3, 0.0) { \textbf{\mcexp }};
        \end{tikzpicture}
    \end{subfigure} &
    % \hspace{70pt}%
    \begin{subfigure}{0.14\textwidth}
        \centering
        \begin{tikzpicture}[remember picture]
            \node at (0,0) {\includegraphics[scale=0.14]{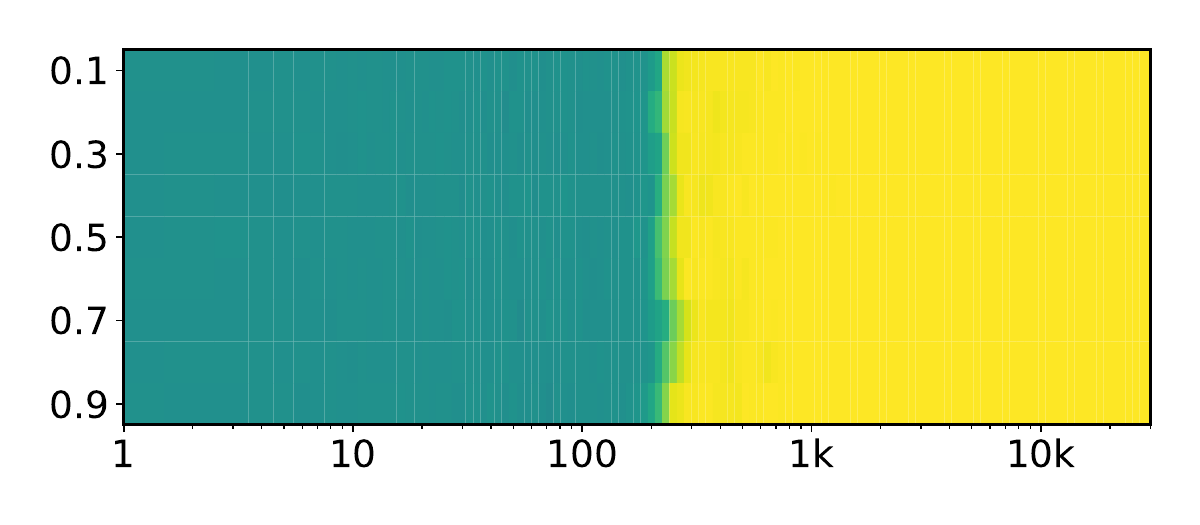}};
            % \node at (0,0) {\includegraphics[scale=0.14]{figures/kl_masked_completion_GT_in_dist_MC10Pos1_order1_L4H4d32_T50_heatmap_avg.pdf}};
            \node[scale=0.8] at (0.0, 0.8) { \textbf{ ID}};
        \end{tikzpicture}
    \end{subfigure} &
    \begin{subfigure}{0.14\textwidth}
        \centering
        \begin{tikzpicture}[remember picture]
            \node at (0,0) {\includegraphics[scale=0.14]{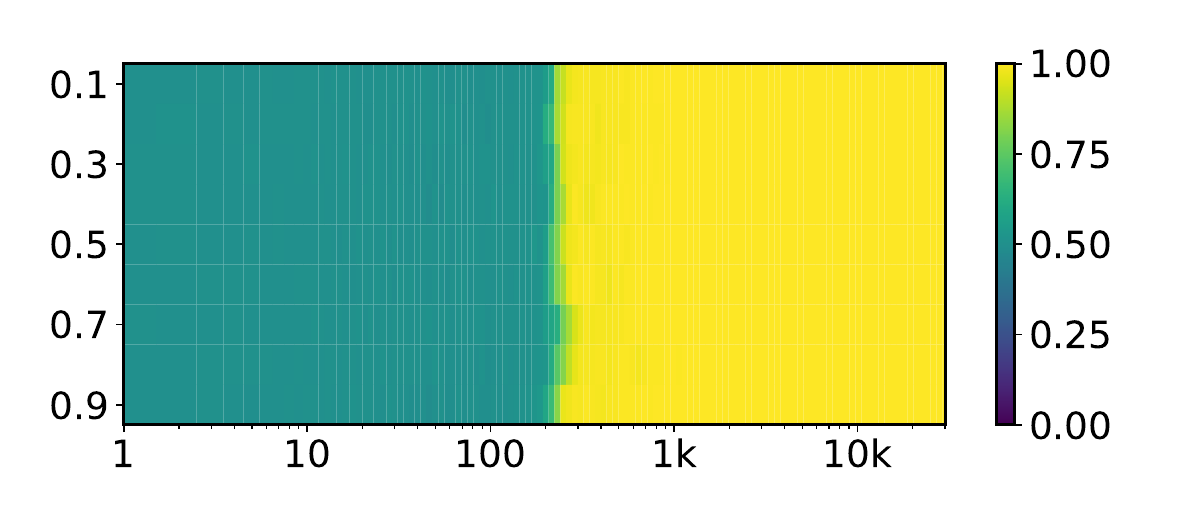}};
            % \node at (0,0) {\includegraphics[scale=0.14]{figures/kl_masked_completion_GT_out_dist_MC10Pos1_order1_L4H4d32_T50_heatmap_avg.pdf}};
            \node[scale=0.8] at (-0.3, 0.8) { \textbf{ OOD}};
            % \node[scale=0.8,rotate=90] at (-2.3, 0.0) { \textbf{\mcexp }};
        \end{tikzpicture}
    \end{subfigure} &
    % \hspace{70pt}%
    \begin{subfigure}{0.14\textwidth}
        \centering
        \begin{tikzpicture}[remember picture]
            \node at (0,0) {\includegraphics[scale=0.14]{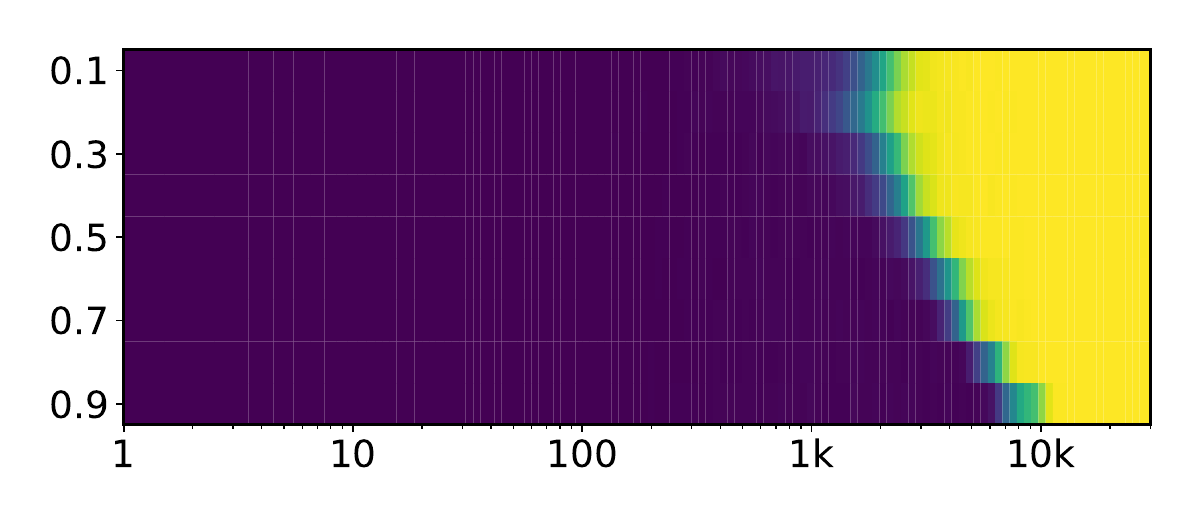}};
            % \node at (0,0) {\includegraphics[scale=0.14]{figures/at_pos_is_bi_rate_in_dist_MC10Pos1_order1_L4H4d32_T50_heatmap_avg.pdf}};
            \node[scale=0.8] at (0.0, 0.8) { \textbf{ ID}};
            % \node[scale=0.8,rotate=90] at (-2.3, 0.0) { \textbf{\mcexp }};
        \end{tikzpicture}
    \end{subfigure} &
    % \hspace{50pt}%
    \begin{subfigure}{0.14\textwidth}
        \centering
        \begin{tikzpicture}[remember picture]
            \node at (0,0) {\includegraphics[scale=0.14]{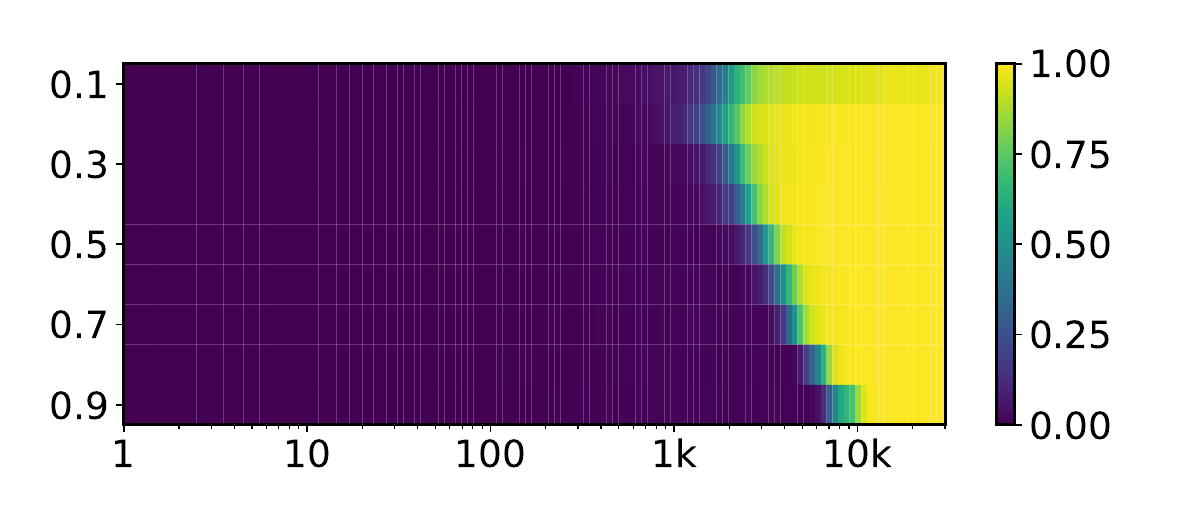}};
            % \node at (0,0) {\includegraphics[scale=0.14]{figures/at_pos_is_bi_rate_out_dist_MC10Pos1_order1_L4H4d32_T50_heatmap_avg.pdf}};
            \node[scale=0.8] at (0.0, 0.8) { \textbf{ OOD  }};
            % \node[scale=0.8] at (0.0, -1.2) { \textbf{ }};
        \end{tikzpicture}
    \end{subfigure}\\[50pt]
    % \vspace{2pt} % vertical gap between rows
    % Second row 
    % \hspace{-25pt}
	% ─── row 2 ───────────────────────────────
    \begin{subfigure}{0.14\textwidth}
        \centering
        \begin{tikzpicture}[remember picture]
            \node at (0,0) {\includegraphics[scale=0.14]{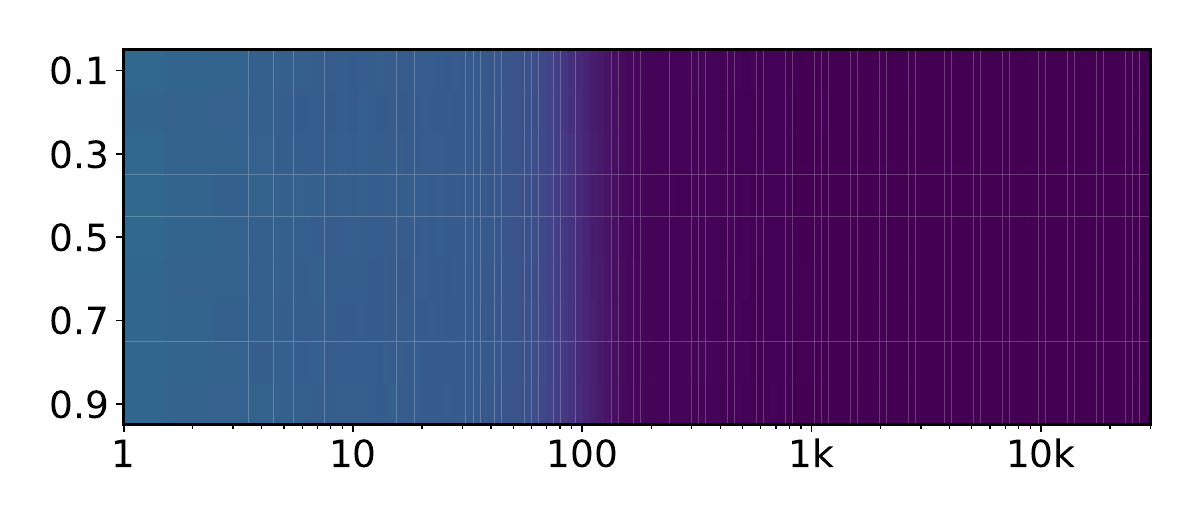}};
            % \node at (0,0) {\includegraphics[scale=0.14]{figures/kl_masked_completion_GT_in_dist_MC1Pos10_order1_L4H4d32_T50_heatmap_avg.pdf}};
            \node[overlay,scale=0.7,rotate=90] at (-2.4, 0.0) { \new{\textbf{Pos Var}}};
            \node[overlay, scale=0.5,rotate=90] at (-2.0, 0.0) { \textbf{(\posexp{10}) }};
            
            \node[scale=0.8] at (0.0, -1.2) { \textbf{ }};
            \node[scale=0.8,rotate=90] at (-1.5, 0.05) { \textbf{{$\dvr$}}};

        \end{tikzpicture}
    \end{subfigure} &%
    \begin{subfigure}{0.14\textwidth}
        \centering
        \begin{tikzpicture}[remember picture]
            \node at (0,0) {\includegraphics[scale=0.14]{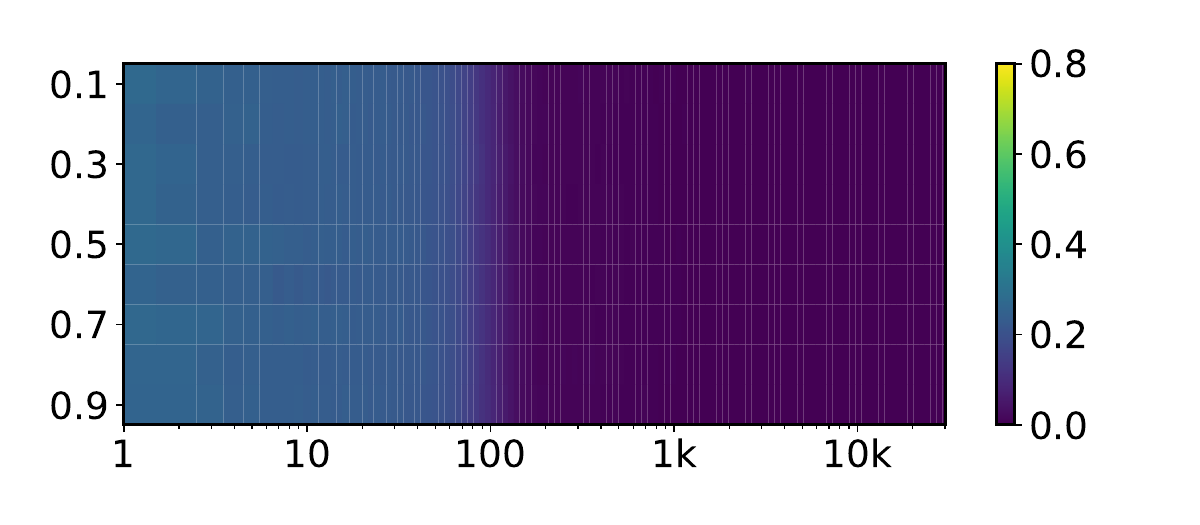}};
            % \node at (0,0) {\includegraphics[scale=0.14]{figures/kl_masked_completion_GT_out_dist_MC1Pos10_order1_L4H4d32_T50_heatmap_avg.pdf}};
            % \node[scale=0.8,rotate=90] at (-2.7, 0.0) { \textbf{\posexp{10} }};            
            \node[scale=0.8] at (0.0, -1.2) { \textbf{ }};
            
        \end{tikzpicture}
    \end{subfigure} &%
    % \hspace{70pt}%
    % \end{minipage}%
    \begin{subfigure}{0.14\textwidth}
        \centering
        \begin{tikzpicture}[remember picture]
            \node at (0,0) {\includegraphics[scale=0.14]{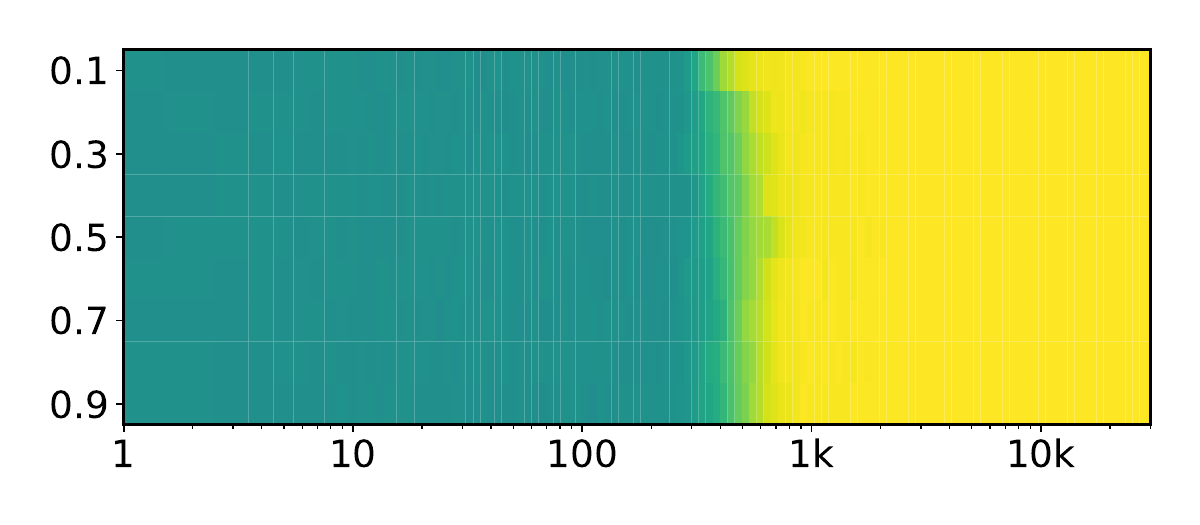}};
            % \node at (0,0) {\includegraphics[scale=0.14]{figures/kl_masked_completion_GT_in_dist_MC1Pos10_order1_L4H4d32_T50_heatmap_avg.pdf}};
            % \node[scale=0.45,rotate=90] at (-2.3, 0.0) { \textbf{(\posexp{10} )}};  
            % \node[scale=0.75,rotate=90] at (-2.6, 0.0) {\new{ \textbf{Pos Var }}};
            \node[scale=0.8] at (0.0, -1.2) { \textbf{ }};
            % \node[scale=0.6,rotate=90] at (-1.9, 0.15) { \textbf{{$\dvr$}}};            

        \end{tikzpicture}
    \end{subfigure} &%
    \begin{subfigure}{0.14\textwidth}
        \centering
        \begin{tikzpicture}[remember picture]
            \node at (0,0) {\includegraphics[scale=0.14]{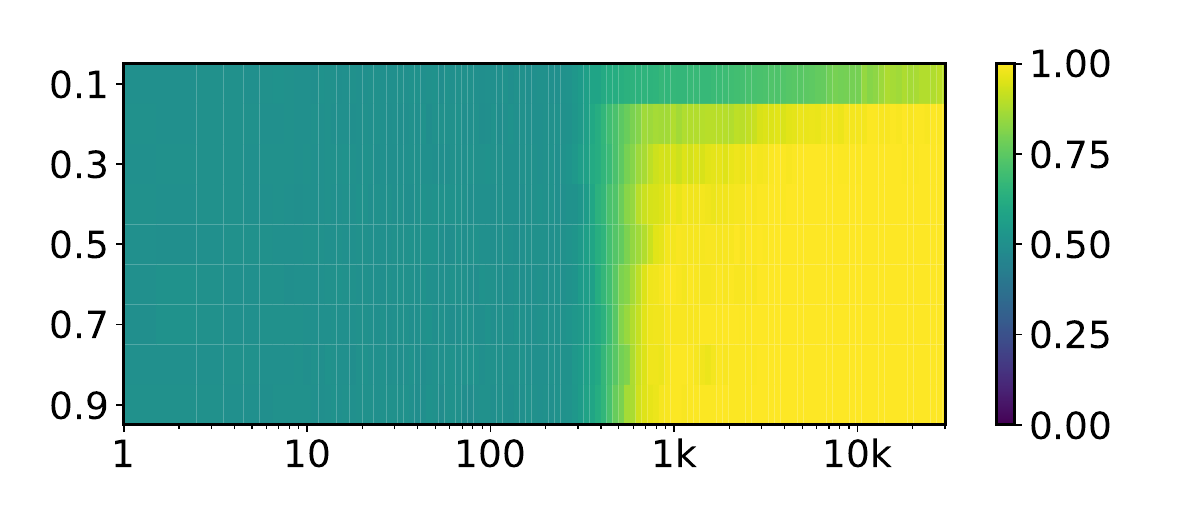}};
            % \node at (0,0) {\includegraphics[scale=0.14]{figures/kl_masked_completion_GT_out_dist_MC1Pos10_order1_L4H4d32_T50_heatmap_avg.pdf}};
            % \node[scale=0.8,rotate=90] at (-2.7, 0.0) { \textbf{\posexp{10} }};            
            \node[scale=0.8] at (0.0, -1.2) { \textbf{ }};
            
        \end{tikzpicture}
    \end{subfigure} &%
    % \hspace{70pt}%
    \begin{subfigure}{0.14\textwidth}
        \centering
        \begin{tikzpicture}[remember picture]
            \node at (0,0) {\includegraphics[scale=0.14]{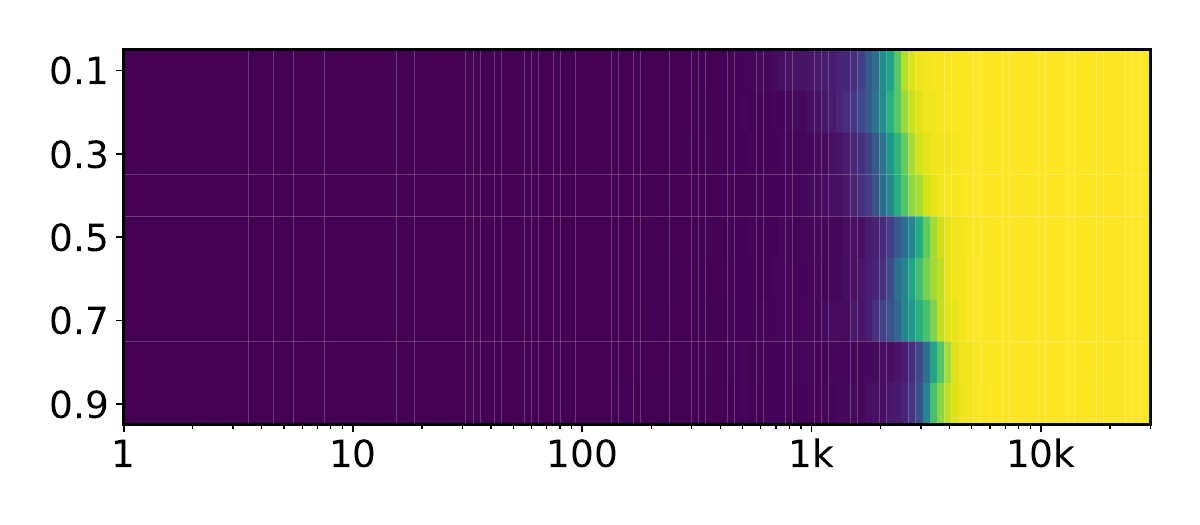}};
            % \node at (0,0) {\includegraphics[scale=0.14]{figures/at_pos_is_bi_rate_in_dist_MC1Pos10_order1_L4H4d32_T50_heatmap_avg.pdf}};
            % \node[scale=0.8,rotate=90] at (-2.7, 0.0) { \textbf{\posexp{10} }};            
            \node[scale=0.8] at (0.0, -1.2) { \textbf{ }};
            
        \end{tikzpicture}
    \end{subfigure} &%
    % \hspace{50pt}%
    \begin{subfigure}{0.14\textwidth}
        \centering
        \begin{tikzpicture}[remember picture]
            \node at (0,0) {\includegraphics[scale=0.14]{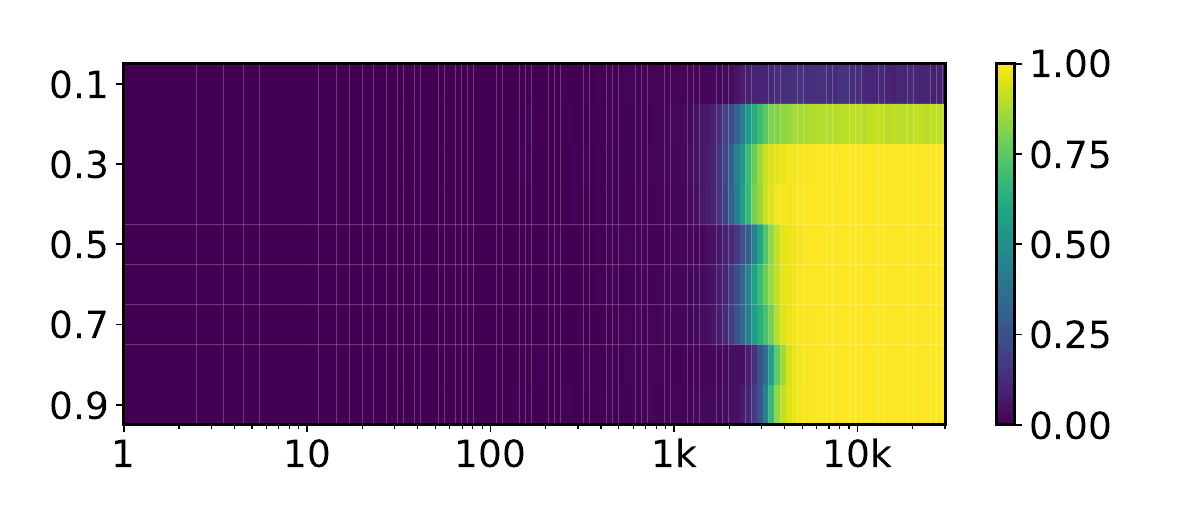}};            \node[scale=0.8] at (0.0, -1.2) {\textbf{}};
            % \node at (0,0) {\includegraphics[scale=0.14]{figures/at_pos_is_bi_rate_out_dist_MC1Pos10_order1_L4H4d32_T50_heatmap_avg.pdf}};            \node[scale=0.8] at (0.0, -1.2) {\textbf{}};
            
        \end{tikzpicture}
    \end{subfigure}\\[-20pt]
% 
    % third row 
    % \hspace{-25pt}
    % \hspace{70pt}%
    % \end{minipage}%
	% ─── row 3 ───────────────────────────────
    \begin{subfigure}{0.14\textwidth}
        \centering
        \begin{tikzpicture}[remember picture]
            \node at (0,0) {\includegraphics[scale=0.14]{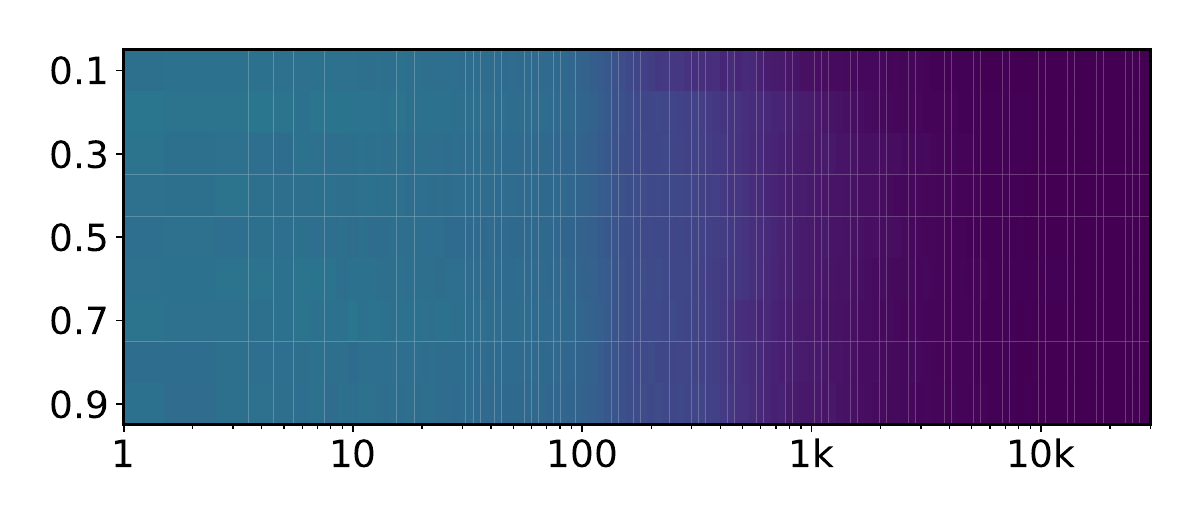}};
            % \node at (0,0) {\includegraphics[scale=0.14]{figures/kl_masked_completion_GT_in_dist_MC10Pos10_order1_L4H4d32_T50_heatmap_avg.pdf}};
            \node[overlay,scale=0.7,rotate=90] at (-2.4, 0.0) { \new{\textbf{StatS-Pos Var}}};
            \node[overlay, scale=0.5,rotate=90] at (-2.0, 0.0) { \textbf{(\mixexp{10}) }};
            
            \node[scale=0.8,rotate=90] at (-1.5, 0.05) { \textbf{{$\dvr$}}};            
            
            \node[scale=0.8] at (0.0, -1.2) { \textbf{ }};
            \node[scale=0.8] at (0.0, -0.9) {Iterations};
            
        \end{tikzpicture}
    \end{subfigure} &%
    \begin{subfigure}{0.14\textwidth}
        \centering
        \begin{tikzpicture}[remember picture]
            \node at (0,0) {\includegraphics[scale=0.14]{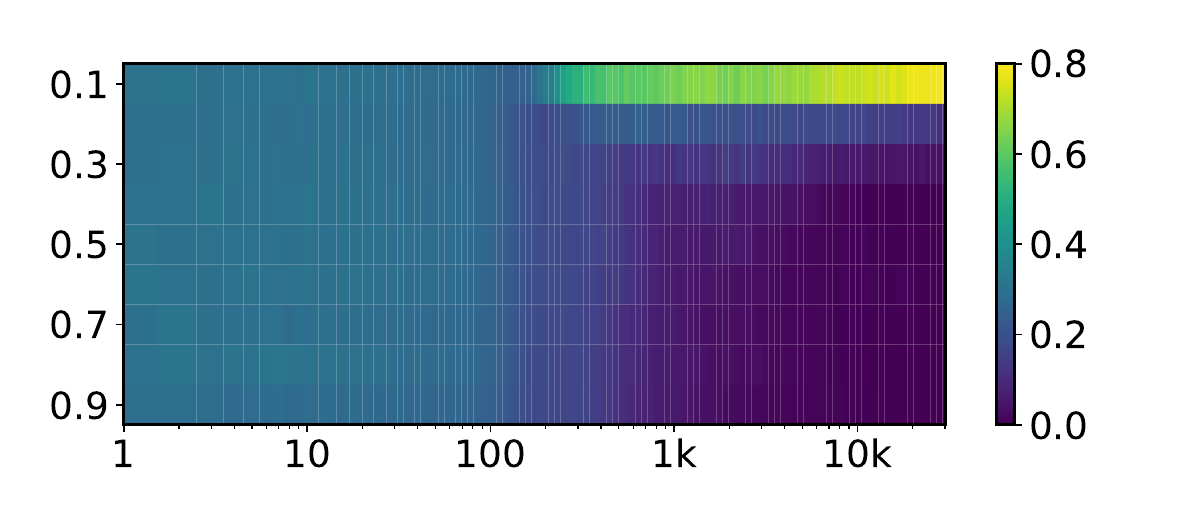}};
            % \node at (0,0) {\includegraphics[scale=0.14]{figures/kl_masked_completion_GT_out_dist_MC10Pos10_order1_L4H4d32_T50_heatmap_avg.pdf}};
            % \node[scale=0.8,rotate=90] at (-2.7, 0.0) { \textbf{\tina{$\mixexp$} }};  
            \node[scale=0.8] at (0.0, -1.2) { \textbf{ }};
            \node[scale=0.8] at (0.0, -0.9) {Iterations};
            
        \end{tikzpicture}
    \end{subfigure} &%
    \begin{subfigure}{0.14\textwidth}
        \centering
        \begin{tikzpicture}[remember picture]
            \node at (0,0) {\includegraphics[scale=0.14]{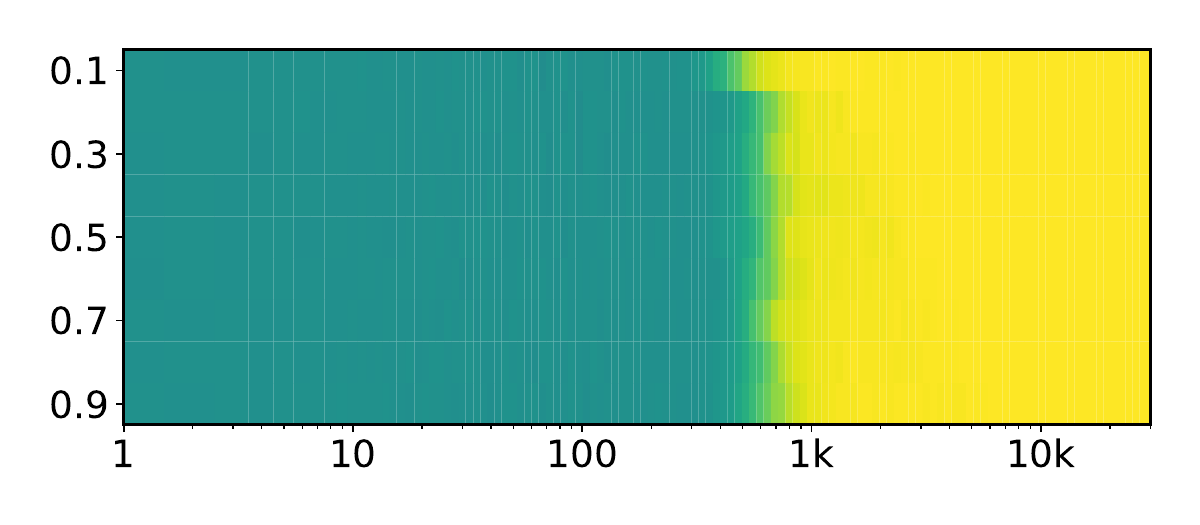}};
            % \node at (0,0) {\includegraphics[scale=0.14]{figures/kl_masked_completion_GT_in_dist_MC10Pos10_order1_L4H4d32_T50_heatmap_avg.pdf}};
            % \node[scale=0.75,rotate=90] at (-2.6, 0.0) {\new{ \textbf{Stat-Pos Var} }};
            % \node[scale=0.45,rotate=90] at (-2.3, 0.0) { \textbf{{($\mixexp{10}$)} }}; 
            % \node[scale=0.6,rotate=90] at (-1.9, 0.15) { \textbf{{$\dvr$}}};            
            
            \node[scale=0.8] at (0.0, -1.2) { \textbf{ }};
            \node[scale=0.8] at (0.0, -0.9) {Iterations};
            
        \end{tikzpicture}
    \end{subfigure} &%
    \begin{subfigure}{0.14\textwidth}
        \centering
        \begin{tikzpicture}[remember picture]
            \node at (0,0) {\includegraphics[scale=0.14]{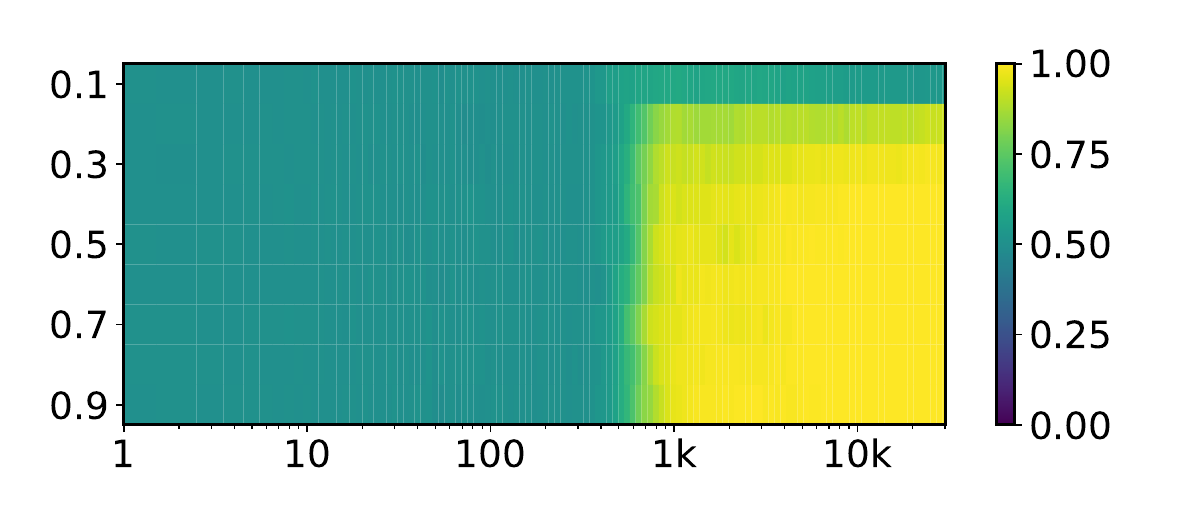}};
            % \node at (0,0) {\includegraphics[scale=0.14]{figures/kl_masked_completion_GT_out_dist_MC10Pos10_order1_L4H4d32_T50_heatmap_avg.pdf}};
            % \node[scale=0.8,rotate=90] at (-2.7, 0.0) { \textbf{\tina{$\mixexp$} }};  
            \node[scale=0.8] at (0.0, -1.2) { \textbf{ }};
            \node[scale=0.8] at (0.0, -0.9) {Iterations};
            
        \end{tikzpicture}
    \end{subfigure} &%
    \begin{subfigure}{0.14\textwidth}
        \centering
        \begin{tikzpicture}[remember picture]
            \node at (0,0) {\includegraphics[scale=0.14]{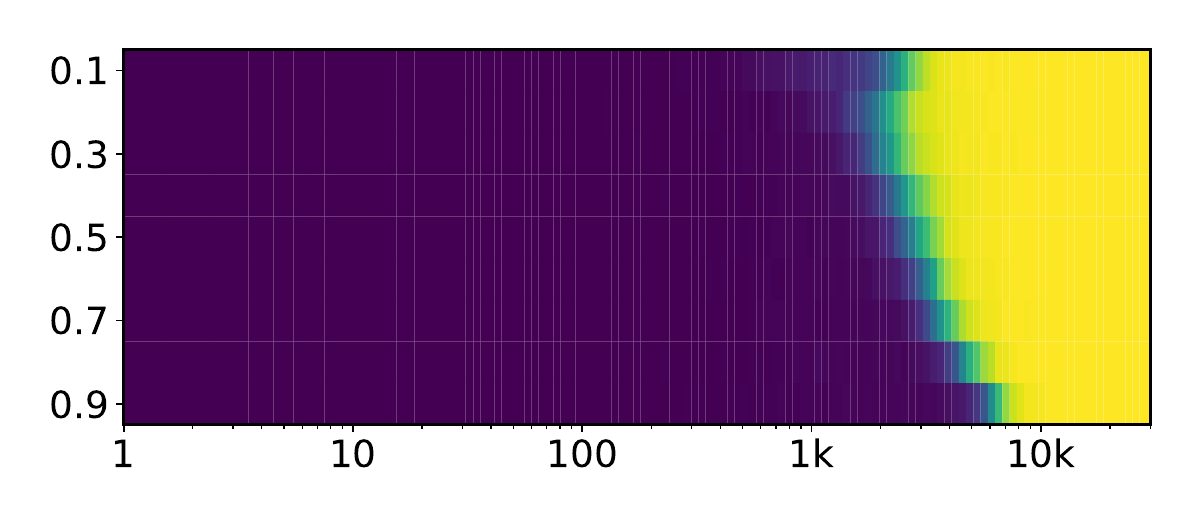}};
            % \node at (0,0) {\includegraphics[scale=0.14]{figures/at_pos_is_bi_rate_in_dist_MC10Pos10_order1_L4H4d32_T50_heatmap_avg.pdf}};
            % \node[scale=0.8,rotate=90] at (-2.7, 0.0) { \textbf{\tina{$\mixexp$} }};  
            \node[scale=0.8] at (0.0, -1.2) { \textbf{ }};
            \node[scale=0.8] at (0.0, -0.9) {Iterations};

        \end{tikzpicture}
    \end{subfigure} &%
    % \hspace{50pt}%
    \begin{subfigure}{0.14\textwidth}
        \centering
        \begin{tikzpicture}[remember picture]
            \node at (0,0) {\includegraphics[scale=0.14]{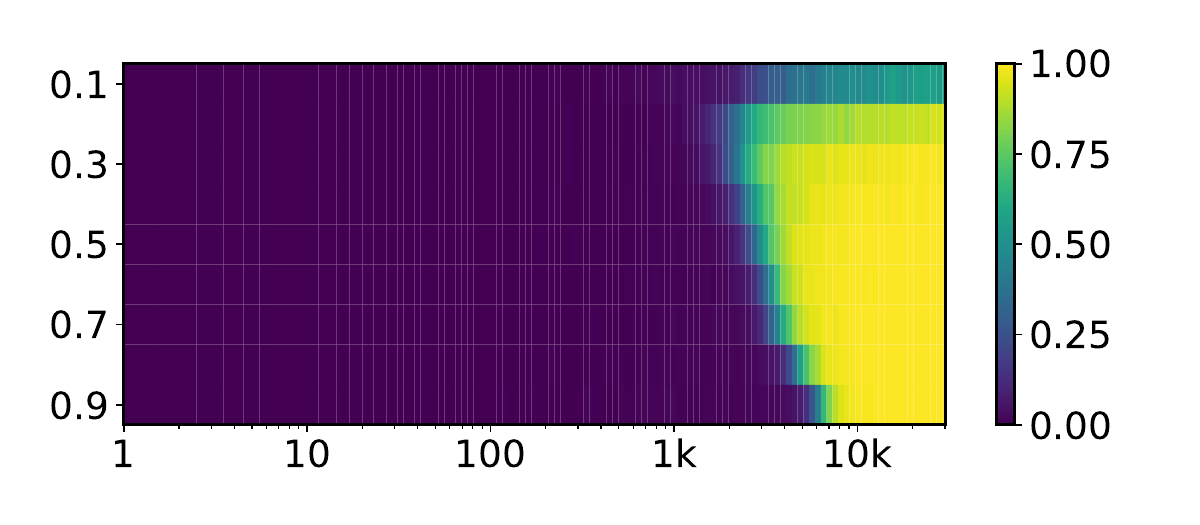}};            \node[scale=0.8] at (0.0, -1.2) {\textbf{}};
            % \node at (0,0) {\includegraphics[scale=0.14]{figures/at_pos_is_bi_rate_out_dist_MC10Pos10_order1_L4H4d32_T50_heatmap_avg.pdf}};            \node[scale=0.8] at (0.0, -1.2) {\textbf{}};
            \node[scale=0.8] at (0.0, -0.9) {Iterations};
            
        \end{tikzpicture}
    \end{subfigure}
  \end{tabular}
  \vspace{-0.0in}
  \caption{
  \new{\textbf{Generalization performance for varying diversity levels and context structures. }}
    Heatmaps show \textbf{(a)} $\KL$, \textbf{(b)} $\posacc$ and \textbf{(c)} $\factacc$ over training iterations (horizontal axis) and diversity level (vertical axis) for the three contextual structures defined in Sec.~\ref{sec:templates}: (top) $\mcexp{10}$ with only statistical variations, (middle) $\posexp{10}$ with only positional variations and (bottom) $\mixexp{10}$ with both types of variations across templates.
    %\mcexp{10}, \posexp{10}, and \mixexp{10} 
    % (see Sec.~\ref{sec:templates}).  
    Each metric is displayed for in‑distribution (ID, left column) and out‑of‑distribution (OOD, right column) fact–template pairs.  
    Brighter colors denote higher \emph{loss} for $\KL$ and higher \emph{accuracy} for $\posacc$ and $\factacc$. Increasing diversity can slow down convergence, whereas very low diversity leads to OOD failure on one or multiple metrics. %\emph{statistical or/and factual} OOD failure. 
    See Sec.~\ref{sec:evaluation_metrics} for metrics and Sec.~\ref{sec:results} for details. \looseness=-1
  % \textbf{Statistical and factual generalization in the synthetic setup of Sec.~\ref{sec:setup}.} \new{Heatmaps of \textbf{(a)} statistical and  \textbf{(b)} factual generalization over training iterations (horizontal axis) and diversity level (vertical axis) for the template families \mcexp{10}, \posexp{10} and \mixexp{10} defined in Sec.~\ref{sec:templates}, for sequences drawn from (left) in‑distribution (ID) and (right) out‑of‑distribution (OOD) fact-tempelate combinations. We measure statistical generalization by $\KL$ and factual recall by $\factacc$. Brighter colors indicate higher accuracy. Higher diversity systematically delays convergence, while very low diversity causes catastrophic OOD failure. See \ref{sec:results} for detailed discussion.}   
    }
    \vspace{-0.155in}
  \label{fig:heatmaps_main}
\end{figure}

%\vspace{-0.3in}
\subsection{Summary of methodology and findings}
%Motivated by these,  we introduce a 
% \vspace{-0.1in}
Our data framework factorizes sequences into a \emph{statistical} stream and a \emph{factual} stream.
We model the statistical stream as a mixture of $N$ \emph{template} distributions over token sequences. Each template {generates a sequence from a template-dependent Markov Chain (MC) distribution} and reserves two designated positions for fact insertion; these positions may vary across templates to capture different structural patterns. The factual stream consists of \emph{source--target} token pairs drawn from distinct vocabularies that are inserted in a template's designated positions to generate complete sequences. This gives precise and independent control along two axes: (a) \textbf{Diversity level}—how many different templates each fact appears in during training, and (b) \textbf{Contextual structure}—how template formats vary in their statistical and positional properties. See Fig.~\ref{fig:setup}-(A,B).\looseness=-1

We train small transformers on this data and evaluate generalization through carefully designed prompts, testing three distinct aspects:
\textbf{(1) Factual recall accuracy}—the model's ability to produce the correct fact target for a given source in the prompt; \textbf{(2) 
{\Structure} accuracy}—adherence to the specific positional patterns of the prompt template; and \textbf{(3) Statistical accuracy}—
adherence to the statistical distribution of the prompt template. We evaluate performance on both \textbf{in-distribution (ID)} {data}, where fact-template combinations match training data, and \textbf{out-of-distribution (OOD)} {data}, where facts appear in novel templates. {See Sec. \ref{sec:setup} for framework details.} 
%\ct{just an idea to have the findings very telegraphic with added references to Sec and figs. don't care about format}
%
We summarize our key findings  \new{(also see Fig.~\ref{fig:setup}-(C,D))}: \looseness=-1

\vspace{5pt}
\begin{tcolorbox}[breakable, colback=gray!2, colframe=gray!40, boxrule=0.3pt, left=3mm, right=3mm, top=1mm, bottom=1mm]
% \textbf{Key Findings:}
\textbf{Impact of diversity}. {The diversity of the training data has several effects on generalization: \looseness=-1 %\tina{4 is not diversity related...}
}
\vspace{5pt}
\begin{enumerate}[leftmargin=20pt, topsep=2pt, itemsep=1pt,label=\textbf{\arabic*.}]
    \item Higher diversity can slow ID factual recall while leaving \new{ID} statistical accuracy largely unaffected. \textcolor{gray!80}{ (Fig.~\ref{fig:heatmaps_main} and Sec.~\ref{sec:ID}).}
    \item OOD effects depend on contextual structure: Low diversity can severely impact factual recall, statistical accuracy, or both. \textcolor{gray!80}{(Figs. \ref{fig:setup}-\ref{fig:heatmaps_main}, and Sec. \ref{sec:OOD}).}
    % \item Diversity level creates temporal trade-offs: intermediate levels optimize limited training budgets, while high diversity achieves superior long-term performance.  \textcolor{gray!80}{Fig.~\ref{fig:div_tradeoff} and Sec.~\ref{sec:OOD}.}
    \item {Data diversity introduces a cost-benefit dynamic: Higher diversity guarantees better performance but only when provided with sufficient training budget. With limited training time, intermediate diversity levels are optimal.\looseness=-1} \textcolor{gray!80}{(Figs.~\ref{fig:heatmaps_main}-\ref{fig:div_tradeoff} and Sec.~\ref{sec:OOD})}.
    % \new{Diversity level introduces a trade-off: with limited training budget, intermediate diversity levels achieve the optimal performance, while high diversity has guaranteed best performance, albeit only with longer training. \textcolor{gray!80}{(Fig.~\ref{fig:div_tradeoff} and Sec.~\ref{sec:OOD}).}}
    
    \item {The learning dynamics of the two streams \new{(statistical and factual)} are coupled: Increased complexity in one stream, delays learning of both streams.}
    % Factual and statistical learning are intertwined: factual recall begins only after positional learning, \new{and a more complex statistical or factual stream, delays the learning of the other component as well.}
    % and more statistical complexity delays factual learning.  
    \textcolor{gray!80}{(Fig.~\ref{fig:params} and Sec.~\ref{sec:param_ablation}).}
\end{enumerate}

\vspace{10pt}
\textbf{Optimization bottlenecks.}
\new{Architectural parameters that achieve perfect OOD generalization exist regardless of diversity level. However, diversity shapes the optimization path, determining whether and how quickly training reaches those parameters. Through targeted interventions, we localize the bottlenecks to specific architectural modules.
% While diversity does not affect whether a generalizing solution exists, it shapes the optimization path: it determines whether, and how quickly, training reaches that solution, and module-wise interventions localize the bottlenecks to specific components.
% Although a generalizing minimizer exists for all diversity levels, training the models with controlled interventions reveals that the varied impacts of diversity can be traced to the optimization difficulty of different model components.
% Retraining the models with a series of controlled component-wise interventions reveals that the model's different generalization failures can be traced back to distinct modules. 
 \textcolor{gray!80}{(Figs.~\ref{fig:prob}-\ref{fig:intervention} and Sec.~\ref{sec:bottleneck}).} %Specifically:
 }
\vspace{5pt}
\begin{enumerate}[leftmargin=20pt, topsep=2pt, itemsep=1pt,label=\textbf{\arabic*.}, start=5]
    \item {Training on low-diversity data produces poor last-layer features; retraining  \new{only} the unembedding layer on high-diversity data fails to restore generalization. 
    % Training on low-diversity data produces corrupted last-layer features: Retraining the final unembedding layer on high-diversity data cannot recover generalization. 
    }

    \item \new{
    %The poor feature quality for statistical and positional generalization stems from the optimization difficulty of the attention and embedding modules. 
    \new{Poor optimization of the attention and embedding modules degrades statistical and positional generalization.}
    % The poor feature quality for statistical/position generalization primarily stems from the model's struggle to learn effective attention patterns and token embeddings.
    % traces back to the poor optimization of the attention patterns and/or the token embedding matrix. 
    }

    \item {
    %Factual recall required both well-optimized features and unembedding layer. Improving either in isolation does not recover factual accuracy.  
    The embedding and unembedding layers are jointly responsible for factual-recall failures; improving either in isolation fails to recover performance, \new{but joint intervention on both helps with generalization.} %but jointly optimizing both succeeds. 
    % feature-only nor unembedding-only interventions are sufficient to recover factual recall; pairing clean features with a well-optimized unembedding restores facts across diversity levels.
    % 
    % 
    % \new{For factual accuracy, both the last-layer features and the unembedding layer are poorly optimized on low-diversity data: training interventions on feature-producing modules alone are insufficient, but when paired with a frozen, well-trained unembedding, model achieves boosted factual accuracy irrespective of diversity level. \tina{}
    % Training on low-diversity data produces both poor features and a poor unembedding layer for factual recall: interventions that target only the feature-producing modules do not improve factual recall. However,  paired with a well-trained unembedding layer, perfect factual accuracy can be achieved, regardless of the training data's diversity level.
    % No interventions on the features alone can fix the training difficulty on low diversity for factual recall. The optimization bottleneck is the joint optimiztion of features and the unembedding layer.
    % 
    % A well-trained unembedding layer is necessary but not sufficient for full factual recall. However, joint optimization of the unembedding and the features on low-diversity data does not learn one.
    }
    
    \item \new{The unembedding layer causes the slowdown in factual recall at high diversity; training features alone with fixed unembeddings shows no such slowdown.
    %The learning slowdown observed at high diversity is traceable to the unembedding layer's optimization; training the features alone shows no such slowdown at any diversity level.
    % The learning slowdown observed at high diversity can be traced to the optimization of the unembedding layer; training the features alone does not show any slowdown in learning factual recall regardless of data diversity.
    }
\end{enumerate}

\end{tcolorbox}

\section{{Data setup}}\label{sec:setup}

\vspace{-0.05in}
% \subsection{Composition of facts and templates}
% \vspace{-0.1in}
We consider sequences $\xb=(x_1,...,x_\seqlen)$ of length $\seqlen$, produced by mixing two independent sources: a \emph{factual} and a \emph{statistical} stream, with each token $x_t$ belonging to a global vocabulary $\vset=[\vocabsize]$. 
% \newpage

\paragraph{Factual stream}
We specify $\factsize$ atomic facts $\factset:=\{(\source_k,\target_k)\}_{k\in[K]}$ each given as an ordered \emph{source-target} pair of distinct tokens $\source_k,\,\target_k\in\vset$. 
The vocabulary of facts is $\vk= \{\source_k\}_{k\in[K]}\bigcup\{\target_k\}_{k\in[K]}$. %We denote $f:\sourceset\to\targetset$ the deterministic one-to-one function of associations, so each source token uniquely specifies its corresponding target $f(\source_k)=\target_k$.

\paragraph{Statistical stream (Templates)} 
Background sequences $\zb=(z_1,\ldots,z_T)$ to which facts are later embedded are sampled from a mixture of  $\tmplnum$ template distributions: $\zb\; \sim\; \dist=\frac1\tmplnum\sum_{n=1}^N\dist_n$. Formally,  a template $\dist_n$ is a probability distribution over length-$\seqlen$ sequences drawn from the \emph{generic vocabulary} $\vmc:=\vset\setminus\vk$ {of size $\vmsize=|\vmc|$}, with two marked positions $\pospair=(\posSource,\posTarget)\in [\seqlen]\times[\seqlen]$, $\posSource \, {\leq}\, \, T/2 < \posTarget$ reserved for the source/target fact tokens. %\footnote{Each $\dist_{n}$ represents a distinct syntactic pattern that when put together, they give the model a variety of “contexts’’ without introducing any factual information of their own.}
For concreteness, {unless otherwise stated,} we assume the generic tokens are produced by a first-order Markov chain (MC).
% \footnote{%\tina{if time add higher order to app}
% %The choice of statistical $n$-gram generation in the form of MC is convenient for measuring statistical accuracy. 
% One could adopt variants of traditional MCs (e.g., the contextual bigram \cite{ren2024learning}) without affecting our main findings. Similarly, we choose order 1 to maintain reasonable model size and training lengths.
% %, as higher orders would not provide additional insights for our analysis.
% }
{Thus, a template $\dist_n$ is defined by: \textbf{(i)} a {transition matrix} $\transitionmat_{n}\in\Delta^{\vmsize\times\vmsize}$, and \textbf{(ii)} a placeholder {position pair} $\pospair_n$. We sample  from $\dist_n$ {by generating a length $T-2$ MC sequence from $\transitionmat_n$, and reserving the positions $\pospair_n$ for facts.}} 
% Thus, a template $\dist_n$ is characterized by: \textbf{(i)} a collection of \textbf{transition matrices} $\transitionset_n=\{\Pb_{n,i}\}_i\subset\Delta^{|\vmc|\times|\vmc|}$, and \textbf{(ii)} a set of placeholder \textbf{position pairs} $\positionset_n=\{\pospair_{n,j}\}_j$. We sample sequence $\zb$  from $\dist_n$ {by drawing a transition matrix $\transitionmat\sim\transitionset_n$, generating an MC sequence of length $T-2$ according to $\transitionmat$, and selecting position pair $\pospair \sim \positionset_n$ to specify the two positions reserved for fact tokens.}
%by drawing a transition matrix $\transitionmat\sim\transitionset_n$, a MC sequence realization of length $T-2$ according to $\transitionset_n$, and $\pospair \sim \positionset_n$ to specify the two position tokens reserved for facts. %The positions $\pospair=(\posSource,\posTarget)$ are reserved for the subsequent insertion of a source-target fact pair in $\zb$, while the rest of the tokens in the sequence are generated by the order-one MC with transition matrix $\transitionmat$ over the generic vocabulary $\vmc$. \tina{simplify a bit later...}

\paragraph{Final sequence generation} We draw sequences $\xb$ that combine facts with templates. First, {we select a template $n$  with (say) position $\pospair=(\posSource,\posTarget)$},  and  sample  $\zb\sim\dist_n$\,.
%we sample a sequence $\zb\sim\dist$ by picking a random template $n\in[N]$ and drawing $\zb\sim\dist_n$. 
% Recall that the template from which we sample is associated with a position $\pospair=(\posSource,\posTarget)$. 
{Then, we insert a chosen fact $(\source,\target)$ in positions} 
%draw a random
% fact $(\source,\target)\sim\factset$ and insert the source and target tokens $(\source,\target)$ in those designated positions 
$(\posSource,\posTarget)$ to form the final sequence $\xb$, i.e., $x_{\posSource}\leftarrow\source$, $x_{\posTarget}\leftarrow\target$ and $x_t\leftarrow z_t$ for $t\notin \{\posSource,\posTarget\}$.\looseness=-1

%\ct{maybe highlight once more that template does not mean fixed sequence}

% \vspace{-0.1in}
\subsection{Level of diversity}\label{sec:div}
% \vspace{-0.1in}
% Diversity level refers to the number of different templates in which each fact is encountered during training: 
The \textbf{diversity-level parameter $\dvr\in(0,1)$} is such that every fact-pair \new{$(\source_k,\target_k),\,k\in[K]$} appears embedded in $\dvr \cdot N$ different templates during training. Larger $\dvr$ indicates a higher diversity level.
To describe template-fact pairings  
during training, we define binary \emph{in-distribution (ID) exposure mask}
$
\exposmat_{\rm in}\in\{0,1\}^{\tmplnum\times\factsize}$ with 
$\exposmat_{\rm in}[n,k]\;=\;
\mathbbm{1}\!\bigl[(\source_k,\target_k)\;\text{occurs in training in a sequence drawn from}\;\dist_n\bigr].
$ Thus, the $k$-th column lists all templates in which the
$k$‑th fact appears during training. 
% The number of non‑zero entries in this column measures that fact's \emph{diversity level}.
 We set $\dvr \cdot N$ entries of the \(k\)-th column of \(\exposmat_{\rm in}\) to~1 uniformly at random. The rest 
 \((1-\dvr)\tmplnum\) templates are unseen at training for the specific $k$-th fact,  forming its
OOD set. {See Fig.~\ref{fig:setup}-B for an example with $\dvr$ = 0.2}.%\new{See Figure \ref{fig:setup} for an example of high- and low- diversity exposure mask $\exposmatin$.}%\ct{If never use Eout, I would remove:}We thus let denote the \emph{out‑of‑distribution (OOD) exposure matrix}
\subsection{Contextual structure}\label{sec:templates}
% \vspace{-0.1in}
{We define contextual structure as the specific configuration of templates, particularly how their two key constituents vary: fact placeholder and statistical contexts.}  We consider the following setups, {all using exactly $\tmplnum$ templates, to  allow their direct comparison across identical $\dvr$ values}: \looseness=-1
% ct{this part is difficult to understand: drop the whole thing and go to structures directly?} To isolate the effects of structure from diversity, we create setups that vary structural characteristics while maintaining identical diversity levels. We achieve this by systematically manipulating position pairs $\{\positionset_n\}$ and transition matrices $\{\transitionset_n\}$. 
%
% Concretely, we  associate each template $n \in [\tmplnum]$  with exactly one transition matrix and one position pair ($|\transitionset_n| = |\positionset_n| = 1$) and define the following complementary setups with distinct contextual structures:

%We define contextual structure as the specific configuration of how fact pairs are embedded within templates, comprising both the positional arrangement (where facts appear {specified by $\positionset_n$}) and the statistical context (the statistical patterns of surrounding tokens {specified by $\transitionset_n$}). To isolate the effects of structure from diversity, we create experimental setups that vary structural characteristics while maintaining identical diversity levels $\dvr$. We achieve this by  trading-off the two defining characteristics of templates: transition matrices {\{$\transitionset_n$\}} and position pairs {\{$\positionset_n$\}}.
%

\paragraph{1) Position-varied Structure (\posexp{\tmplnum})} All templates use the same statistical distribution, i.e., $\transitionmat_n = \transitionmat$ for all $n \in [\tmplnum]$, while fact placeholders $\pospair_n$ vary across templates $n$.
% In this setup, we maintain similar statistical structure across templates by using an identical transition matrix ($\transitionmat_n = \transitionmat$ for all $n \in [\tmplnum]$), while varying the composition format of facts in each template by assigning distinct position placeholders $\pospair_n$ for each $n$. \looseness=-1

\paragraph{2) Statistical-varied Structure (\mcexp{\tmplnum})} Each template has its own  distinct transition matrix $\transitionmat_n$, while using the same fact placeholders ($\pospair_n = \pospair$ for all $n \in [\tmplnum]$).
%In this setup, we keep the insertion pattern fixed while varying the underlying \emph{statistical context} in which facts are embedded. All templates use the same placeholder pair ($\pospair_n = \pospair$ for all $n \in [\tmplnum]$), but each has its own distinct transition matrix $\transitionmat_n$.

{\paragraph{3) Mixed-varied Structure (\mixexp{\tmplnum})} Each template has its own distinct transition matrix $\transitionmat_n$ and distinct position placeholder $\pospair_n$.
% Here, we vary both the statistical and compositional rule by assigning a distinct transition matrix $\transitionmat_{n}$ \emph{and} a distinct placeholder pair $\pospair_{n}$ to each template $n\in[\tmplnum]$. 
% Thus the statistical context surrounding a fact and the slot in which the fact is inserted both change from one template to the next. 
% This lets us test the model when both the positional and contextual cues shift simultaneously.
}

%\ct{these are the two extremes, we also study intermediate settings (Fig 1 row c see appendix)}
% In our experiments, we simplify the template design by letting each template $n \in [\tmplnum]$ be associated with exactly one transition matrix and one position pair. That is, for every $n$, we set $|\transitionset_n| = |\positionset_n| = 1$, and we write $\transitionmat_n$ and $\pospair_n$ to refer to the unique transition matrix and placeholder position pair associated with template $n$.

%  We consider two complementary setups: \tina{if time add ablation to app and refer to it}
 
% \paragraph{1) \posexp \tina{}} In this setup, we choose similar statistical structure across templates, i.e., $\transitionmat_n =: \transitionmat$ for all $n \in [\tmplnum]$, while varying the composition format of facts in each template by choosing distinct position placeholders $\positionset_n$ for each $n$.

% \paragraph{2) \mcexp \tina{} } In this setup, we keeps the insertion pattern fixed while varying the underlying \emph{statistical context} in which facts are embedded. That is all templates use the same placeholder pair, i.e., $\pospair_n =: \pospair$ for all $n \in [\tmplnum]$, but each has its own distinct transition matrix $\transitionmat_n$. 

\tina{fact and stata fig}
\subsection{{Evaluation metrics: In- and out-of-distribution performance}}
\label{sec:evaluation_metrics}
% \vspace{-0.in}
{

To systematically evaluate the model, we measure its generalization using controlled probing. For sequence $\xb$, drawn from template $n\in[\tmplnum]$, that contains a given fact pair $(\source,\target)$, we prompt the model with its first half, $\xb_{1:\seqlen/2}$, \new{containing the source token $\source$}, and have it auto-regressively complete the second half, i.e., $\hat{\xb}_{\seqlen/2+1:\seqlen}$. A successful completion requires the model to simultaneously demonstrate: (1) \textbf{structural understanding}, by placing a fact token only at the designated position $\posTarget_{n}$ and generic tokens elsewhere; (2) \textbf{factual recall}, by generating the correct target token $\target$ in the completion; and (3) \textbf{statistical consistency}, by ensuring the generated generic tokens follow the template's statistical rules.}

{
We quantify each of these aspects with the following metrics. For positional and factual evaluation, we focus on simple zero-one accuracy in the main text and defer entropy-based counterparts to App.~\ref{app:loss_metrics}.
}
 {
For each metric, we distinguish between in-distribution (ID) and out-of-distribution (OOD) performance: We evaluate the model on sequences from two disjoint sets of template-fact pairs $(n,k)$: those seen during training (ID, where ${\exposmatin}[n,k]=1$) and those held out during training (OOD, where ${\exposmatin}[n,k]=0$).\looseness=-1
% For each metric, we distinguish between in-distribution (ID) and out-of-distribution (OOD) performance. We evaluate the model on template-fact pairs $(n,k)$ that were seen during training (where ${\exposmatin}[n,k]=1$) separately from the template-fact pairs that were held out (where ${\exposmatin}[n,k]=0$).
}
% 
% {To systematically evaluate the model's learning, we measure its generalization across three distinct capabilities: (1) understanding the sequence structure, (2) recalling facts, and (3) following statistical patterns. We assess these abilities using a controlled probing task where, for a given template $n\in[\tmplnum]$ and a fact pair $(\source,\target)$ and a sequence $\xb$ drawn from these two, we prompt the model with the first half of the sequence, $\xb_{1:\seqlen/2}$, and have it auto-regressively generate the second half, $\hat{\xb}_{\seqlen/2+1:\seqlen}$. A successful completion requires the model to: (i) place the correct target token $\target$ at the position specified by the template, $\posTarget_{n}$, and (ii) generate generic tokens that follow the statistical rules of that template.}
% 
% We define three metrics to quantify these aspects. For positional and factual evaluation, we focus on simple accuracy metrics in the main text and defer their continuous, loss-based counterparts to Appendix~\ref{app:loss_metrics}.
% 
\begin{enumerate}[leftmargin=0pt, itemindent=1.2em,noitemsep]
    \item \textbf{Position accuracy}  {assesses learning the composition of the two streams: placing a fact token from $\vk$ \emph{only} at the designated target position and generic tokens $\vmc$ everywhere else. 
    Formally:} %\tina{consider in-line}
    % 
    % We define positional accuracy as the average success rate across two conditions: correctly placing a fact token from $\vk$ at the target position, and correctly placing generic tokens from $\vmc$ at all other positions in the generated sequence. 
    \begin{align}\label{eq:postion_acc}
    % $        
    \posacc := \tfrac{1}{2} \big( \mathbbm{1} [\hat{x}_{\posTarget_{n}} \in \vk] + \tfrac{1}{\seqlen/2-1} \textstyle\sum\nolimits_{t > \seqlen/2, t \neq \posTarget_{n}} \mathbbm{1} [\hat{x}_{t} \in \vmc] \big).
    % $
    \end{align}
    
    \item \textbf{Factual accuracy} 
     {measures factual recall, requiring that the only fact token in the completion be the correct target $\target$. Concretely,} %\tina{changed metric}
    % 
    % This metric directly measures the model's ability for factual recall by checking whether the correct target token $\target$ appears in the completion $\hat{\xb}_{>\seqlen/2}$, and no other fact tokens are present. Concretely, 
    \begin{align}\label{eq:fact_acc} 
        % $
        \factacc := \mathbbm{1} \big[ \{\hat{x}_t\}_{t > \seqlen/2} \cap \vk = \{\target\} \big].
        % $
    \end{align}
    % \begin{align}\label{eq:fact_acc}
    %     \factacc := \mathbbm{1}\left[\exists t,\,\hat{x}_{t}=b, \text{and} \forall t, \hat{x}_{t} \not\in\vk/\{b\}\right].
    % \end{align}
    % \begin{align}\label{eq:fact_acc}
        % \factacc := \mathbbm{1}\left[\hat{x}_{\posTarget_{n}} = \target\right].
    % \end{align}
    
    \item \textbf{Statistical loss}
     {evaluates statistical generalization.  We compute the KL divergence between the model's predictions and the ground-truth, averaged over the set of positions $\mathcal{G}$ where a generic token is generated. For each $t \in \mathcal{G}$, we compare the model’s distribution over generic tokens (after masking the fact tokens), ${\pb}_t\in\Delta^{\vmsize}$, against the ground-truth $\pb^*_t$, the row in the template's transition matrix $\transitionmat_n$ specified by the previous generic tokens in the sequence.} \looseness=-1
    \begin{align}\label{eq:KL}
        % $
        \KL := \tfrac{1}{|\mathcal{G}|} \textstyle\sum_{t \in \mathcal{G}} \mathrm{KL}\big( {\pb}_t \,\|\, \pb^*_t \big),
        % $
    \end{align}
    % To evaluate how well the model captures a template's statistical structure, we compute the KL divergence between the model's predictions and the template's ground-truth distribution. This is averaged over the set of positions $\mathcal{G}$ where a generic token was generated (i.e., $\hat{x}_t \in \vmc$). For each position $t \in \mathcal{G}$, we compare the model’s predicted distribution over generic tokens (after masking the fact tokens) $\tilde{\pb}_t$ to the true distribution $\pb^*_t$, specified by the template transition matrix $\transitionmat_n$ and the previous generic tokens in the completion: 
% 
% 
% 
    % To evaluate how well the model captures a template's statistical structure, we measure the divergence between its predicted distribution over generic tokens (after masking fact tokens) and the ground-truth distribution specified by $\transitionmat_n$. For all positions $t$ in the completion that $\mathcal{G}=\{\hat{x}_t\in\vmc\}$, we compute the KL divergence between the model's predicted next-token distribution $\tilde{\pb}_t$, and the true template distribution $\pb^*_t$. The statistical loss is the average KL divergence across these positions:
    % \begin{align}\label{eq:KL}
    %     \KL := \frac{1}{|\mathcal{G}|} \sum\nolimits_{t \in \mathcal{G}} \mathrm{KL}\big( \tilde{\pb}_t \,\|\, \pb^*_t \big).
    % \end{align}
    % where $G_n$ is the set of generic token positions in the generated sequence.
\end{enumerate}

%\vspace{-0.2in}
\section{{Impacts of diversity}}\label{sec:results}
%\vspace{-0.15in}
% ------------------------------------------------------------------
\tina{posacc fig}

{
\paragraph{Experimental setup} We use a 4‑layer decoder-only Transformer, \citep{radford2018improving} trained auto-regressively with the standard next‑token prediction loss. Each training sequence has length $\seqlen=50$. We use a template pool of size $\tmplnum=10$ and a fact set of $\factsize=100$ source-target pairs. For the MC, we choose a generic vocabulary set of size $\vmsize=3$. We sweep diversity  $\dvr$ from $0.1-0.9$ and train the model for \new{$30k$ iterations}. Unless otherwise noted, metrics are averaged over three random initializations over both model initialization, and data generation \new{(including template definition and ID/OOD splits)}.
% Models were trained on a single Tesla V100-SXM2 GPU (16GB memory).
For each template configuration of Sec.\ref{sec:templates}, $\KL$, $\posacc$ and $\factacc$ are reported in Fig.~\ref{fig:heatmaps_main}. %, and $\posacc$ in Fig.~\ref{fig:position_acc}. %\tina{make one fig??}.
 Each figure shows the heatmap of the metrics across a grid of diversity levels $\dvr\in(0,1)$ and training iterations, for sequences randomly drawn from ID/OOD template-fact distributions. %We discuss the results below. %In the following sections, we will have a more fine-grained look at these heatmaps.
 \new{In App.\ref{app:exp_size}, we discuss the impact of model size by showing similar results on 1- and 10-layer models.} 
}

% ------------------------------------------------------------------------------------------------------

% ------------------------------------------------------------------------------------------------------
%\vspace{-0.1in}
\subsection{Impact of diversity on in-distribution performance}\label{sec:ID}
%\vspace{-0.1in}
% \tina{move this somewhere} We analyze how context diversity influences the model's structural and factual accuracy in both ID and OOD settings. For structural accuracy, we track performance on both generic token positions and fact token positions (Figure~\ref{fig:structural_accuracy}); for factual accuracy, we track the recall metric (Figure~\ref{fig:factual_recall}). As before, we study both contextual structures, {\mcexp{10}} and {\posexp{10}}, this time across a grid of diversity levels $\dvr\in(0,1)$ and training iterations. 

%We analyse how context diversity influences the model’s structural and factual accuracy.  For structure, we track the \tina{generic positions} and \tina{fact position} (Fig. \ref{fig}); for factual accuracy, we track the recall metric \tina{} (Fig.\ref{fig}). As before, we study both contextual structures, $\mcexp$ and $\posexp{10}$, this time over a grid of diversity levels $\dvr\in(0,1)$ and training iterations, allowing a direct comparison of how increased context variety accelerates or impedes learning ID and OOD.

% \paragraph{ID performance: Statistical and structural learning is diversity-neutral, factual recall is diversity-sensitive}
Across contextual structures and full range of diversity levels, the model ultimately reaches perfect ID performance on all three metrics. Yet, the rate varies.
%distinctly among the metrics as a function of diversity.
%
As Figs.~\ref{fig:heatmaps_main}-(a,b) %and \ref{fig:position_acc}  
show, both $\KL$ and $\posacc$ converge to 0 and 1, respectively at nearly the same rate, largely unaffected by $\dvr$.  In contrast, $\factacc$ is sensitive to diversity: the more templates in which a fact appears in training, the longer the model requires to disentangle the correct source–target mapping from contextual patterns. Fig.~\ref{fig:heatmaps_main}-(c) illustrates this delay clearly: as $\dvr$ increases (lower rows in heat map), the yellow band marking perfect recall shifts rightward, signifying that additional training is needed to reach full accuracy in higher-diversity settings. \new{However, the severity of this slowdown depends on the contextual structure, with the \posexp{10} setup showing only a minimal effect.}%requiring additional training to reach full accuracy.

\subsection{Impact of diversity on out-of-distribution performance}\label{sec:OOD}
%\vspace{-0.1in}
On unseen template-fact distributions, the effect of diversity level depends critically on contextual structure: how fact positions and token statistics vary across templates.\looseness=-1
%\ct{}—how the two streams are composed. 

\vspace{2pt}
\paragraph{\Structure accuracy (Fig.~\ref{fig:heatmaps_main}-(b))} 
%As Fig.~\ref{fig:position_acc} shows, 
In the \mcexp{10} setup, where all templates share the same placeholder positions, the model separates generic and fact positions at nearly the same rate for every diversity level $\dvr$. In the \posexp{10} and \mixexp{10} setups, where there are more variations across templates on the position assignments, diversity plays a crucial role:
In extreme low-diversity settings (i.e., $\dvr = 0.1$ or $0.2$), progress in $\posacc$ slows down noticeably or fails. 
However, once diversity is moderate or high, the learning speed differences across $\dvr$ become negligible.

% As displayed in Fig.~\ref{fig:position_acc}, in the \mcexp{10} setup—where placeholder positions are fixed across templates—distinguishing the generic vs fact positions happens consistently at the same speed for all diversity levels. However, in the two other cases \posexp{10} and \mixexp{10} where there are more variations across templates on the position assignments, diversity becomes crucial for learning the structure efficiently. With extremely low levels of diversity like $\dvr=0.1,0.2$, the speed of improvement in $\posacc$ slows down significantly. \tina{in sm, we show that with much longer training time, it can eventually recover, however the speed is too slow....}. For moderate and high levels of diversity though the difference in speed of recovery becomes negligible.

\vspace{2pt}
\paragraph{Statistical loss (Fig.~\ref{fig:heatmaps_main}-(a))}
{In the \posexp{10} setup, where all templates share the same token statistics (i.e., identical transition matrices $\Pb_n$), the statistical loss is largely unaffected by the diversity level and converges at a consistent rate regardless of $\dvr$. In contrast, for the \mcexp{10} and \mixexp{10} setups, which feature statistical variations across templates, diversity becomes a critical factor. At low diversity levels ($\dvr<0.3$), prolonged training causes the model's predicted distribution to diverge from the ground truth, as displayed by the light bars in the top-right corners of the OOD heatmaps in Fig.~\ref{fig:heatmaps_main}-{(a)}}
% {The statistical loss in the \posexp{10} setup is almost unaffected by diversity level: all templates share the same statistics over the tokens (same transition matrices $\Pb_n$). Thus, despite the positional variations across templates, with high or low diversity training, the statistical loss always reaches zero at roughly the same rate. In contrast, in the \mcexp{10} and \mixexp{10} setups, \new{with statistical variations across the templates}, diversity becomes a critical factor for learning the underlying distributions: With low diversity $\dvr<0.3$, the model's distribution drifts away from the ground-truth with long-training as displayed by the light bars in the top-right corners of the OOD heatmaps in Fig.~\ref{fig:heatmaps_main}-{(a)}}

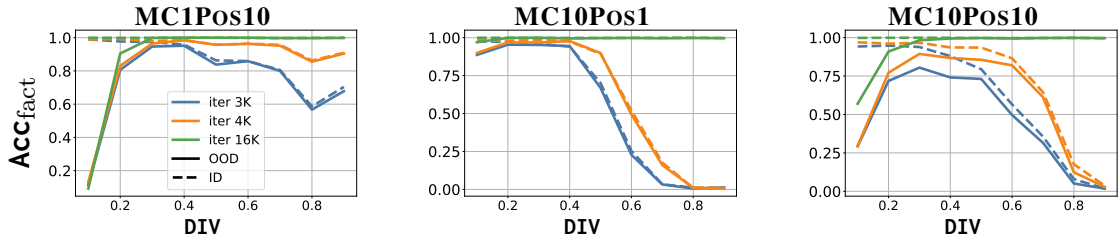
\begin{figure}[t]
  \centering
  \begin{tabular}{@{\hspace{-55pt}}c@{\hspace{50pt}}c@{\hspace{40pt}}c@{}}
  % ─── header row: one cell for col 1, one spanning cols 2–3 ─────────
  % \multicolumn{2}{c}{\fontsize{9pt}{8pt}\selectfont\textbf{(a) $\dvr=0.8$}}
  % & \multicolumn{2}{c}{\fontsize{9pt}{8pt}\selectfont\textbf{(b) $\dvr=0.1$}} \\[8pt]
  % ─── two rows of subfigures ───────────────────────────────
    % \hspace{-25pt}
    \begin{subfigure}{0.22\textwidth}
        \centering
        \begin{tikzpicture}[remember picture]
            \node at (0,0) {\includegraphics[scale=0.22]{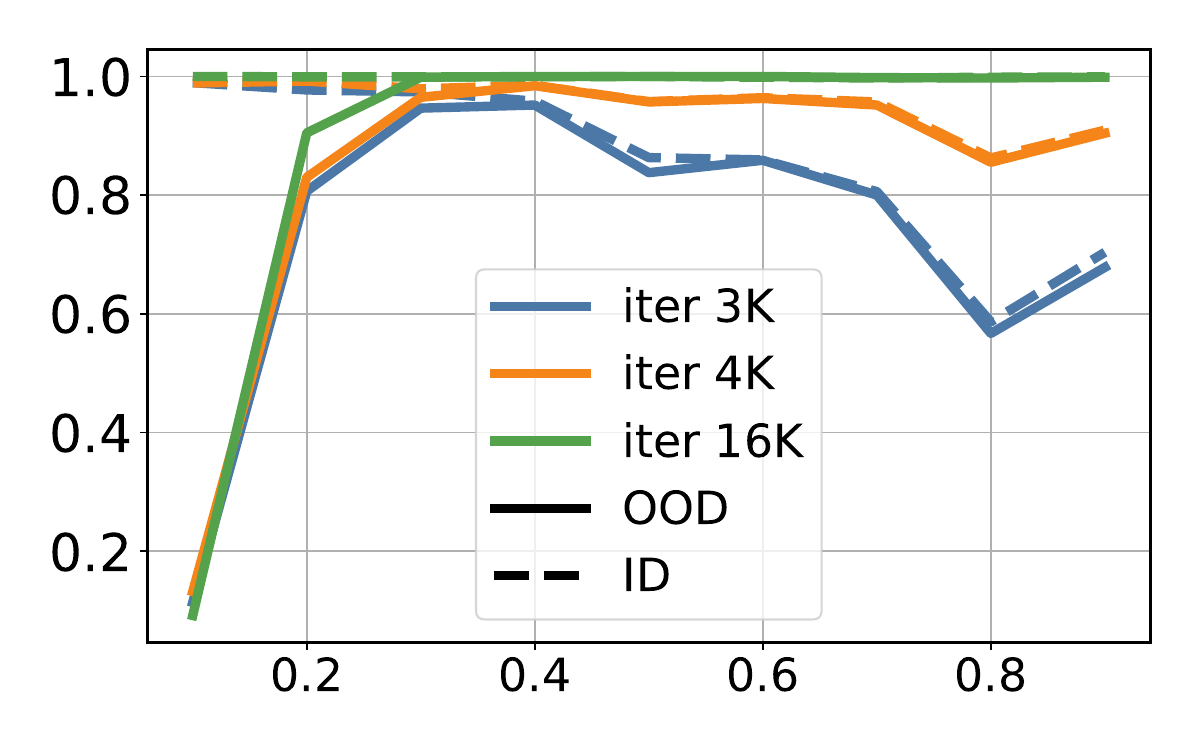}};
            % \node at (0,0) {\includegraphics[scale=0.22]{figures/at_pos_is_bi_rate_MC1Pos10_order1_L4H4d32_T50_monotonic.pdf}};
            % 
            % \node[scale=0.8] at (0.0, 1.8) { \textbf{(a) ID mask $\exposmatin$ }};
            \node[scale=0.9] at (0.0, -1.4) { \textbf{$\dvr$ }};
            \node[scale=1.2,rotate=90] at (-2.4,0.0) { \textbf{$\factacc$ }};
            \node[scale=0.9] at (0.0, 1.4) { \textbf{\posexp{10} }};
        \end{tikzpicture}
    \end{subfigure} &
    % \hspace{70pt}%
    \begin{subfigure}{0.22\textwidth}
        \centering
        \begin{tikzpicture}[remember picture]
            \node at (0,0) {\includegraphics[scale=0.22]{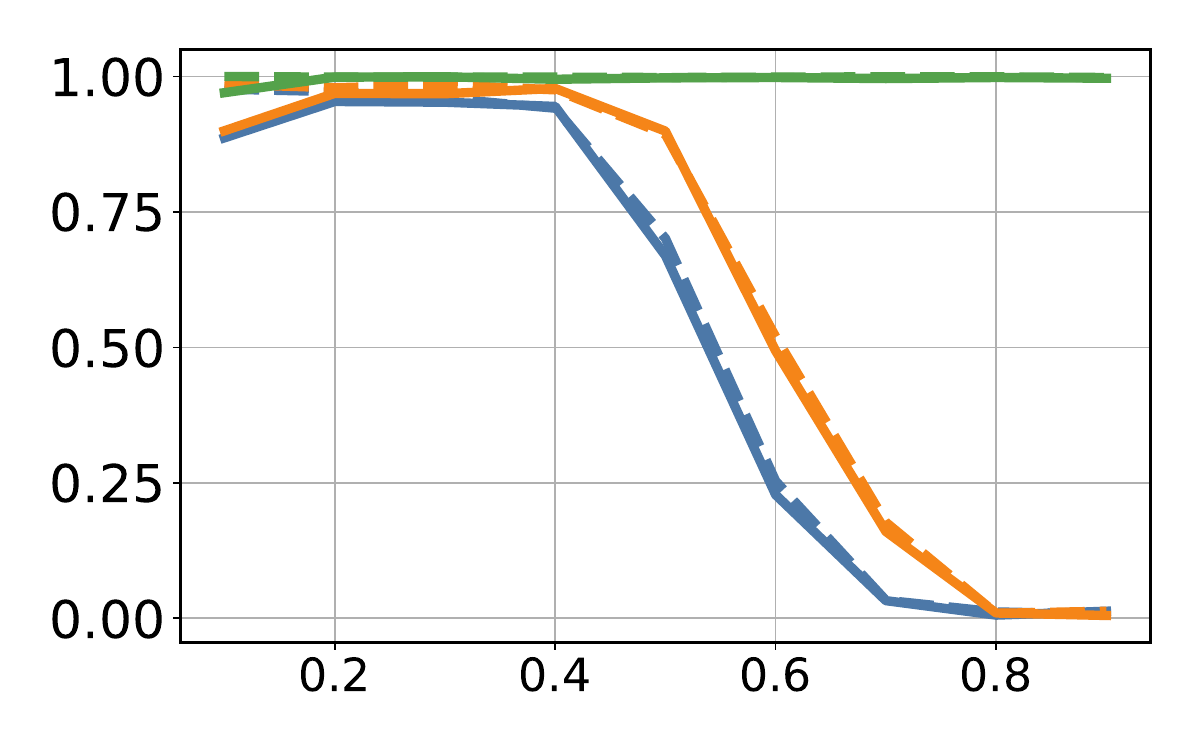}};
            % \node at (0,0) {\includegraphics[scale=0.22]{figures/at_pos_is_bi_rate_MC10Pos1_order1_L4H4d32_T50_monotonic.pdf}};
            \node[scale=0.9] at (0.0, -1.4) { \textbf{$\dvr$ }};
            \node[scale=0.9] at (0.0, 1.4) { \textbf{\mcexp{10}}};
        \end{tikzpicture}
    \end{subfigure} &
    % \hspace{50pt}%
    \begin{subfigure}{0.22\textwidth}
        \centering
        \begin{tikzpicture}[remember picture]
            \node at (0,0) {\includegraphics[scale=0.22]{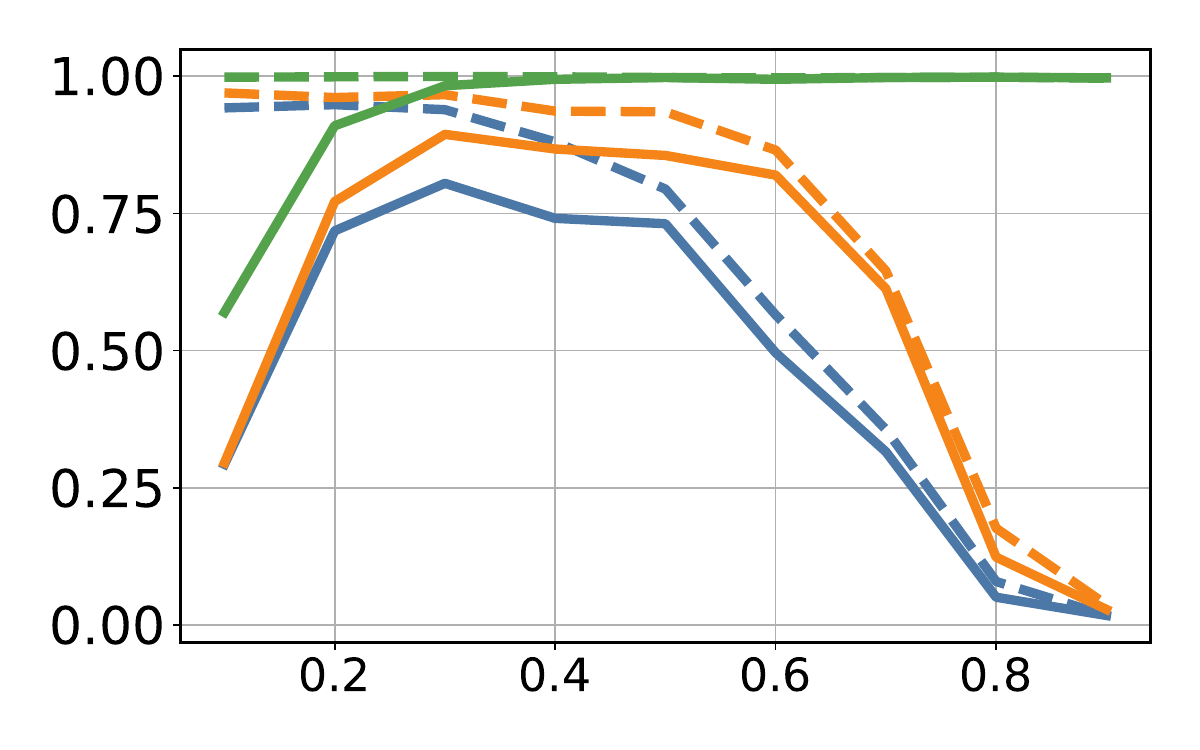}};
            % \node at (0,0) {\includegraphics[scale=0.22]{figures/at_pos_is_bi_rate_MC10Pos10_order1_L4H4d32_T50_monotonic.pdf}};
            \node[scale=0.9] at (0.0, 1.4) { \textbf{{\mixexp{10}} }};
            \node[scale=0.9] at (0.0, -1.4) { \textbf{$\dvr$ }};
        \end{tikzpicture}
    \end{subfigure}\\[2pt]
    % 
    % \begin{subfigure}{0.25\textwidth}
    %     \centering
    %     \begin{tikzpicture}[remember picture]
    %         \node at (0,0) {\includegraphics[scale=0.25]{figures/kl_masked_completion_GT_MC1Pos10_order1_L4H4d32_T50_monotonic.pdf}};
    %         % 
    %         % \node[scale=0.8] at (0.0, 1.8) { \textbf{(a) ID mask $\exposmatin$ }};
    %         % \node[scale=0.8] at (0.0, 1.8) { \textbf{\posexp{10} }};
    %     \end{tikzpicture}
    % \end{subfigure} &
    % % \hspace{70pt}%
    % \begin{subfigure}{0.25\textwidth}
    %     \centering
    %     \begin{tikzpicture}[remember picture]
    %         \node at (0,0) {\includegraphics[scale=0.25]{figures/kl_masked_completion_GT_MC10Pos1_order1_L4H4d32_T50_monotonic.pdf}};
    %         % \node[scale=0.8] at (0.0, 1.8) { \textbf{$\mcexp$}};
    %     \end{tikzpicture}
    % \end{subfigure} &
    % % \hspace{50pt}%
    % \begin{subfigure}{0.25\textwidth}
    %     \centering
    %     \begin{tikzpicture}[remember picture]
    %         \node at (0,0) {\includegraphics[scale=0.25]{figures/kl_masked_completion_GT_MC10Pos10_order1_L4H4d32_T50_monotonic.pdf}};
    %         % \node[scale=0.8] at (0.0, 1.8) { \textbf{  \tina{$\mixexp$} }};
    %     \end{tikzpicture}
    % \end{subfigure}
  \end{tabular}
  \vspace{-0.1in}
  \caption{\textbf{Diversity pays off with time.}~{Factual accuracy versus diversity level after $3$K, $4$K and $16$K iterations. Depending on the template structure, very low diversity can remain unrecoverable; no amount of additional training can restore OOD performance. High-diversity runs begin with lower accuracy but continue to improve until they surpass low-diversity models, demonstrating that diversity incurs an initial cost yet yields long-term benefits.
  {Each bold curve is the average over three individual runs for OOD (solid) and ID (dashed) data.} 
  % no amount of increase in training budget can recover the performance in low diveristy settings. In high diversity, earlier iterations show worse performance, but longer training helps recover the performance and diversity pays off. 
  }
  % \tina{effect of $\dvr$ is not monotonic on the performance and depends on training time...} \tina{play with the three checkpoints to see if it looks better} \tina{add the reference heatmaps ans highlight the iters?} \tina{MC10pos1 !!!!! wtf} %\tina{second row id kl -- keep? remove i believe. doesn't add anything}
  }
  \vspace{-0.1in}
    \label{fig:div_tradeoff}
\end{figure}

\vspace{2pt}
\paragraph{Factual accuracy (Fig.~\ref{fig:heatmaps_main}-(c))} {The effect of diversity on factual recall is more nuanced. At $\dvr = 0.9$, $\factacc$ improves more slowly, just as it does on ID data, whereas at $\dvr \leq 0.2$ the model suffers severe factual errors. The severity of this low-diversity failure, however, depends on the contextual structure. 
% Contextual structure shapes the degree of severity. 
In \posexp{10} setup, where all templates share the same transition matrix, the slowdown under high diversity is modest, yet the low‑diversity failure is severe. In contrast, in \mcexp{10}, low diversity slightly delays reaching high $\factacc$ rather than causing complete failure, as the fixed position signal across templates acts as a strong positional cue, making the model less sensitive to the lack of contextual variety.} \new{Lastly, the behavior for \mixexp{10} is a hybrid of \mcexp{10} and \posexp{10}: It fails at low diversity and shows slower convergence at high diversity}.\looseness=-1

% \puneesh{"In contrast, in \mcexp{10}, low diversity slightly delays"—the delay is in high diversity and there is no complete failure in low div, aren't these two separate points? also better to say }

% \puneesh{rewrite suggestion: as the fixed position signal across templates acts as a strong positional cue, making the model less sensitive to the lack of contextual variety}. 

% \puneesh{comment on MC10POS10? Suggestion: Lastly, the behavior for \mixexp{10} is a hybrid of \mcexp{10} and \posexp{10}: it fails at low diversity and shows slower convergence at high diversity, though both effects are less pronounced. }

\vspace{2pt}
\paragraph{Diversity trade-off for factual accuracy (Fig.~\ref{fig:div_tradeoff})} %\new{In Fig.~\ref{fig:div_tradeoff}, 
To understand how the impact of diversity on factual recall depends on training duration, we plot $\factacc$ versus diversity level $\dvr$ at three training checkpoints -- roughly 
% $20$k, $30$k and $50$k 
\new{$3$K, $4$K, and $16$K} iterations. 
With a short budget ($3$K steps), \emph{intermediate} diversity is best: each fact appears in enough contexts to isolate the associations between fact tokens, yet the model is not overloaded with context variations that would stall learning. With more updates ($4$K steps), factual recall in the high-diversity regimes keeps improving, and in the later checkpoints ($16$K steps) the performance in high diversity setting recovers and the extra training time compensates for its slower start. 
{Intermediate diversity also benefits from longer training, but extremely low diversity does not recover: except in the \mcexp{10} setup, that has a strong position signal for the facts, low‑diversity runs \new{fail to make significant progress with additional training time.}
% lag throughout and additional iterations offer little help. 
Thus, the sweet spot of diversity depends on both training length and how the statistical and factual streams interact in each template.
}
% while the setups that lagged with low‑diversity earlier in training do not recover. Thus, 
% the sweet spot of diversity depends on the length of training and on how the two streams of data interact in the final sequences. 
Finally, when low diversity hurts OOD recall (solid lines), the ID recall (dashed lines) remains nearly perfect. However, in high‑diversity runs, poor OOD performance is accompanied by similarly poor ID performance, suggesting that training length, not the distribution shift, is the limiting factor.\looseness=-1

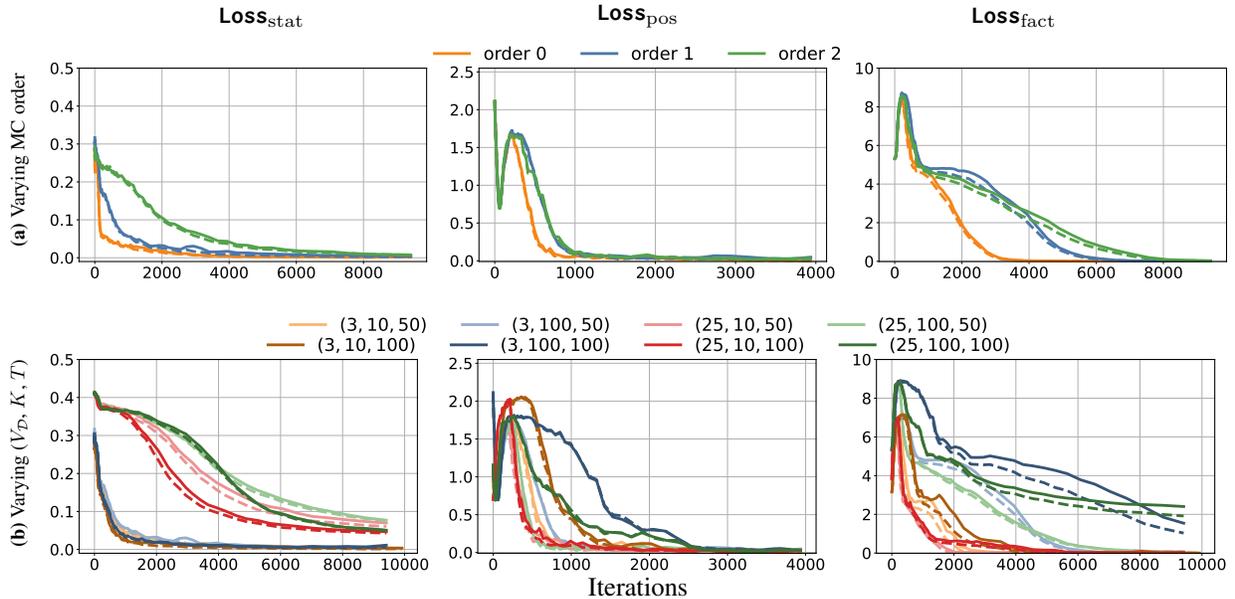
\begin{figure}[t]
  \centering

  % \begin{subfigure}{\linewidth}
  %   \centering
  %   \includegraphics[width=0.9\linewidth]{figures/temp/vocab.png}
  %   \caption{training dynamics -- varying vocab sizes and context length}
  %   \label{fig:vocab}
  % \end{subfigure}

  \vspace{-0.5em}
  
  \begin{subfigure}{\linewidth}
    \centering
    \begin{tikzpicture}
    
      % Place the image in a node named 'image'
      % \node (image) {\includegraphics[width=0.95\linewidth]{figures/temp/order.png}};
      \node (image) {\includegraphics[width=0.95\linewidth]{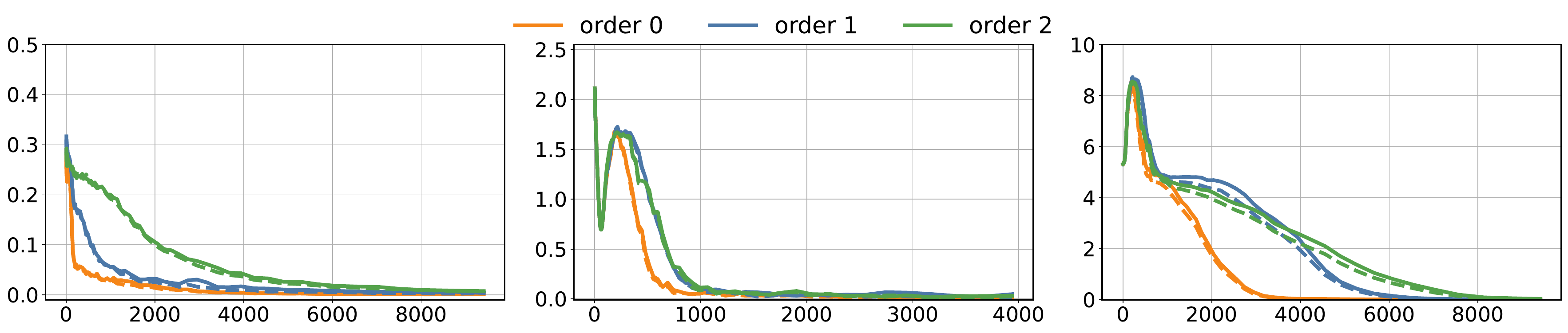}};
      
      % Add the x-label below the image
      \node[above=-5pt of image, font=\small, xshift=0.0cm,] {$\posloss$};
      \node[above=-5pt of image, font=\small, xshift=5.0cm,] {$\factloss$};
      \node[above=-5pt of image, font=\small, xshift=-5.0cm,] {$\KL$};
      
      \node[above=-1.8cm of image, font=\small, xshift=-8.cm, rotate=90, scale=0.75] {\new{\textbf{(a)} Varying MC order}};
      
      % Add the y-label to the left, rotated 90 degrees
      % \node[left=-1pt of image, rotate=90, anchor=center, font=\small] {Accuracy};
      % \node[right=-1pt of image, rotate=90, anchor=center, font=\small] {Loss};
    \end{tikzpicture}
    % \includegraphics[width=0.9\linewidth]{figures/temp/order.png}
    % \caption{training dynamics -- varying order of MC}
    \label{fig:order}
  \end{subfigure}

  \vspace{-.1em}

  \begin{subfigure}{\linewidth}
    \centering
    \begin{tikzpicture}
      % Place the image in a node named 'image'
      % \node (image) {\includegraphics[width=0.95\linewidth]{figures/temp/vocab_ex.png}};
      % \node (image) {\includegraphics[width=0.95\linewidth]{figures/temp/L4H4.pdf}};
      \node (image) {\includegraphics[width=0.95\linewidth]{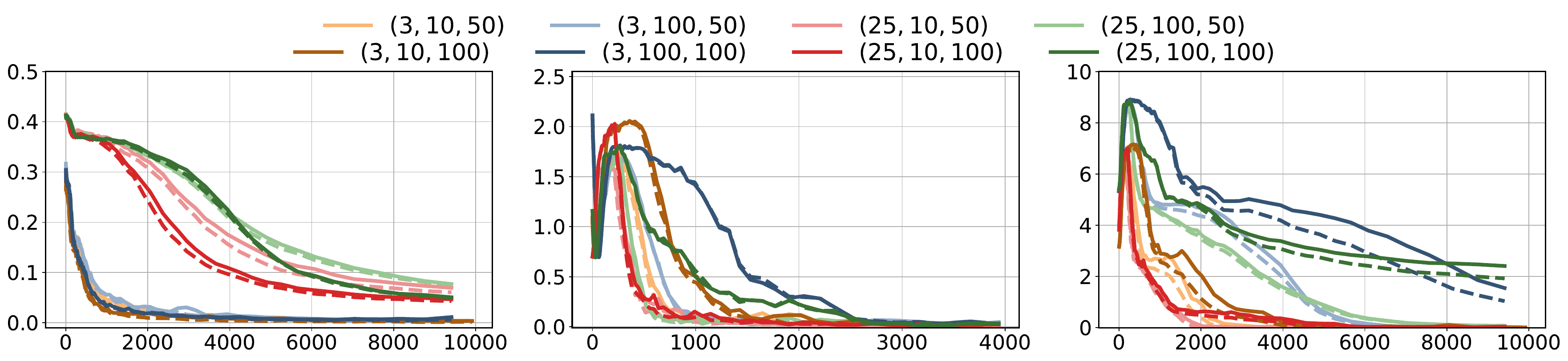}};
      
      % Add the x-label below the image
      \node[below=-8pt of image, font=\small]  {Iterations};

      \node[above=-2.1cm of image, font=\small, xshift=-8.cm, rotate=90, scale=0.75] {\new{\textbf{(b)} Varying ($\vmsize$, $\factsize$, $\seqlen$)}};

      % Add the y-label to the left, rotated 90 degrees
      % \node[left=-1pt of image, rotate=90, anchor=center, font=\small] {Accuracy};
      % \node[right=-1pt of image, rotate=90, anchor=center, font=\small] {Loss};
    \end{tikzpicture}
    % \includegraphics[width=0.9\linewidth]{figures/temp/vocab_ex.png}
    % \caption{replace above if adding 25-10 -- fix the colors --\tina{reorder: stat pos fact} \tina{only keep average lines? it's too crowded}}
    \label{fig:vocab}
  \end{subfigure}

\vspace{-7pt}
  \caption{\new{\textbf{The complexity of one stream impacts learning of the other.}} Speed of learning the statistical, position, and factual metrics \new{as data complexity is varied in a high-diversity setting,} by varying \textbf{(top)} the MC order, and \textbf{(bottom)} the generic vocabulary size $\vmsize$, number of facts $\factsize$, and sequence length $\seqlen$. See Sec.\ref{sec:param_ablation} for discussion and App.\ref{app:loss_metrics} for metrics.\looseness=-1}
  \label{fig:params}
  \vspace{-0.1in}
\end{figure}

\vspace{-0.in}
\subsection{\new{Interplay between streams' complexity and  learning dynamics}}\label{sec:param_ablation}
\vspace{-0.05in}
% 
% 
% \puneesh{Interplay between stream complexities and learning dynamics?}

{Here, we study how learning dynamics are affected by systematically varying the complexity of the two streams. We conduct our analysis in a high-diversity regime ($\dvr=0.8$), where all three metrics are learnable with sufficient training. We track $\KL$ along with $\posloss$ and $\factloss$, the entropy-based analogues of position and factual accuracy (see App.~\ref{app:loss_metrics}), while varying four key factors: the generic vocabulary size ($\vmsize=3, 25$), the number of facts ($\factsize=10, 100$), the sequence length ($\seqlen=50, 100$), and the order of the MC (0 to 2). Results are shown in Fig.~\ref{fig:params}. We observe a strong dependence between how fast each metric can be learned. Specifically:\looseness=-1
% 
 % four key factors: the generic vocabulary size ($\vmsize=3, 25$), the number of facts ($\factsize= 100, 10$), the sequence length ($\seqlen=50, 100$), and the order of the MC (from 0 to 2). Here, we track metric $\KL$ along with $\posloss$ and $\factloss$, the entropy-based analogues of position and factual accuracy (see App.~\ref{app:loss_metrics}). We focus on high-diversity ($\dvr=0.8$), where all three metrics are learnable with sufficient training. Results are shown in Fig.~\ref{fig:params}. We observe that there is strong dependecy between the metrics and how fast they can be learned. Specifically:\looseness=-1
}

\paragraph{$\KL$}
% The rate at which the model learns the statistical stream depends primarily on the stream's own complexity: 
\new{
As expected, increasing the complexity of the statistical stream slows the decay of $\KL$. This effect is clear when either increasing the order of the MC (Fig.~\ref{fig:params}-(a)) or increasing the generic vocabulary size from $\vmsize=3$ to $\vmsize=25$ for both sequence lengths (Fig.~\ref{fig:params}-(b)). However, this is not the only factor, as the increased complexity of the factual stream (a larger $\factsize$) also negatively impacts learning, albeit to a lesser degree. Interestingly, we observe a slight acceleration in statistical learning with increased sequence length ($\seqlen$). This effect is likely because longer sequences provide more training examples of the short-range dependencies due to the Markovian assumption. \looseness=-1
% the increased complexity of the actual stream (larger $\factsize$) can also slow down learning the statistical stream; however, the impact of increasing $\factsize$ is minor compared to $\vmc$. We also see that statistical learning can become minorly faster with increased sequence length $\seqlen$, an effect likely attributed to our setup's short-range dependencies dictated by the Markovian assumption and the fact that each sequence has now more samples of this short range dependency for the model to learn.
}
% As expected, increasing the statistical vocabulary size from $\vmsize=3$ to $\vmsize=25$ significantly slows the decay of $\KL$, an effect observed for both sequence lengths. Similarly, increasing the order of the MC also slows learning. The complexity of the factual stream ($\factsize$) can also slow down learning the statistical stream; however, the impact of increasing $\factsize$ is minor compared to $\vmc$. We also find that statistical learning is largely independent of the sequence length $\seqlen$, likely attributed to our setup's short-range dependencies dictated by the Markovian assumption.

{\paragraph{$\posloss$}
\new{The speed of positional learning is governed by two factors: sequence length ($\seqlen$) and the relative sizes of the vocabulary sets $\vmc$ and $\vk$. Longer sequences generally delay the model's ability to identify the fact positions, an effect visible in Fig.~\ref{fig:params}-(b) by comparing the slower convergence of the darker curves (longer sequences) to the lighter ones.   However, when the factual vocabulary is small and comparable to that of the statistical stream (e.g., $\factsize=10$ vs. $\vmsize=25$), the model learns the fact positions rapidly even with longer sequences, as demonstrated by the red curve in Fig.~\ref{fig:params}-(b).}
}

{\paragraph{$\factloss$} {The dynamics of factual learning are more nuanced, showing stronger dependencies on the other two learning aspects. First, factual learning is contingent on positional learning; the model learns \emph{where} to place facts before it learns \emph{what} facts to generate, consistent with observations in \citet{zucchet2025language}. Consequently, any factor that slows positional learning, such as increasing the sequence length, slows down factual learning and often has an amplified negative impact. {Second}, factual learning is not independent of the statistical stream. Even when other factors are held constant, increasing only the statistical stream's complexity (larger $\vmsize$ or higher order) can also slow down the model's ability to learn facts. \looseness=-1}
% The dynamics of factual learning are more nuanced.  as before, having a more complex factual stream (larger number of facts), slows down learning the factual associations. However, we observe a more nuanced dynamic. First factual learning always happens after positions are learned. A similar observation has been made in \citet{zucchet2025language}. So, any factor that delays position, delays this too. We see this more clearly when moving from $\seqlen=50$ to 100 as the slow down of the position metric, significantly delays factual recall and the impact is usually more pronounced. Second, the learning of facts is not independent from the statistical component as we had seen also on stats being not indep of facts: when everything else is held fixed and only the statistical stream’s complexity increases, factual learning slows as well.
}

\section{{Localizing the optimization bottleneck}}\label{sec:bottleneck}
% \subsection{Probing the Optimization Bottleneck via Interventions}

\begin{figure}[t]
  \centering

  \begin{subfigure}{0.33\linewidth}
    \centering
    \begin{tikzpicture}
      % Place the image in a node named 'image'
      \node (image) {\includegraphics[width=0.95\linewidth]{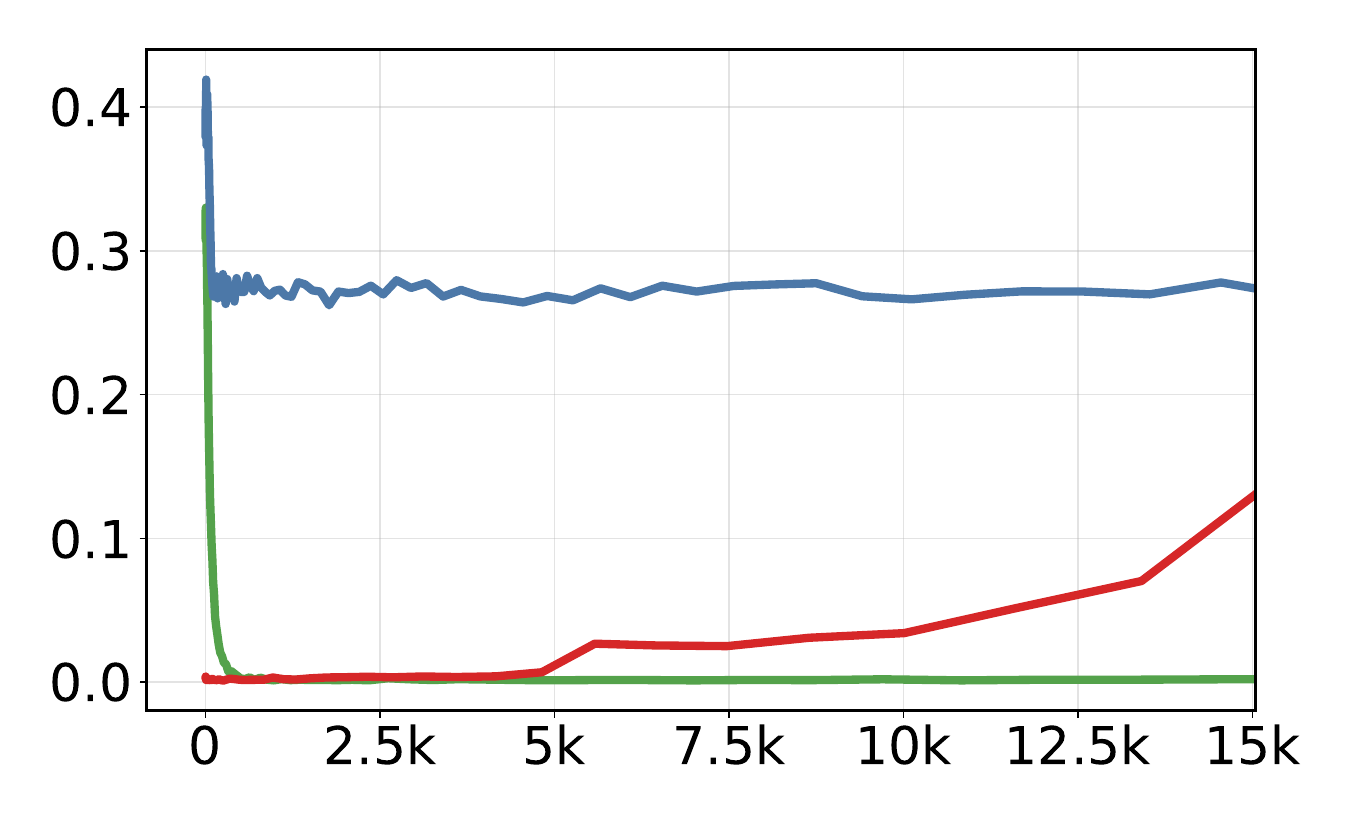}};
      
      % Add the x-label below the image
      \node[below=-7pt of image, font=\small, scale=0.9] {Iterations };
      
      % Add the y-label to the left, rotated 90 degrees
      \node[above=-1pt of image, rotate=0, anchor=center, font=\small] {$\KL$};
      % \node[right=-1pt of image, rotate=90, anchor=center, font=\small] {Loss};
    \end{tikzpicture}
    % \caption{}
    % \label{fig:left_large}
  \end{subfigure}\hfill
  \begin{subfigure}{0.33\linewidth}
    \centering
    \begin{tikzpicture}
      \node (image) {\includegraphics[width=0.95\linewidth]{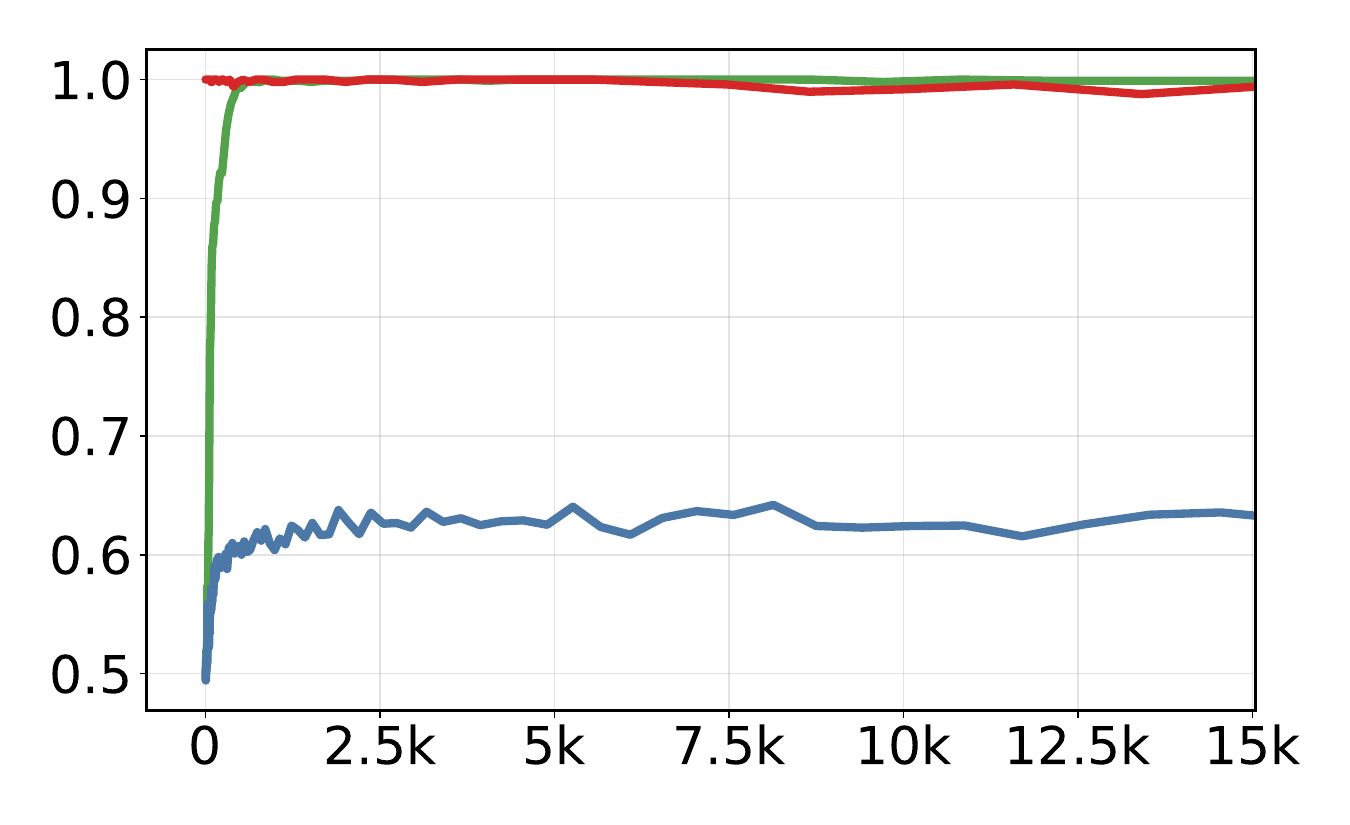}};
      
      % Add the x-label below the image
      \node[below=-7pt of image, font=\small, scale=0.9] {Iterations };
      
      % Add the y-label to the left, rotated 90 degrees
      \node[above=-1pt of image, rotate=0, anchor=center, font=\small] {$\posacc$};
      
      % You can add a y-label here too if it's different
      % \node[left=0.2cm of image, rotate=90, anchor=center] {Your Y-axis Label};
    \end{tikzpicture}
    % \caption{}
    % \label{fig:right_small}
  \end{subfigure}
 \begin{subfigure}{0.33\linewidth}
    \centering
    \begin{tikzpicture}
      \node (image) {\includegraphics[width=0.95\linewidth]{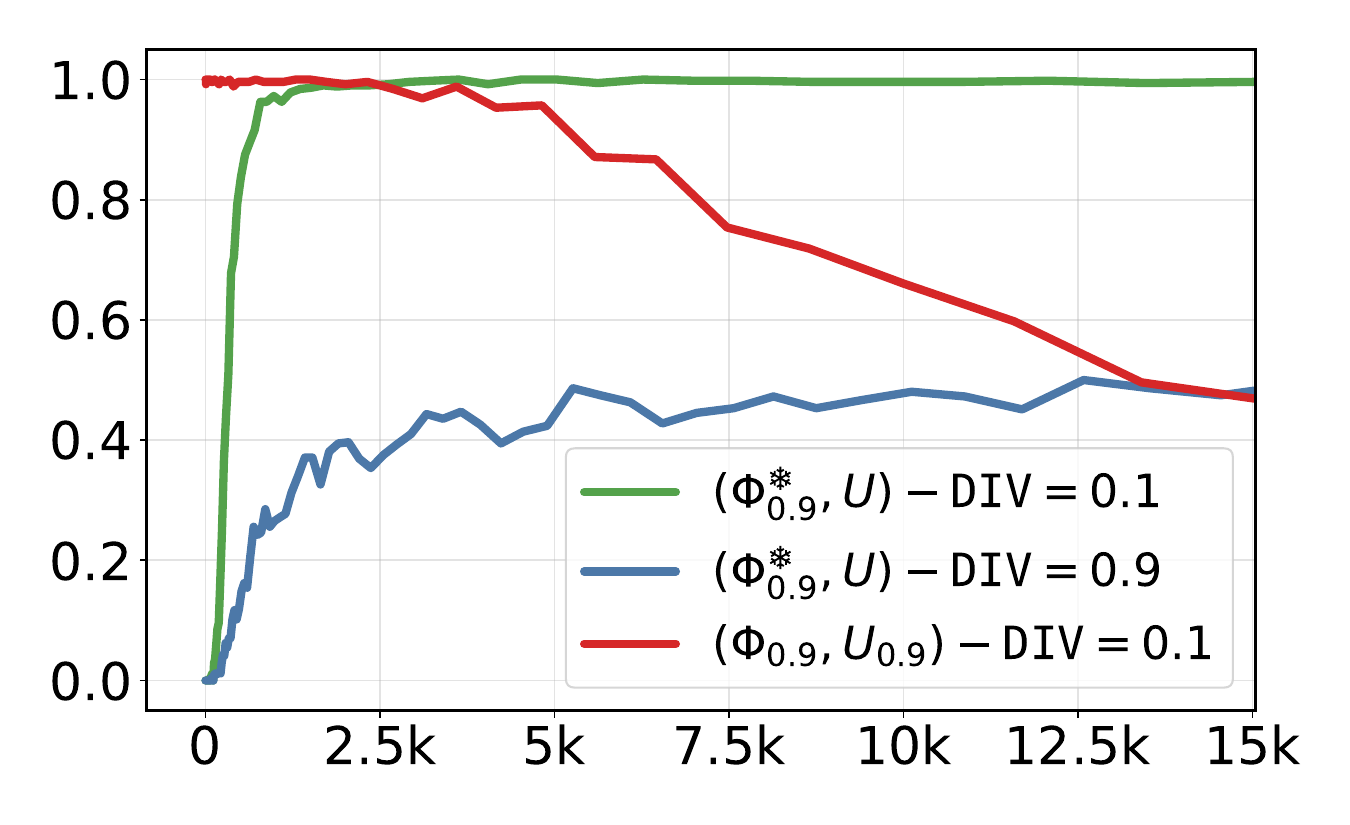}};
      
      % Add the x-label below the image
      \node[below=-7pt of image, font=\small, scale=0.9] {Iterations };
      
      % Add the y-label to the left, rotated 90 degrees
      \node[above=-1pt of image, rotate=0, anchor=center, font=\small] {$\factacc$};
      
      % You can add a y-label here too if it's different
      % \node[left=0.2cm of image, rotate=90, anchor=center] {Your Y-axis Label};
    \end{tikzpicture}
    % \caption{}
    % \label{fig:right_small}
  \end{subfigure}
  \vspace{-8pt}
  \caption{\textbf{Probing the Quality of Last-Layer Features.} 
    \new{Learning curves for the unembedding layer ($\mathbf{U}$) retrained on {last-layer} features ($\Phi$) from models pre-trained on high- or low-diversity data. When using features from a low-diversity model, retraining $\mathbf{U}$ even on high-diversity data shows little improvement over the baseline (blue). Conversely, retraining $\mathbf{U}$ on low-diversity data leads to successful generalization when using features from a high-diversity model (green). However, these high-quality features degrade if they are not frozen and are updated along with $\mathbf{U}$ on the low-diversity data (red). In the legend, $\frz$ denotes that the parameters are frozen, and the subscript indicates the diversity level ($\dvr$) on which those parameters were originally trained. See Sec.~\ref{sec:prob} for details. 
    % $\Phi_{\frz}^{\dvr}$ denotes features ($\Phi$) from a model whose body was frozen ($\frz$) after being trained with a diversity level of $\dvr$.
    } 
    }
  % \caption{\textbf{Probing last-layer features} \new{Retraining the unembedding layer $\mathbf{U}$ on features $\Phi$ learned on low-/high- diversity data. 
  % \textbf{(left)}  With features learned on low-diversity data, training the unembedding layer on diverse data barely ourperforms the original poor generalization of the full model trained on low-div. For features trained on diverse data though, unembedding does not need diverse data to generalize. Hoever if the good features are trained along U they will degrade. $\frz$ means the parameters are frozen and the subscript means the parameters are set to that of learned by training the full model on on which $\dvr$ level. $\Phi(0.1)$ and $\Phi(0.9)$   learned from low- / high-diversity data, while being trainable or frozen ($\frz$).}
  % \tina{update legend}
  % } 
  \vspace{-0.1in}
  \label{fig:prob}
\end{figure}
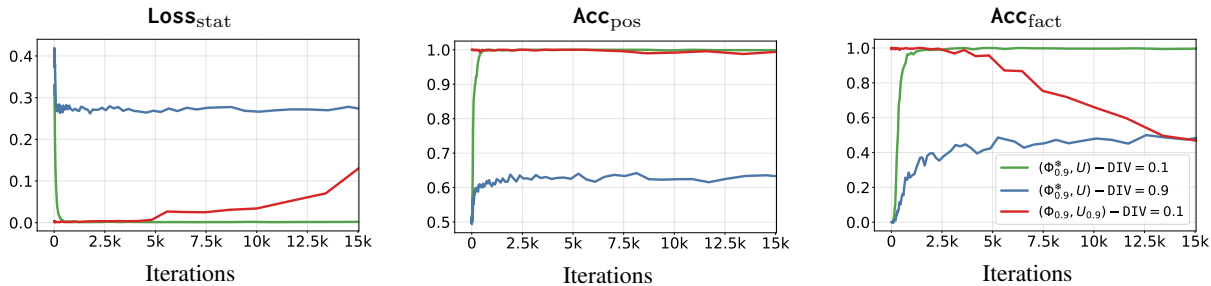

\new{
Previous sections showed that data diversity affects OOD generalization in two ways: sufficient diversity is needed to generalize to unseen template-fact pairs, while very high diversity can slow convergence to the generalizing solution. This impact is not about the amount of exposure to each of the streams themselves: all individual facts and templates appear with roughly the same frequency at training time for any $\dvr$. Instead, $\dvr$ controls which \emph{combinations} of facts and templates are seen during training, and thus sets the test-time distribution shift.
% Previous sections have shown that data diversity has a twofold impact on OOD generalization: sufficient diversity is required for the model to generalize to unseen template-fact combinations, while very high diversity can slow convergence to this generalizing state. This challenge does not stem from a lack of exposure to each of the data streams, as all individual facts and templates are observed with roughly the same frequency during training, irrespective of $\dvr$. Instead, $\dvr$ exclusively controls which combinations of facts and templates are accessible at training time, thereby defining the distribution shift.
}

\new{
The effect of $\dvr$ is also clearly not about model expressivity. Success in mid– to high–diversity regimes shows that a generalizing solution is well within the model's capacity. The issue is one of \emph{optimization}: the set of generalizing solutions remains a set of global minimizers of the training loss for every $\dvr$. Diversity therefore changes whether, and how quickly, training finds one of these solutions, rather than removing them from the landscape (see App.~\ref{app:minimal} for a formal discussion).
}

In this section, we ask which components create a bottleneck to reaching a generalizing solution at a given diversity level, and how this differs across our three aspects of generalization.

\subsection{Probing the last-layer features}\label{sec:prob}
% \new{
% In this section, we investigate which model components introduce a bottleneck for finding a generalizing solution if trained on a specific level of diversity and how these differ for each aspect of generalization. 
% We start with the following quick observations:
% }

% \subsection{Last-layer features}
\paragraph{Low diversity corrupts the learned features} 
\new{Let $\lowmodel$ be the model trained with $\dvr=0.1$. We probe the learned features by freezing the whole network except the unembedding layer $\mathbf{U}$, then retrain $\mathbf{U}$ from a random initialization on diverse data with $\dvr=0.9$. This allows us to isolate the effect of diversity only on the last-layer features. As shown by the blue curves in Fig.~\ref{fig:prob}, even after retraining $\mathbf{U}$, performance improves only marginally over $\lowmodel$. Thus, simple retraining of the last layer on high-quality data \citep{kirichenko2022last} does not fix the problem, suggesting that $\lowmodel$ failed to learn the  features required for generalization.} %\puneesh{core fits in their terminology not ours imo, id say \emph{good}}

% \new{Let the model trained on data with $\dvr=0.1$ be $\lowmodel$. We probe the learned features by freezing all the parameters except the unembedding layer $\mathbf{U}$, and retrain the unembedding from random initialization on diverse data $\dvr=0.9$. This allows us to isolate the impact of diversity on the last-layer features. As shown in Figure~\ref{}\tina{}, with the updated $\mathbf{U}$, the model only marginally improves over $\lowmodel$, suggesting that simple retraining of the last layer on high-quality data does not fix the performance \citep{} and the $\lowmodel$ has failed to learn the core features needed for generalization.}

\new{\paragraph{High diversity promotes good feature learning} 
Now let $\highmodel$ be the model trained with $\dvr=0.9$. We freeze all feature-producing modules and retrain $\mathbf{U}$ on low-diversity data with $\dvr=0.1$. 
The green curves in Fig.~\ref{fig:prob} reach perfect performance, which hints that features learned under high diversity disentangle template and fact effectively in the feature space such that $\mathbf{U}$ trained only on low-diversity data can still recover the right mapping. These features are, however, fragile: when we unfreeze the body of $\highmodel$ and continue training at $\dvr=0.1$ together with $\mathbf{U}$, performance drops (red curves), suggesting that continued exposure to low-diversity data corrupts the initially good features. %\puneesh{here corrupts is the perfect word, but not in previous heading. Low diversity fails to learn \emph{good} features makes more sense to me}
% 
% As shown by the green curves in Fig.~\ref{fig:emb_new}, the unembedding layer trained on only low-diveristy data generalizes, hinting at the fact that the features learned on diverse data cleanly disentangle templates and facts information. These features are, however, fragile: if the body of $\highmodel$ is unfrozen and trained further on $\dvr=0.1$ together with $\mathbf{U}$, performance degrades (red curves), suggesting continued exposure to low-diversity data corrupts the initially good features.
}

\new{We further analyze how training diversity affects the clustering properties of the model's internal embeddings in App.~\ref{sec:representation_analysis}.} \new{We next investigate which modules introduce a bottleneck for effective feature learning.} %\tina{Add ref to app on clustering}

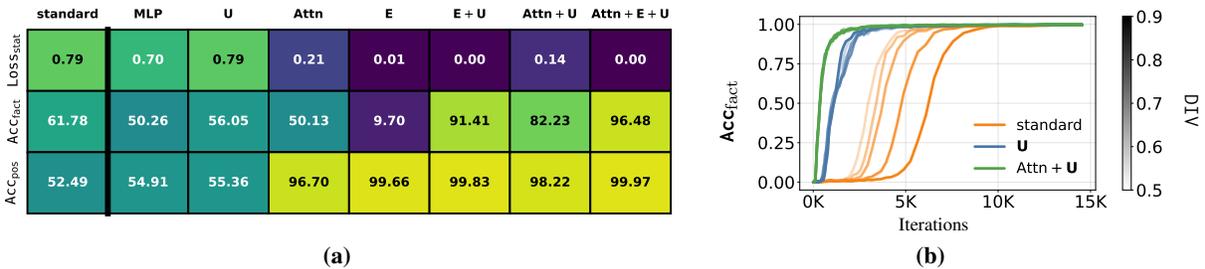
\begin{figure}[t]
  \hspace{-40pt}
  \centering
  \begin{subfigure}{0.6\linewidth}
    \centering
    \begin{tikzpicture}
      % Place the image in a node named 'image'
      \node (image) {\includegraphics[width=\linewidth]{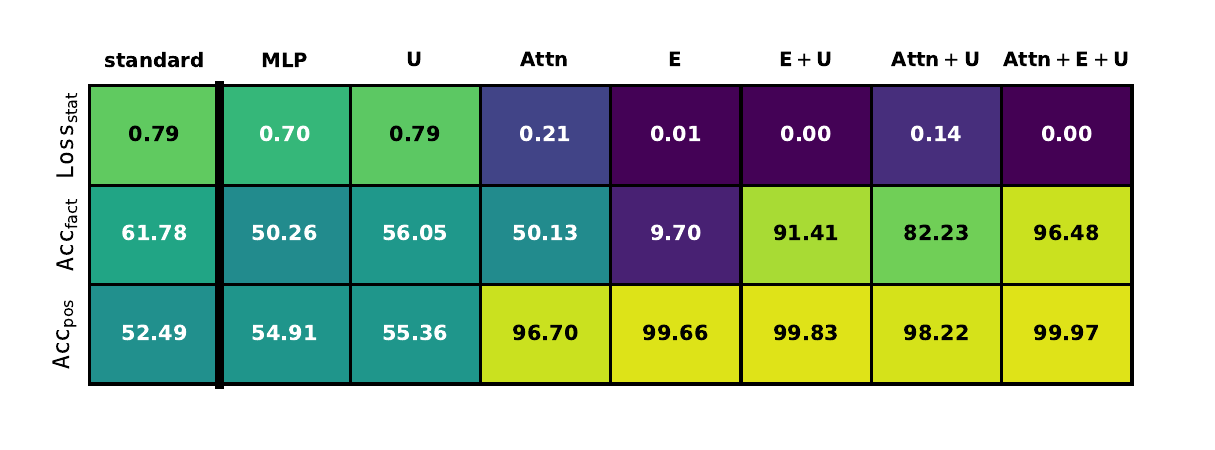}};
      % \node (image) {\includegraphics[width=\linewidth]{figures/otherSec5/L4H4_table_ood1_iter105_.pdf}};
     \node[below=-9pt of image, font=\small] {};

      % Add the x-label below the image
      % \node[below=-7pt of image, font=\small] {Iterations};
      
      % Add the y-label to the left, rotated 90 degrees
      % \node[left=-1pt of image, rotate=90, anchor=center, font=\small] {Accuracy};
      % \node[right=-1pt of image, rotate=90, anchor=center, font=\small] {Loss};
    \end{tikzpicture}
    \vspace{-15pt}
    \caption{}
    % \label{fig:left_large}
  \end{subfigure}%\hfill
  % \hspace{-10pt}
  \begin{subfigure}{0.35\linewidth}
    \centering
    \begin{tikzpicture}
      \node (image) {\includegraphics[width=\linewidth]{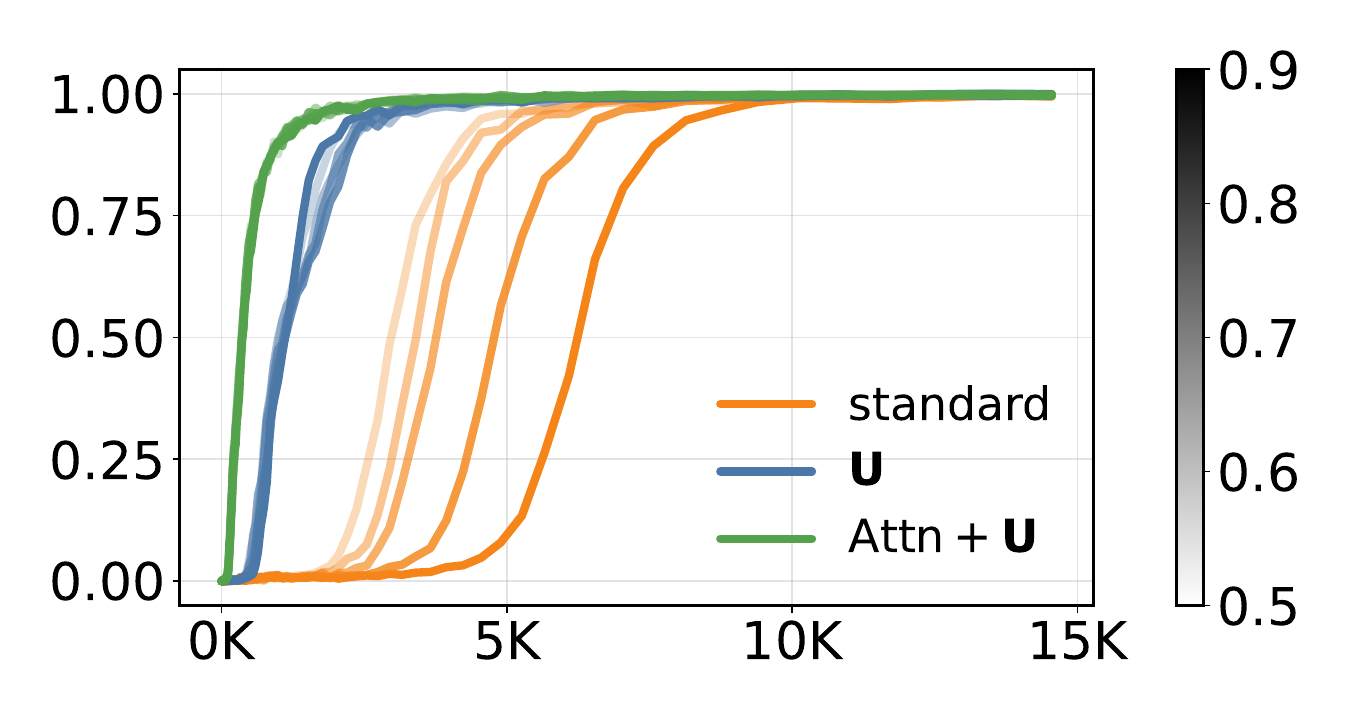}};
      
      % Add the x-label below the image
     \node[below=-10.pt of image, font=\small, scale=0.7, xshift=-0.5cm] {Iterations};
      \node[left=1pt of image, rotate=90, anchor=center, scale=0.8] {$\factacc$};
      \node[right=1pt of image, rotate=-90, anchor=center, scale=0.8] {$\dvr$};
      
      % You can add a y-label here too if it's different
      % \node[left=0.2cm of image, rotate=90, anchor=center] {Your Y-axis Label};
    \end{tikzpicture}
    \vspace{-5pt}
    \caption{}
    % \label{fig:right_small}
  \end{subfigure}
  \vspace{-5pt}
  \caption{\new{\textbf{Tracing the optimization failure to individual model components.}} Models trained with a selected module (MLP, unembedding ($\mathbf{U}$), token embedding ($\mathbf{E}$), or attention ($\textbf{Attn}$)), replaced by its frozen counterpart from a well-trained model on high-diversity data. \emph{Standard} refers to training the model, end-to-end with no interventions. All reported values are averaged over three independent runs. \textbf{(a)} Performance on all metrics after 30K iterations on low-diversity $(\dvr=0.1)$ data. Note that for $\KL$ darker and for $\posacc$ and $\factacc$ brighter colors are optimal. \textbf{(b)} Factual accuracy over time, for different diversity values (wiht darker curves corresponding to higher diversity), showing the convergence speed to perfect recall for various interventions. See Sec.~\ref{sec:intervention} for details and App.~\ref{app:exp_intervention} for more results. \looseness=-1} 
  \label{fig:intervention}
  \vspace{-0.1in}
\end{figure}

\subsection{Module-wise interventions {for locating the bottleneck}}\label{sec:intervention}

\new{
We now retrain models on different diversity levels, not end-to-end, but after employing controlled interventions on specific modules. For each intervention, we replace a module with its counterpart from a well-trained model ($\highmodel$, trained on $\dvr=0.9$). We target four modules: token embeddings ($\mathbf{E}$), attention ($\textbf{Attn}$), the MLP, and the unembedding layer ($\mathbf{U}$) (see App.~\ref{app:exp} for more ablations). For all these modules, except the \textbf{Attn}, we copy and freeze their corresponding weights at initialization. The special case where all modules except the unembedding layer are frozen is equivalent to the linear probing discussed in the previous section. Our intervention method differs for attention, where we patch the pre-computed attention patterns from $\highmodel$ during each forward pass \citep{zucchet2025language}. 
% Our intervention method differs for attention, where we patch the pre-computed attention patterns from $\highmodel$ during each forward pass \citep{zucchet2025language}. For the remaining modules, we copy and freeze their corresponding weights at initialization. The special case where all modules except the unembedding layer are frozen is equivalent to the linear probing discussed in the previous section.
% 
% We retrain the models on different diversity levels, not end-to-end, but after employing controlled interventions on specific modules. For each (subset of) module(s), we replace it with its counterpart from a well-trained model ($\highmodel$, trained on diverse data with $\dvr=0.9$). We target specifically four modules: token embeddings \new{($\mathbf{E}$)}, attention \new{($\textbf{Attn}$)},  MLP \new{($\textbf{MLP}$)}, and unembedding layer \new{($\mathbf{U}$)} (see App.~\ref{app:} for more ablations). Our intervention method differs for attention, where, for each sequence and at each layer, we substitute the pre-computed attention patterns from $\highmodel$ during each forward pass \citep{zucchet2025language}. For the remaining modules, at initialization, we copy and freeze the corresponding weights from $\highmodel$, training only the rest of the network. The special case where all modules except the unembedding layer are frozen boils down to linear probing on the last-layer features discussed above. 
}

Here, we focus our analysis on the \mixexp{10} setup trained at a low diversity of $\dvr=0.1$, as it fails on both factual and statistical learning, with results for this setting summarized in the table in Fig.~\ref{fig:intervention}. Comprehensive results across all diversity levels, along with corresponding experiments on a one-layer Transformer, are provided in App.~\ref{app:exp}. Our key observations are as follows:\looseness=-1

\new{\paragraph{Stats and position fail due to poor \textbf{Attn} and/or $\mathbf{E}$ optimization}
Transferring either the attention patterns ($\textbf{Attn}$) or the token embeddings ($\mathbf{E}$) from $\highmodel$ allows the model to learn effective features for statistical and positional generalization, even when trained on low-diversity data.
% 
% When we train on low-diversity data while transferring either \textbf{Attn} patterns or $\mathbf{E}$ from $\highmodel$, 
% \new{the model learns features that do well on the staitistical and position learning. }
% feature learning recovers for learning the statistics and position information. 
% \puneesh{"feature learning recovers" sounds weird—the model learns to do well on statistical and positional metrics?}% metrics improve sharply. 
Using \textbf{Attn} patterns from $\highmodel$, essentially telling the model the correct position to focus in context, enables obtaining near-perfect $\posacc$ and a large drop in $\KL$. Freezing $\mathbf{E}$ produces the same effect: near-perfect $\posacc$ and a statistical loss close to zero. 
% With \textbf{Attn} from $\highmodel$, which essentially tells the model the correct position to focus in context, we obtain near-perfect $\posacc$ and a large drop in $\KL$. Freezing $\mathbf{E}$ produces the same effect: near-perfect $\posacc$ and a statistical loss close to zero. 
% \puneesh{Using Attn patterns from $\highmodel$ essentially tell the model....?}
% 
% Optimizing the model on low-diversity data, while the $\textbf{Attn}$ patterns or $\mathbf{E}$ are transferred from $\highmodel$, exclusively helps with feature learning for recovering the statsitical and position learning. With the attention patterns from $\highmodel$, effectively telling the model where to focus in the context, the model achieves near-perfect $\posacc$ and a significantly lower $\KL$. Similarly, freezing the token embeddings yields near-perfect $\posacc$, while this time, reducing the statistical loss to nearly zero.
}

\new{\paragraph{Factual recall needs a well-optimized $\mathbf{U}$ \emph{and} good features} 
Intervening on feature-producing modules alone 
% \puneesh{this reads like if you freeze all of them to good and just train U, it fails, which is not true}
(any subset of component not including $\mathbf{U}$) does not restore factual recall. However, intervening on $\mathbf{U}$ together with \textbf{Attn} or $\mathbf{E}$ boosts factual accuracy along with improved statistical generalization. Importantly, intervening on $\mathbf{U}$ alone, while features are learned from low-diversity data, does not help factual recall. Thus, the bottleneck for factual generalization is joint: $\mathbf{U}$ must be trained together with the features. %\puneesh{clean? umm, good? dirty disgusting features :P}
% 
% No such kind of interventions on the feature-producing modules (any component other than the unembedding layer) alone provide a similar benefit to factual recall. However, intervention on the unembedding layer along with $\textbf{Attn}$ / $\mathbf{E}$ boosts factual recall while preserving the feature quality for statistical generalization. Improtantly, intervention on $\mathbf{U}$ alone, while learning the features freely on low diversity data shows no improvement in factual recall, showing optimization of the unembedding layer alone is not the bottleneck for factual generalization.
% Similar to that we saw that no intervention on the feature-producing modules fixed the features for optimal factual performance, if we transfer the unembedding layer $\mathbf{U}$ for $\highmodel$, while training the rest of the model on low-diversity data, we see no major improvemnts on the factual preformance, noting that features are bad. However, when combining the intervention on the unembedding layer with the effective interventions on the features, we see that along with stat and pos, fact also reaches near perfect accuracy. all these observations hint at the fact that the joint optimization of the features and the unembedding layer introduce an optimization bottleneck. 
}

\new{\paragraph{The slowdown in factual learning traces to optimizing $\mathbf{U}$} 
As shown in Fig.~\ref{fig:intervention}-(b), while standard end-to-end training shows progressively slower convergence as diversity increases from $\dvr=0.5$ to $0.9$, when we only train the features and keep the unembedding layer frozen to that of well-trained $\highmodel$, factual recall improves at a consistent pace regardless of diversity levels. This suggests the optimization of the unembedding layer as the cause of the slow generalization with higher diversity. Interestingly, this speed-up is also observed in the low-diversity regime, although intervention on $\mathbf{U}$ alone is not enough for recovering from the failure caused by lack of data diversity (see Figs.~\ref{fig:app_interventions} and \ref{fig:intervention_10k} in the App). } 
{
}

\vspace{-0.01in}
\section{Conclusion and limitations}\label{sec:conclusion}
% \tina{??}

% \tina{add explicit questions? takeaways and outlooks for future works}

\vspace{-0.1in}
We have introduced a synthetic framework to systematically investigate the generalization abilities of language models pre-trained on sequences that combine statistical patterns and factual information. The simplicity of our testbed is intentional, offering several advantages: \textbf{(1)} it enables fine-grained control over diversity level, while independently varying contextual structure; \textbf{(2)} it requires minimal computational overhead; and \textbf{(3)} it captures different aspects of language model generalization in both ID and OOD settings.
Using this, we have uncovered several illuminating behaviors regarding the interplay between statistical and factual generalization as functions of diversity and structure. 

% \tina{gemini's draft: needs another round of edits}
\new{Our findings, derived from this controllable and inexpensive testbed, open several avenues for deeper investigation into the learning dynamics of language models often obscured in large-scale regimes. Building upon our results in Sec.~\ref{sec:bottleneck}, future works can formalize our observations, such as how data diversity shapes the optimization landscape and how different modules come to serve specialized roles.\looseness=-1
}
% \new{Our findings, derived from this controllable and inexpensive testbed, open several avenues for deeper investigation into the learning dynamics of language models. First, this simplified setting can serve as a more suitable ground for developing a concrete theoretical and mechanistic understanding of phenomena often obscured in large-scale models. Building up on our results in Sec.~\ref{sec:freeze}, future works can formalize how data diversity impacts the optimization landscape, design targeted strategies to mitigate its negative effects, and more precisely determine how individual modules contribute to different aspects of generalization.}

\new{Furthermore, our setup provides a controlled baseline for studying the model's behavior under more realistic and challenging data conditions, such as the noisy, long-tailed, and imbalanced distributions characteristic of web-scale corpora.}
% \new{The controlled nature of our setup also provides a baseline for studying more realistic and challenging data conditions. For instance, while our experiments used a balanced exposure to facts, real-world knowledge is famously long-tailed \citep{??}. Our framework allows for a systematic study of how such data imbalance affects learning. Similarly, one could inject a controlled amount of noisy or contradictory facts to probe the model's ability to distill correct information from the wrong ones, mimicking the uncurated fact representation in the web-scraped data. This testbed can also be used to disentangle knowledge acquired during pre-training from that acquired during fine-tuning more formally.}

\new{Finally, the minimalist nature of our testbed was a deliberate choice to enable controlled and low-cost analysis. A key challenge for future work is navigating the trade-off between this analyzability and linguistic realism. While bottom-up refinements are necessary to better capture the complex structures of natural language, it is crucial to maintain a balance that allows for the clear, mechanistic insights that simple testbeds provide.}

\section*{Acknowledgments}
\vspace{-10pt}
This work was partially funded by the NSERC Discovery Grant No. 2021-03677 and the Alliance Grant ALLRP 581098-22. TB and PD were also supported by the UBC 4YF Doctoral Fellowship. The authors also acknowledge use of the
Sockeye cluster by UBC Advanced Research Computing and thank the anonymous reviewers for their helpful feedback.

\bibliography{refs,refs_language}

%%%%%%%%%%%%%%%%%%%%%%%%%%%%%%%%%%%%%%%%%%%%%%%%%%%%%%%%%%%%
% \section*{Contents} 
% \addcontentsline{toc}{section}{Contents of Appendix} % Optional: adds the title to the main ToC

% % Start recording entries for a new ToC named 'app'
% \startcontents[app] 

% % Print the ToC for the 'app' list, starting from section level 1
% \printcontents[app]{}{1}{} 

\newpage
\appendix
\textbf{Other notations.} We denote matrices/vectors/scalars as $\Ab$/$\ab$/$a$ respectively. We view token-sequences as vectors and denote  $\ab_{t_1:t_2}$ as the subsequence from index $t_1$ to $t_2$. We denote $\Ab[i,j]$ the $(i,j)$-th entry of matrix $\Ab$, and respectively for vectors. We let $\sft{\cdot}$ denote the softmax map, $\Delta$ the simplex, and ${\rm{KL}}( {\pb_1} \,\|\, \pb_2 )$ the Kullback–Leibler (KL) divergence between probability vectors $\pb_1, \pb_2$. We denote $[N]:=\{1,\ldots,N\}$ and use $\mathbbm{1}[\mathcal{C}]$ for the indicator function (1 if condition $\mathcal{C}$ is satisfied, 0 otherwise). \looseness=-1

% \section{Overview of supplementary material}\label{app:overview}

\paragraph{Overview of supplementary material}~Section \ref{app:related_works} expands on the literature discussion in Section \ref{sec:intro}. Section \ref{app:loss_metrics} introduces additional metrics used in Section ~\ref{sec:param_ablation}. Section \ref{app:exp} provides additional experimental details and results to complement Sections \ref{sec:results} and \ref{sec:bottleneck}. Section \ref{app:minimal} introduces a minimal toy setting -- as a starting point for theoretical analysis. Finally, Section \ref{sec:representation_analysis} examines the model’s internal embeddings and how training diversity affects them. \looseness=-1

% In Section \ref{app:related_works} we expand on the related works that we briefly discussed in Section \ref{sec:related_works}. Section \ref{app:exp} presents additional experimental results and discussion. In Section \ref{app:minimal} we outline a minimal toy setting (briefly discussed in Section \ref{sec:additional}) to propose a direction for a theoretical study. Finally, Section \ref{sec:representation_analysis} inspect the internal embeddings of the model to see the impact of diversity on them.

% \tina{fix code and github}

% \tina{add the discussion on extra metrics}

\begin{table}[t]
\centering
\caption{Table of notations.}
\renewcommand{\arraystretch}{1.05}
\begin{tabular}{@{}ll@{}}
\toprule
\textbf{Symbol} & \textbf{Meaning} \\
\midrule
$\seqlen$                          & Sequence length. \\
$\mathbf{x}=(x_1,\dots,x_T)$ & Token sequence produced by mixing two streams. \\
$\mathcal{V}=[V]$            & Global vocabulary. \\[4pt]

\multicolumn{2}{@{}l}{\textbf{Factual stream}}\\
$K$                          & Number of atomic facts. \\
$\mathcal{K}=\{(a_k,b_k)\}_{k=1}^{K}$ & Set of $(\mathrm{source},\mathrm{target})$ pairs. \\
$\mathcal{A}=\{a_k\}$,\; $\mathcal{B}=\{b_k\}$ & Source and target vocabularies. \\
$\mathcal{V}_{\!K}=\mathcal{A}\cup\mathcal{B}$ & Fact vocabulary. \\
% $f:\mathcal{A}\!\to\!\mathcal{B}$       & One‑to‑one mapping, $f(a_k)=b_k$. \\[2pt]

\multicolumn{2}{@{}l}{\textbf{Statistical stream (templates)}}\\
$N$                          & Number of templates. \\
$\mathcal{D}_n$              & Distribution associated with template $n$. \\
$\mathcal{V}_{\dist}=\mathcal{V}\setminus\mathcal{V}_{\!K}$ & Set of generic vocabularies (the MC vocabulary). \\
$\pospair_n=(\posSource_n,\posTarget_n)$ & Placeholder positions for fact pairs in template $n$ ($\posSource_n\le T/2<\posTarget_n$). \\
% $\mathcal{P}_n$              & Set of transition matrices of template $n$. \\
% $\mathcal{Q}_n$              & Set of position pairs of template $n$. \\
$\mathbf{P}_n$               & The MC transition matrix associated with template $n$. \\[4pt]

\multicolumn{2}{@{}l}{\textbf{Exposure and diversity}}\\
$\exposmatin\in\{0,1\}^{N\times K}$ & ID exposure mask: $(n,k)$ entry is~1 if fact $k$ occurs in template $n$ during training. \\
% $\mathrm{E}_{\text{out}}=\mathbf{1}-\mathrm{E}_{\text{in}}$ & OOD exposure mask. \\
$\dvr\in(0,1]$ & Diversity level: fraction of templates in which each fact appears. \\[4pt]

\multicolumn{2}{@{}l}{\textbf{Metrics}}\\
 $\posacc$  & Structural (position) accuracy (Eq.~\eqref{eq:postion_acc}). 
\\
% $\posacck$  & position accuracy at the target position. \\
$\factacc$ & Factual accuracy (Eq.~\eqref{eq:fact_acc}). 
\\
$\KL$  & Statistical loss (Eq.~\eqref{eq:KL}). 
% (KL divergence between the model's prediction and the reference MC, Eq.~\eqref{eq:KL}). 
\\
$\posloss$  & Entropy analogue of $\posacc$ (Eq.~\eqref{eq:postion_loss}). 
\\
$\factloss$  & Entropy analogue of $\factacc$ (Eq.~\eqref{eq:fact_loss}). 
\\

\bottomrule
\end{tabular}
\label{tab:notation}
\end{table}

\section{Additional Details on Related Works}\label{app:related_works}
LLMs have been observed to pack a substantial amount of knowledge in their parameters during pretraining, allowing them to answer real‑world questions without consulting external resources \citep{petroni2019language,roberts2020much}, raising the question of whether they can replace conventional knowledge bases \citep{omar2023chatgpt,sun2023head}.
% 
% A growing body of mechanistic‑interpretability work explores \emph{where} LMs store knowledge \citep{meng2023locating,dai2021knowledge,geva2020transformer} and \emph{how} they recall the correct fact \citep{geva2023dissecting,lv2024interpreting,choe2025all} by inspecting the activation values at different layers and different modules. 
By inspecting the internal activations of different modules and layers, a growing body of mechanistic interpretability work seeks to understand where LMs store knowledge \citep{meng2023locating,dai2021knowledge,geva2020transformer} and how they recall it \citep{geva2023dissecting,lv2024interpreting,choe2025all}. Other studies trace each learned fact back to the pre‑training data to investigate which corpus patterns enable its acquisition \citep{elazar2022measuring,akyurek2022towards,li2022pre} and demonstrate that recall accuracy depends on the number of exposures to that fact in the pretraining corpus \citep{kandpal2023large,allen2023physics}. Studies on learning dynamics similarly find that different knowledge types are learned at different rates \citep{liu2021probing}. \citet{chang2024large} probe the factual recall dynamics by injecting fictional facts during pre‑training and tracking their probabilities over time, observing that knowledge accumulates through many small ``micro‑updates'' that gradually decay unless the fact is periodically reinforced to avoid forgetting.\looseness=-1

Recent works have initiated systematic exploration of factual recall in language models using \emph{controlled synthetic setups}. Typically, these works model each fact as a triplet (source, \emph{relation}, target) embedded in a context; the model is then probed with a context containing a (source, relation) and must produce the corresponding target \citep{allen2023physics,nichani2024understanding,zucchet2025language}. We adopt a similar framework but omit the \textit{relation} token. The simpler (source, target) setup suffices for our purpose of studying how models acquire and recall factual associations along with other aspects of generalization from first principles. \looseness=-1

\citet{allen2023physics} use a controlled synthetic biography dataset:
% with fixed-sentence templates to examine factual recall
each biography entry is a multi-sentence paragraph about an individual, the \emph{source}, where each sentence represents a (relation, target) chosen from a set of fixed‑sentence templates, e.g., ``\textit{<Name> was born in <City>}''. 
They specifically consider question-answer (QA) formats  for probing knowledge, e.g., ``\textit{What is <Name>'s city of birth? <City>}'', which are shown at training time for only a subset of individuals. We treat these QA forms as just another template family and evaluate by probing the model with any template that was \emph{unseen} for a given fact during training.  %since we can simply treat the QA templates as another family of templates. 
Focusing on factual recall performance, they show that factual recall after fine-tuning on question templates is only possible if facts are observed
% in diverse contexts, i.e., 
in several templates during pre‑training and 
% (i) when question templates are introduced \emph{only} at the fine-tuning stage, factual recall is impossible unless each fact was seen in diverse contexts, i.e., in several templates, during pre‑training, and (ii) even when questions are already present in pre‑training data, 
more paraphrase diversity markedly boosts recall. In other words, successful recall requires \emph{varied} exposure to each fact, not mere repetition in a single template. 
%Our data‑generation scheme draws on this insight\new{, but comes with several unique features}.

% {While our data-generation scheme draws on this core insight, our synthetic testbed  offers a more abstracted and finely controlled framework.}
% By abstracting the biography setup further while retaining its template–fact structure, we gain more fine-grained control over other aspects like the statistical and structural composition{, which were fixed in their work}. 
Our data-generation scheme extends this core insight by further abstracting the template–fact structure. This provides fine-grained control over the statistical and structural composition of the data, aspects that were fixed in the original work. We focus exclusively on pre-training experiments here, although the same abstract framework could also be used to study fine-tuning. {Yet, we show that even without finetuning and even in minimal settings like those described in Sec.~\ref{app:minimal} low-diversity can impair generalization. Our analysis of the impacts of diversity on the statistical aspects of generalization as well as the systematic categorization of different contextual structures are also unique to our study compared to this prior work.}

Recent work by \citet{zucchet2025language} analyzes the dynamics of factual recall in a synthetic biography setup, similar to that of \citet{allen2023physics}, and reports a stage-wise dynamic. By inspecting the model’s predictions at the target position across training checkpoints, the model first restricts its choice to the fact vocabulary and only later learns the correct mapping from the the specific (source, relation) present in the context to the correct fact token. 
We make similar observation between factual and position accuracy Sec.~\ref{sec:param_ablation}. 
Our evaluation, however, is broader: we probe the model with an incomplete prompt and grade the \emph{entire} completion from different aspects -- whether the correct fact token appears in the correct position, whether the remaining positions are filled with generic tokens, and whether those tokens follow the template’s statistical pattern. {As \citet{zucchet2025language} follow similar setup as \citet{allen2023physics}, the unique and distinctive characteristics of our setup mentioned above—particularly the explicit statistical stream enabling a joint study of statistical and factual generalization aspects, alongside systematic control over contextual structure and diversity—apply equally as differentiators here.}

Both of the above referenced studies also vary the data distribution, exploring how ``celebrity'' entries -- individuals whose biographies appear in many templates during pre‑training -- affect factual recall for less‑frequent entries and alter learning dynamics. 
{While our current work focuses only on overall context diversity, our flexible framework can readily accommodate such experiments on the impact of data distribution by appropriately designing the template-fact exposure matrix $\exposmatin$ to vary fact frequencies across templates. We leave this as interesting future work.}
{Perhaps the most closely-related work in terms of model abstraction, although coming from differing motivations, appears in \citet{nichani2024understanding}. While they focus on capacity and storage tradeoffs for factual recall, we investigate the impacts of diversity and tradeoffs between statistical and factual accuracy.}
% \citet{nichani2024understanding} study factual recall in shallow Transformers through the lens of associative memory, focusing primarily on capacity and storage trade‑offs. 
In their synthetic data setup, they sample the (source, relation, target) mappings randomly by choosing them from a fact set. Each sequence is generated by placing the (source, relation) at two random position, appending the target at the end and filling the remaining positions with tokens uniformly drawn from a disjoint \emph{noise} vocabulary (functionally identical to our generic tokens).
Training then minimizes the loss at the target position, focusing on the fact storage capacity of the model. 
% Within this abstract setting they prove theoretically that a single‑layer transformer can memorize facts if the model size scales appropriately and they quantify how the capacity can be allocated between attention heads and MLP weights. 
Similarly in our setup, we preserve the separation of facts from background tokens but introduce a \emph{structured} statistical stream: generic tokens are generated by a Markov process and fact placeholders occupy slots that can vary across different templates. {This richer design allows us to introduce and systematically investigate how contextual structure and diversity affect performance, while extending the analysis beyond factual recall to include statistical and structural generalization, aspects not explored by \citet{nichani2024understanding}.} {Through our effort to identify minimal toy settings where key tradeoffs, such as the impact of diversity on factual recall, are maintained, it might be possible to leverage some of the theoretical ideas from \citet{nichani2024understanding} to analyze our findings. However, this would require various non-trivial extensions, particularly incorporating the impact of diversity and adapting for an autoregressive generation setting rather than their last-token prediction.}

{We also review a growing body of recent works that have used Markov chains, as we do here to model the statistical stream,} to study various behavioral aspects of transformers in next-token prediction tasks. \citet{attention_with_markov} study the loss landscape properties of a single-layer transformer trained on sequences drawn from a fixed order-1 Markov chain, characterizing the influence of transition probabilities and architectural choices on the loss landscape.  \citet{statistical_induction_heads} demonstrates that transformers trained on sequences generated from random order-$1$ Markov chains develop the ability to perform in-context inference on unseen Markov chains by outputting bigram statistics inferred from the context. \citet{algorithmic_phases} extend this framework by examining the regime where training sequences are drawn from a fixed, finite set of Markov chains.  \citet{rajaraman2024transformers} has analyzed the representational capacity of transformers for in-context learning of order-$k$ Markov chains. {None of these works combines MCs with factual information, as we do here.} %\puneesh{need a connecting line here, relating MCs to our setup} 
% 
% \tina{cognitive stuff?}
% 
% \tina{mechanistic}
% 
% \tina{there were one or two relevant papers that i missed last time}
% 
More broadly, synthetic tasks have proven valuable for dissecting LM behavior beyond our focus, including self-attention mechanisms  \cite{tian2023scan,li2024mechanics,ren2024learning,makkuva2024attention}, in-context learning \citep{garg2022can,bietti2023birth,edelman2024evolution,algorithmic_phases}, and distributional associations \cite{chendistributional}.

\new{Additionally, our work is broadly related to prominent neuroscientific theories of language processing. For instance, some theories propose separate cortical circuits dedicated to syntactic and semantic processing, suggesting a functional segregation in the brain for handling language structure versus meaning \citep{Matchin2020}. Other research points to a stage-wise progression in human language comprehension, where an initial, rapid phase of syntactic analysis is thought to precede and scaffold deeper semantic integration \citep{Friederici2017}.}

\section{Metric overview}\label{app:loss_metrics}

% \subsection{Evaluation metrics}\label{app:loss_metrics}
\new{In this section, we review the metrics in Sec.~\ref{sec:evaluation_metrics} along with introducing the entropy-based metrics $\factloss$ and $\posloss$ used in Sec.~\ref{sec:param_ablation} to study the learning dynamics. Recall that,} whether ID or OOD, we measure the model's: (i) adherence to the composition rule between the two streams, (ii) accuracy of factual recall, and (iii) ability to follow the statistical patterns of the background template.
%
%{To formalize these metrics, consider $\prompt:= \xb_{\leq T/2}$ from template $n$ including token $\source$ from the factual pair $(\source, \target) \in \factset$ at position $\pos_{n,\source}$.

\new{For each ID/OOD template-fact combination $(n,k)$, we draw a sequence $\xb_{1:\seqlen}$, use its first half ($\xb_{1:\seqlen/2}$) as a prompt, and have the model generate the second half. We refer to the} generated tokens at positions $t>\seqlen/2$ by $(\hat{x}_{\seqlen/2+1}, \dots, \hat{x}_\seqlen)$. Let $\hat{\ellb}_t(\cdot)\in\R^V$ be the model's predicted \emph{logits} at position $t\leq\seqlen$ conditioned on input $(x_1,.\cdots, x_{\seqlen/2}, \hat{x}_{T/2+1},\cdots,\hat{x}_{t-1})$. Also, let $\hat{\pb}_t(\cdot) \in \Delta^{|\vset|}$ be the softmax \emph{probability} at this position. 
Without loss of generality, assume the index of the generic tokens is $[|\vmc|]$.

% To formalize these metrics, let $\hat{\pb}_t(\cdot) \in \Delta^{|\vset|}$ be the model's predictive distribution at position $t\leq\seqlen$, i.e., the output probability distribution conditioned on input tokens up to position ${t-1}$. To distinguish generated tokens at positions $t>\seqlen/2$ from prompt tokens at positions $t\leq \seqlen/2$, we denote by $(\hat{x}_{\seqlen/2+1}, \dots, \hat{x}_\seqlen)$ the tokens of the generated completion.

\begin{enumerate}[leftmargin=0pt, itemindent=1.2em]

    \item \textbf{\Structure accuracy{/loss}:}
   To obey the composition rule between the factual and statistical streams, the generated sequence should contain a fact tokens  from $\vk$ \emph{only} at the target position $\posTarget_{n}$, and all other positions should contain tokens from the generic vocabulary $\vmc$. Formally, we define \emph{\structure accuracy}  as: 
   % and \emph{loss} as:  \looseness=-1
   % 
    \begin{align}
        \posacc = \tfrac{1}{2} \big( \mathbbm{1} [\hat{x}_{\posTarget_{n}} \in \vk] + \tfrac{1}{\seqlen/2-1} \textstyle\sum\nolimits_{t > \seqlen/2, t \neq \posTarget_{n}} \mathbbm{1} [\hat{x}_{t} \in \vmc] \big).
    \end{align}
    % 
    % \begin{align}\label{eq:postion_acc}
    %     \posacc =\mathbbm{1} [\hat{x}_{\pos_{n,\target}} \in \vk]+ \frac{1}{T/2-1} \sum\nolimits_{t=\seqlen/2, t\neq\pos_{n,\target}}^\seqlen \, \mathbbm{1} [\hat{x}_{t} \in \vmc] \,,
    % \end{align}
    %The first term verifies that the model places a fact token in the designated template slot, while the second term confirms that every remaining position is filled exclusively with tokens from the MC's generic vocabulary $\vmc$. 
    \new{We also define the \emph{position loss} by measuring how much of the probability distribution at each position is accumulated on the correct subset of the vocabulary. Specifically,}
    % $        \posloss =-\log(\sum_{v \in \vk} \hat{\pb}_{\pos_{n,\target}}(v))- \frac{1}{T/2-1} \sum\nolimits_{t=\seqlen/2, t\neq\pos_\target}^\seqlen \, \log(\sum_{v\in\vmc}\hat{\pb}_{t}(v)))\,.$
    \begin{align}\label{eq:postion_loss}
        \posloss =-\tfrac{1}{2}\big(\log(\textstyle\sum\nolimits_{v \in \vk} \hat{\pb}_{\posTarget_n}(v))+ \frac{1}{T/2-1} \sum\nolimits_{t=\seqlen/2+1, t\neq\posTarget_n}^\seqlen \, \log(\sum\nolimits_{v\in\vmc}\hat{\pb}_{t}(v)))\big).
    \end{align}
   \item \textbf{Factual accuracy{/loss}:} 
    We define \emph{factual accuracy} as the existence of the correct prediction of the target in the prompt completion generated by the model:
    \begin{align}
        \factacc := \mathbbm{1} \big[ \{\hat{x}_t\}_{t > \seqlen/2} \cap \vk = \{\target_k\} \big].
    \end{align}
    % \begin{align}
    %     \factacc := \mathbbm{1}\left[\hat{x}_{\pos_{n,\target}} = \target\right].
    % \end{align}
    We accordingly define the \emph{factual loss} as 
    % $\factloss := -\log\left(\hat{\pb}_{\pos_{n,\target}}(f(a))\right)\,.$
    \begin{align}\label{eq:fact_loss}
        \factloss := -\log\left(\hat{\pb}_{\posTarget_n}(\target_k)\right)\,.
    \end{align}
    % \begin{align}\label{eq:fact_loss}
    %     \factloss := -\log\left(\hat{\pb}_t(f(a))\right)\,.
    % \end{align}

    \item \textbf{Statistical {loss}:}
Denote $\mathcal{G} \subseteq [\seqlen/2+1, \seqlen]$ the set of positions in the generated completion that are filled with generic tokens from $\vmc$. For each such position $t$, we compare the model’s distribution over generic tokens with the ground‑truth MC distribution ($\transitionmat_n$) of the template.
Concretely, keep the first $|\vmc|$ coordinates of the logit $\tilde{\ellb}_t = \ellb_t[\vmc]$ corresponding to the generic tokens and compute the model's distribution over $\vmc$ as $\tilde{\pb}_t=\sft{\tilde{\ellb_t}}$. Let $\pb^*_t$ be the row of $\transitionmat_n$ that corresponds to the preceding generic token of $\hat{x}_t$. We measure \emph{statistical loss} as:
%the KL of the model's probability $\tilde{\pb}_t$ and the ground-truth transition probability $\pb^*$, \new{averaged over positions of generic tokens in $\mathcal{G}$}:
% 
% completion that should be filled with generic tokens from $\vmc$ according to template $n$.
% \ct{Fix to ground-truth} 
% Denote $p_{{\rm{bi}},t}$ be the empirical bi-gram distribution over the generic tokens in the sequence $(x_1,\ldots,x_{T/2},\hat{x}_{T/2+1},\ldots,\hat{x}_{t-1})$. We use the KL divergence between this and the model output to assess how well the model follows the statistical pattern dictated by the MC structure:
    \begin{align}
        \KL := \frac{1}{|\mathcal{G}|} \sum\nolimits_{t \in \mathcal{G}} {\rm{KL}}\big( \tilde{\pb}_t \,\|\, \pb^*_t \big)\,.
    \end{align}

\end{enumerate}

% \subsection{Metric heatmaps for the experiments of section \ref{sec:}}

\section{Additional experiment results}\label{app:exp}

\subsection{\new{Experiment details of Sec.~\ref{sec:results}}}
In the experiments of Secs.~\ref{sec:results}, we use a decoder-only Transformer \citep{radford2018improving} with 4 layers and 4 heads trained auto-regressively with the standard next‑token prediction loss. Unless stated otherwise, each training sequence has length $\seqlen=50$. We use a template pool of size $\tmplnum=10$. We choose a generic vocabulary set of size $\vmsize=3$ for the MC and a fact set of $\factsize=100$ source-target pairs, which gives a total of $\vocabsize=203$ vocabularies. \new{We allocate the first $2\factsize$ tokens in the vocabulary to the fact stream and build up the fact set $\factset$ by randomly pairing these tokens together. We build the statistical stream by generating $\tmplnum$ random transition matrices for each template, and $\tmplnum$ distinct position placeholders $\pospair_n=(\posSource_n,\posTarget_n)$ along the sequence.} We train the models for \new{$30k$ iterations} and a batch size of 64, with  AdamW \citep{loshchilov2017decoupled} and a fixed \new{learning rate of $10^{-3}$}. 

\new{
Unless otherwise specified, all reported metrics are averaged over three independent runs, each with different random seeds for both data generation (which controls the random generation of $\factset$, $\{\transitionmat_n\}_{n\in[\tmplnum]}$ and $\{\pospair_n\}_{n\in[\tmplnum]}$) and model initialization. For experiments that sweep across the diversity level $\dvr$ (e.g., Fig.~\ref{fig:heatmaps_main}), we fix these seeds for a given run and vary only $\dvr$ from 0.1 to 0.9.}

% \tina{Model size}

% \tina{Long training?? -> didn't work :/}

% \tina{OOD V2!}

\subsection{Impact of model size}\label{app:exp_size}
\begin{figure}[htbp]
  \centering
  \begin{tabular}{@{\hspace{-45pt}}c@{\hspace{40pt}}c@{\hspace{35pt}}c@{}}

  % ─── header row: one cell for col 1, one spanning cols 2–3 ─────────
  % \multicolumn{1}{c@{\hspace{70pt}}}{\fontsize{9pt}{8pt}\selectfont\textbf{(a) ID mask ($\exposmatin$)}}
  % & \multicolumn{2}{c}{\fontsize{9pt}{8pt}\selectfont\textbf{(b) Factual accuracy}} \\[8pt]
  
  % ─── KL ───────────────────────────────
    \begin{subfigure}[t]{0.25\textwidth}
        \centering
        \begin{tikzpicture}[remember picture]
            \node at (0,0) {\includegraphics[scale=0.25]{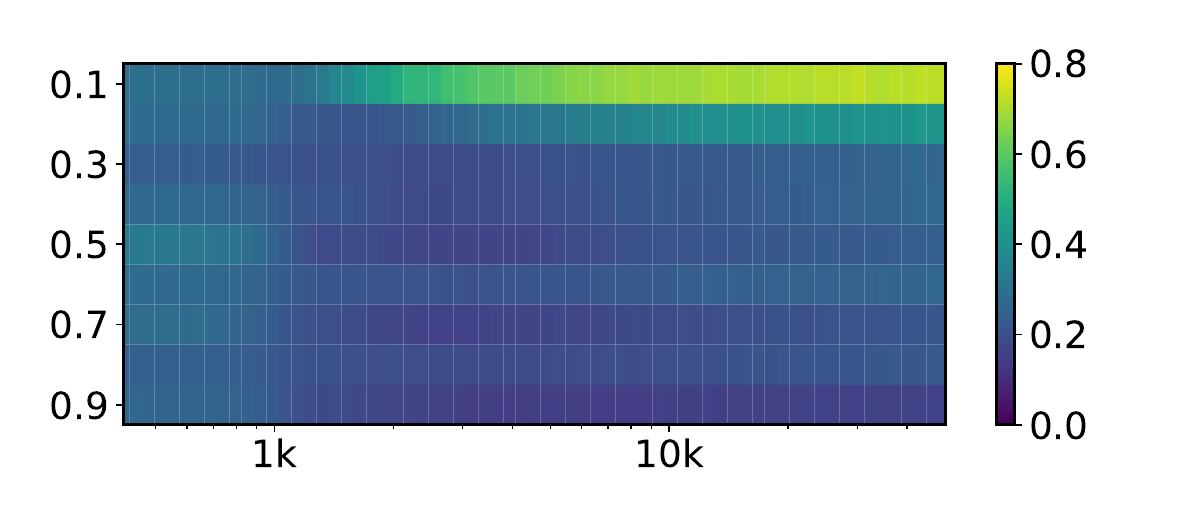}};
            % \node[scale=0.8] at (0.0, 1.3) { \textbf{ }};
            \node[scale=0.9,rotate=90] at (-2.6, 0.1) { \textbf{1 Layer}};
            % \node[scale=0.8] at (0.0, -1.2) {iteration ($\times 10^3$)};
            \node[scale=0.8] at (0.0, 1.0) { \textbf{ \mcexp{10}}};
            \node[scale=1.2] at (0.0, 1.5) { };

        \end{tikzpicture}
    \end{subfigure} &
    % \hspace{70pt}%
    % \end{minipage}%
\colTwoThreeShift{
    \begin{subfigure}[t]{0.25\textwidth}
        \centering
        \begin{tikzpicture}[remember picture]
            \node at (0,0) {\includegraphics[scale=0.25]{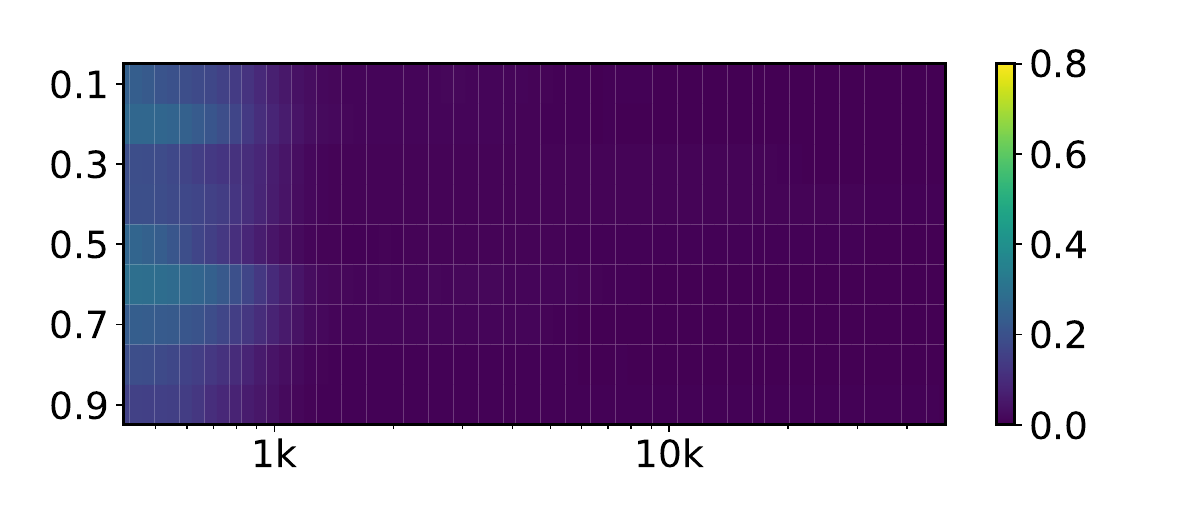}};
            % \node[scale=0.8,rotate=90] at (-2.7, 0.0) { \textbf{\posexp{10} }};            \node[scale=0.8] at (0.0, -1.2) { \textbf{ }};
            % \node[scale=0.8] at (0.0, -1.2) {iteration ($\times 10^3$)};
            \node[scale=0.8] at (0.0, 1.0) { \textbf{ \posexp{10}}};
            \node[scale=1.] at (0.0, 1.5) { \textbf{(a) Statistical Loss}};

        \end{tikzpicture}
    \end{subfigure}
    }&%
    % \hspace{50pt}%
\colTwoThreeShift{    
    \begin{subfigure}[t]{0.25\textwidth}
        \centering
        \begin{tikzpicture}[remember picture]
            \node at (0,0) {\includegraphics[scale=0.25]{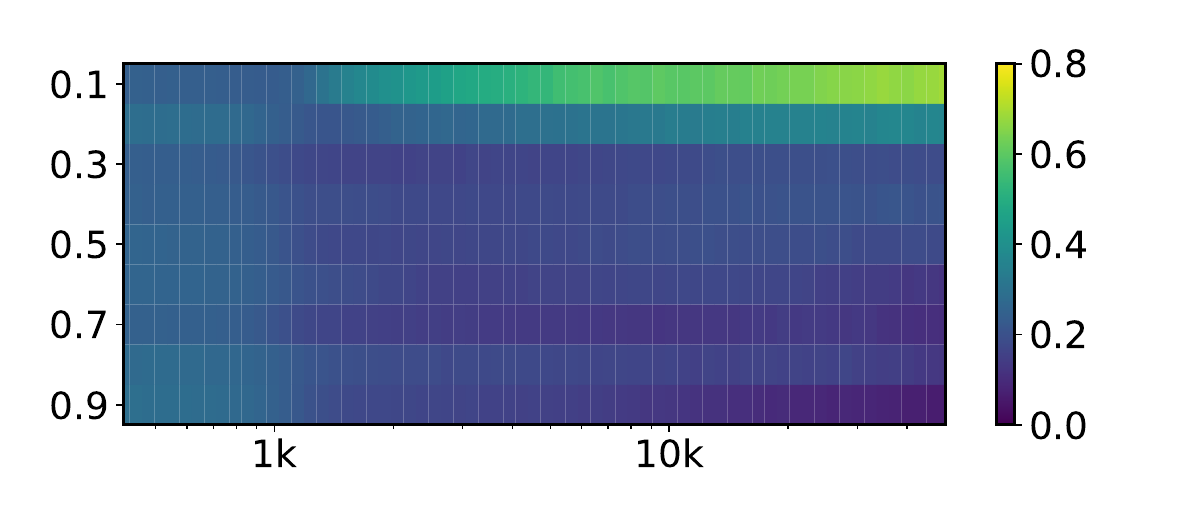}};            
            % \node[scale=0.8] at (0.0, -1.2) {iteration ($\times 10^3$)};
            \node[scale=0.8] at (0.0, 1.0) { \textbf{ \mixexp{10}}};
            
            \node[scale=1.2] at (0.0, 1.6) { };
            
        \end{tikzpicture}
    \end{subfigure}}\\[-17pt]
       \begin{subfigure}[t]{0.25\textwidth}
        \centering
        \begin{tikzpicture}[remember picture]
            \node at (0,0) {\includegraphics[scale=0.25]{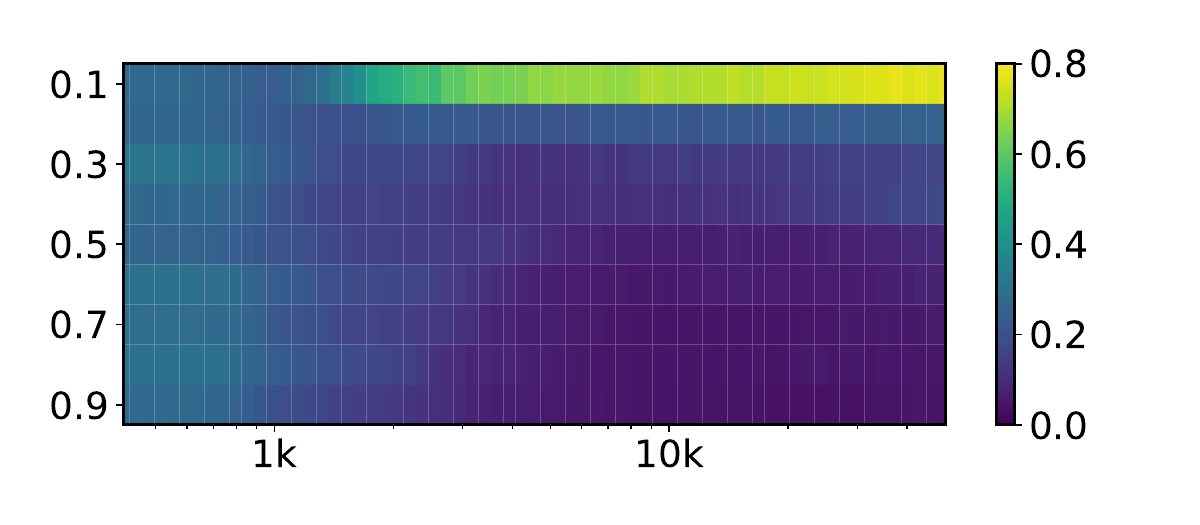}};
            % \node[scale=0.8] at (0.0, 1.3) { \textbf{ }};
            \node[scale=0.9,rotate=90] at (-2.6, 0.1) { \textbf{4 Layers}};
            % \node[scale=0.8] at (0.0, -1.2) {iteration ($\times 10^3$)};
            
        \end{tikzpicture}
    \end{subfigure} &
    % \hspace{70pt}%
    % \end{minipage}%
    \begin{subfigure}[t]{0.25\textwidth}
        \centering
        \begin{tikzpicture}[remember picture]
            \node at (0,0) {\includegraphics[scale=0.25]{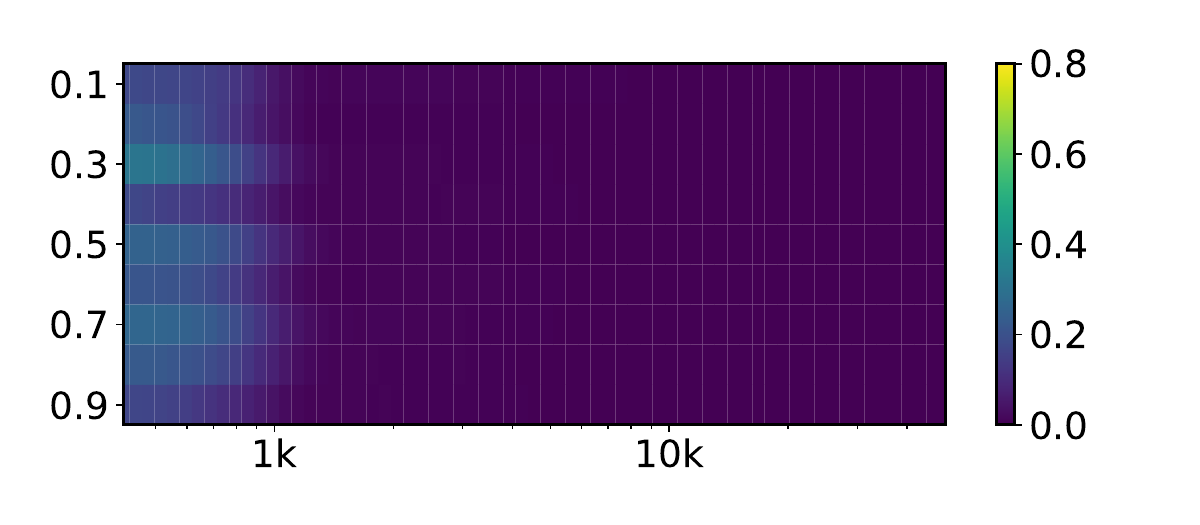}};
            % \node[scale=0.8,rotate=90] at (-2.7, 0.0) { \textbf{\posexp{10} }};            \node[scale=0.8] at (0.0, -1.2) { \textbf{ }};
            % \node[scale=0.8] at (0.0, -1.2) {iteration ($\times 10^3$)};

        \end{tikzpicture}
    \end{subfigure} &%
    % \hspace{50pt}%
    \begin{subfigure}[t]{0.25\textwidth}
        \centering
        \begin{tikzpicture}[remember picture]
            \node at (0,0) {\includegraphics[scale=0.25]{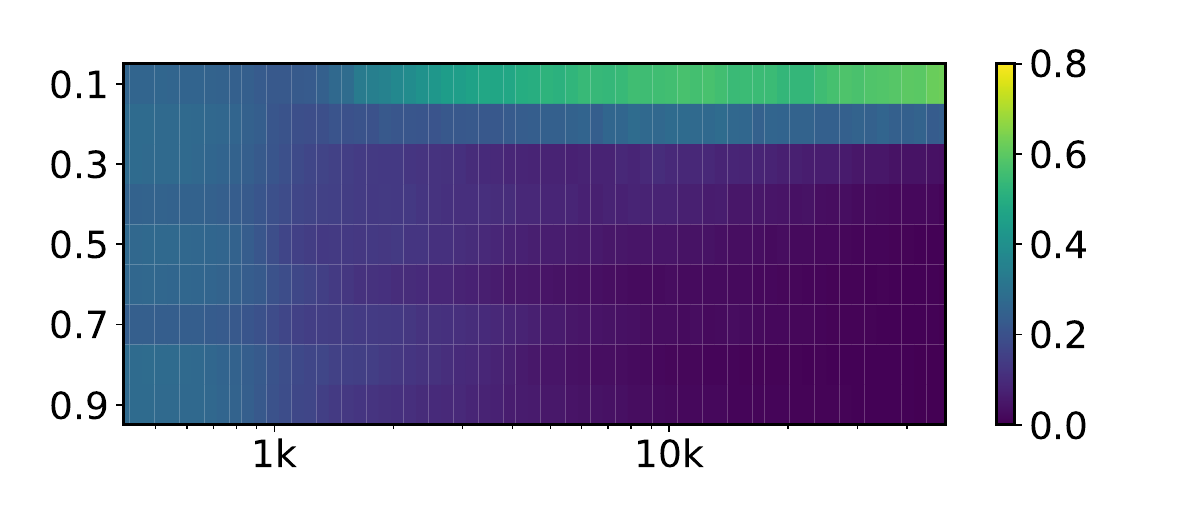}};            
            % \node[scale=0.8] at (0.0, -1.2) {iteration ($\times 10^3$)};
            
        \end{tikzpicture}
    \end{subfigure}\\[-17pt]
       \begin{subfigure}[t]{0.25\textwidth}
        \centering
        \begin{tikzpicture}[remember picture]
            \node at (0,0) {\includegraphics[scale=0.25]{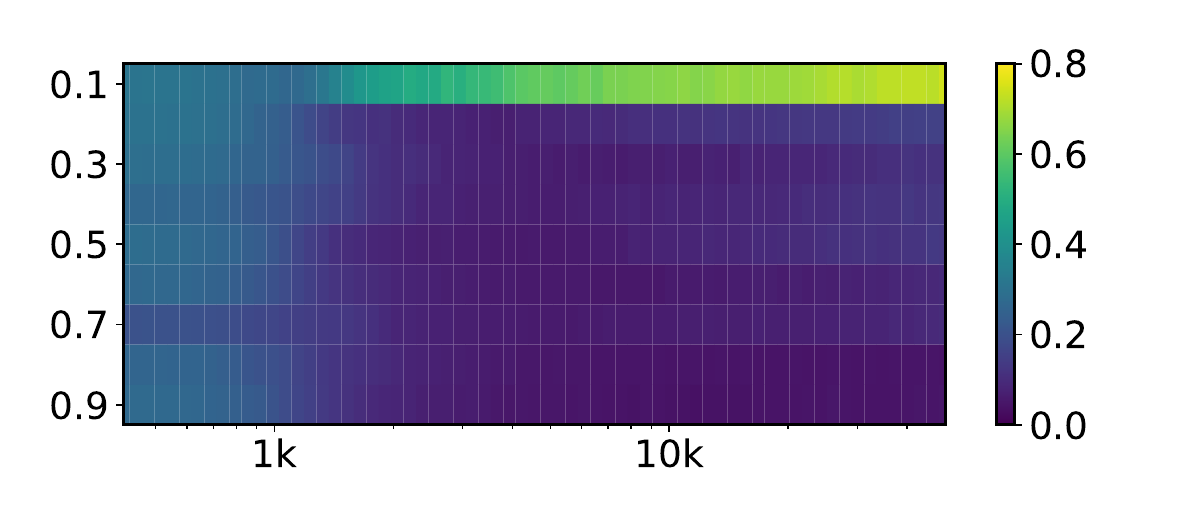}};
            % \node[scale=0.8] at (0.0, 1.3) { \textbf{ }};
            \node[scale=0.9,rotate=90] at (-2.6, 0.1) { \textbf{10 Layers}};
            % \node[scale=0.8] at (0.0, -1.2) {iteration ($\times 10^3$)};
            
        \end{tikzpicture}
    \end{subfigure} &
    % \hspace{70pt}%
    % \end{minipage}%
    \begin{subfigure}[t]{0.25\textwidth}
        \centering
        \begin{tikzpicture}[remember picture]
            \node at (0,0) {\includegraphics[scale=0.25]{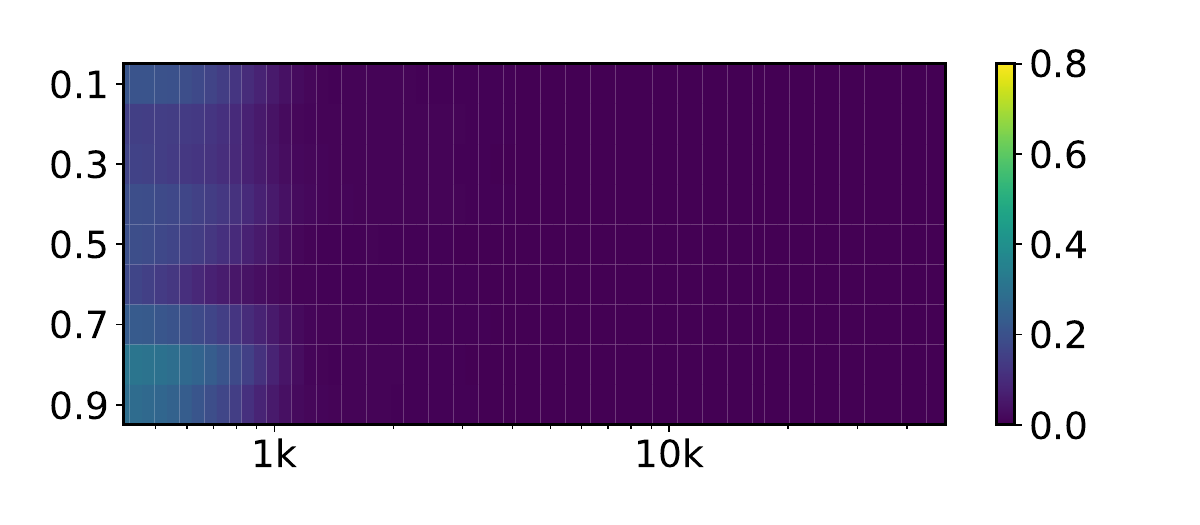}};
            % \node[scale=0.8,rotate=90] at (-2.7, 0.0) { \textbf{\posexp{10} }};            \node[scale=0.8] at (0.0, -1.2) { \textbf{ }};
            % \node[scale=0.8] at (0.0, -1.2) {iteration ($\times 10^3$)};

        \end{tikzpicture}
    \end{subfigure} &%
    % \hspace{50pt}%
    \begin{subfigure}[t]{0.25\textwidth}
        \centering
        \begin{tikzpicture}[remember picture]
            \node at (0,0) {\includegraphics[scale=0.25]{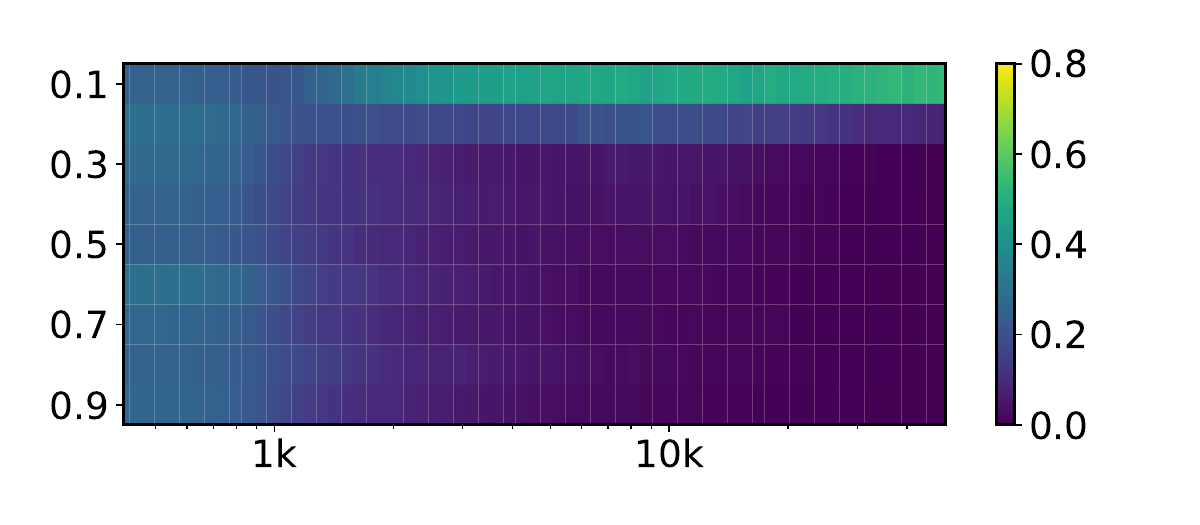}};            
            % \node[scale=0.8] at (0.0, -1.2) {iteration ($\times 10^3$)};
            
        \end{tikzpicture}
    \end{subfigure}\\[-1pt]
  % ─── POSITION ACC ───────────────────────────────
  
    \begin{subfigure}[t]{0.25\textwidth}
        \centering
        \begin{tikzpicture}[remember picture]
            \node at (0,0) {\includegraphics[scale=0.25]{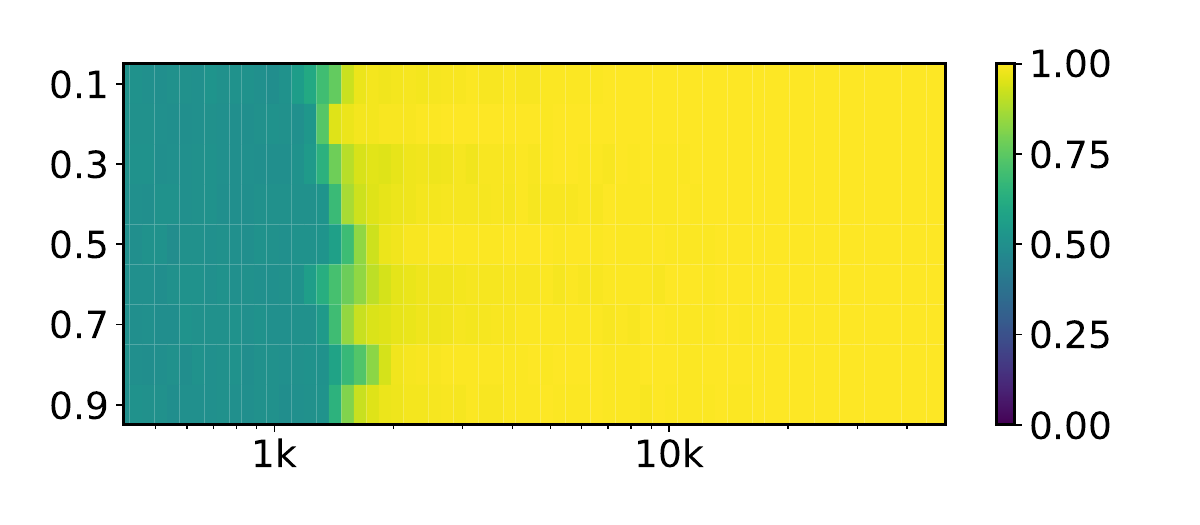}};
            % \node[scale=0.8] at (0.0, 1.3) { \textbf{ }};
            \node[scale=0.9,rotate=90] at (-2.6, 0.1) { \textbf{1 Layer}};
            % \node[scale=0.8] at (0.0, -1.2) {iteration ($\times 10^3$)};
            % \node[scale=0.9] at (0.0, 1.1) { \textbf{ \mcexp{10}}};
            \node[scale=1.2] at (0.0, 1.2) { };

        \end{tikzpicture}
    \end{subfigure} &
    % \hspace{70pt}%
    % \end{minipage}%
    \colTwoThreeShift{
    \begin{subfigure}[t]{0.25\textwidth}
        \centering
        \begin{tikzpicture}[remember picture]
            \node at (0,0) {\includegraphics[scale=0.25]{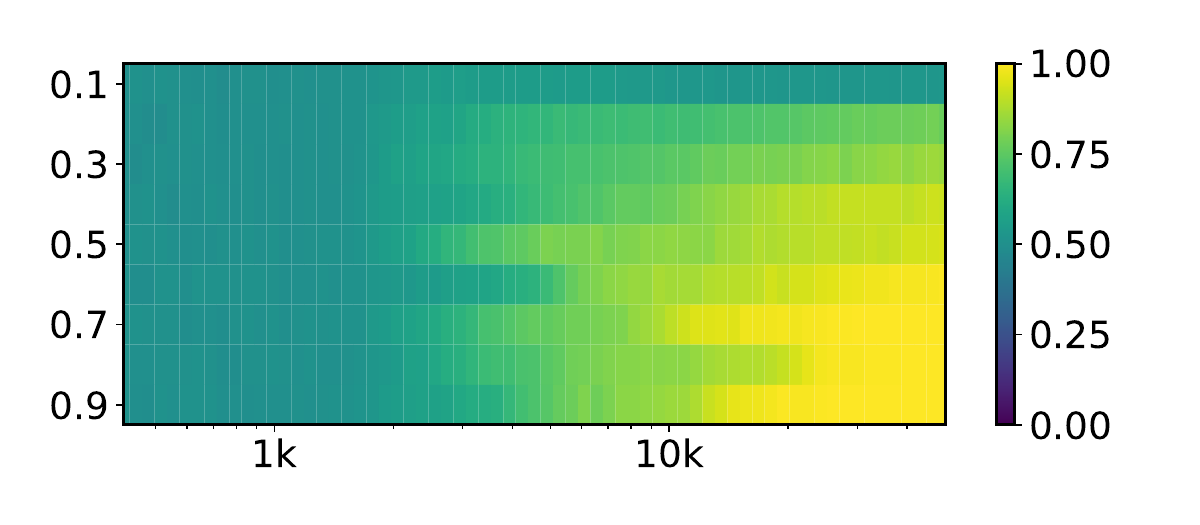}};
            % \node[scale=0.8,rotate=90] at (-2.7, 0.0) { \textbf{\posexp{10} }};            \node[scale=0.8] at (0.0, -1.2) { \textbf{ }};
            % \node[scale=0.8] at (0.0, -1.2) {iteration ($\times 10^3$)};
            % \node[scale=0.9] at (0.0, 1.1) { \textbf{ \posexp{10}}};
            \node[scale=1.] at (0.0, 1.2) { \textbf{(b) Position Accuracy}};

        \end{tikzpicture}
    \end{subfigure}}&%
    % \hspace{50pt}%

    \colTwoThreeShift{
    \begin{subfigure}[t]{0.25\textwidth}
        \centering
        \begin{tikzpicture}[remember picture]
            \node at (0,0) {\includegraphics[scale=0.25]{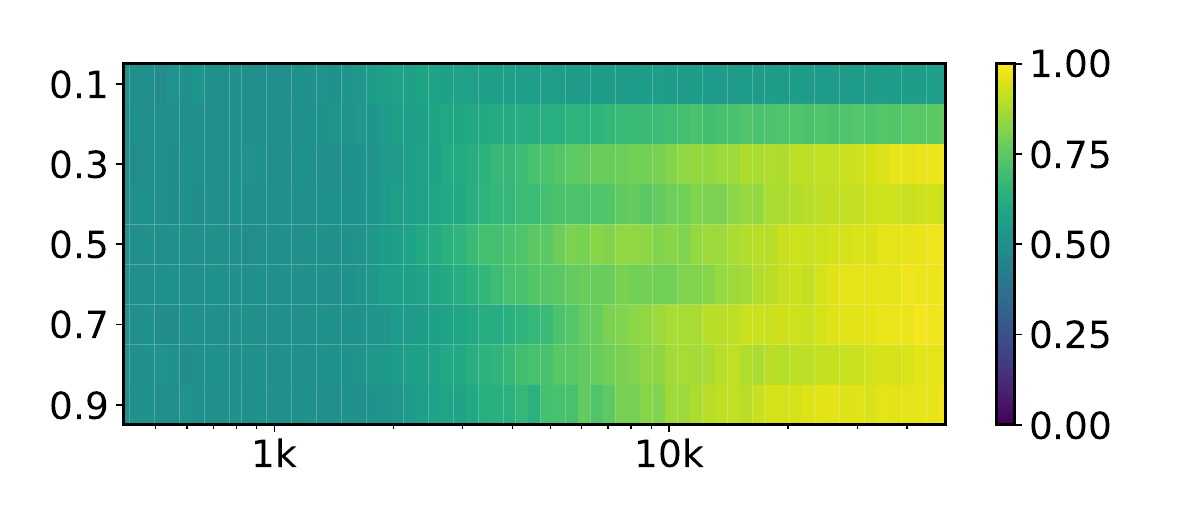}};            
            % \node[scale=0.8] at (0.0, -1.2) {iteration ($\times 10^3$)};
            % \node[scale=0.9] at (0.0, 1.1) { \textbf{ \mixexp{10}}};
            \node[scale=1.2] at (0.0, 1.3) { };

        \end{tikzpicture}
    \end{subfigure}}\\[-17pt]
       \begin{subfigure}[t]{0.25\textwidth}
        \centering
        \begin{tikzpicture}[remember picture]
            \node at (0,0) {\includegraphics[scale=0.25]{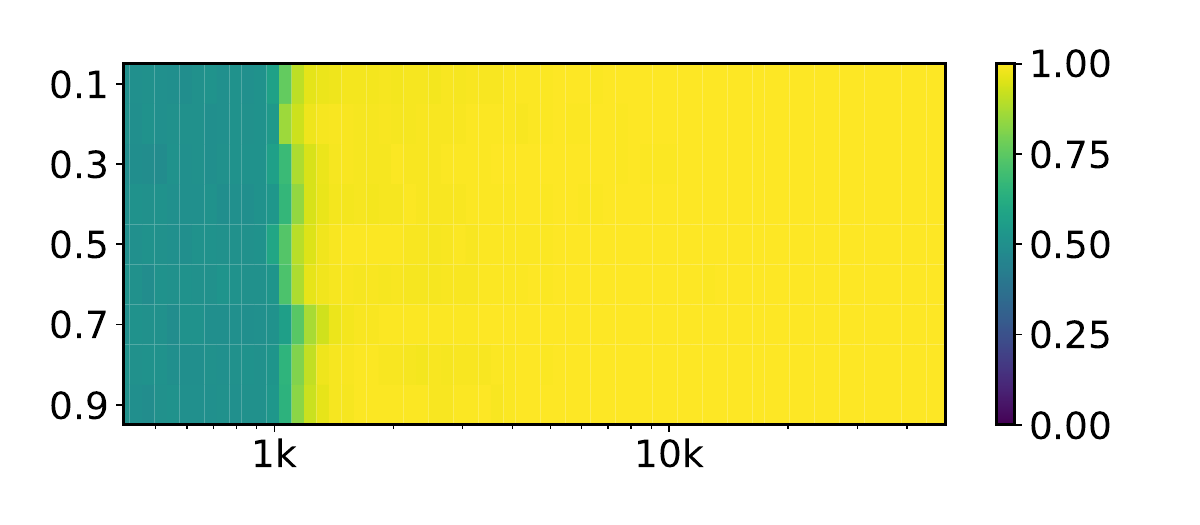}};
            % \node[scale=0.8] at (0.0, 1.3) { \textbf{ }};
            \node[scale=0.9,rotate=90] at (-2.6, 0.1) { \textbf{4 Layers}};
            % \node[scale=0.8] at (0.0, -1.2) {iteration ($\times 10^3$)};
            
        \end{tikzpicture}
    \end{subfigure} &
    % \hspace{70pt}%
    % \end{minipage}%
    \begin{subfigure}[t]{0.25\textwidth}
        \centering
        \begin{tikzpicture}[remember picture]
            \node at (0,0) {\includegraphics[scale=0.25]{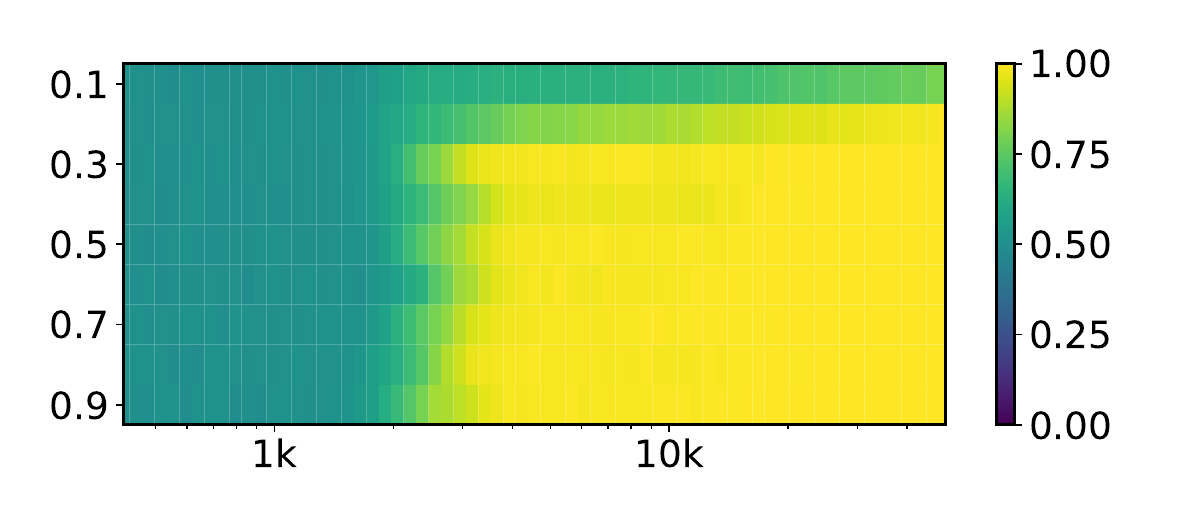}};
            % \node[scale=0.8,rotate=90] at (-2.7, 0.0) { \textbf{\posexp{10} }};            \node[scale=0.8] at (0.0, -1.2) { \textbf{ }};
            % \node[scale=0.8] at (0.0, -1.2) {iteration ($\times 10^3$)};

        \end{tikzpicture}
    \end{subfigure} &%
    % \hspace{50pt}%
    \begin{subfigure}[t]{0.25\textwidth}
        \centering
        \begin{tikzpicture}[remember picture]
            \node at (0,0) {\includegraphics[scale=0.25]{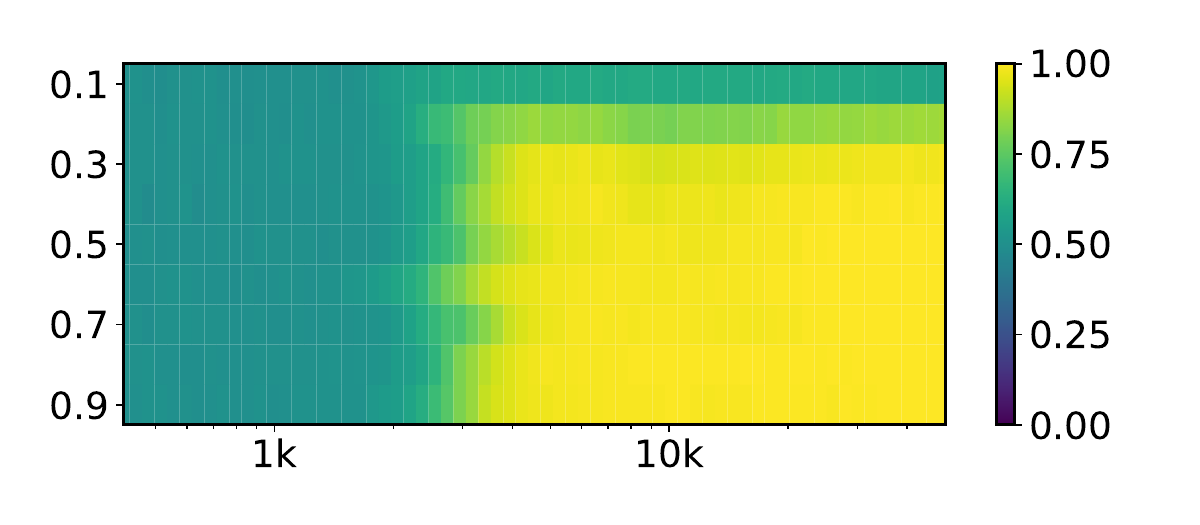}};            
            % \node[scale=0.8] at (0.0, -1.2) {iteration ($\times 10^3$)};
            
        \end{tikzpicture}
    \end{subfigure}\\[-17pt]
       \begin{subfigure}[t]{0.25\textwidth}
        \centering
        \begin{tikzpicture}[remember picture]
            \node at (0,0) {\includegraphics[scale=0.25]{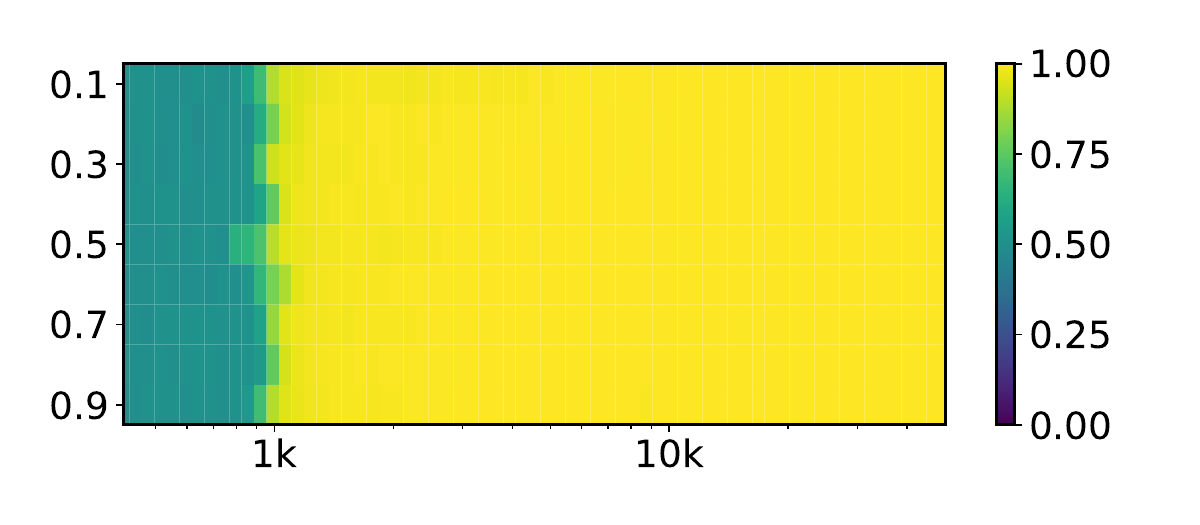}};
            % \node[scale=0.8] at (0.0, 1.3) { \textbf{ }};
            \node[scale=0.9,rotate=90] at (-2.6, 0.1) { \textbf{10 Layers}};
            % \node[scale=0.8] at (0.0, -1.2) {iteration ($\times 10^3$)};
            
        \end{tikzpicture}
    \end{subfigure} &
    % \hspace{70pt}%
    % \end{minipage}%
    \begin{subfigure}[t]{0.25\textwidth}
        \centering
        \begin{tikzpicture}[remember picture]
            \node at (0,0) {\includegraphics[scale=0.25]{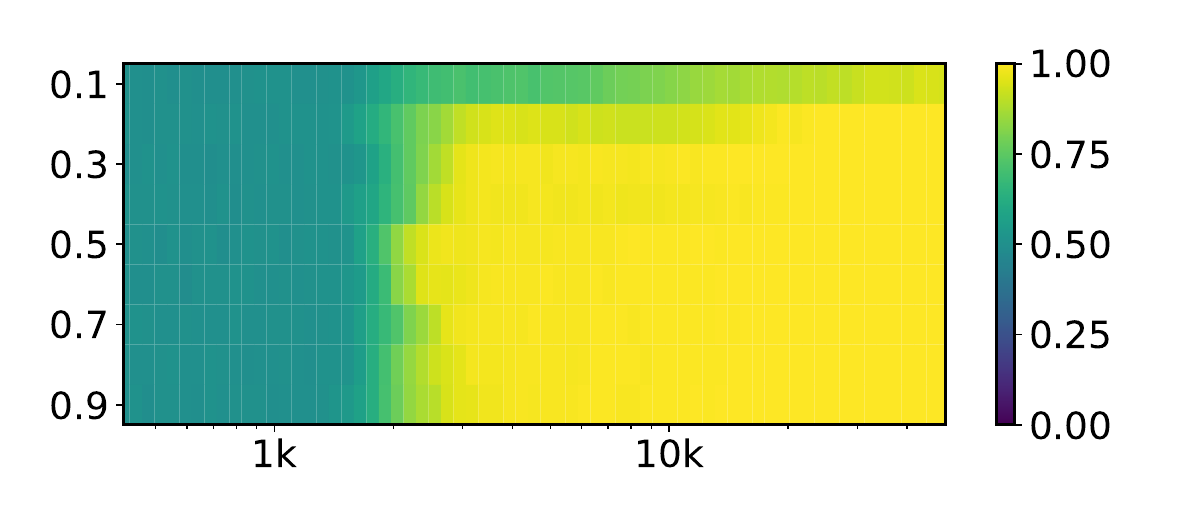}};
            % \node[scale=0.8,rotate=90] at (-2.7, 0.0) { \textbf{\posexp{10} }};            \node[scale=0.8] at (0.0, -1.2) { \textbf{ }};
            % \node[scale=0.8] at (0.0, -1.2) {iteration ($\times 10^3$)};

        \end{tikzpicture}
    \end{subfigure} &%
    % \hspace{50pt}%
    \begin{subfigure}[t]{0.25\textwidth}
        \centering
        \begin{tikzpicture}[remember picture]
            \node at (0,0) {\includegraphics[scale=0.25]{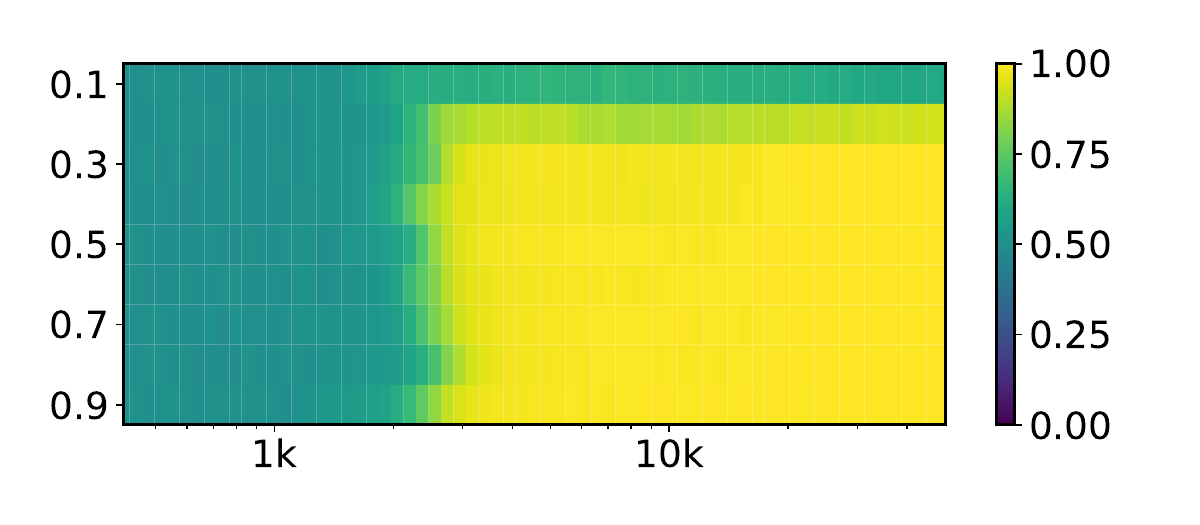}};            
            % \node[scale=0.8] at (0.0, -1.2) {iteration ($\times 10^3$)};
            
        \end{tikzpicture}
    \end{subfigure}\\[-10pt]
    
  % ─── FactACC ───────────────────────────────
    \begin{subfigure}[t]{0.25\textwidth}
        \centering
        \begin{tikzpicture}[remember picture]
            \node at (0,0) {\includegraphics[scale=0.25]{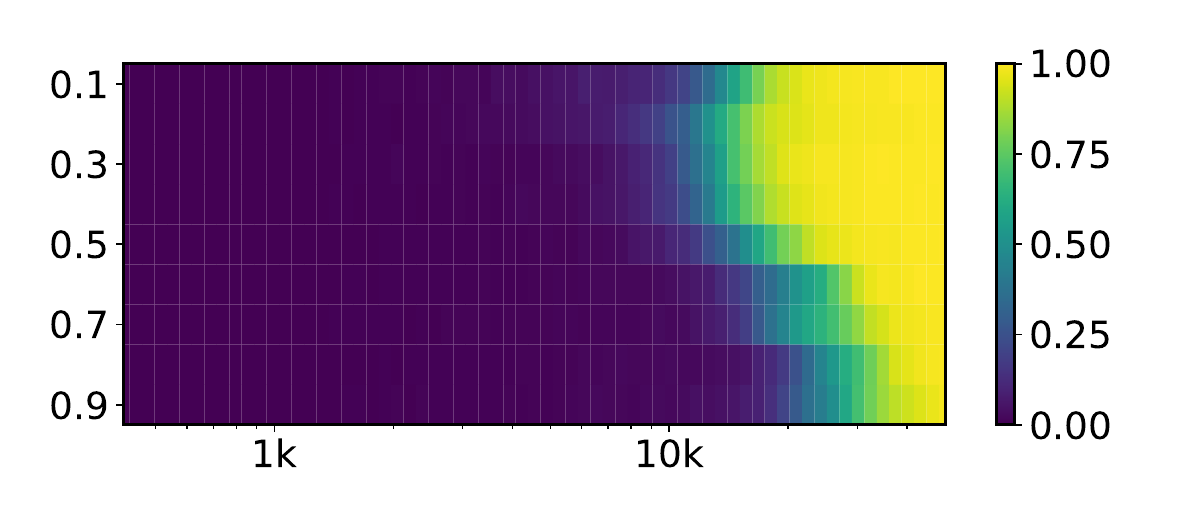}};
            % \node[scale=0.8] at (0.0, 1.3) { \textbf{ }};
            \node[scale=0.9,rotate=90] at (-2.6, 0.1) { \textbf{1 Layer}};
            % \node[scale=0.8] at (0.0, -1.2) {iteration ($\times 10^3$)};
            \node[scale=1.0] at (0.0, 1.4) { };

        \end{tikzpicture}
    \end{subfigure} &
    % \hspace{70pt}%
    % \end{minipage}%
    \begin{subfigure}[t]{0.25\textwidth}
        \centering
        \begin{tikzpicture}[remember picture]
            \node at (0,0) {\includegraphics[scale=0.25]{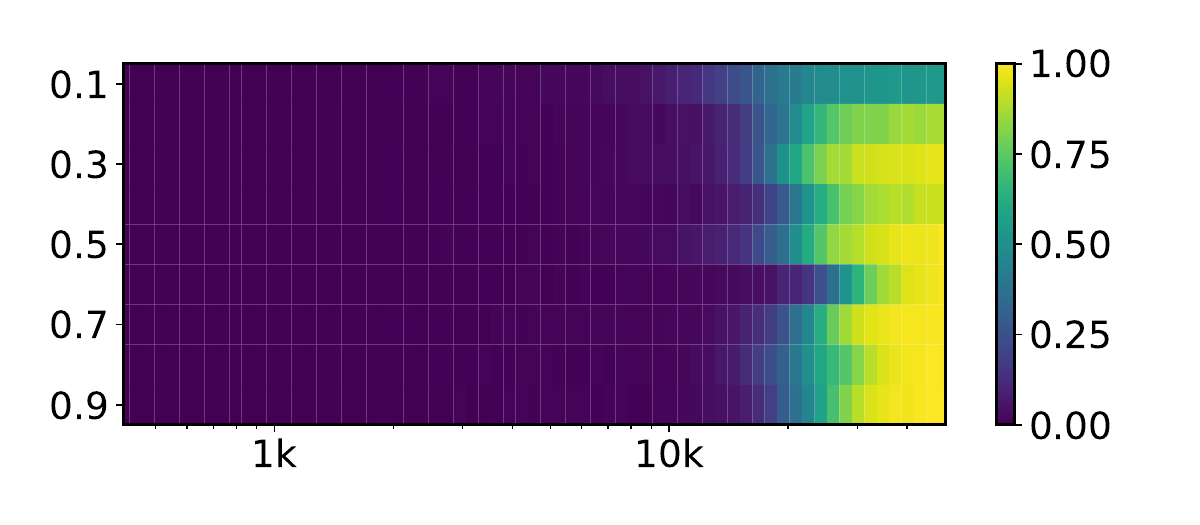}};
            % \node[scale=0.8,rotate=90] at (-2.7, 0.0) { \textbf{\posexp{10} }};            \node[scale=0.8] at (0.0, -1.2) { \textbf{ }};
            % \node[scale=0.8] at (0.0, -1.2) {iteration ($\times 10^3$)};
            % \node[scale=0.9] at (0.0, 1.1) { \textbf{ \posexp{10}}};
            \node[scale=1.0] at (0.0, 1.2) { \textbf{(c) Factual Accuracy}};

        \end{tikzpicture}
    \end{subfigure} &%
    % \hspace{50pt}%
    \begin{subfigure}[t]{0.25\textwidth}
        \centering
        \begin{tikzpicture}[remember picture]
            \node at (0,0) {\includegraphics[scale=0.25]{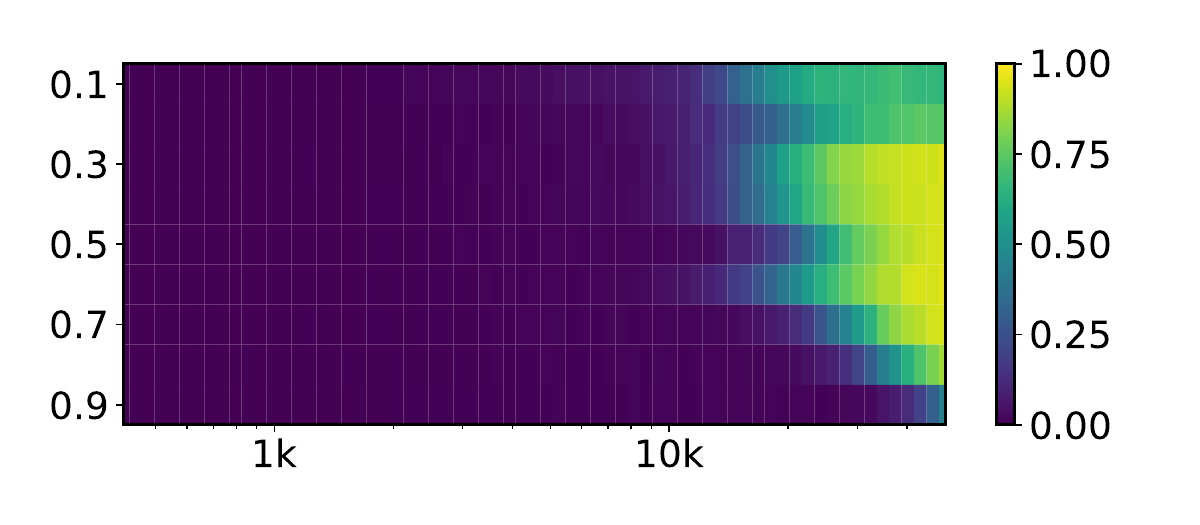}};            
            % \node[scale=0.8] at (0.0, -1.2) {iteration ($\times 10^3$)};
            % \node[scale=0.9] at (0.0, 1.1) { \textbf{ \mixexp{10}}};
            \node[scale=1.0] at (0.0, 1.4) { };

        \end{tikzpicture}
    \end{subfigure}\\[-17pt]
       \begin{subfigure}[t]{0.25\textwidth}
        \centering
        \begin{tikzpicture}[remember picture]
            \node at (0,0) {\includegraphics[scale=0.25]{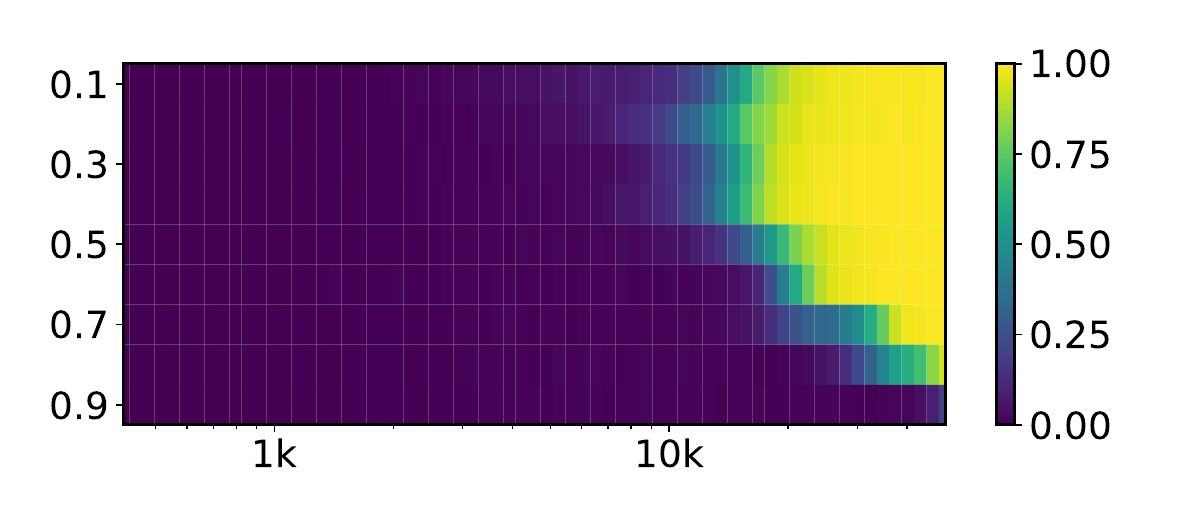}};
            % \node[scale=0.8] at (0.0, 1.3) { \textbf{ }};
            \node[scale=0.9,rotate=90] at (-2.6, 0.1) { \textbf{4 Layers}};
            % \node[scale=0.8] at (0.0, -1.2) {iteration ($\times 10^3$)};
            
        \end{tikzpicture}
    \end{subfigure} &
    % \hspace{70pt}%
    % \end{minipage}%
    \begin{subfigure}[t]{0.25\textwidth}
        \centering
        \begin{tikzpicture}[remember picture]
            \node at (0,0) {\includegraphics[scale=0.25]{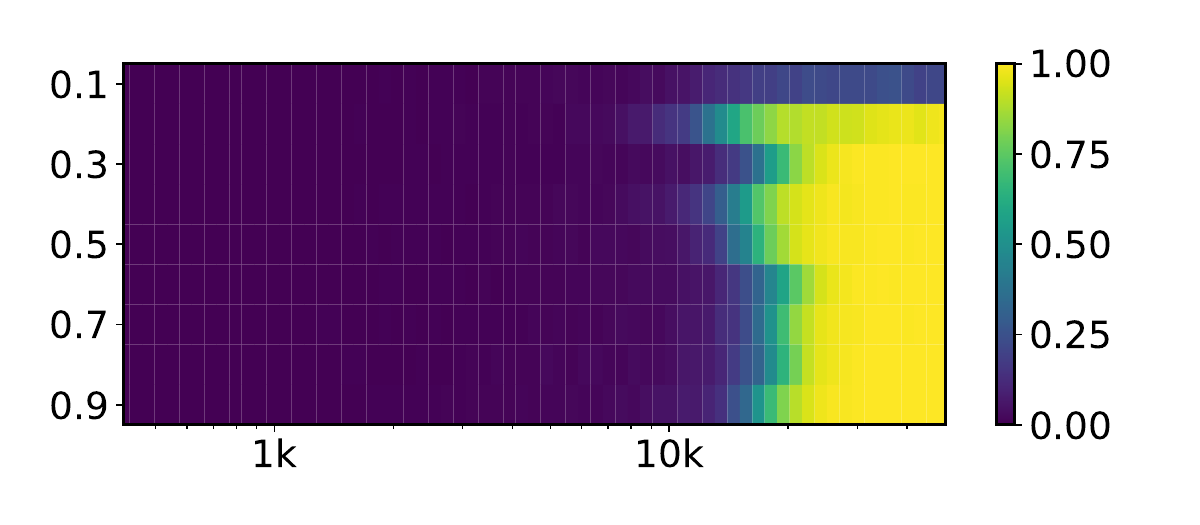}};
            % \node[scale=0.8,rotate=90] at (-2.7, 0.0) { \textbf{\posexp{10} }};            \node[scale=0.8] at (0.0, -1.2) { \textbf{ }};
            % \node[scale=0.8] at (0.0, -1.2) {iteration ($\times 10^3$)};

        \end{tikzpicture}
    \end{subfigure} &%
    % \hspace{50pt}%
    \begin{subfigure}[t]{0.25\textwidth}
        \centering
        \begin{tikzpicture}[remember picture]
            \node at (0,0) {\includegraphics[scale=0.25]{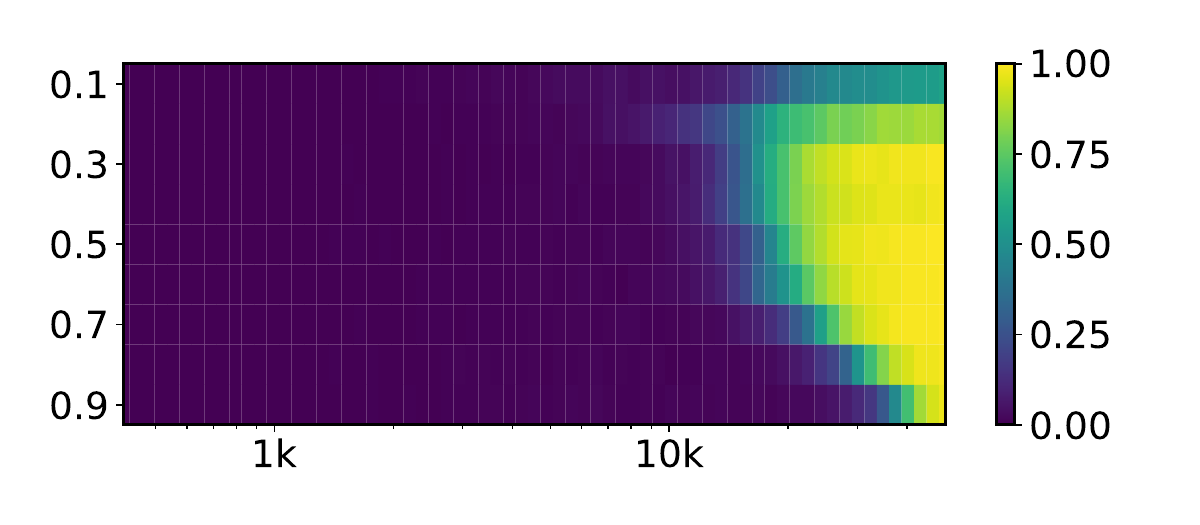}};            
            % \node[scale=0.8] at (0.0, -1.2) {iteration ($\times 10^3$)};
            
        \end{tikzpicture}
    \end{subfigure}\\[-17pt]
       \begin{subfigure}[t]{0.25\textwidth}
        \centering
        \begin{tikzpicture}[remember picture]
            \node at (0,0) {\includegraphics[scale=0.25]{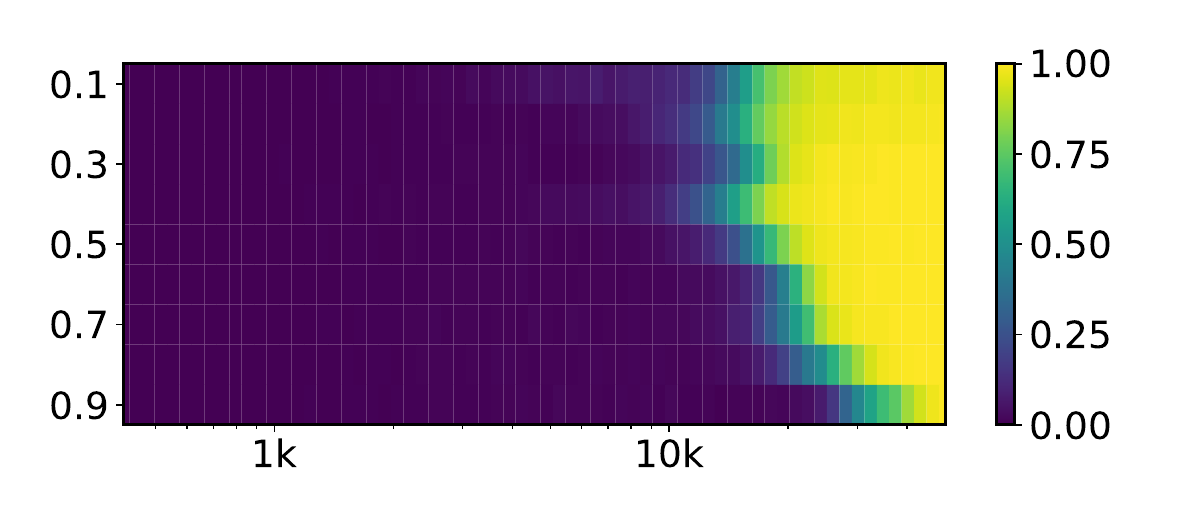}};
            % \node[scale=0.8] at (0.0, 1.3) { \textbf{ }};
            \node[scale=0.9,rotate=90] at (-2.6, 0.1) { \textbf{10 Layers}};
            \node[scale=0.8] at (0.0, -1.2) {Iterations};
            
        \end{tikzpicture}
    \end{subfigure} &
    % \hspace{70pt}%
    % \end{minipage}%
    \begin{subfigure}[t]{0.25\textwidth}
        \centering
        \begin{tikzpicture}[remember picture]
            \node at (0,0) {\includegraphics[scale=0.25]{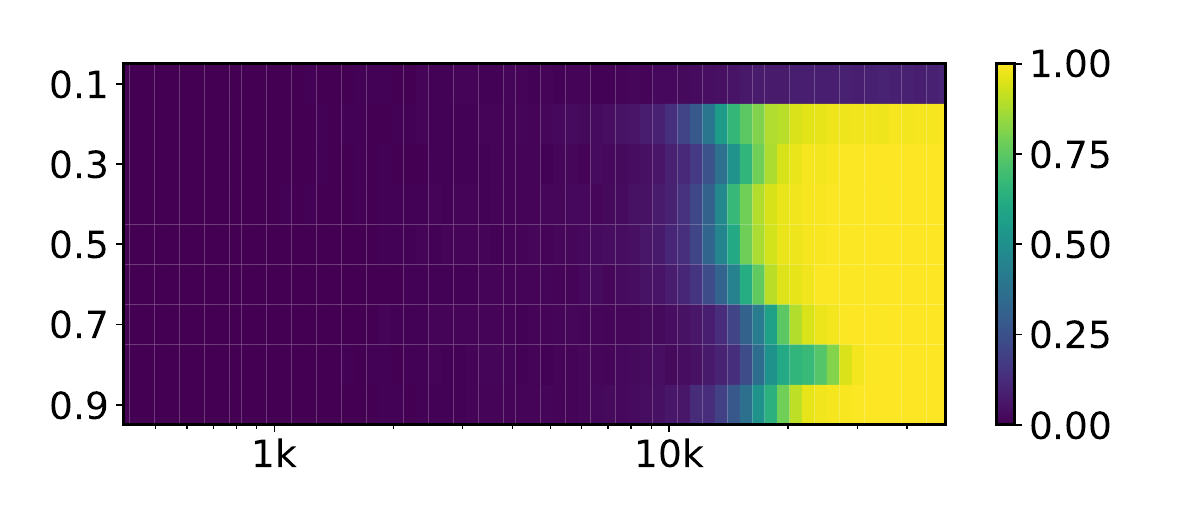}};
            % \node[scale=0.8,rotate=90] at (-2.7, 0.0) { \textbf{\posexp{10} }};            \node[scale=0.8] at (0.0, -1.2) { \textbf{ }};
            \node[scale=0.8] at (0.0, -1.2) {Iterations};

        \end{tikzpicture}
    \end{subfigure} &%
    % \hspace{50pt}%
    \begin{subfigure}[t]{0.25\textwidth}
        \centering
        \begin{tikzpicture}[remember picture]
            \node at (0,0) {\includegraphics[scale=0.25]{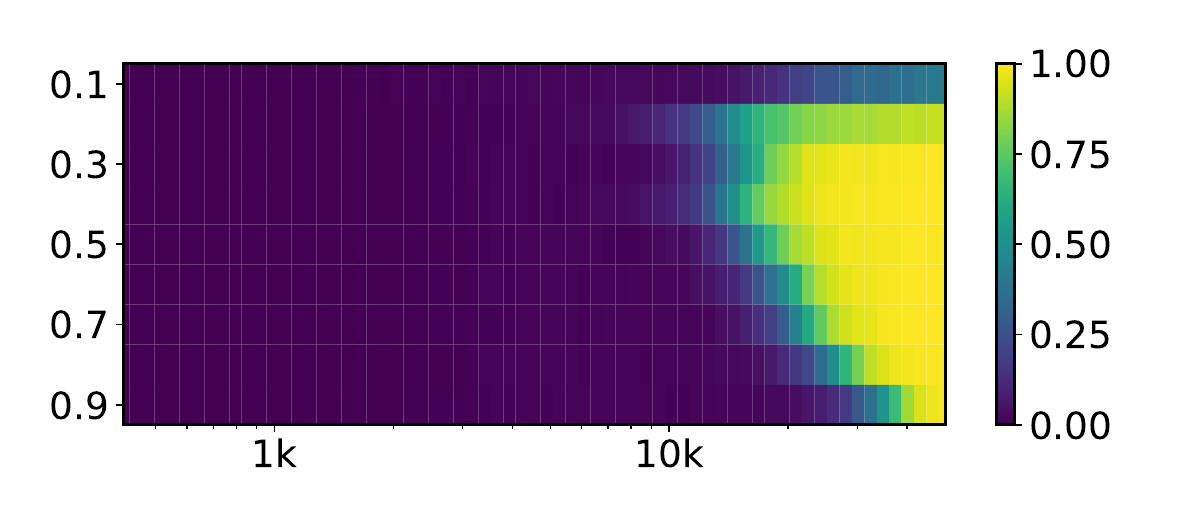}};            
            \node[scale=0.8] at (0.0, -1.2) {Iterations};
            
        \end{tikzpicture}
    \end{subfigure}%\\[-24pt]
    
  \end{tabular}
  % \vspace{-0.1in}
  \caption{\textbf{Impact of model size.} Replication of the experiments in Sec.~\ref{sec:results} (4-layer) with smaller (1-layer) and larger (10-layer) models on OOD sequences. {With increased model capacity, we need fewer iterations to achieve the same level of performance. However, model capacity alone cannot alleviate the failure of factual recall at low diversity levels.%\tina{update fig format if data still saved} 
  }  %\tina{fix alignments}
    }
    % \vspace{-0.2in}
  \label{fig:model_size}
\end{figure}
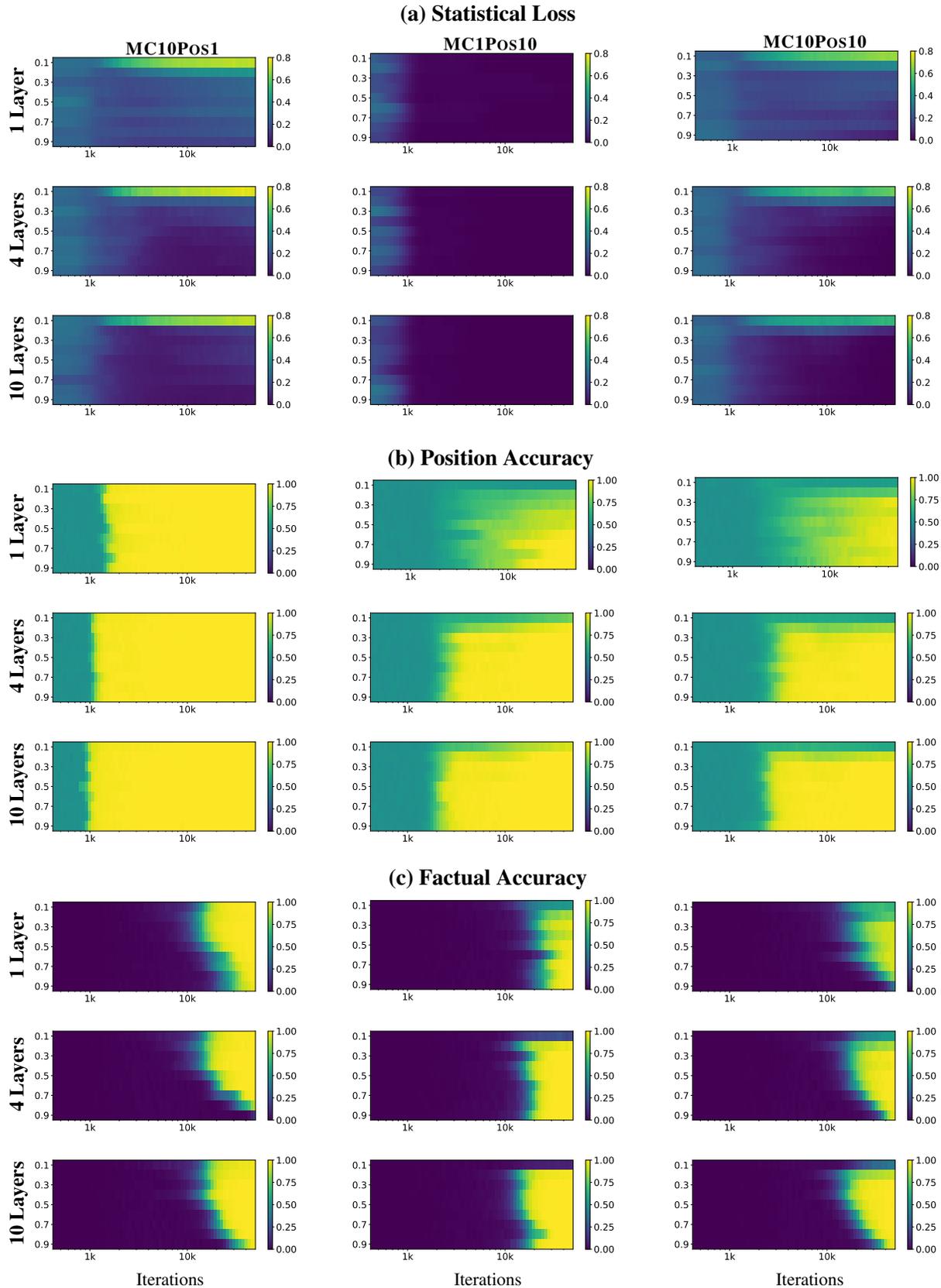

Here, we repeat our main experiments -- originally with 4-layer transformer -- this time with 1-layer and 10-layer models. In this section, all experiments are run with a learning rate $10^{-4}$. In Fig.~\ref{fig:model_size}, we gather heatmaps of (a) $\KL$, (b) $\posacc$, and (c) $\factacc$ on OOD sequences for each model size. \looseness=-1

Across all model sizes, the impact of diversity level and training duration is identical. %\ct{shall we talk here about recall just as an example of other things that also remain same? (cause recall is only one aspect of the study)} 
{For example, in terms of factual recall, low diversity causes a failure, moderate diversity is best for intermediate training length, and high diversity achieves optimal performance only after long training.}
% Low diversity causes a failure in factual recall, moderate diversity peaks recall mid‑training, and high diversity achieves optimal performance only after long training.
% 
% 
The primary impact of depth is on the convergence speed: at any fixed iteration count, the 10‑layer model achieves higher accuracy than the 4‑layer, which in turn outperforms the 1‑layer.  
Notably, larger models help slowly improve \emph{position accuracy} even under low diversity (seen as brighter colors in the upper-right portions of panel b). However, the factual recall failure at extreme low diversity persists across depths, highlighting that model capacity alone cannot substitute for contextual variety in achieving robust OOD generalization. %\ct{very nice! I would add this sentence in the figure caption as well}
% Crucially, however, extreme low‐diversity failure in factual recall is not rescued by additional layers, underscoring that breadth of contextual exposure, rather than model capacity alone, is essential for out‐of‐distribution generalization.

\subsection{\mixexp{10}: \emph{Structural}-OOD data}\label{app:out_struct}

\begin{figure}[t]
  \centering
\begin{tabular}{cc}   % two equal-width columns
%──────────────────────────────── LEFT COLUMN ──────────────────────────────%
\begin{minipage}{0.46\textwidth}
  \centering
  % (a) ────────────────────────────────────────────────────────────────
  \begin{subfigure}{\textwidth}
    \centering
    \begin{tikzpicture}
      \node (img) {\includegraphics[width=\linewidth]{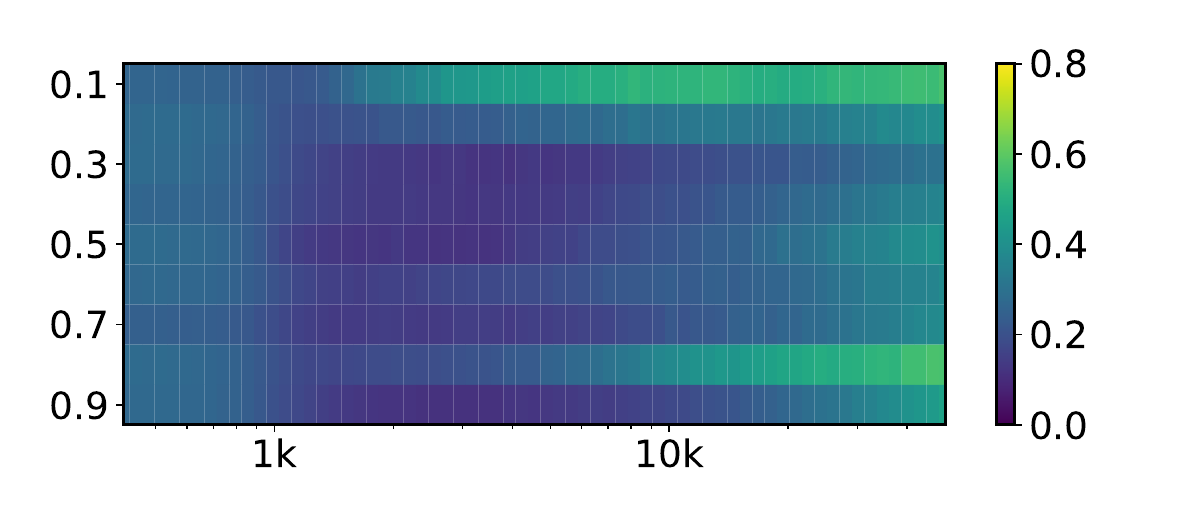}};
      % \node[anchor=north west,font=\bfseries\small,xshift=45pt,yshift=0pt]
      %   at (img.north west){(a) Statistical loss};
             \node[anchor=north west,font=\bfseries\small,xshift=-10pt,yshift=-60pt,rotate=90]
        at (img.north west){(a) $\KL$};
    \end{tikzpicture}
  \end{subfigure}\\%[0.1em]
    \vspace{-10.5mm}
  % (b) ────────────────────────────────────────────────────────────────
  \begin{subfigure}{\textwidth}
    \centering
    \begin{tikzpicture}
      \node (img) {\includegraphics[width=\linewidth]{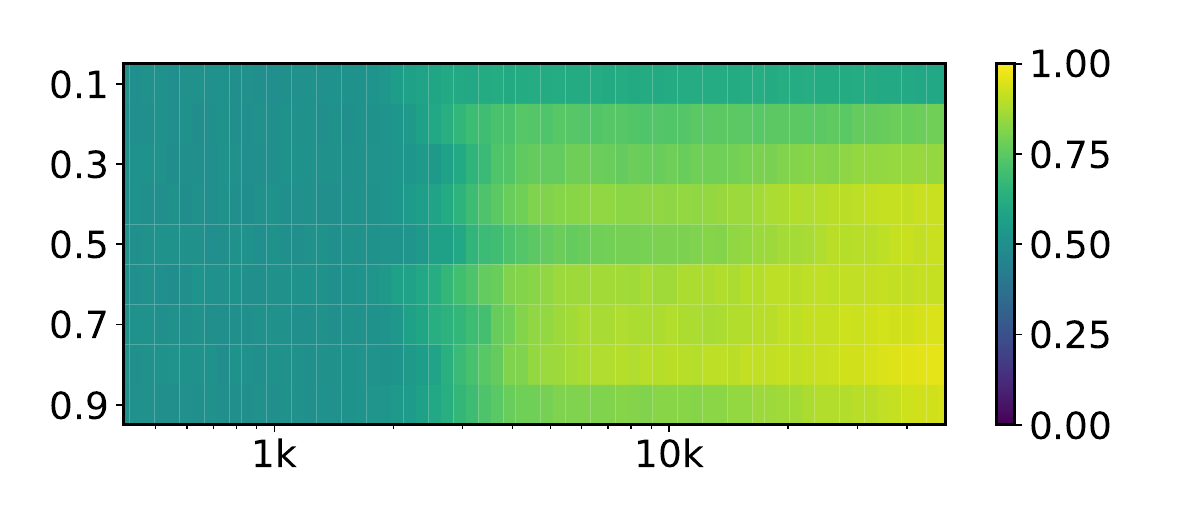}};
      % \node[anchor=north west,font=\bfseries\small,xshift=42pt,yshift=0pt]
        % at (img.north west){(b) Position accuracy};
        
      \node[anchor=north west,font=\bfseries\small,xshift=-10pt,yshift=-60pt,rotate=90]
        at (img.north west){(b) $\posacc$};
    \end{tikzpicture}
  \end{subfigure}\\%[0.4em]
\vspace{-10.5mm}
  % (c) ────────────────────────────────────────────────────────────────
  \begin{subfigure}{\textwidth}
    \centering
    \begin{tikzpicture}
      \node (img) {\includegraphics[width=\linewidth]{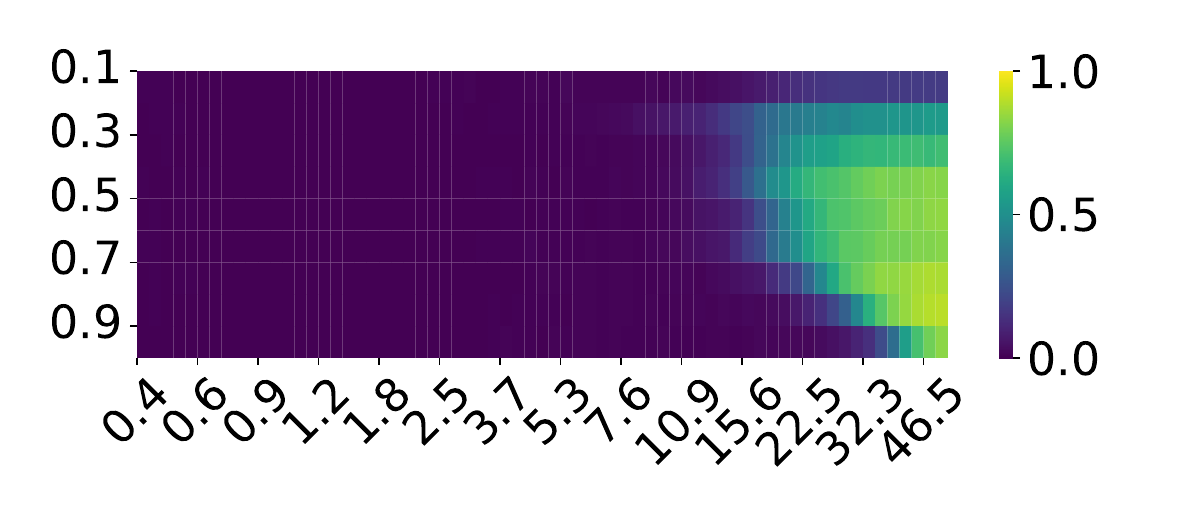}};
      \node (img) {\includegraphics[width=\linewidth]{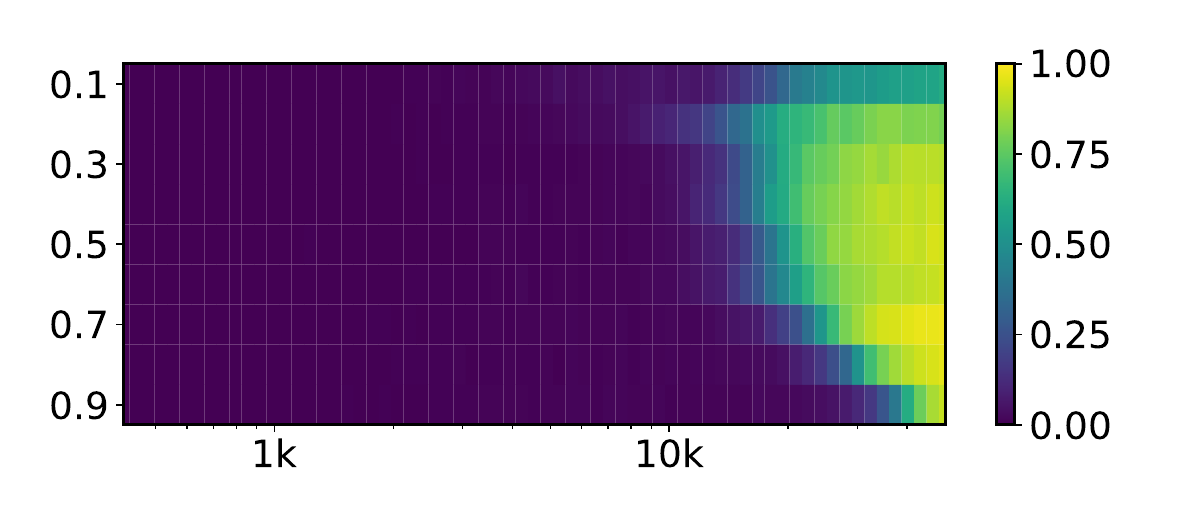}};
      % \node[anchor=north west,font=\bfseries\small,xshift=42pt,yshift=0pt]
      %   at (img.north west){(c) Factual accuracy};
           \node[anchor=north west,font=\bfseries\small,xshift=-10pt,yshift=-60pt,rotate=90]
        at (img.north west){(c) $\factacc$};
    \end{tikzpicture}
  \end{subfigure}
\end{minipage}
&
%──────────────────────────────── RIGHT COLUMN ─────────────────────────────%
\begin{minipage}{0.46\textwidth}
  \centering
  \begin{subfigure}{\textwidth}
    \centering
    \begin{tikzpicture}
      % Set the height to equal *roughly* three heat-map heights.
      % \node (img) {\includegraphics[scale=0.32]{figures/out_struct_at_pos_is_bi_rate_MC10Pos10_order1_L10H4d32_T50_monotonic.pdf}};
      \node (img) {\includegraphics[scale=0.32]{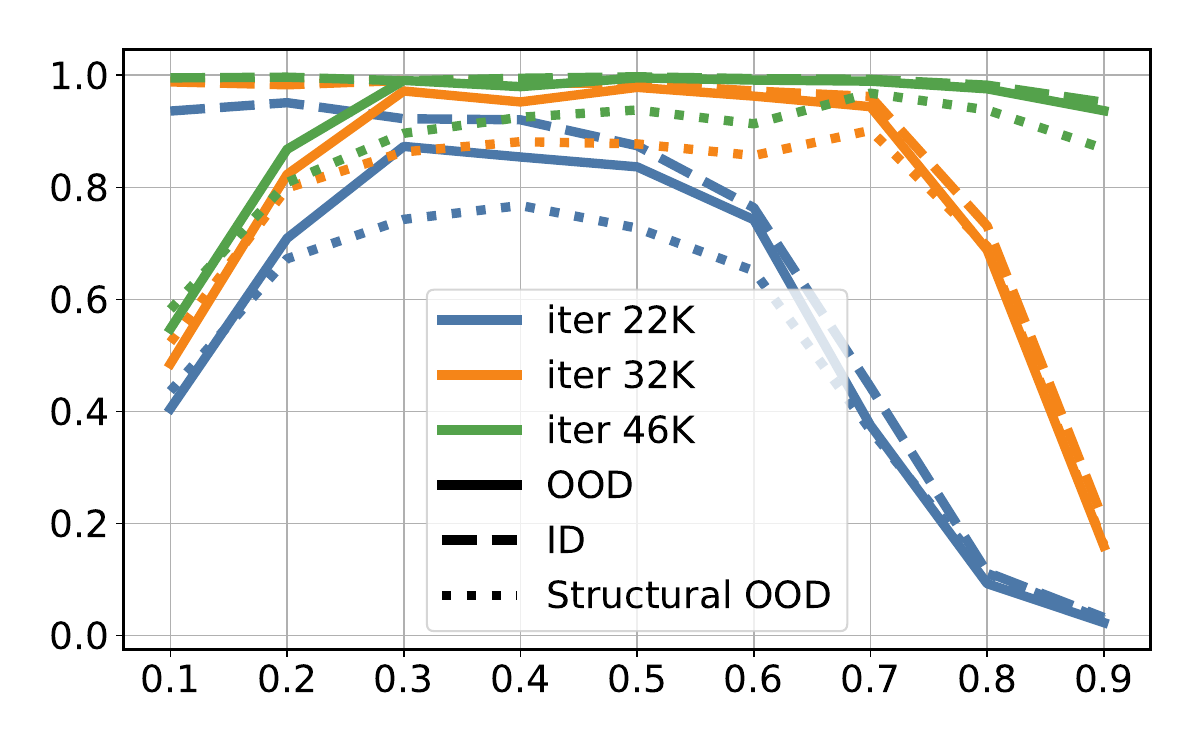}};
      \node[anchor=north west,font=\bfseries\small,xshift=74pt,yshift=6pt]
        at (img.north west){(d) $\factacc$};
      \node[anchor=north west,font=\bfseries\small,xshift=84pt,yshift=-110pt]
        at (img.north west){$\dvr$};
        
    \end{tikzpicture}
  \end{subfigure}
\end{minipage}
\end{tabular}
  \vspace{-0.1in}
  \caption{\textbf{Structural‑OOD performance in the \mixexp{10} setup.} \textbf{(a)} $\KL$, \textbf{(b)} $\posacc$ and \textbf{(c)} $\factacc$ for the experimental setup of Fig.~\ref{fig:heatmaps_main} on sequences drawn from structural-OOD templates, defined in Sec.~\ref{app:out_struct}. Panel \textbf{(d)} replots $\factacc$ versus $\dvr$, at three training checkpoints as in  Fig.~\ref{fig:div_tradeoff}.
    }
    % \vspace{-0.2in}
  \label{fig:out_struct}
\end{figure}

% In case of \mixexp{10}, we can go beyond the usual ID and OOD splits by introducing \emph{structural}‑OOD templates. Recall that in the \mixexp{10} setup, each of the $\tmplnum=10$ templates is specified by a distinct transition matrix and position pair: $(\transitionmat_n, \pospair_n)$, $n\in[\tmplnum]$.

% Similar to the other two contextual structure settings, we have defined ID and OOD templates for a given fact $(\source,\target)$ as those templates among the $\tmplnum$ templates depending on whether they were seen or not in the training set with $(\source,\target)$. But unlike the other two settings, our \mixexp{10} setupt further allow us  to test pure compositional generalization since here we can form new templates by pairing each $\transitionmat_n$ with a position pair $\pospair_{n'}$, for $n\neq n'$. Each mixed template combines a Markov chain and position both familiar in isolation but never jointly encountered and the model must combine familiar subcomponents to be able to generalize to the structural-OOD sequences.

The \mixexp{10} setup enables us to evaluate a particularly challenging form of 
generalization beyond standard ID/OOD splits: \emph{structural}-OOD templates that 
test pure compositional reasoning.

Recall that in \mixexp{10}, each of the 
$\tmplnum=10$ templates is uniquely specified by a transition matrix and position 
pair: $(\transitionmat_n, \pospair_n)$ for $n \in [\tmplnum]$. As in our other 
contextual structure settings, we define ID and OOD templates for a given fact 
$(\source,\target)$ based on whether they appeared with this fact pair during 
training.

The distinctive feature of \mixexp{10} is that it allows us to test pure 
compositional generalization by forming new templates through pairing each 
$\transitionmat_n$ with a position pair $\pospair_{n'}$, for $n\neq n'$. Specifically, in this case, each 
mixed template combines a Markov chain and position both familiar in isolation 
but never jointly encountered, and the model must combine these familiar 
subcomponents to generalize to the structural-OOD sequences. We call sequences generated from such templates \emph{structural}-OOD sequences.

Fig.~\ref{fig:out_struct} plots $\KL$, $\posacc$ and $\factacc$ on  \emph{structural}-OOD sequences in the 4-layer experimental setup of Fig.~\ref{fig:model_size}. 
The qualitative trends mirror our standard OOD findings: {Under low diversity levels, factual recall} fails catastrophically; {mid-training, optimal performance is achieved with intermediate diversity levels}; {and high diversity helps recovering the performance in long training regimes. However, the absolute value of the metrics remains lower than on the original OOD split, hinting the model needs even longer training to master completely novel template combinations. {Position accuracy shows a similar pattern: Sharp failure at low diversity, but little sensitivity to diversity elsewhere. Interestingly, $\KL$ is hit hardest by these novel template compositions at test time. We leave a deeper investigation for future work,} In panel (d), we further visualize, as in Fig.~\ref{fig:div_tradeoff}, that extended training uncovers the long‑term benefits of high diversity, even though it may slow progress during the intermediate phase.

% \subsection{Different order MCs}
% \tina{if time: add higher and lower order}

\subsection{Experiment details of Sec.~\ref{sec:bottleneck}}\label{app:exp_intervention}

\new{
In the controlled experiments of Section~\ref{sec:intervention}, we use the same data and model initialization seeds for each experimental run as those used to train the corresponding well-trained model $\highmodel$. This ensures that the underlying fact and template definitions are identical and that training begins from the exact same initialization point. After initializing the model, we transfer the desired parameter subset from $\highmodel$, freeze it, and then retrain the model on a new dataset that differs only in its diversity level. \looseness=-1
}

\new{While the main text (Sec.~\ref{sec:intervention} and Fig.~\ref{fig:intervention}) focused on the impact of interventions at the lowest diversity level ($\dvr=0.1$), Fig.~\ref{fig:app_interventions}  provides the complete results across all diversity levels and training iterations in a format similar to Fig.~\ref{fig:heatmaps_main}. Here, we only focus on the OOD performance.
For completeness, the figure also includes results for interventions on other modules that had a limited impact and were therefore not discussed in the main text. Table~\ref{fig:intervention_10k} reports metrics at iteration 10k (mid-training), analogous to the final results shown in Fig.~\ref{fig:intervention}-(a). This mid-training snapshot more clearly demonstrates that intervening on $\mathbf{U}$ accelerates learning, even in low-diversity settings. We replicate the trends in Sec.~\ref{sec:bottleneck}, for a one-layer, one-head Transformer, as detailed in Figs.~\ref{fig:app_interventions_L1}–\ref{fig:high_div_init}. }

\begin{figure}[p]
\vspace{-50pt}
\centering
% 4 columns: three plots + right-hand note lane
\begin{tabular}{@{\hspace{-5pt}}c@{\hspace{40pt}}c@{\hspace{35pt}}c@{\hspace{40pt}}R{0.22\textwidth}@{}}

% ───────────────────────── Row 1: Standard ───────────────────────────
\begin{subfigure}[t]{0.2\textwidth}\centering
  \rowstrut
  \begin{tikzpicture}[remember picture]
    \node at (0,0) {\includegraphics[scale=0.2]{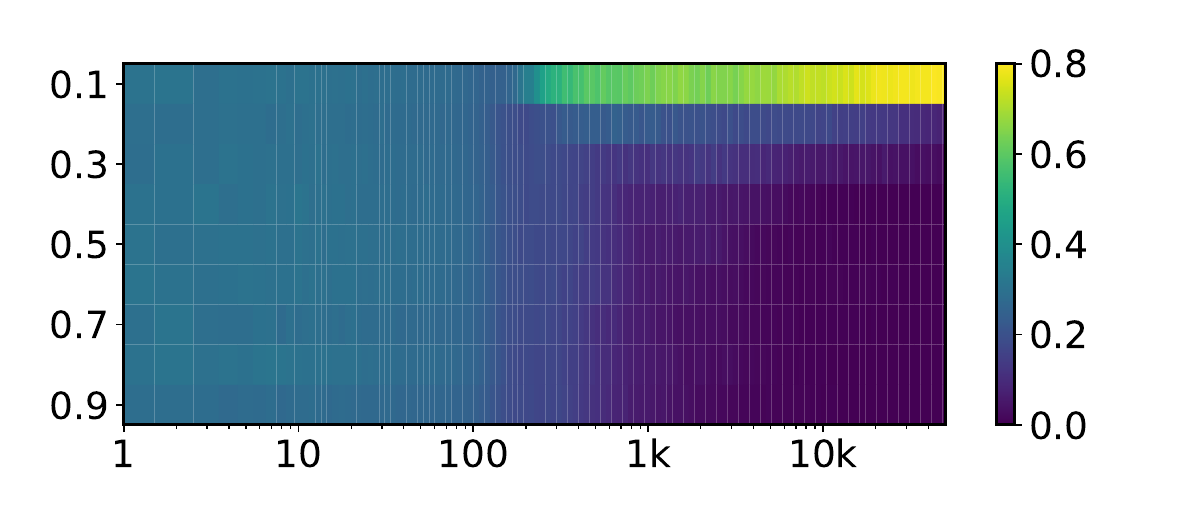}};
    \node[scale=0.75,rotate=90] at (-2.2,0.1) {\textbf{Standard}};
    % \node[scale=0.8] at (0.0,-1.2) {iteration ($\times 10^3$)};
    \node[scale=0.8] at (0.0, 1.0) {\textbf{$\KL$}};
  \end{tikzpicture}
\end{subfigure} &
{
\begin{subfigure}[t]{0.2\textwidth}\centering
  \begin{tikzpicture}[remember picture]
    \node at (0,0) {\includegraphics[scale=0.2]{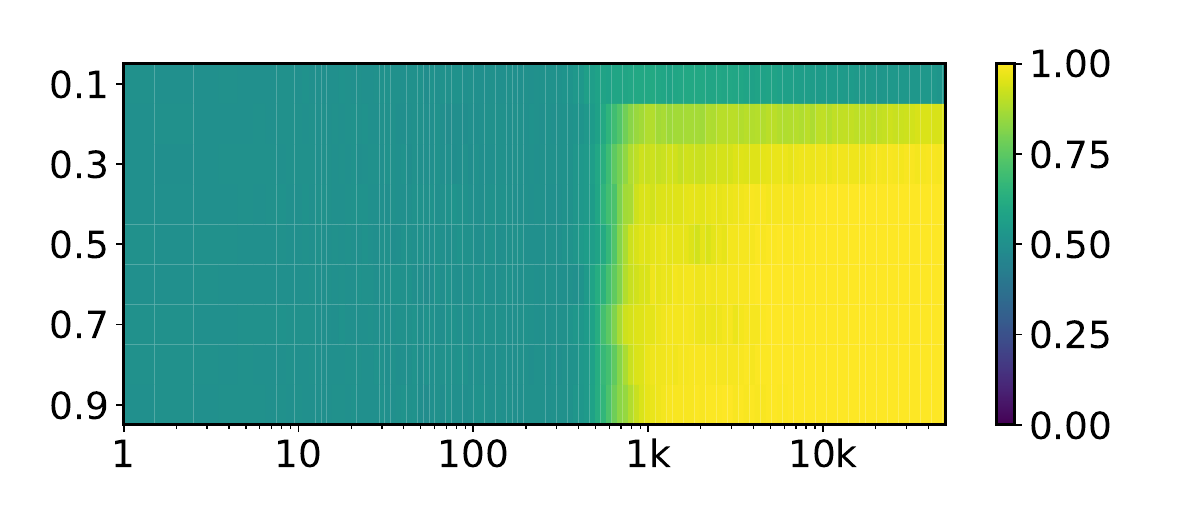}};
    \node[scale=0.8] at (0.0,1.0) {\textbf{$\posacc$}};
  \end{tikzpicture}
\end{subfigure}} &
{
\begin{subfigure}[t]{0.2\textwidth}\centering
  \begin{tikzpicture}[remember picture]
    \node at (0,0) {\includegraphics[scale=0.2]{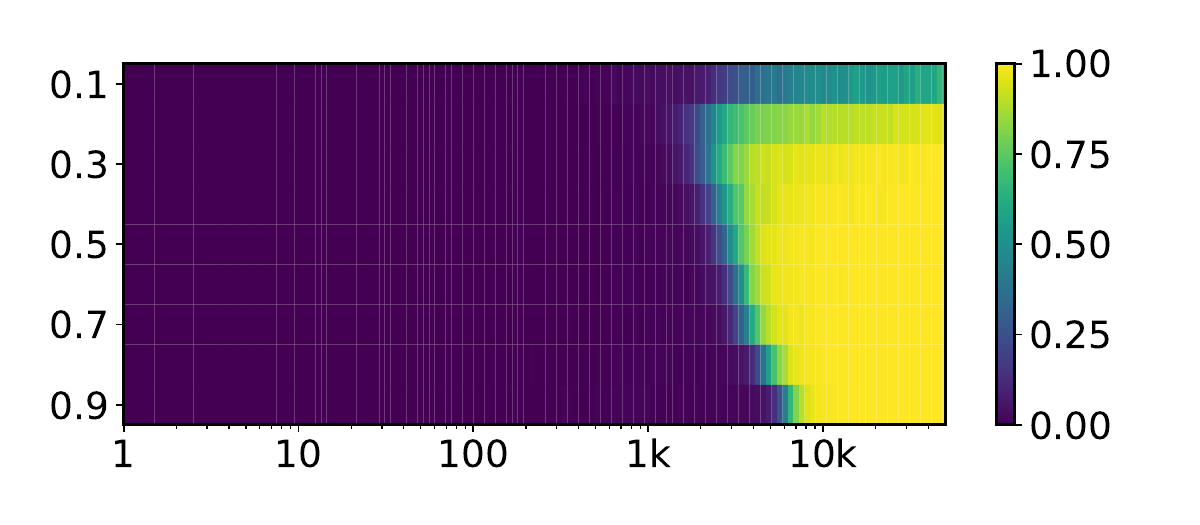}};
    \node[scale=0.8] at (0.0,1.0) {\textbf{$\factacc$}};
  \end{tikzpicture}
\end{subfigure}} &
\rownote{Failure at low diversity. Slowdown with high diversity}
\\[-20pt]
% ───────────────────────── Row 4: Attn (patched) ─────────────────────
\begin{subfigure}[t]{0.2\textwidth}\centering
  \rowstrut
  \begin{tikzpicture}[remember picture]
    \node at (0,0) {\includegraphics[scale=0.2]{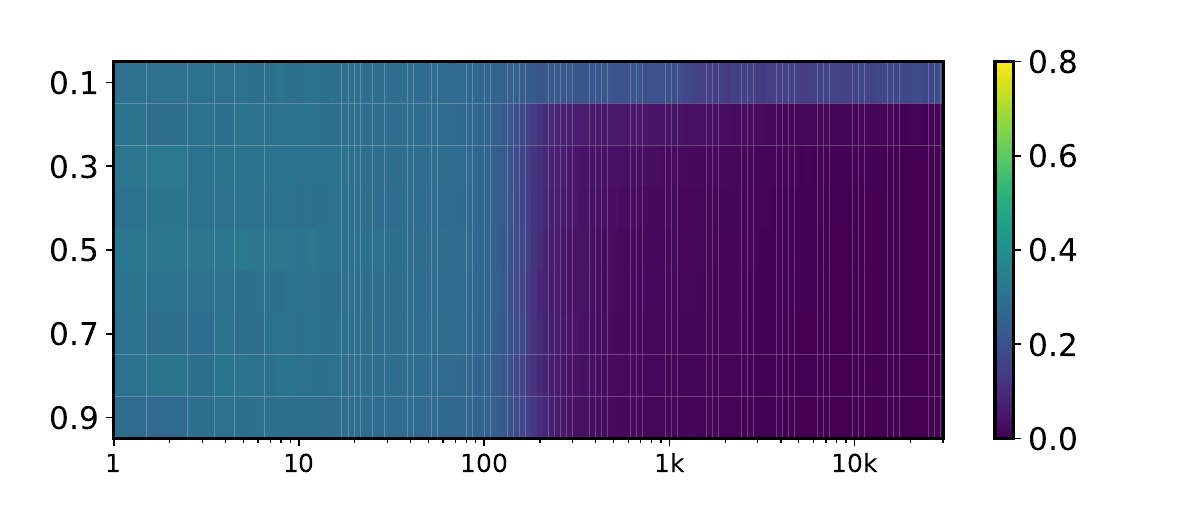}};
    \node[scale=0.75,rotate=90] at (-2.2,0.1) {\textbf{Attn}};
  \end{tikzpicture}
\end{subfigure} &
\begin{subfigure}[t]{0.2\textwidth}\centering
  \begin{tikzpicture}[remember picture]
    \node at (0,0) {\includegraphics[scale=0.2]{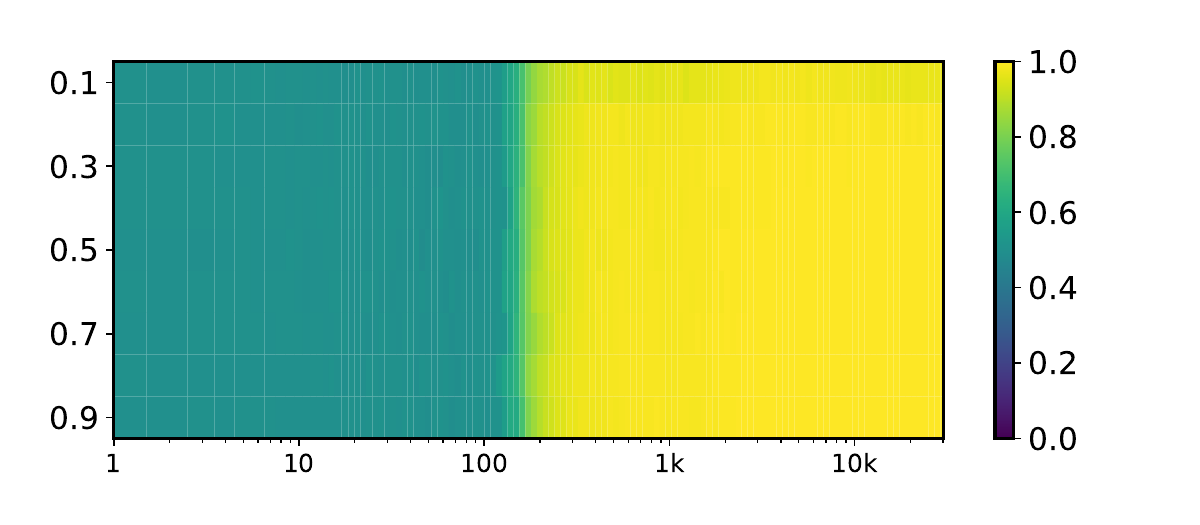}};
  \end{tikzpicture}
\end{subfigure} &
\begin{subfigure}[t]{0.2\textwidth}\centering
  \begin{tikzpicture}[remember picture]
    \node at (0,0) {\includegraphics[scale=0.2]{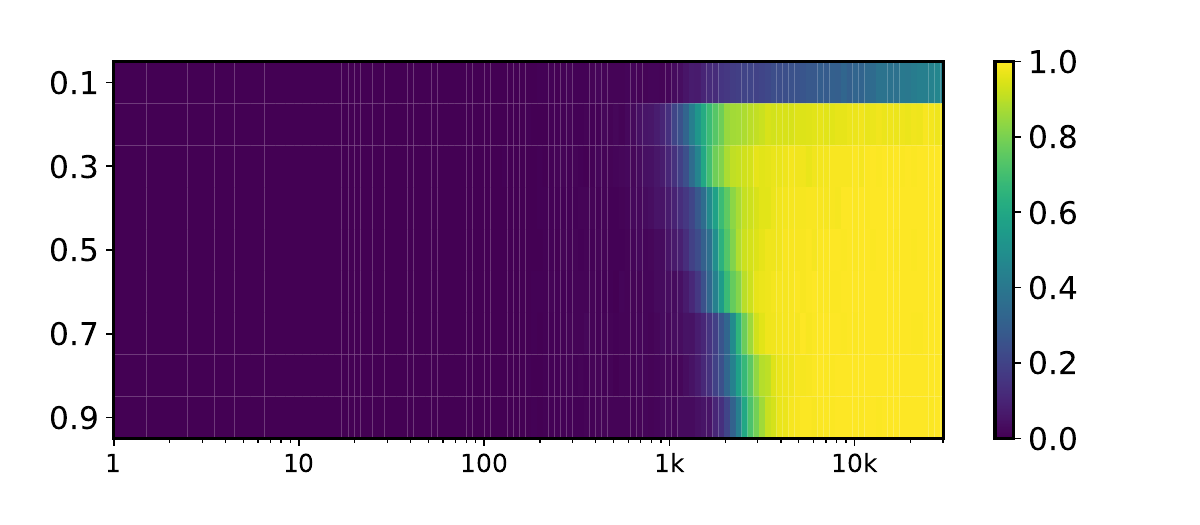}};
  \end{tikzpicture}
\end{subfigure} &
\rownote{\emph{Stat} \& \emph{pos} recover. No impact on \emph{fact}.}
\\[-20pt]

% ───────────────────────── Row 5: E (embeddings) ─────────────────────
\begin{subfigure}[t]{0.2\textwidth}\centering
  \rowstrut
  \begin{tikzpicture}[remember picture]
    \node at (0,0) {\includegraphics[scale=0.2]{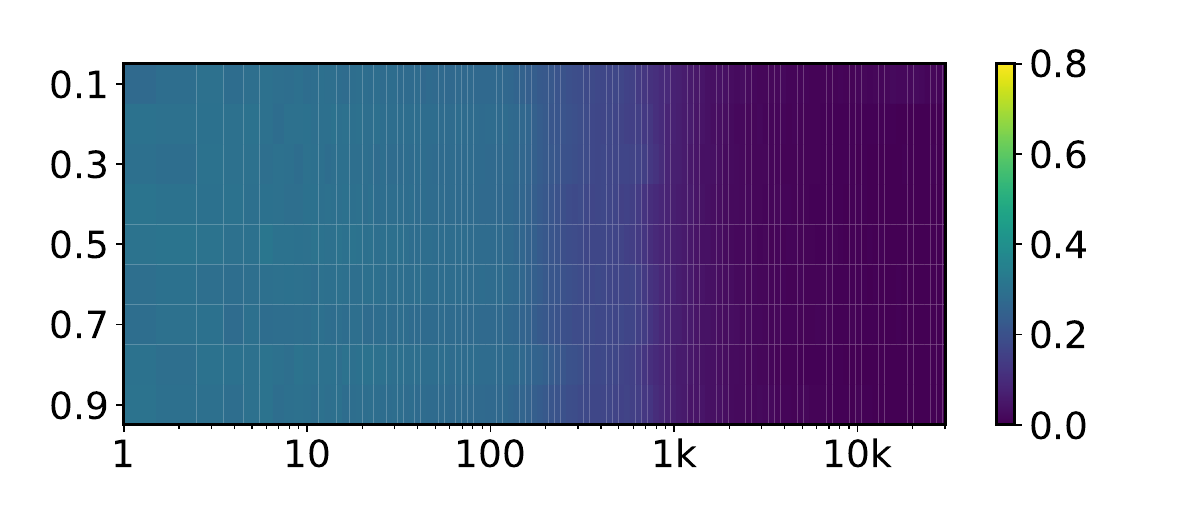}};
    \node[scale=0.75,rotate=90] at (-2.2,0.1) {\textbf{$\mathbf{E}$}};
  \end{tikzpicture}
\end{subfigure} &
\begin{subfigure}[t]{0.2\textwidth}\centering
  \begin{tikzpicture}[remember picture]
    \node at (0,0) {\includegraphics[scale=0.2]{figures/heatmaps/H4L4AttnPatch/combined_non_kb_and_kb_at_pos_out_dist_override_MC10Pos10_order1_L4H4d32_T50_heatmap_avg.pdf}};
  \end{tikzpicture}
\end{subfigure} &
\begin{subfigure}[t]{0.2\textwidth}\centering
  \begin{tikzpicture}[remember picture]
    \node at (0,0) {\includegraphics[scale=0.2]{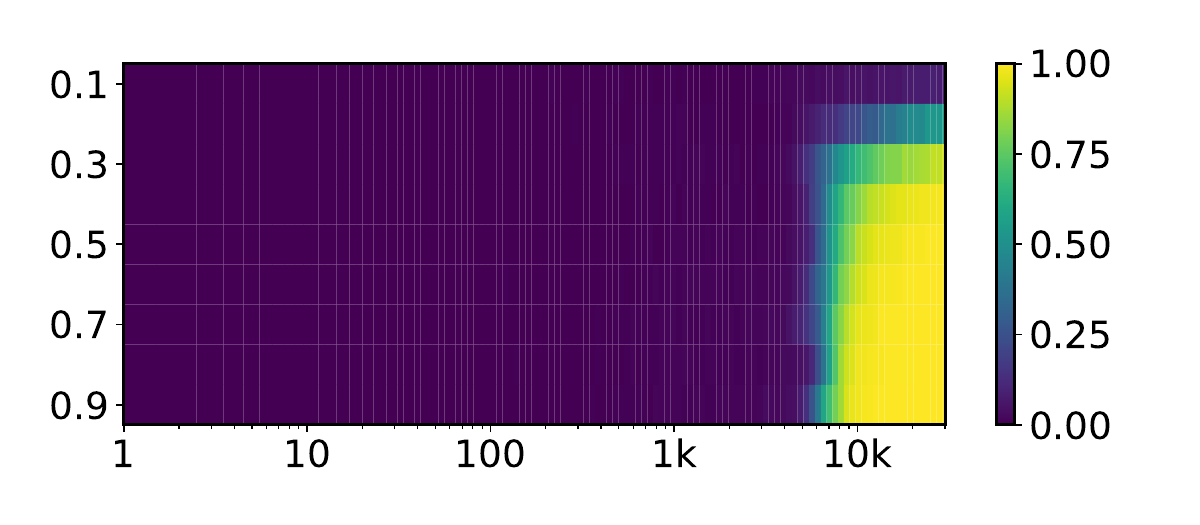}};
  \end{tikzpicture}
\end{subfigure} &
\rownote{}
\\[-20pt]

% ───────────────────────── Row 3: U (head) ───────────────────────────
\begin{subfigure}[t]{0.2\textwidth}\centering
  \rowstrut
  \begin{tikzpicture}[remember picture]
    \node at (0,0) {\includegraphics[scale=0.2]{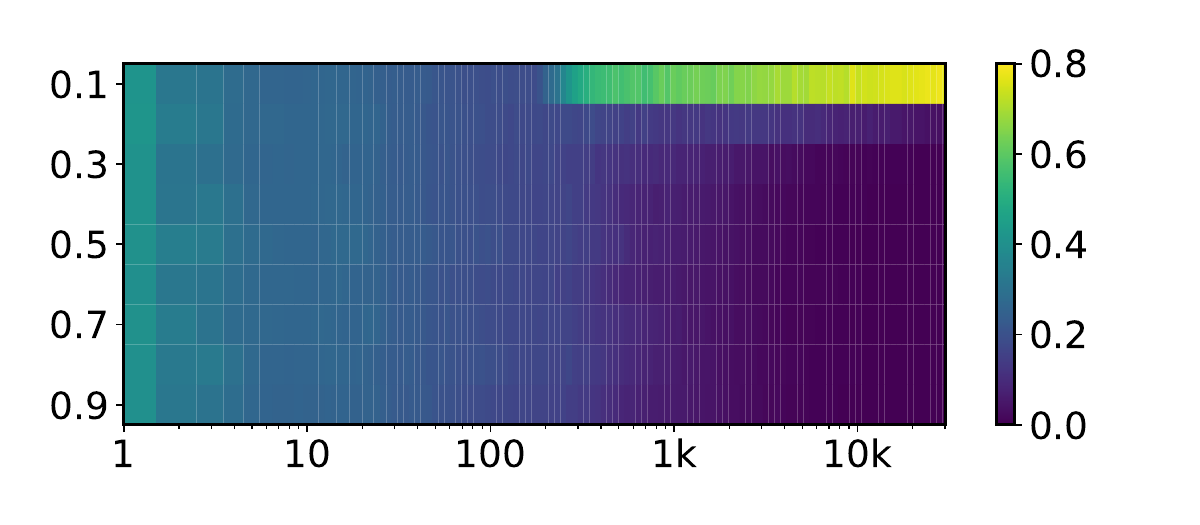}};
    \node[scale=0.75,rotate=90] at (-2.2,0.1) {\textbf{$\mathbf{U}$}};
  \end{tikzpicture}
\end{subfigure} &
\begin{subfigure}[t]{0.2\textwidth}\centering
  \begin{tikzpicture}[remember picture]
    \node at (0,0) {\includegraphics[scale=0.2]{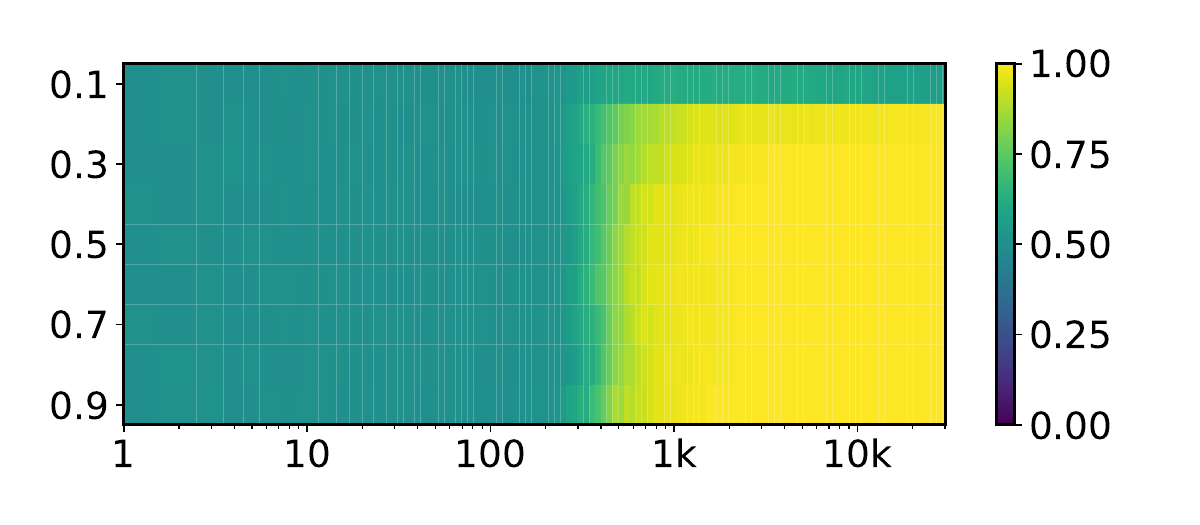}};
  \end{tikzpicture}
\end{subfigure} &
\begin{subfigure}[t]{0.2\textwidth}\centering
  \begin{tikzpicture}[remember picture]
    \node at (0,0) {\includegraphics[scale=0.2]{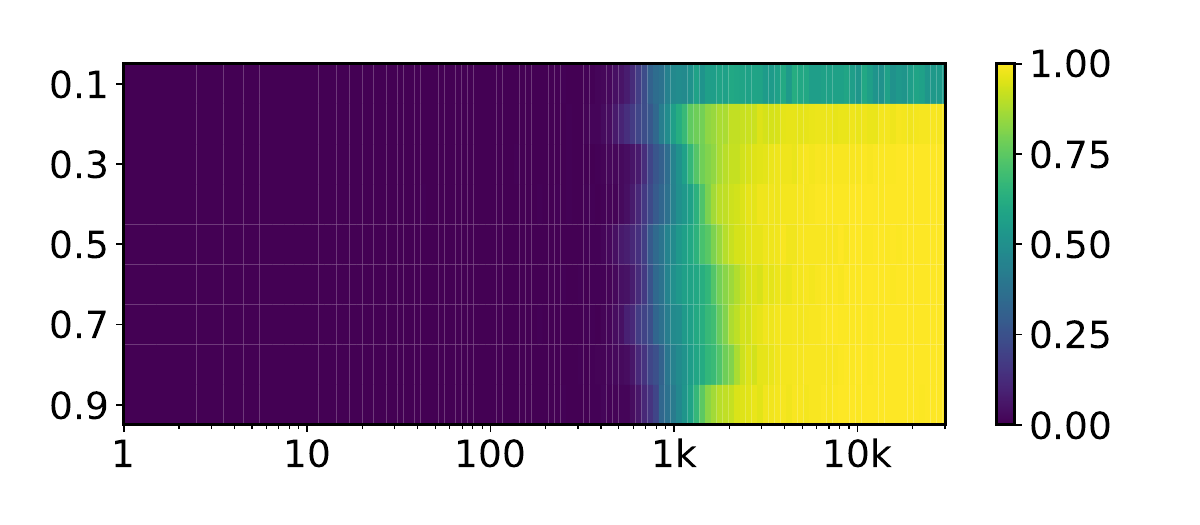}};
  \end{tikzpicture}
\end{subfigure} &
\rownote{No impact on \emph{stat} and \emph{pos}. Remedies slowdown of \emph{fact}.}
\\[-20pt]

% ───────────────────────── Row 6: U + Attn ───────────────────────────
\begin{subfigure}[t]{0.2\textwidth}\centering
  \rowstrut
  \begin{tikzpicture}[remember picture]
    \node at (0,0) {\includegraphics[scale=0.2]{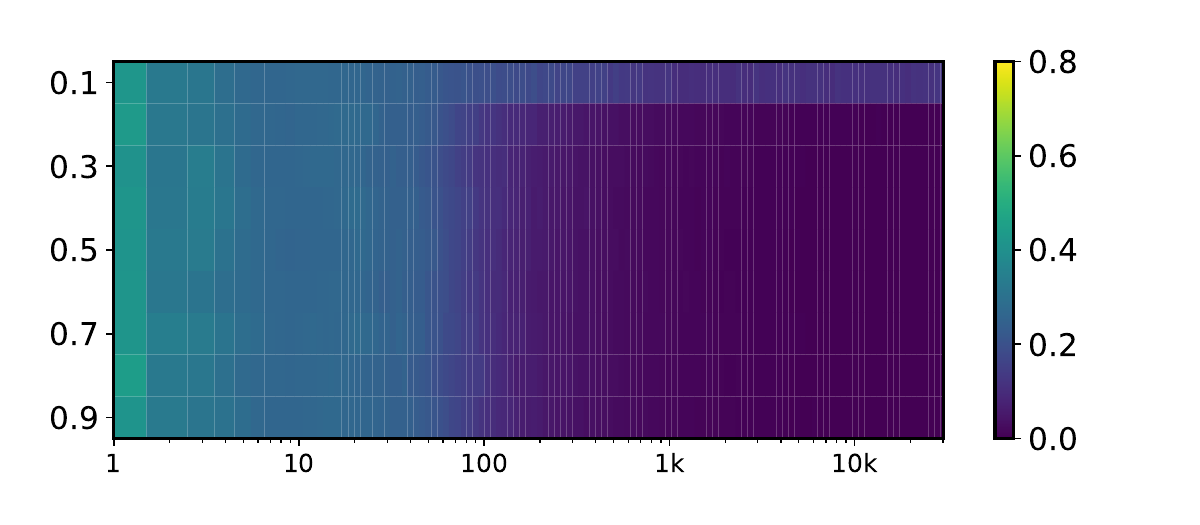}};
    \node[scale=0.75,rotate=90] at (-2.2,0.1) {$\mathbf{U},$\textbf{Attn}};
  \end{tikzpicture}
\end{subfigure} &
\begin{subfigure}[t]{0.2\textwidth}\centering
  \begin{tikzpicture}[remember picture]
    \node at (0,0) {\includegraphics[scale=0.2]{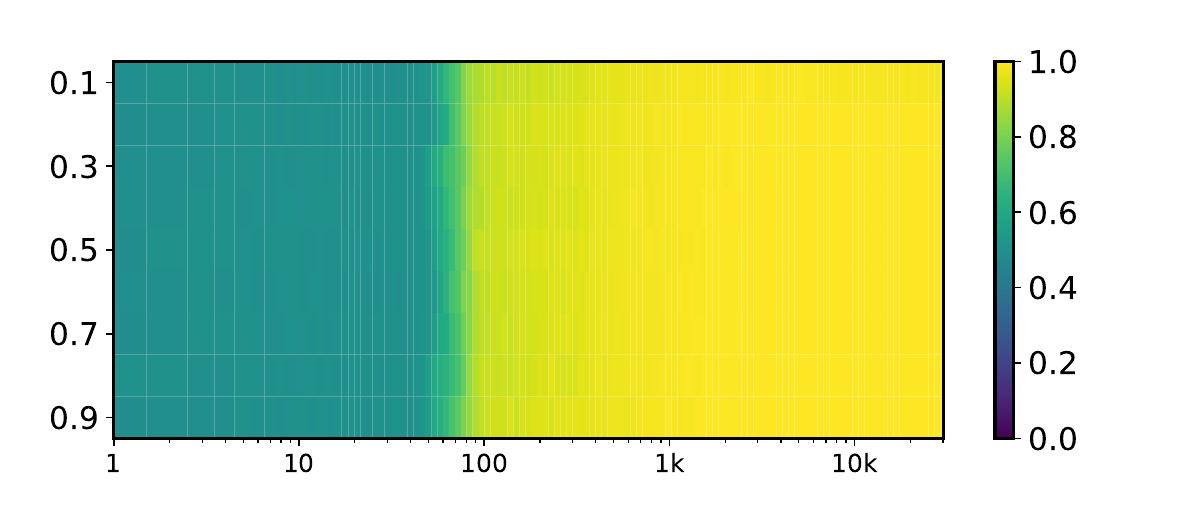}};
  \end{tikzpicture}
\end{subfigure} &
\begin{subfigure}[t]{0.2\textwidth}\centering
  \begin{tikzpicture}[remember picture]
    \node at (0,0) {\includegraphics[scale=0.2]{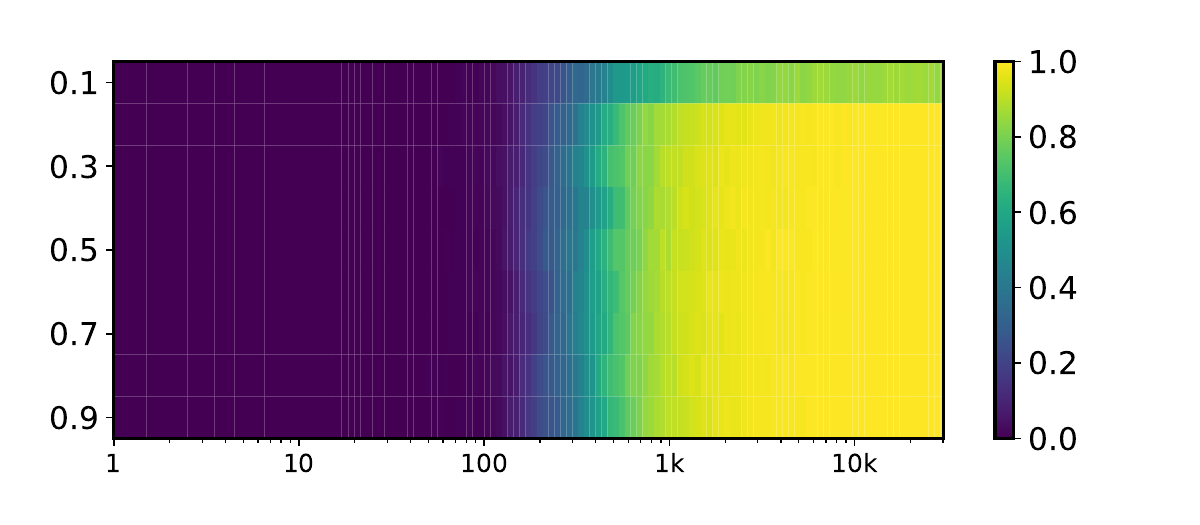}};
  \end{tikzpicture}
\end{subfigure} &
\rownote{Improves all metrics, on both ends of diversity.}
\\[-20pt]

% ───────────────────────── Row 7: U + E ──────────────────────────────
\begin{subfigure}[t]{0.2\textwidth}\centering
  \rowstrut
  \begin{tikzpicture}[remember picture]
    \node at (0,0) {\includegraphics[scale=0.2]{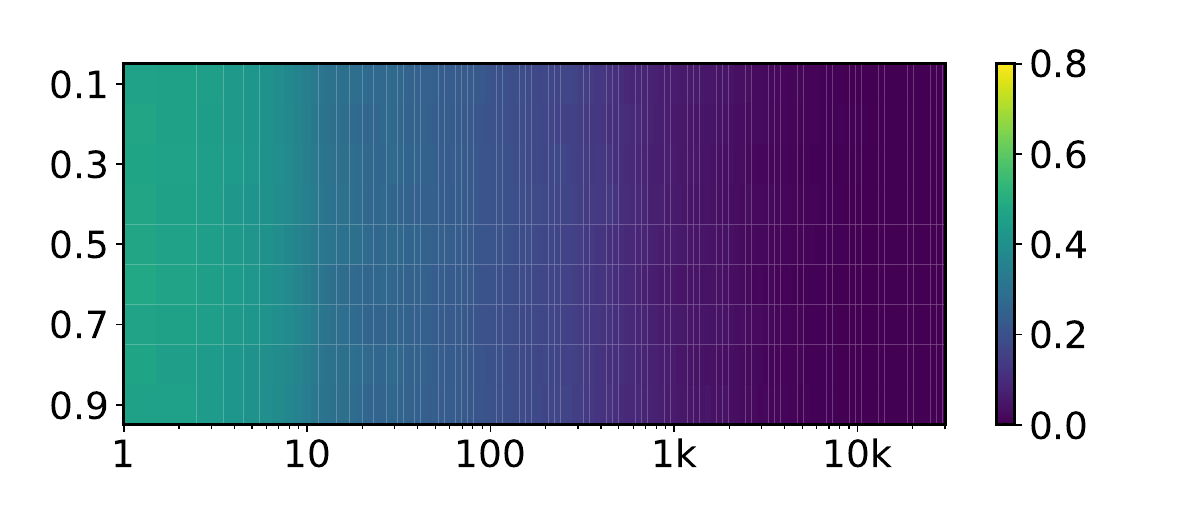}};
    \node[scale=0.75,rotate=90] at (-2.2,0.1) {$\mathbf{U},\mathbf{E}$};
  \end{tikzpicture}
\end{subfigure} &
\begin{subfigure}[t]{0.2\textwidth}\centering
  \begin{tikzpicture}[remember picture]
    \node at (0,0) {\includegraphics[scale=0.2]{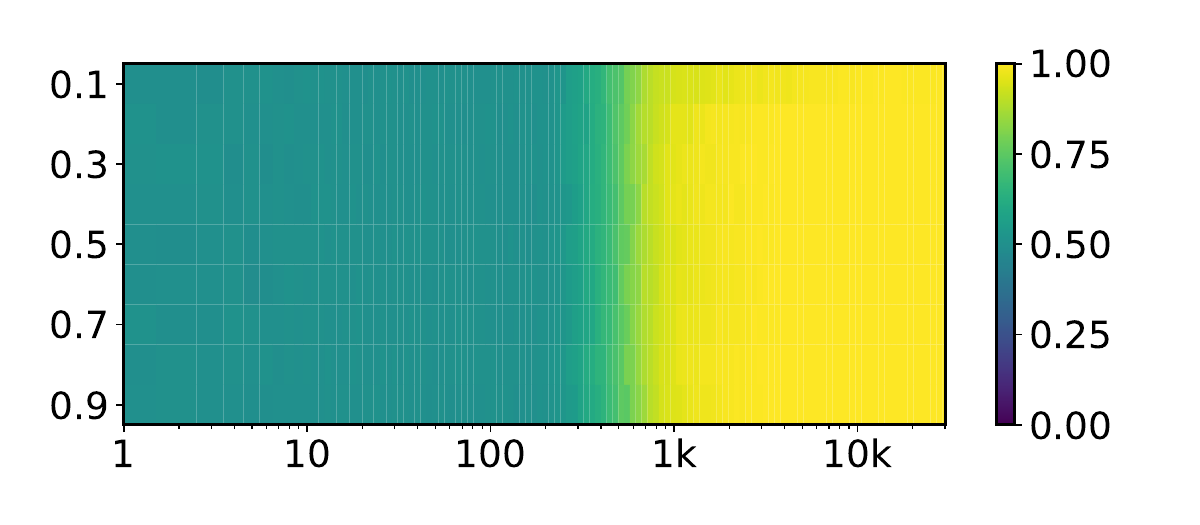}};
  \end{tikzpicture}
\end{subfigure} &
\begin{subfigure}[t]{0.2\textwidth}\centering
  \begin{tikzpicture}[remember picture]
    \node at (0,0) {\includegraphics[scale=0.2]{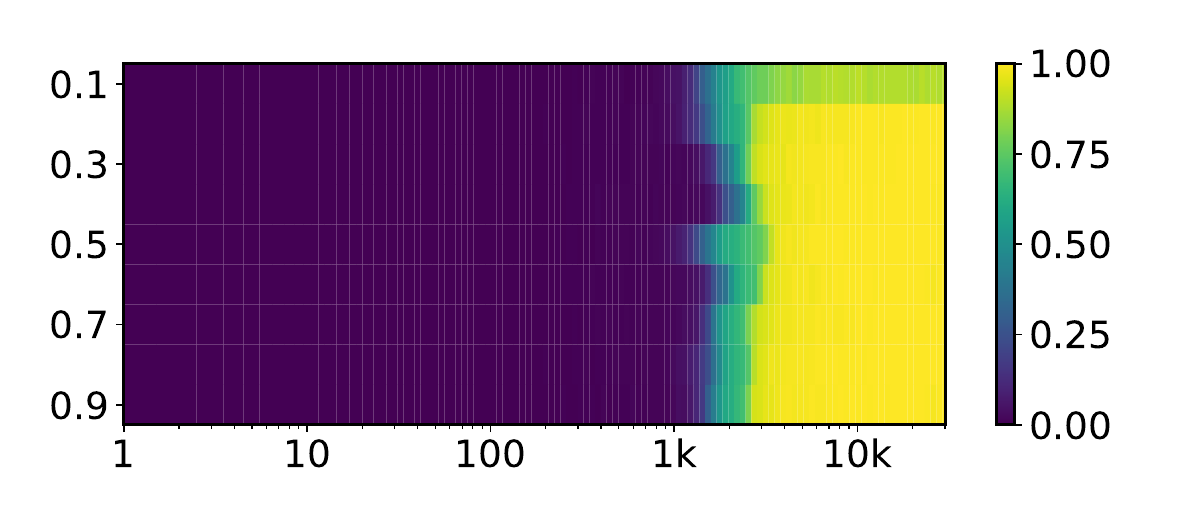}};
  \end{tikzpicture}
\end{subfigure} &
\rownote{}
\\[-20pt]

% ───────────────────────── Row 8: U + E + Attn ───────────────────────
\begin{subfigure}[t]{0.2\textwidth}\centering
  \rowstrut
  \begin{tikzpicture}[remember picture]
    \node at (0,0) {\includegraphics[scale=0.2]{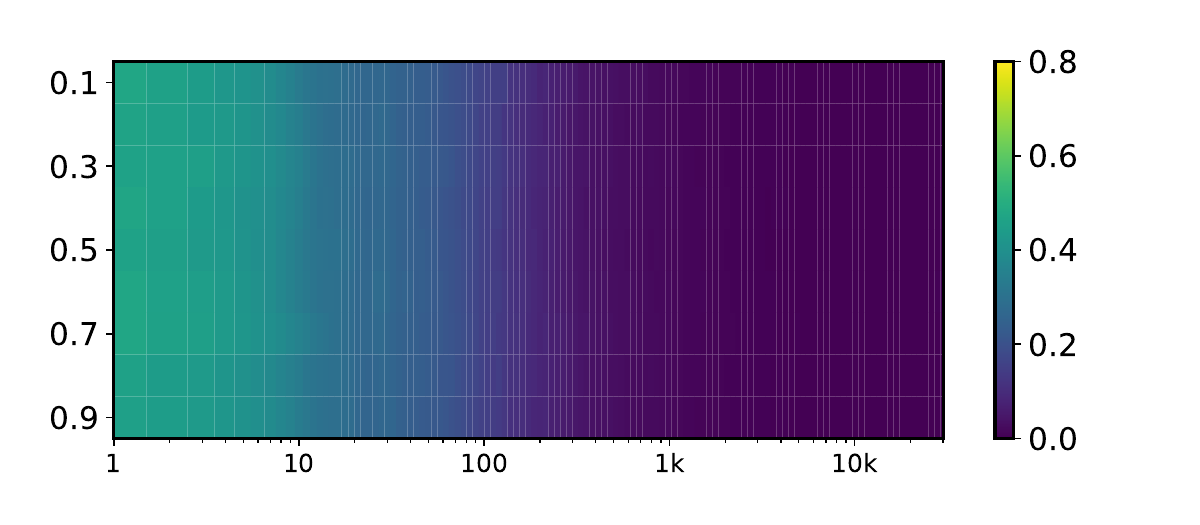}};
    \node[scale=0.75,rotate=90] at (-2.2,0.1) {$\mathbf{U},\mathbf{E},$\textbf{Attn}};
  \end{tikzpicture}
\end{subfigure} &
\begin{subfigure}[t]{0.2\textwidth}\centering
  \begin{tikzpicture}[remember picture]
    \node at (0,0) {\includegraphics[scale=0.2]{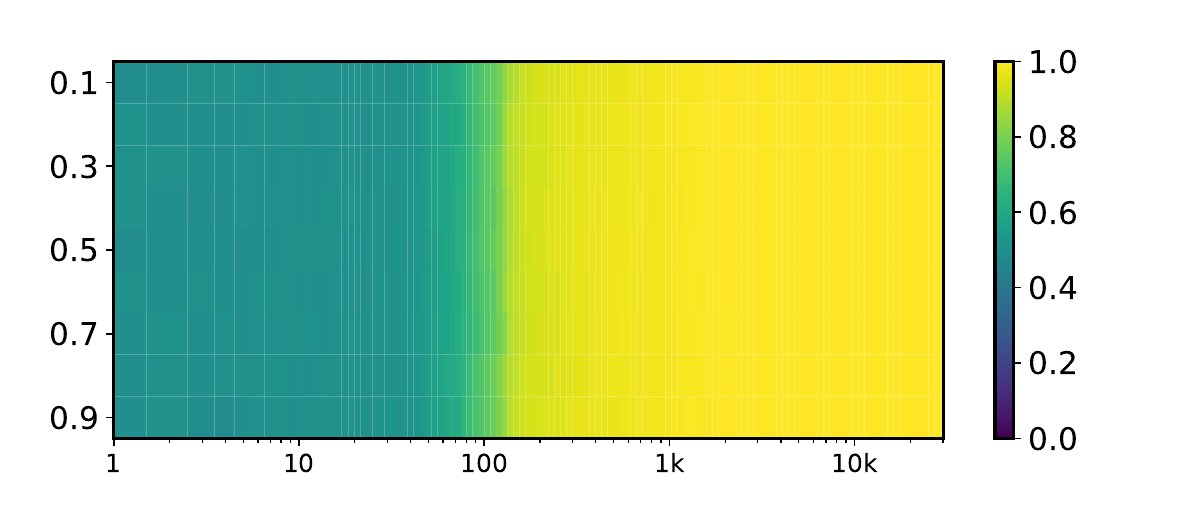}};
  \end{tikzpicture}
\end{subfigure} &
\begin{subfigure}[t]{0.2\textwidth}\centering
  \begin{tikzpicture}[remember picture]
    \node at (0,0) {\includegraphics[scale=0.2]{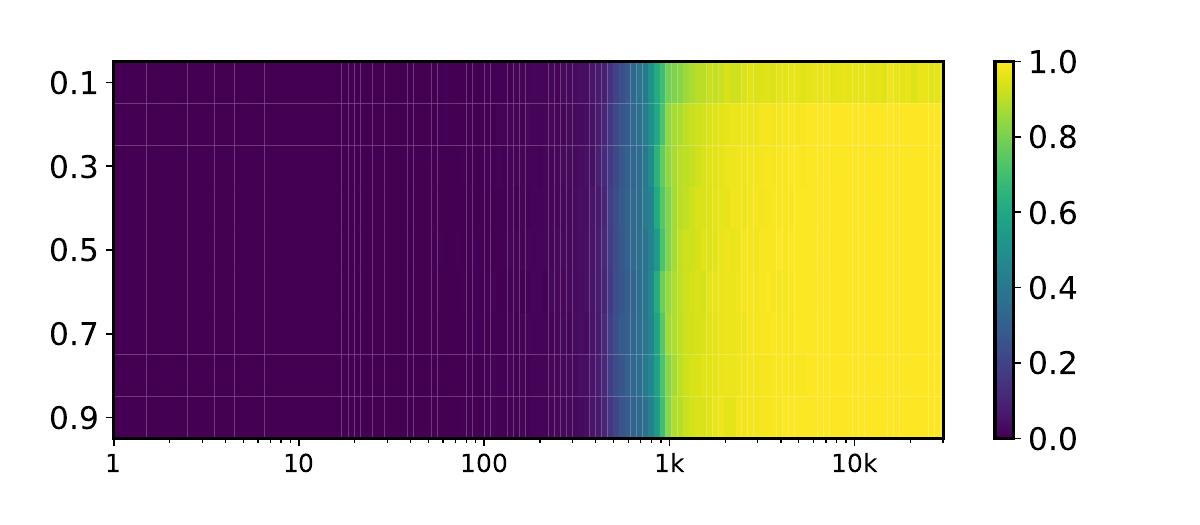}};
  \end{tikzpicture}
\end{subfigure} &
\rownote{}
\\[-20pt]

% 

% ───────────────────────── Row 2: MLP +U────────────────────────────────
% \begin{subfigure}[t]{0.2\textwidth}\centering
%   \rowstrut
%   \begin{tikzpicture}[remember picture]
%     \node at (0,0) {\includegraphics[scale=0.2]{figures/heatmaps/H4L4MLPHead/kl_masked_completion_GT_out_dist_MC10Pos10_order1_L4H4d32_T50_heatmap_avg.pdf}};
%     \node[scale=0.75,rotate=90] at (-2.2,0.1) {\textbf{MLP},\textbf{U} \tina{to do}};
%   \end{tikzpicture}
% \end{subfigure} &
% \begin{subfigure}[t]{0.2\textwidth}\centering
%   \begin{tikzpicture}[remember picture]
%     \node at (0,0) {\includegraphics[scale=0.2]{figures/heatmaps/H4L4MLPHead/combined_non_kb_and_kb_at_pos_out_dist_MC10Pos10_order1_L4H4d32_T50_heatmap_avg.pdf}};
%   \end{tikzpicture}
% \end{subfigure} &
% \begin{subfigure}[t]{0.2\textwidth}\centering
%   \begin{tikzpicture}[remember picture]
%     \node at (0,0) {\includegraphics[scale=0.2]{figures/heatmaps/H4L4MLPHead/only_bi_rate_out_dist_MC10Pos10_order1_L4H4d32_T50_heatmap_avg.pdf}};
%   \end{tikzpicture}
% \end{subfigure} &
% \rownote{No improvements.}
% \\[-20pt]

% ───────────────────────── Row 2: MLP ────────────────────────────────
\begin{subfigure}[t]{0.2\textwidth}\centering
  \rowstrut
  \begin{tikzpicture}[remember picture]
    \node at (0,0) {\includegraphics[scale=0.2]{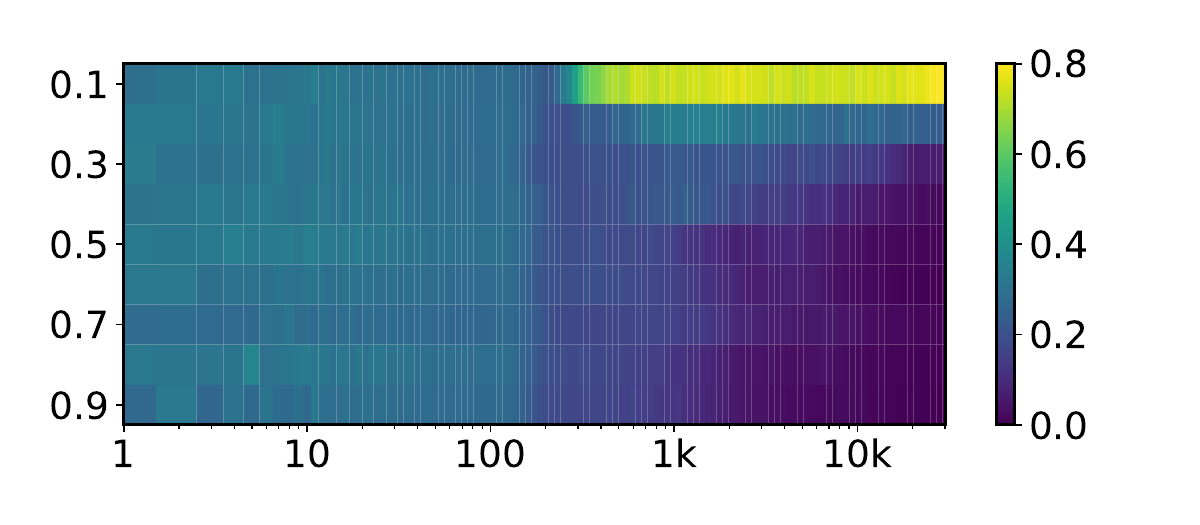}};
    \node[scale=0.75,rotate=90] at (-2.2,0.1) {\textbf{MLP}};
  \end{tikzpicture}
\end{subfigure} &
\begin{subfigure}[t]{0.2\textwidth}\centering
  \begin{tikzpicture}[remember picture]
    \node at (0,0) {\includegraphics[scale=0.2]{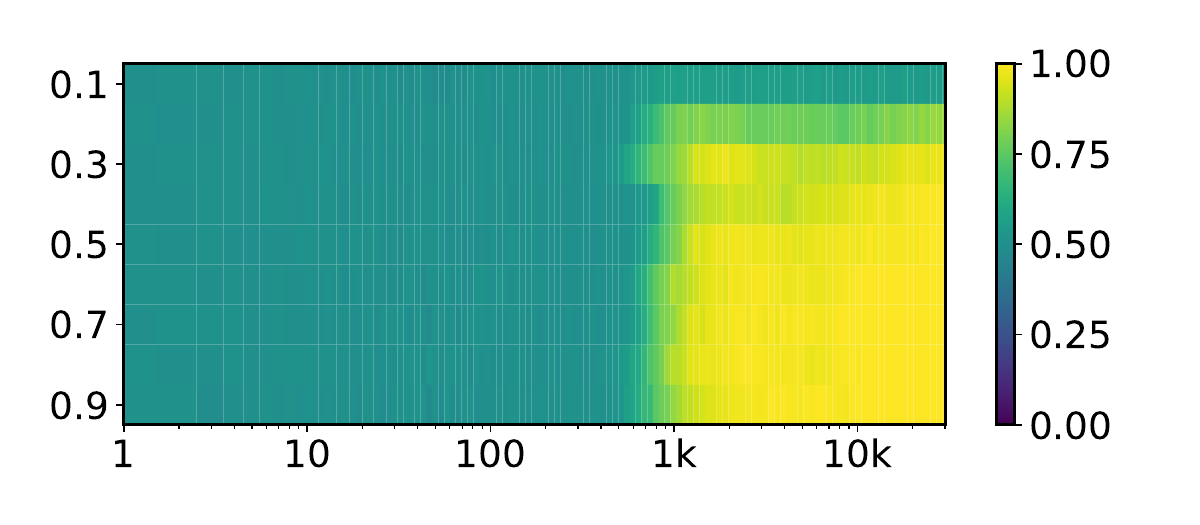}};
  \end{tikzpicture}
\end{subfigure} &
\begin{subfigure}[t]{0.2\textwidth}\centering
  \begin{tikzpicture}[remember picture]
    \node at (0,0) {\includegraphics[scale=0.2]{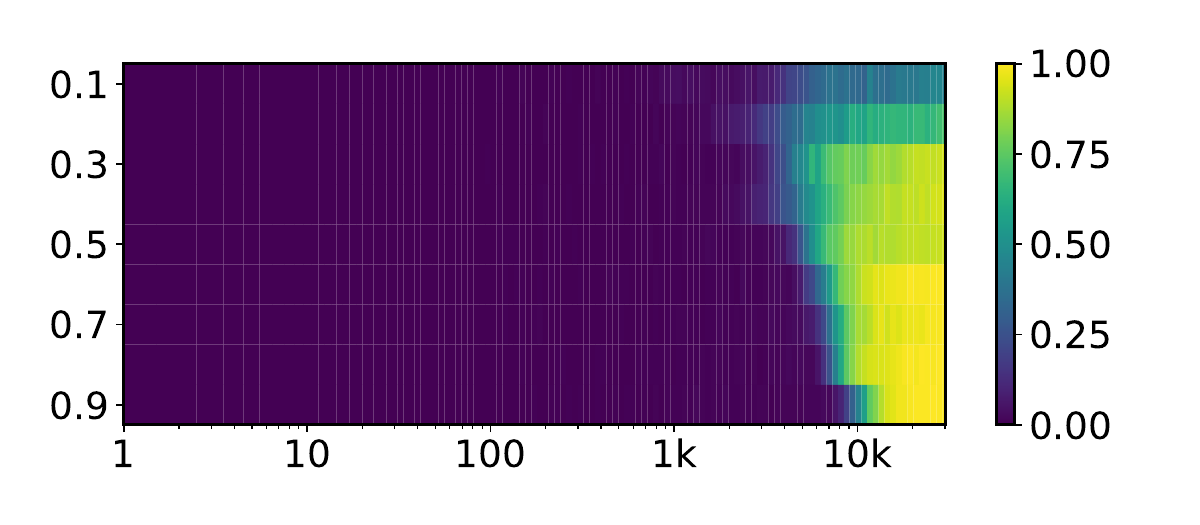}};
  \end{tikzpicture}
\end{subfigure} &
\rownote{No improvements.}
\\[-20pt]

% ───────────────────────── Row 9: V + O ──────────────────────────────
\begin{subfigure}[t]{0.2\textwidth}\centering
  \rowstrut
  \begin{tikzpicture}[remember picture]
    \node at (0,0) {\includegraphics[scale=0.2]{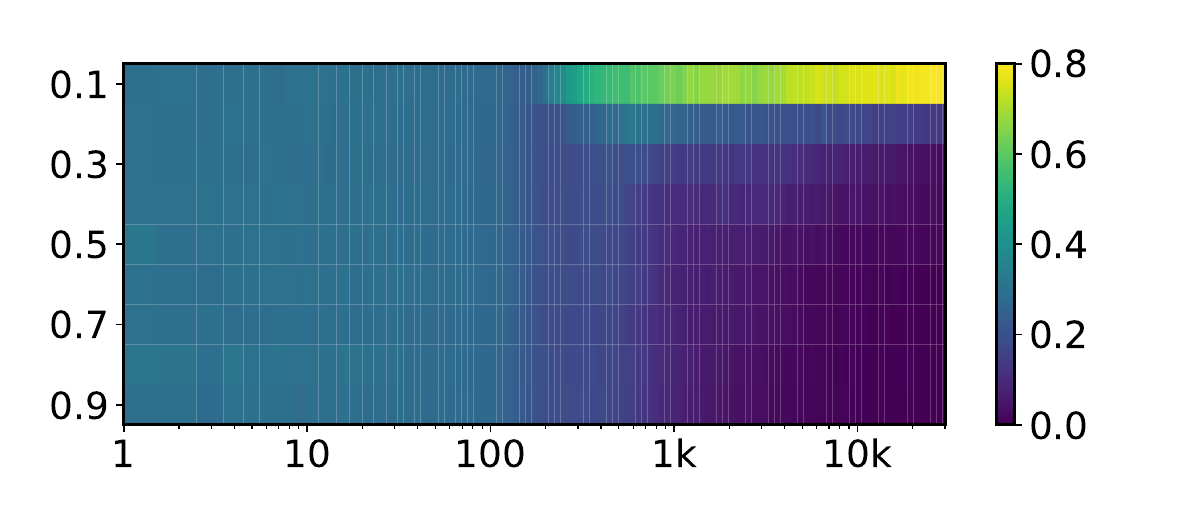}};
    \node[scale=0.75,rotate=90] at (-2.2,0.1) {$\mathbf{V},\mathbf{O}$};
  \end{tikzpicture}
\end{subfigure} &
\begin{subfigure}[t]{0.2\textwidth}\centering
  \begin{tikzpicture}[remember picture]
    \node at (0,0) {\includegraphics[scale=0.2]{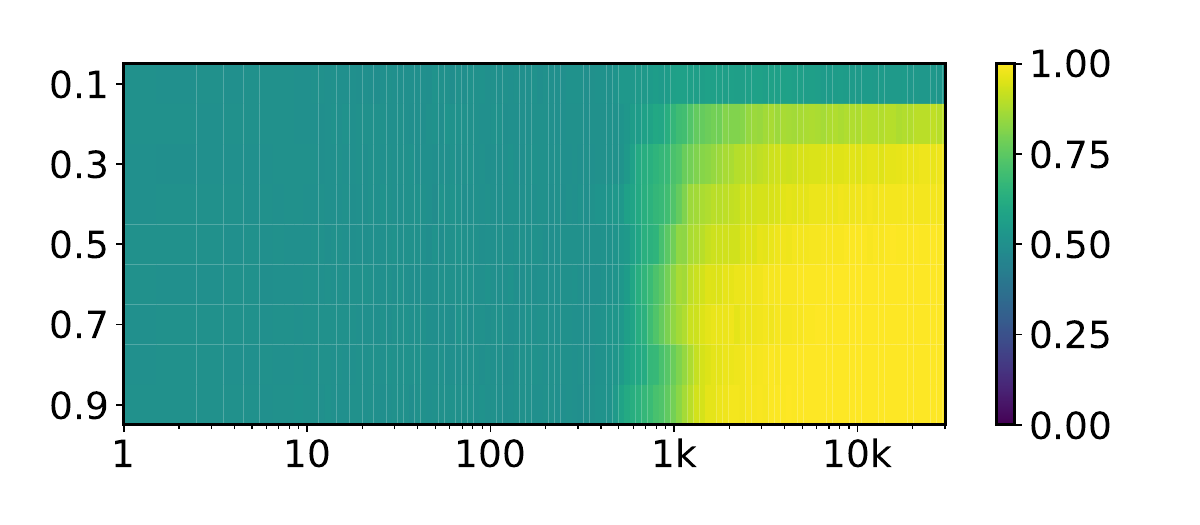}};
  \end{tikzpicture}
\end{subfigure} &
\begin{subfigure}[t]{0.2\textwidth}\centering
  \begin{tikzpicture}[remember picture]
    \node at (0,0) {\includegraphics[scale=0.2]{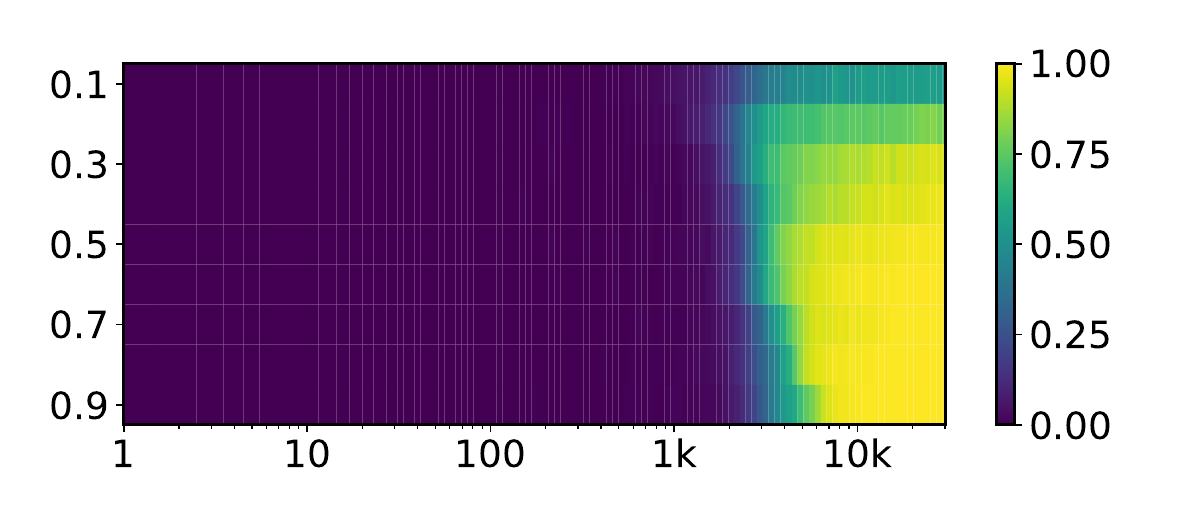}};
  \end{tikzpicture}
\end{subfigure} &
\rownote{}
\\[-20pt]

% ───────────────────────── Row 10: MLP + E ───────────────────────────
\begin{subfigure}[t]{0.2\textwidth}\centering
  \rowstrut
  \begin{tikzpicture}[remember picture]
    \node at (0,0) {\includegraphics[scale=0.2]{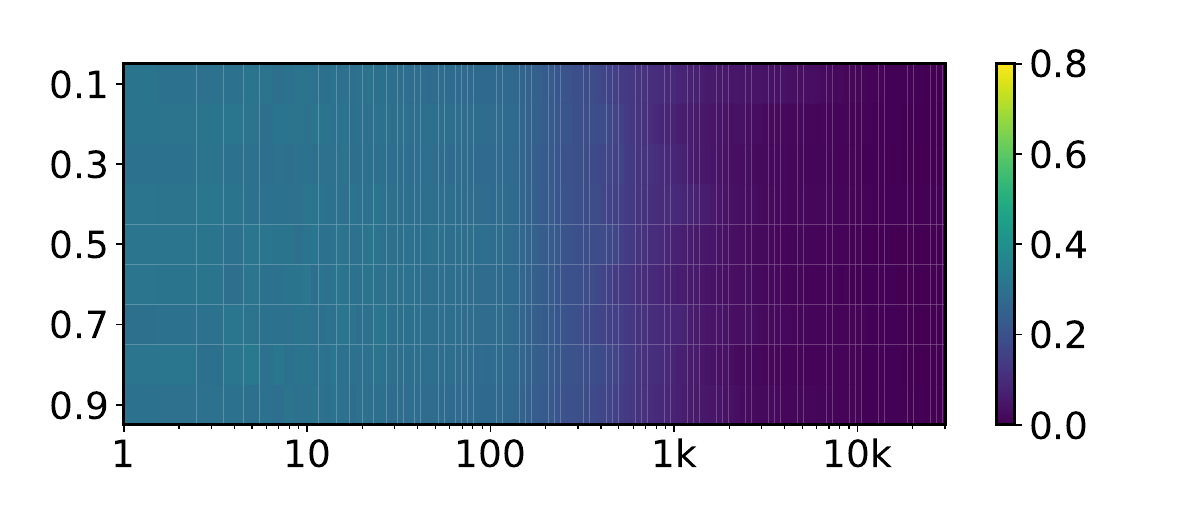}};
    \node[scale=0.75,rotate=90] at (-2.2,0.1) {$\textbf{MLP},\mathbf{E}$};
  \end{tikzpicture}
\end{subfigure} &
\begin{subfigure}[t]{0.2\textwidth}\centering
  \begin{tikzpicture}[remember picture]
    \node at (0,0) {\includegraphics[scale=0.2]{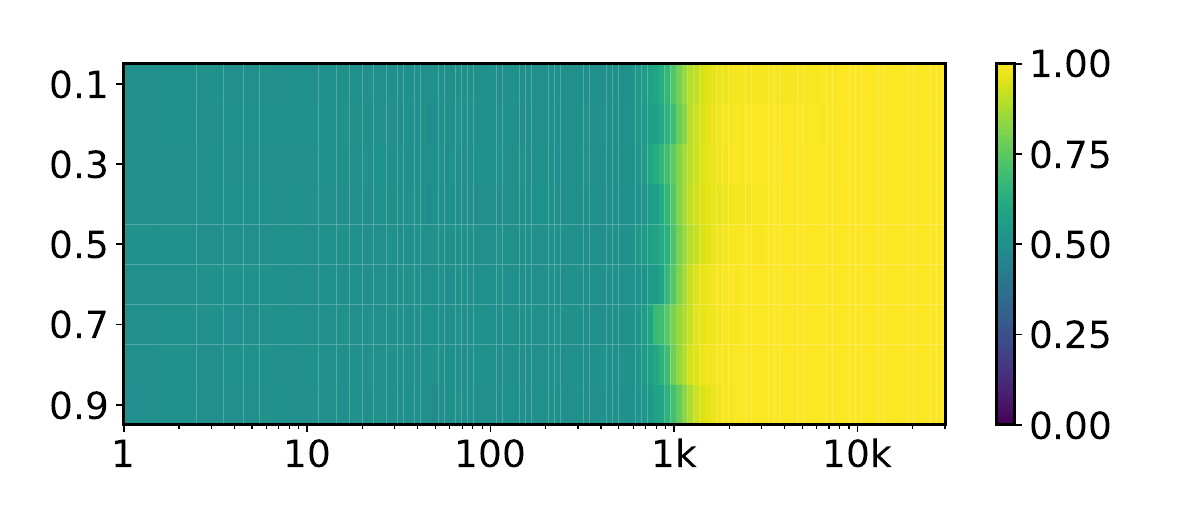}};
  \end{tikzpicture}
\end{subfigure} &
\begin{subfigure}[t]{0.2\textwidth}\centering
  \begin{tikzpicture}[remember picture]
    \node at (0,0) {\includegraphics[scale=0.2]{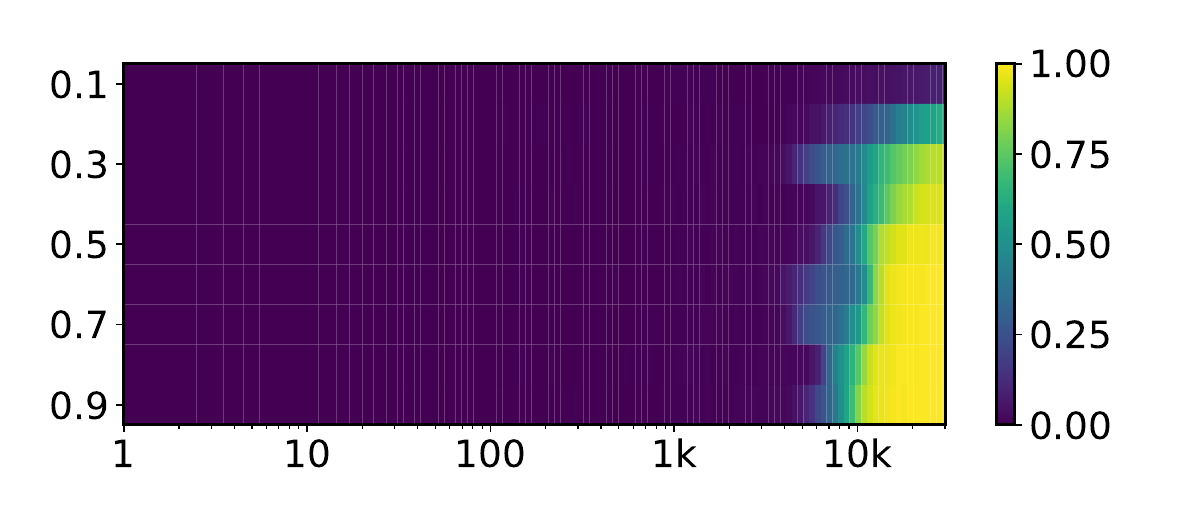}};
  \end{tikzpicture}
\end{subfigure} &
\rownote{}
\\[-20pt]

% ───────────────────────── Row 11: MLP + E + U ───────────────────────
\begin{subfigure}[t]{0.2\textwidth}\centering
  \rowstrut
  \begin{tikzpicture}[remember picture]
    \node at (0,0) {\includegraphics[scale=0.2]{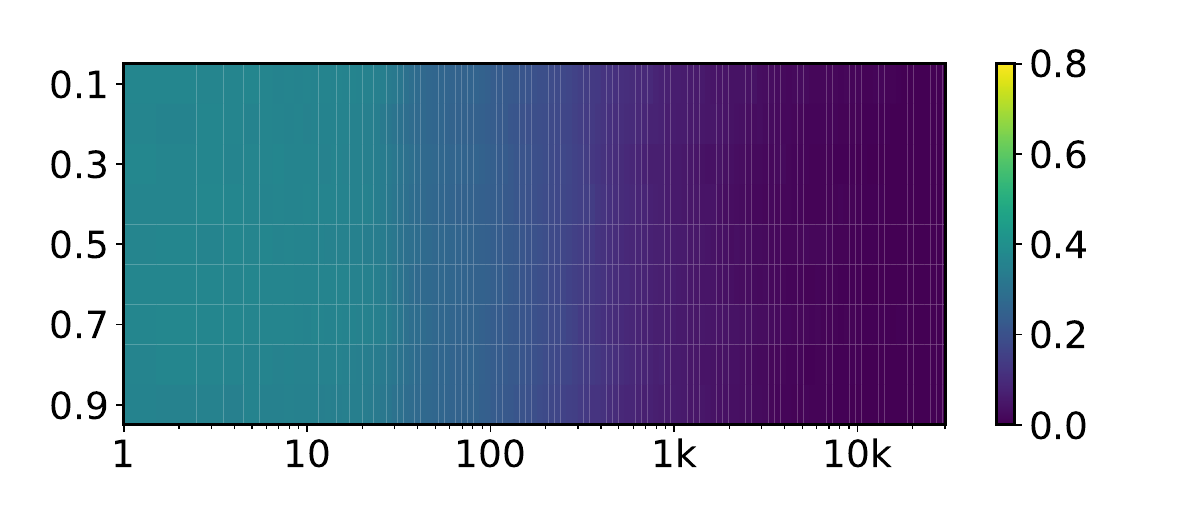}};
    \node[scale=0.75,rotate=90] at (-2.2,0.1) {$\textbf{MLP},\mathbf{E},\mathbf{U}$};
  \end{tikzpicture}
\end{subfigure} &
\begin{subfigure}[t]{0.2\textwidth}\centering
  \begin{tikzpicture}[remember picture]
    \node at (0,0) {\includegraphics[scale=0.2]{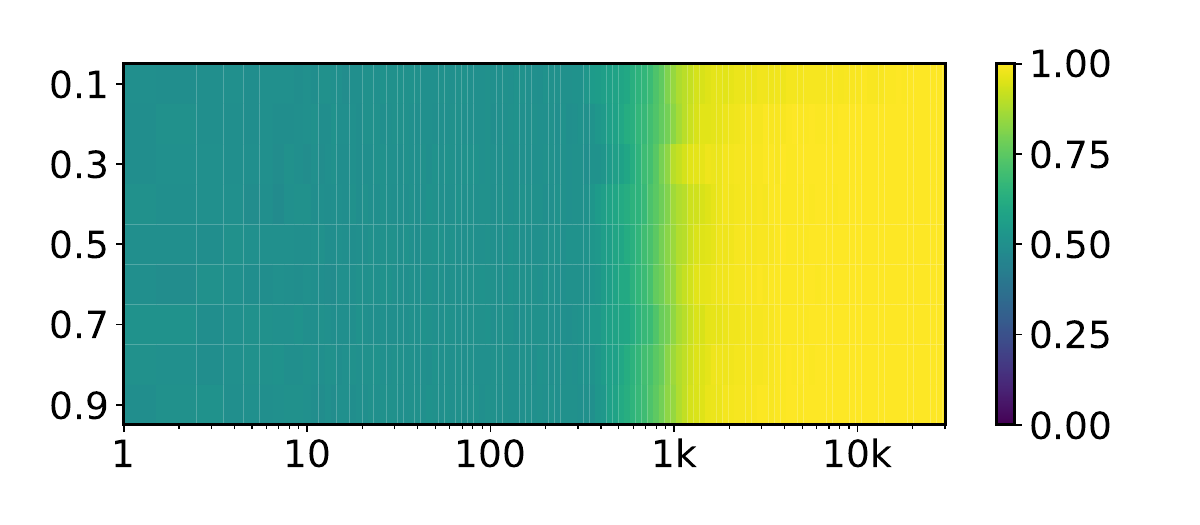}};
  \end{tikzpicture}
\end{subfigure} &
\begin{subfigure}[t]{0.2\textwidth}\centering
  \begin{tikzpicture}[remember picture]
    \node at (0,0) {\includegraphics[scale=0.2]{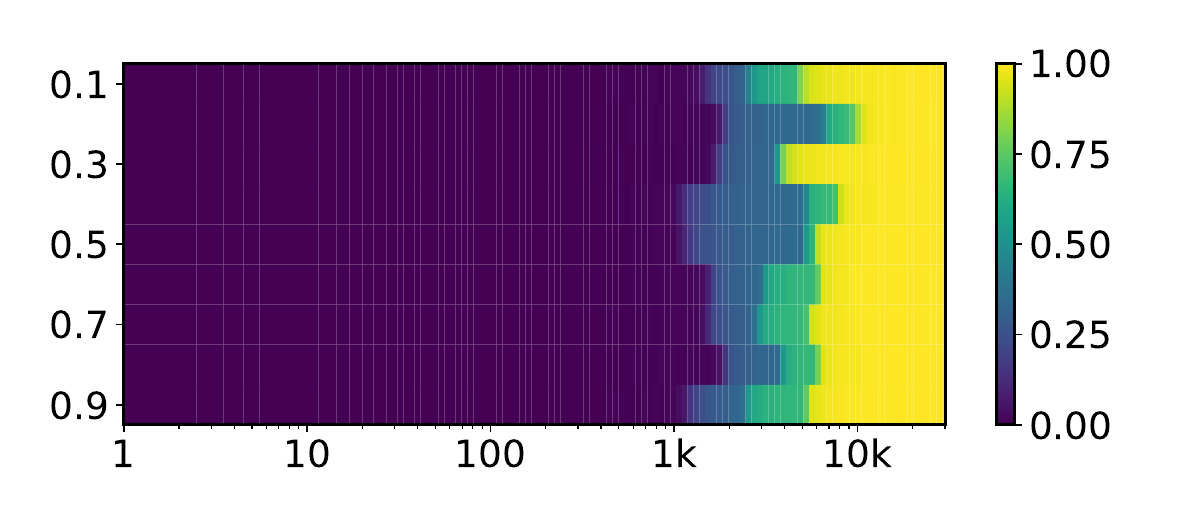}};
  \end{tikzpicture}
\end{subfigure} &
\rownote{}
\\[-20pt]

% ───────────────────────── Row 12: K + Q ─────────────────────────────
\begin{subfigure}[t]{0.2\textwidth}\centering
  \rowstrut
  \begin{tikzpicture}[remember picture]
    \node at (0,0) {\includegraphics[scale=0.2]{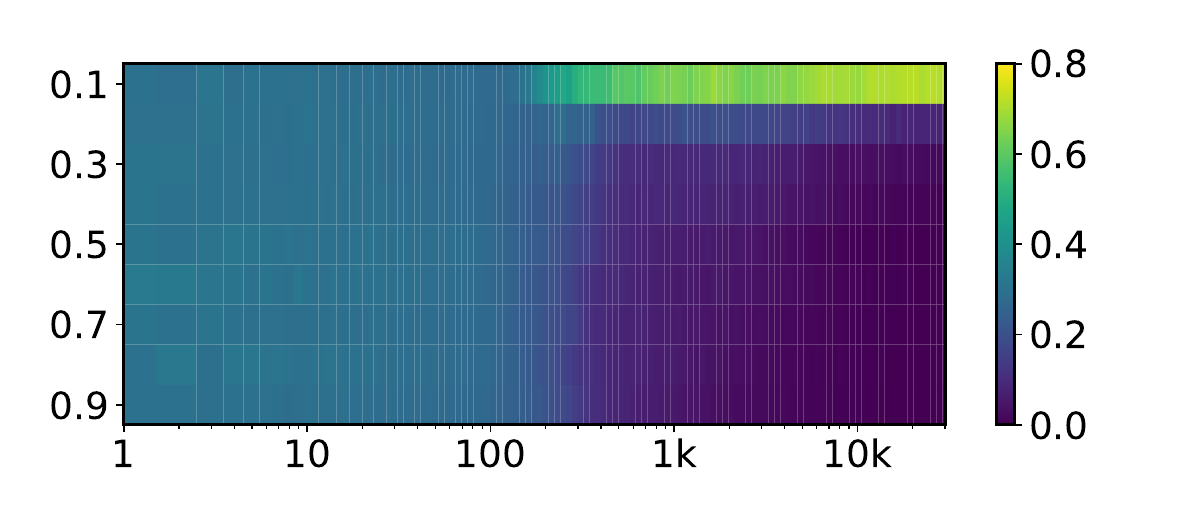}};
    \node[scale=0.75,rotate=90] at (-2.2,0.1) {$\mathbf{K},\mathbf{Q}$};
  \end{tikzpicture}
\end{subfigure} &
\begin{subfigure}[t]{0.2\textwidth}\centering
  \begin{tikzpicture}[remember picture]
    \node at (0,0) {\includegraphics[scale=0.2]{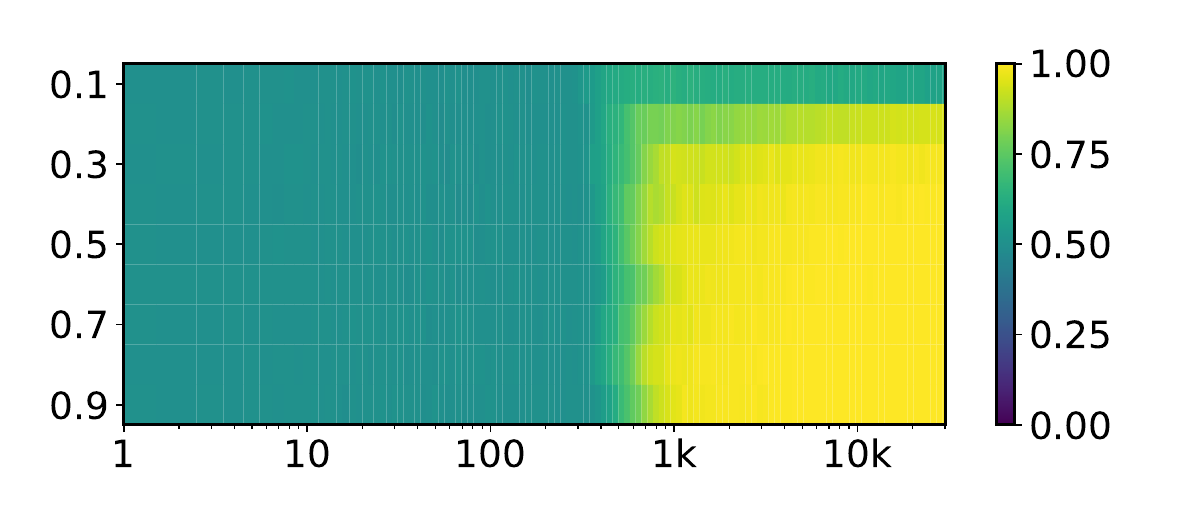}};
  \end{tikzpicture}
\end{subfigure} &
\begin{subfigure}[t]{0.2\textwidth}\centering
  \begin{tikzpicture}[remember picture]
    \node at (0,0) {\includegraphics[scale=0.2]{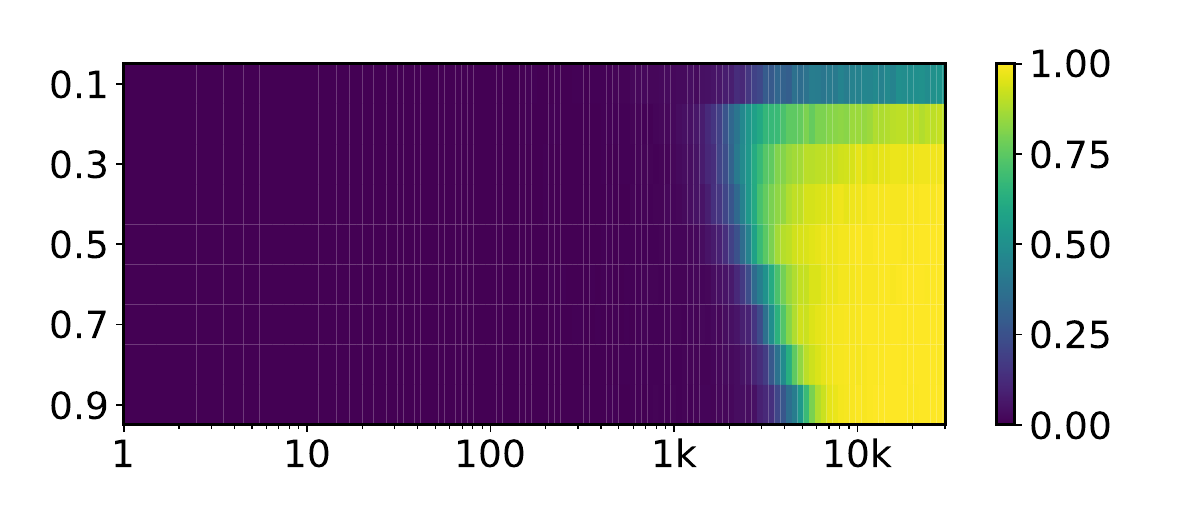}};
  \end{tikzpicture}
\end{subfigure} &
\rownote{}
\\[-20pt]

% ───────────────────────── Row 13: KQVOMLP ───────────────────────────
\begin{subfigure}[t]{0.2\textwidth}\centering
  \rowstrut
  \begin{tikzpicture}[remember picture]
    \node at (0,0) {\includegraphics[scale=0.2]{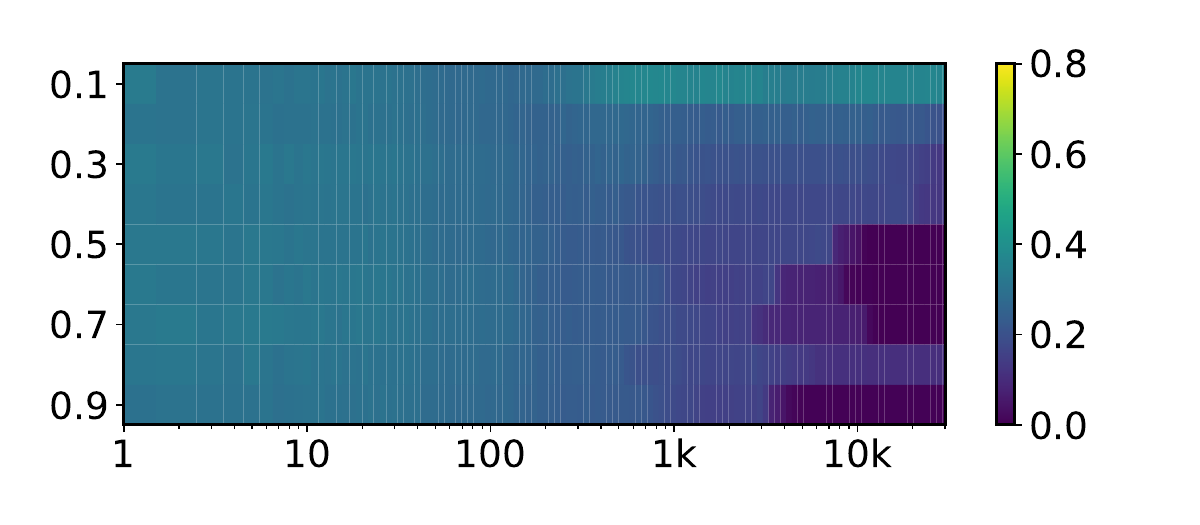}};
    \node[scale=0.65,rotate=90] at (-2.2,0.1) {$\mathbf{K,Q,V,O,MLP}$};
  \end{tikzpicture}
\end{subfigure} &
\begin{subfigure}[t]{0.2\textwidth}\centering
  \begin{tikzpicture}[remember picture]
    \node at (0,0) {\includegraphics[scale=0.2]{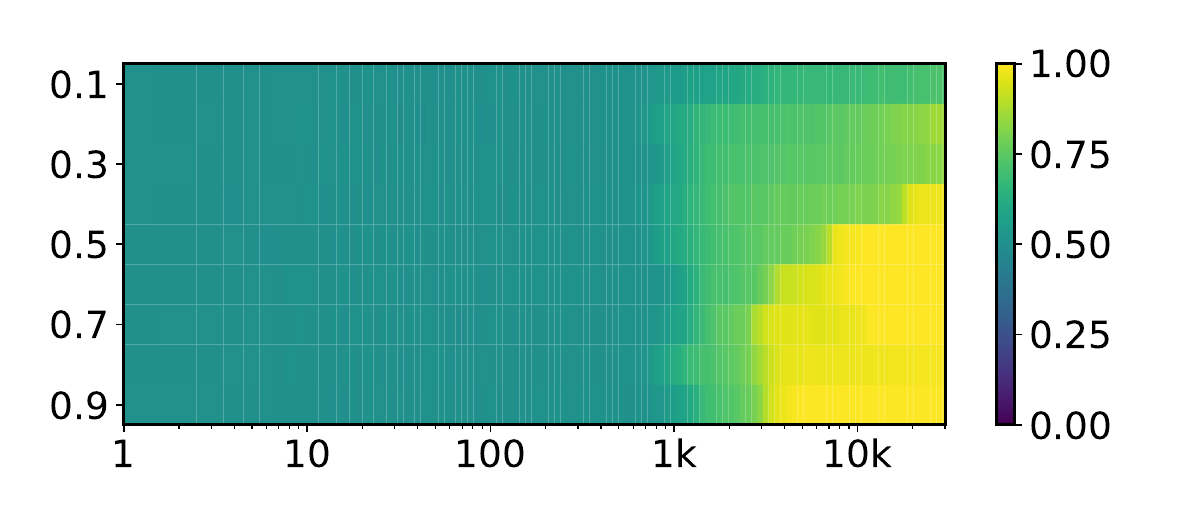}};
  \end{tikzpicture}
\end{subfigure} &
\begin{subfigure}[t]{0.2\textwidth}\centering
  \begin{tikzpicture}[remember picture]
    \node at (0,0) {\includegraphics[scale=0.2]{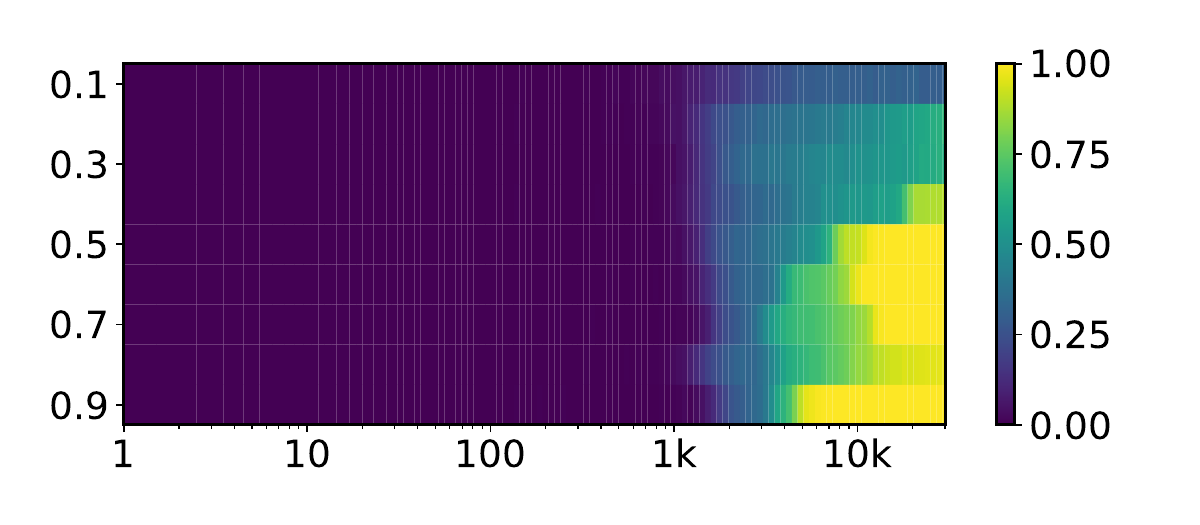}};
  \end{tikzpicture}
\end{subfigure} &
\rownote{}
\\[-0pt]

\end{tabular}
\caption{ Experiments of Sec.~\ref{sec:intervention} across different diversity levels and over training time. Each row label specifies which modules in the model, among the following, are frozen to that of $\highmodel$: softmax attention matrices $\textbf{Attn}$, token embedding / unembedding layer $\mathbf{E}/\mathbf{U}$, the weights of the fully-connected modules at each layer \textbf{MLP}, key, query, value, output matrices in self-attention module $\mathbf{K},\mathbf{Q},\mathbf{V},\mathbf{O}$.}
\label{fig:app_interventions}
\end{figure}

\begin{figure}[p]
\vspace{-70pt}
\centering
% 4 columns: three plots + right-hand note lane
\begin{tabular}{@{\hspace{-5pt}}c@{\hspace{40pt}}c@{\hspace{35pt}}c@{\hspace{40pt}}R{0.22\textwidth}@{}}

% ───────────────────────── Row 1: Standard ───────────────────────────
\begin{subfigure}[t]{0.2\textwidth}\centering
  \rowstrut
  \begin{tikzpicture}[remember picture]
    \node at (0,0) {\includegraphics[scale=0.2]{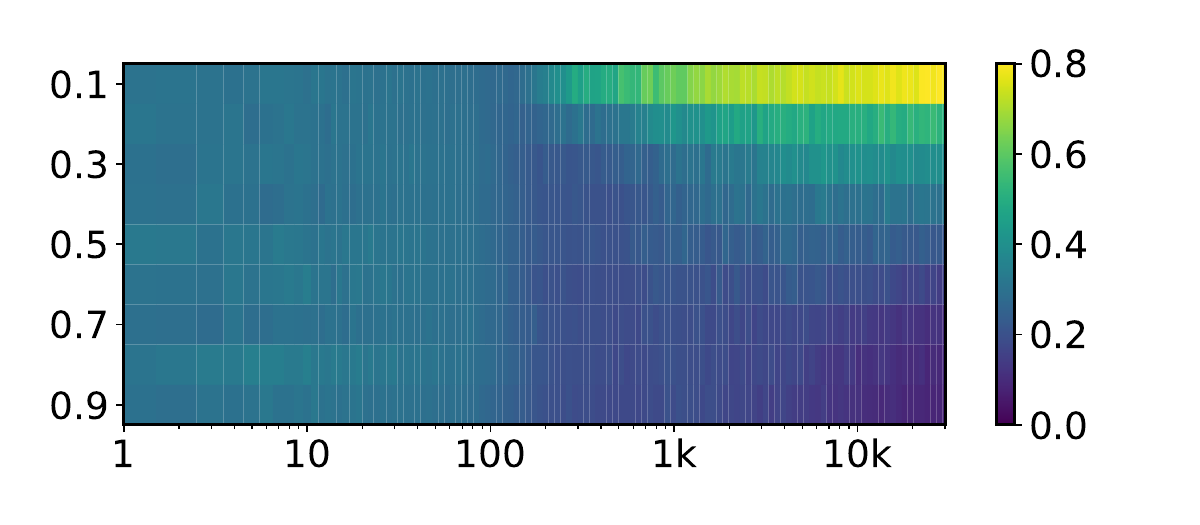}};
    \node[scale=0.75,rotate=90] at (-2.2,0.1) {\textbf{Standard}};
    % \node[scale=0.8] at (0.0,-1.2) {iteration ($\times 10^3$)};
    \node[scale=0.8] at (0.0, 1.0) {\textbf{$\KL$}};
  \end{tikzpicture}
\end{subfigure} &
{
\begin{subfigure}[t]{0.2\textwidth}\centering
  \begin{tikzpicture}[remember picture]
    \node at (0,0) {\includegraphics[scale=0.2]{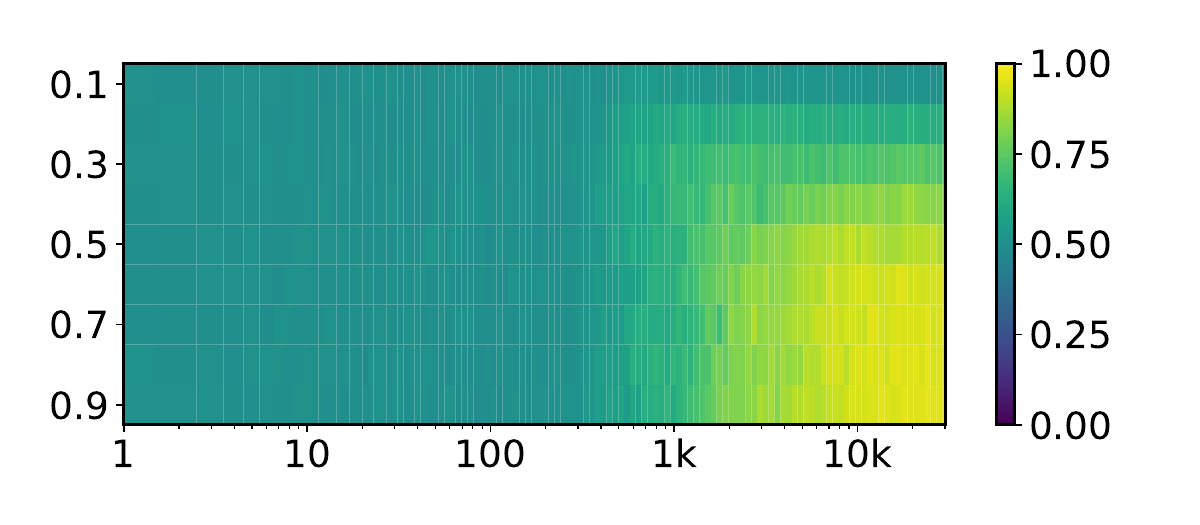}};
    \node[scale=0.8] at (0.0,1.0) {\textbf{$\posacc$}};
  \end{tikzpicture}
\end{subfigure}} &
{
\begin{subfigure}[t]{0.2\textwidth}\centering
  \begin{tikzpicture}[remember picture]
    \node at (0,0) {\includegraphics[scale=0.2]{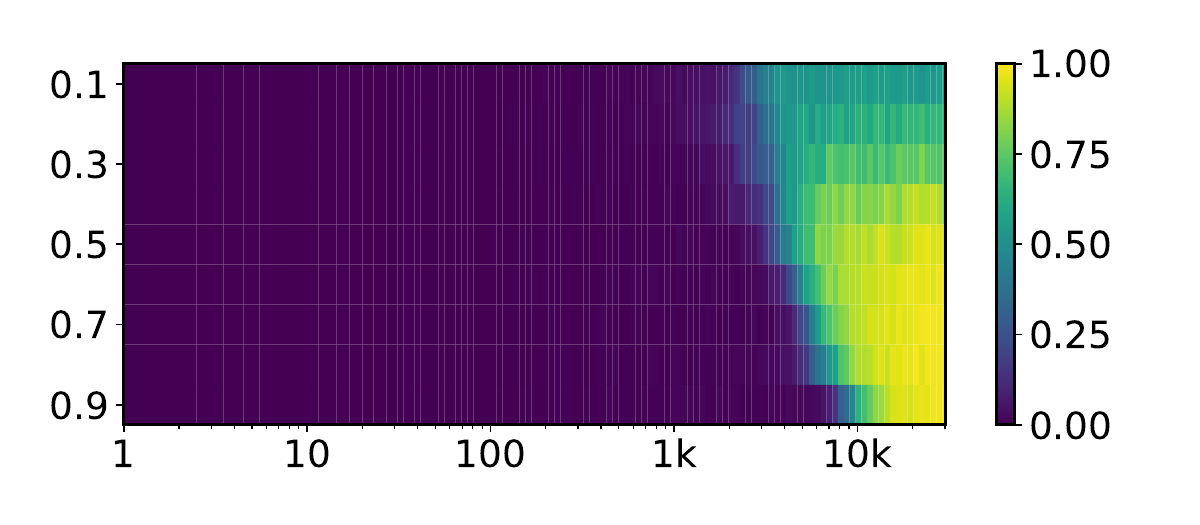}};
    \node[scale=0.8] at (0.0,1.0) {\textbf{$\factacc$}};
  \end{tikzpicture}
\end{subfigure}} &
\rownote{Failure at low diversity. Slowdown with high diversity}
\\[-20pt]
% ───────────────────────── Row 4: Attn (patched) ─────────────────────
\begin{subfigure}[t]{0.2\textwidth}\centering
  \rowstrut
  \begin{tikzpicture}[remember picture]
    \node at (0,0) {\includegraphics[scale=0.2]{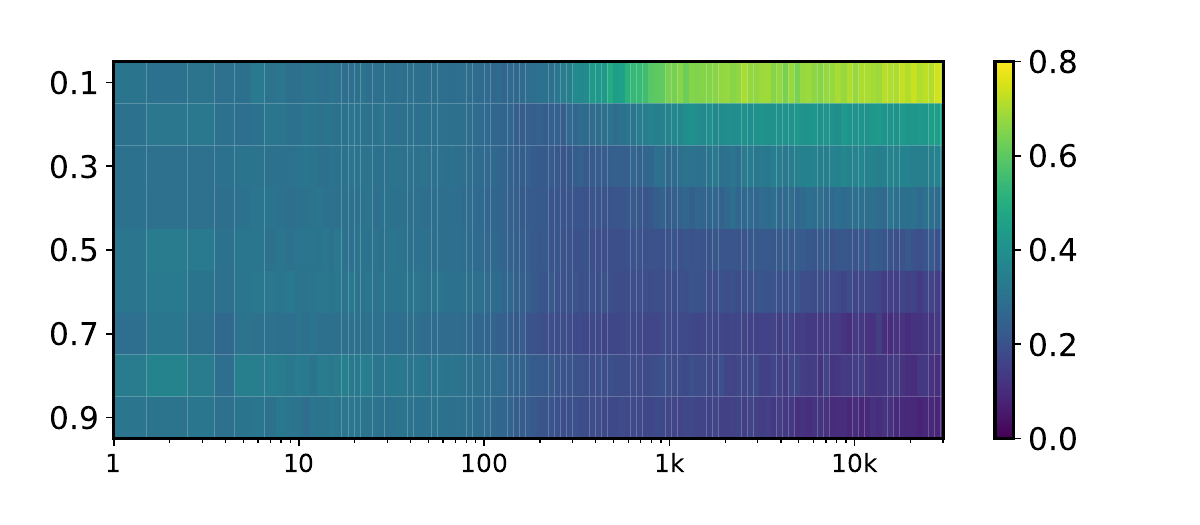}};
    \node[scale=0.75,rotate=90] at (-2.2,0.1) {\textbf{Attn}};
  \end{tikzpicture}
\end{subfigure} &
\begin{subfigure}[t]{0.2\textwidth}\centering
  \begin{tikzpicture}[remember picture]
    \node at (0,0) {\includegraphics[scale=0.2]{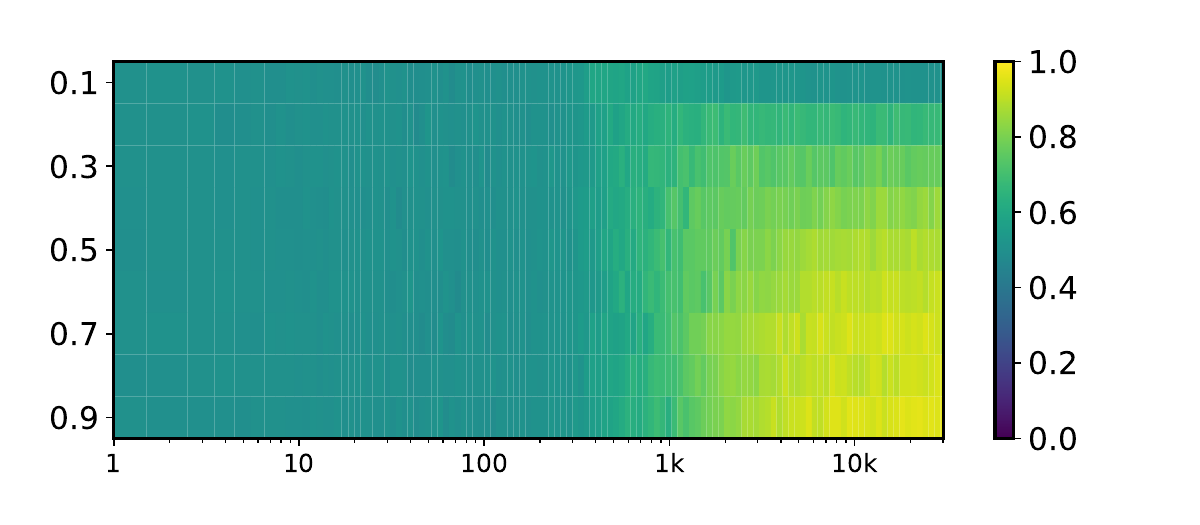}};
  \end{tikzpicture}
\end{subfigure} &
\begin{subfigure}[t]{0.2\textwidth}\centering
  \begin{tikzpicture}[remember picture]
    \node at (0,0) {\includegraphics[scale=0.2]{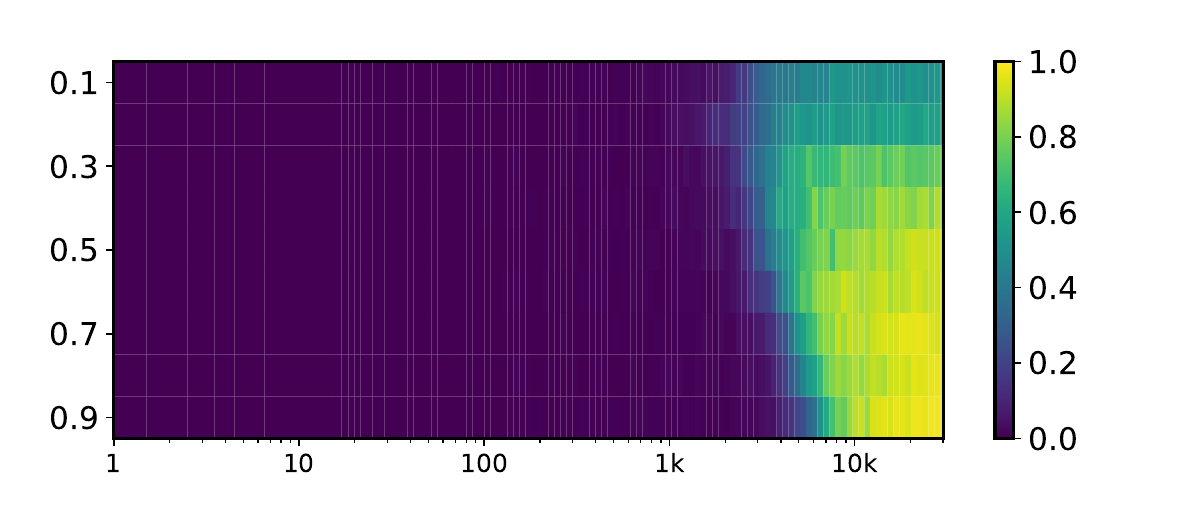}};
  \end{tikzpicture}
\end{subfigure} &
\rownote{\emph{Stat} \& \emph{pos} improve. No impact on \emph{fact}.}
\\[-20pt]

% ───────────────────────── Row 5: E (embeddings) ─────────────────────
\begin{subfigure}[t]{0.2\textwidth}\centering
  \rowstrut
  \begin{tikzpicture}[remember picture]
    \node at (0,0) {\includegraphics[scale=0.2]{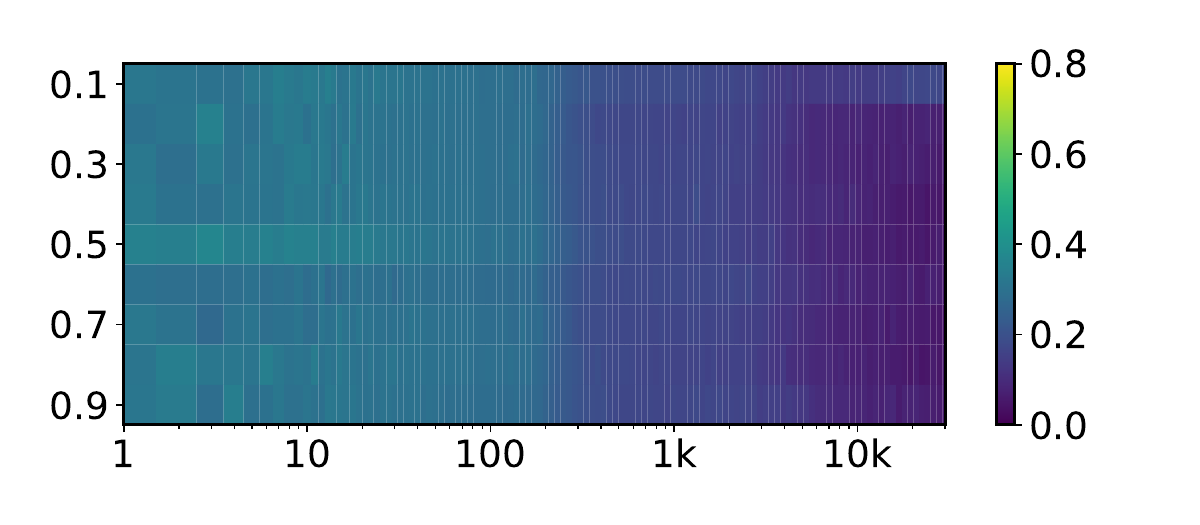}};
    \node[scale=0.75,rotate=90] at (-2.2,0.1) {\textbf{$\mathbf{E}$}};
  \end{tikzpicture}
\end{subfigure} &
\begin{subfigure}[t]{0.2\textwidth}\centering
  \begin{tikzpicture}[remember picture]
    \node at (0,0) {\includegraphics[scale=0.2]{figures/heatmaps/H1L1AttnPatch/combined_non_kb_and_kb_at_pos_out_dist_override_MC10Pos10_order1_L4H4d32_T50_heatmap_avg.pdf}};
  \end{tikzpicture}
\end{subfigure} &
\begin{subfigure}[t]{0.2\textwidth}\centering
  \begin{tikzpicture}[remember picture]
    \node at (0,0) {\includegraphics[scale=0.2]{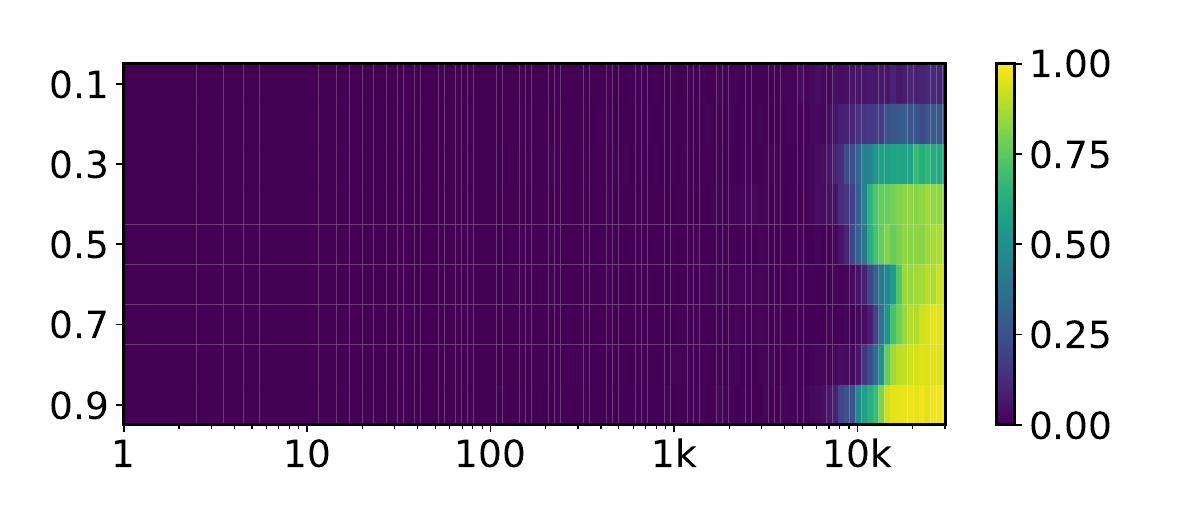}};
  \end{tikzpicture}
\end{subfigure} &
\rownote{}
\\[-20pt]

% ───────────────────────── Row 3: U (head) ───────────────────────────
\begin{subfigure}[t]{0.2\textwidth}\centering
  \rowstrut
  \begin{tikzpicture}[remember picture]
    \node at (0,0) {\includegraphics[scale=0.2]{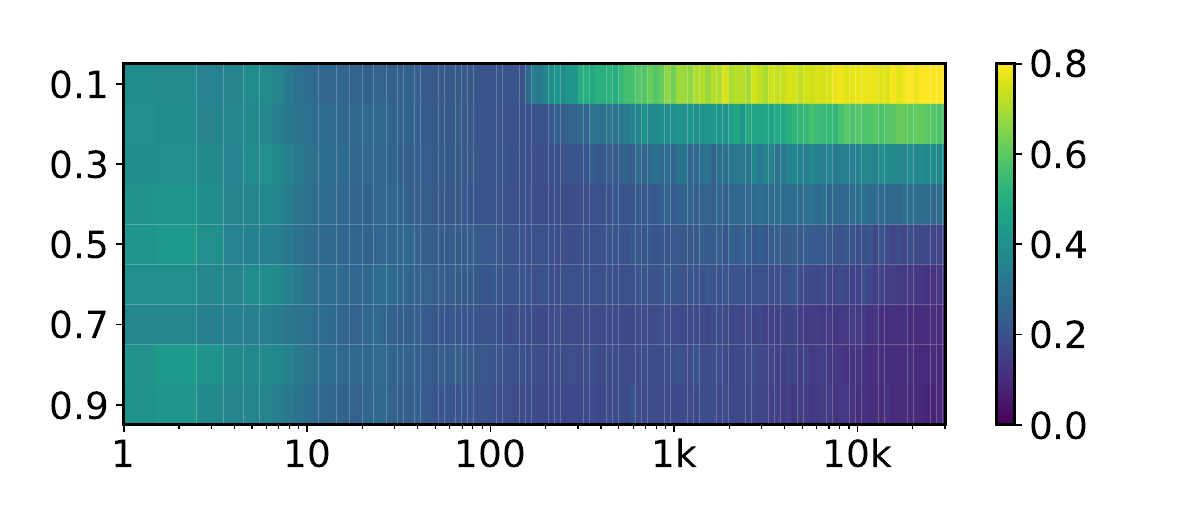}};
    \node[scale=0.75,rotate=90] at (-2.2,0.1) {\textbf{$\mathbf{U}$}};
  \end{tikzpicture}
\end{subfigure} &
\begin{subfigure}[t]{0.2\textwidth}\centering
  \begin{tikzpicture}[remember picture]
    \node at (0,0) {\includegraphics[scale=0.2]{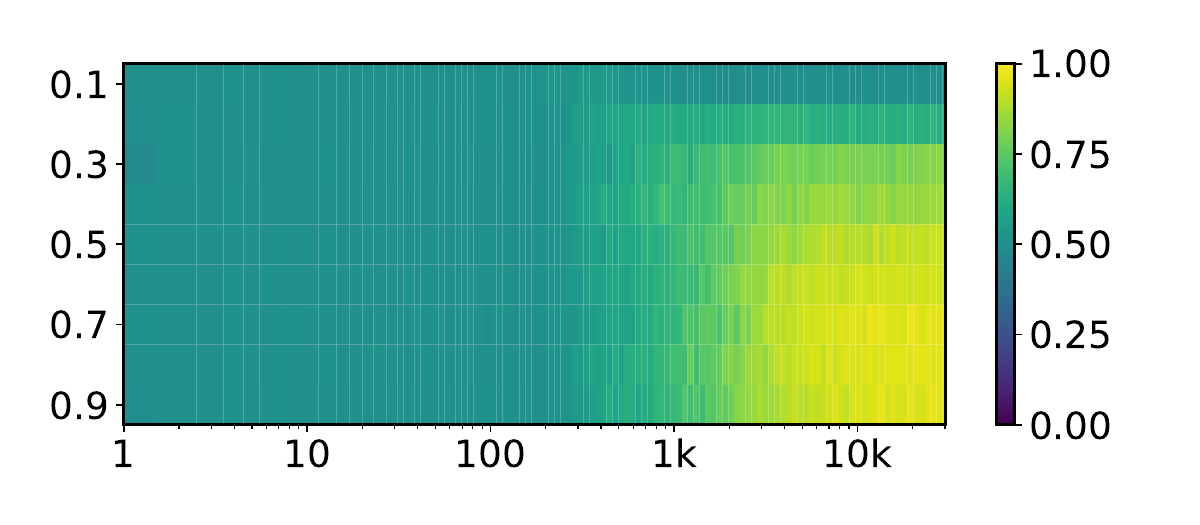}};
  \end{tikzpicture}
\end{subfigure} &
\begin{subfigure}[t]{0.2\textwidth}\centering
  \begin{tikzpicture}[remember picture]
    \node at (0,0) {\includegraphics[scale=0.2]{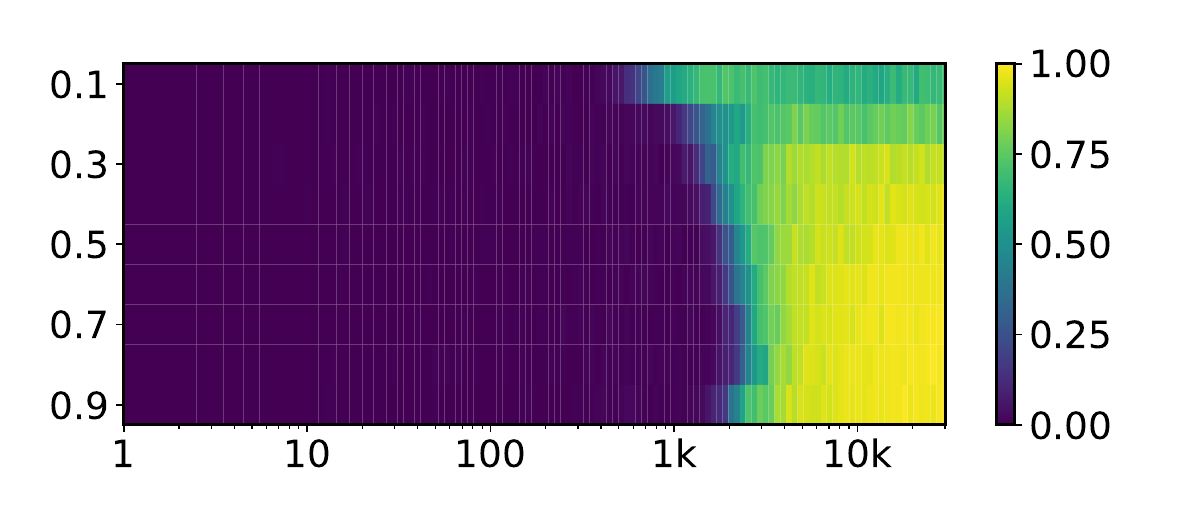}};
  \end{tikzpicture}
\end{subfigure} &
\rownote{No impact on \emph{stat} and \emph{pos}. Remedies slowdown of \emph{fact}.}
\\[-20pt]

% ───────────────────────── Row 6: U + Attn ───────────────────────────
\begin{subfigure}[t]{0.2\textwidth}\centering
  \rowstrut
  \begin{tikzpicture}[remember picture]
    \node at (0,0) {\includegraphics[scale=0.2]{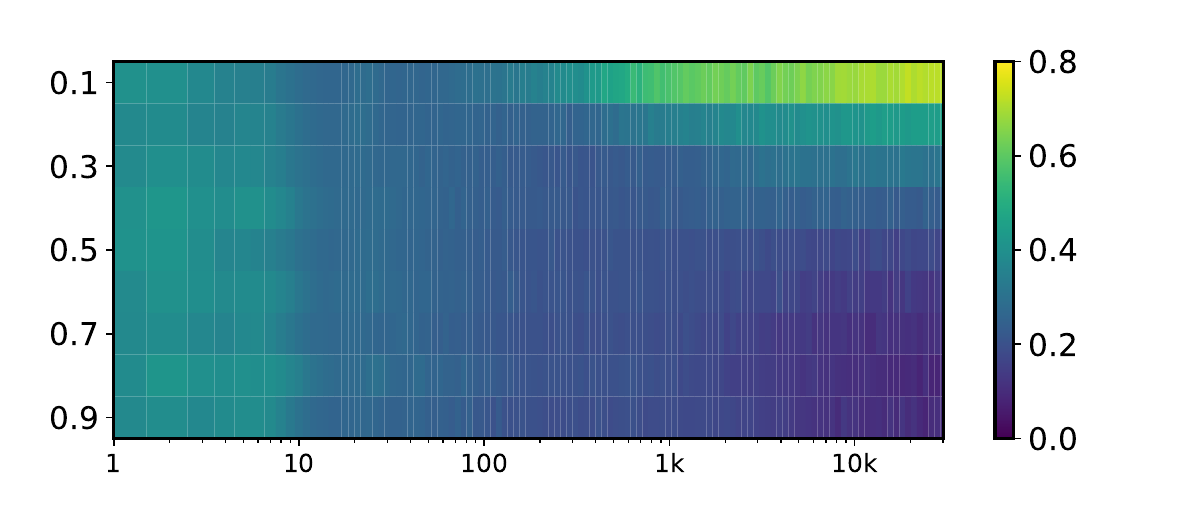}};
    \node[scale=0.75,rotate=90] at (-2.2,0.1) {$\mathbf{U},$\textbf{Attn}};
  \end{tikzpicture}
\end{subfigure} &
\begin{subfigure}[t]{0.2\textwidth}\centering
  \begin{tikzpicture}[remember picture]
    \node at (0,0) {\includegraphics[scale=0.2]{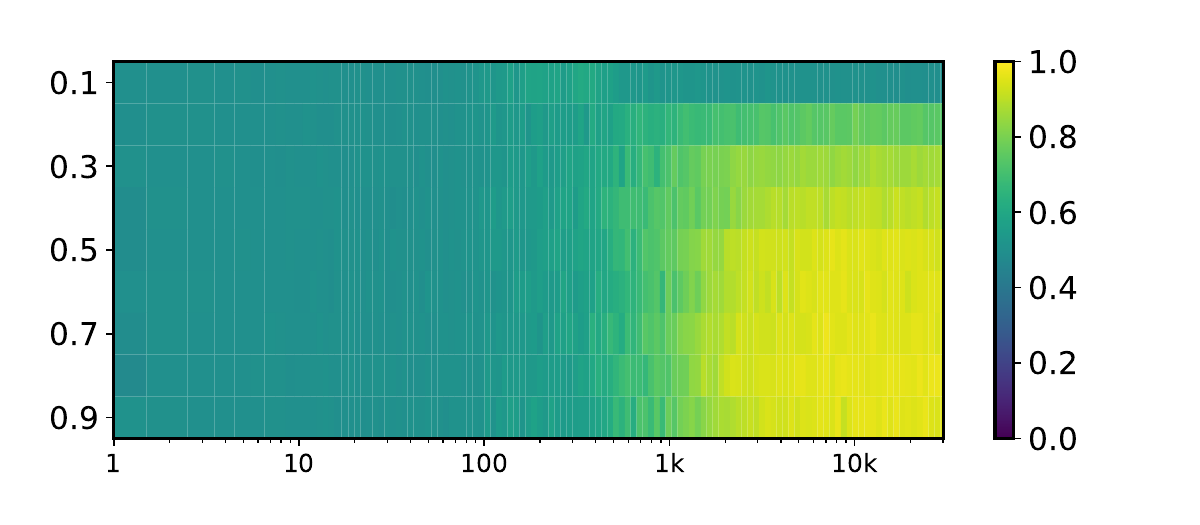}};
  \end{tikzpicture}
\end{subfigure} &
\begin{subfigure}[t]{0.2\textwidth}\centering
  \begin{tikzpicture}[remember picture]
    \node at (0,0) {\includegraphics[scale=0.2]{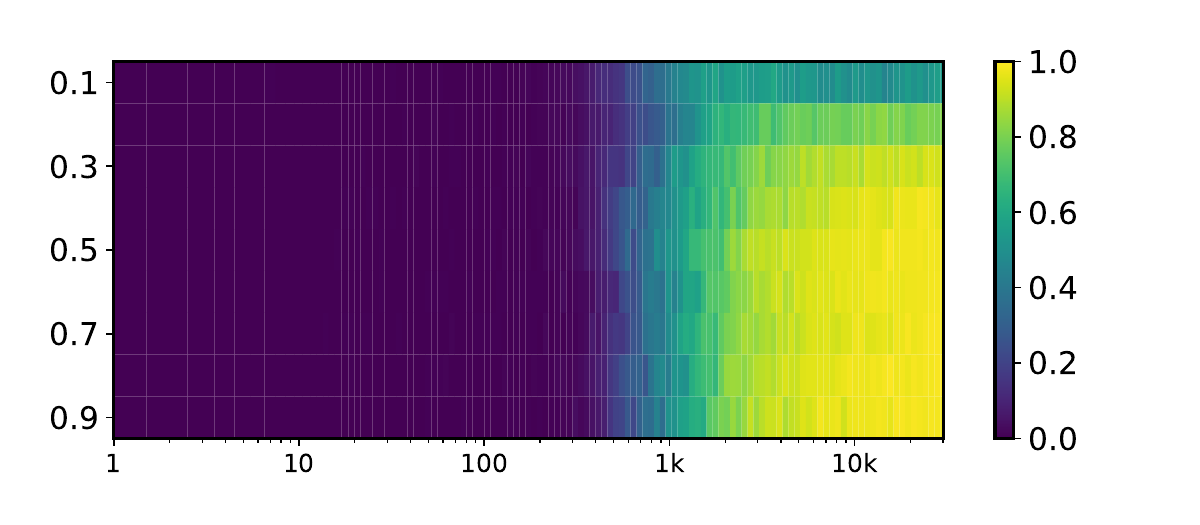}};
  \end{tikzpicture}
\end{subfigure} &
\rownote{Improves all metrics, on both ends of diversity.}
\\[-20pt]

% ───────────────────────── Row 7: U + E ──────────────────────────────
\begin{subfigure}[t]{0.2\textwidth}\centering
  \rowstrut
  \begin{tikzpicture}[remember picture]
    \node at (0,0) {\includegraphics[scale=0.2]{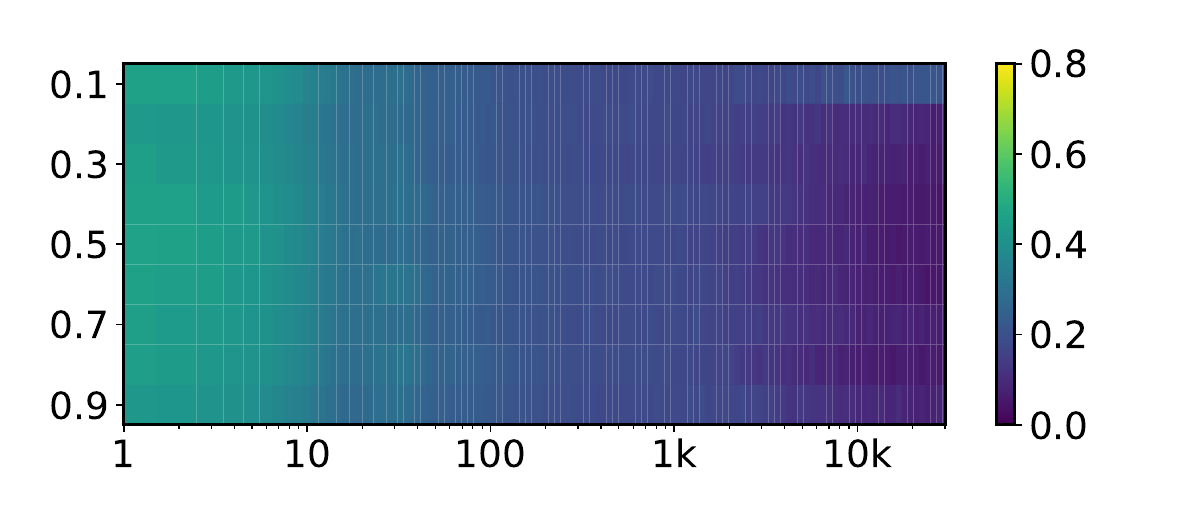}};
    \node[scale=0.75,rotate=90] at (-2.2,0.1) {$\mathbf{U},\mathbf{E}$};
  \end{tikzpicture}
\end{subfigure} &
\begin{subfigure}[t]{0.2\textwidth}\centering
  \begin{tikzpicture}[remember picture]
    \node at (0,0) {\includegraphics[scale=0.2]{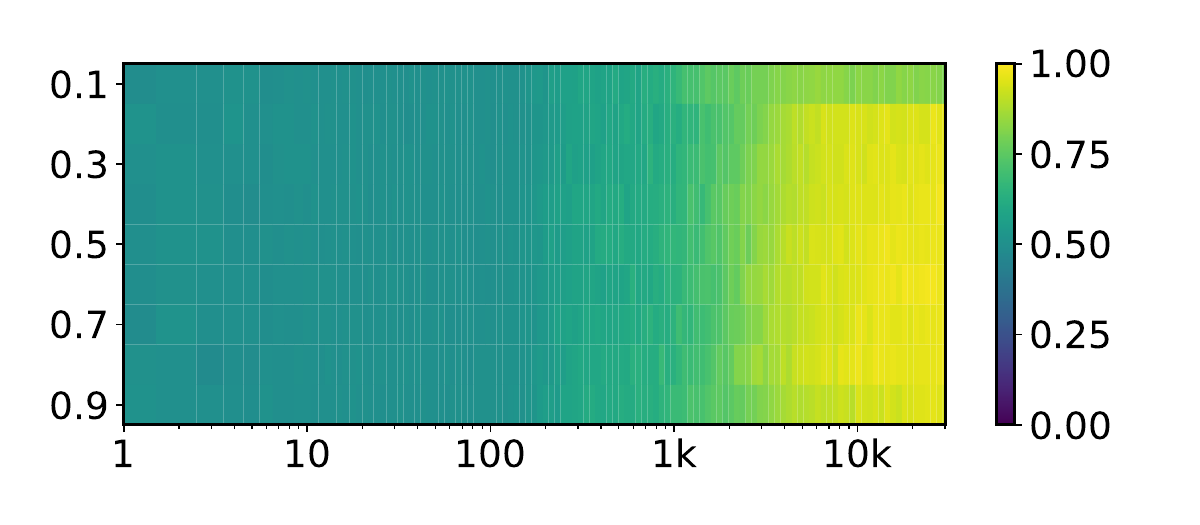}};
  \end{tikzpicture}
\end{subfigure} &
\begin{subfigure}[t]{0.2\textwidth}\centering
  \begin{tikzpicture}[remember picture]
    \node at (0,0) {\includegraphics[scale=0.2]{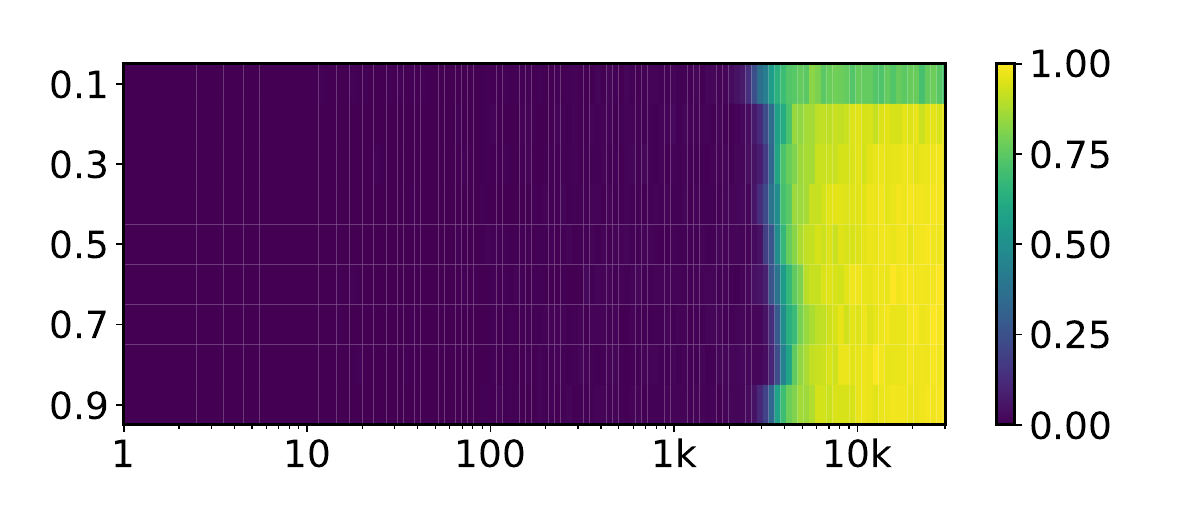}};
  \end{tikzpicture}
\end{subfigure} &
\rownote{}
\\[-20pt]

% ───────────────────────── Row 8: U + E + Attn ───────────────────────
\begin{subfigure}[t]{0.2\textwidth}\centering
  \rowstrut
  \begin{tikzpicture}[remember picture]
    \node at (0,0) {\includegraphics[scale=0.2]{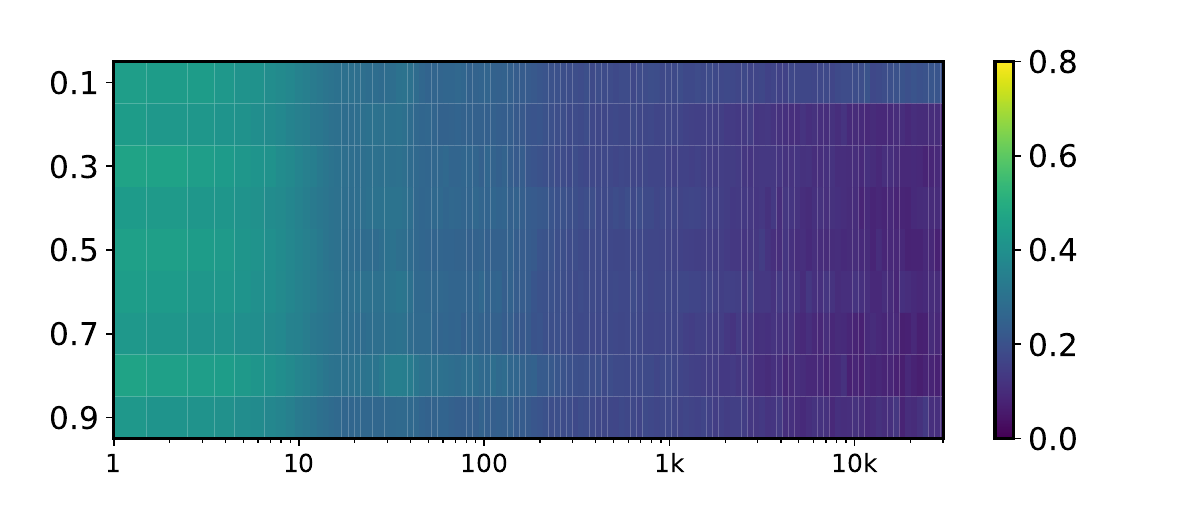}};
    \node[scale=0.75,rotate=90] at (-2.2,0.1) {$\mathbf{U},\mathbf{E},$\textbf{Attn}};
  \end{tikzpicture}
\end{subfigure} &
\begin{subfigure}[t]{0.2\textwidth}\centering
  \begin{tikzpicture}[remember picture]
    \node at (0,0) {\includegraphics[scale=0.2]{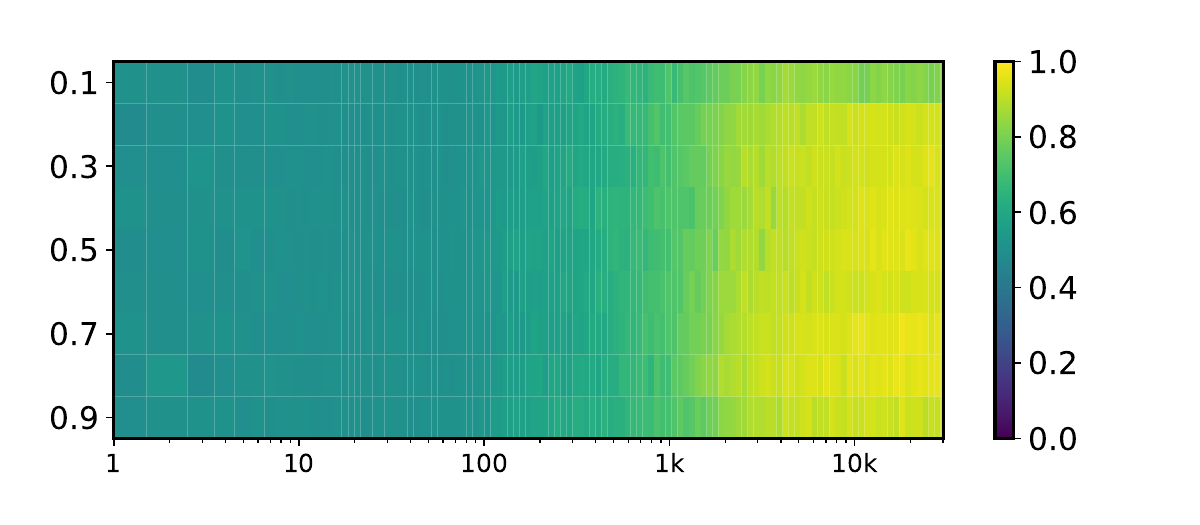}};
  \end{tikzpicture}
\end{subfigure} &
\begin{subfigure}[t]{0.2\textwidth}\centering
  \begin{tikzpicture}[remember picture]
    \node at (0,0) {\includegraphics[scale=0.2]{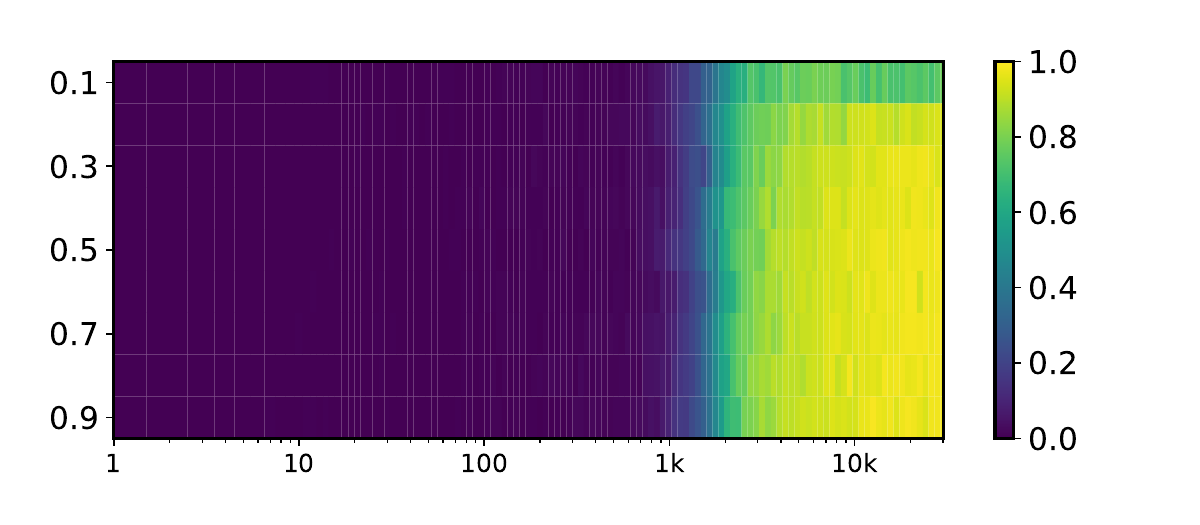}};
  \end{tikzpicture}
\end{subfigure} &
\rownote{}
\\[-20pt]

% 

% ───────────────────────── Row 2: MLP +U────────────────────────────────
% \begin{subfigure}[t]{0.2\textwidth}\centering
%   \rowstrut
%   \begin{tikzpicture}[remember picture]
%     \node at (0,0) {\includegraphics[scale=0.2]{figures/heatmaps/H1L1MLPHead/kl_masked_completion_GT_out_dist_MC10Pos10_order1_L4H4d32_T50_heatmap_avg.pdf}};
%     \node[scale=0.75,rotate=90] at (-2.2,0.1) {\textbf{MLP},\textbf{U} \tina{to do}};
%   \end{tikzpicture}
% \end{subfigure} &
% \begin{subfigure}[t]{0.2\textwidth}\centering
%   \begin{tikzpicture}[remember picture]
%     \node at (0,0) {\includegraphics[scale=0.2]{figures/heatmaps/H1L1MLPHead/combined_non_kb_and_kb_at_pos_out_dist_MC10Pos10_order1_L4H4d32_T50_heatmap_avg.pdf}};
%   \end{tikzpicture}
% \end{subfigure} &
% \begin{subfigure}[t]{0.2\textwidth}\centering
%   \begin{tikzpicture}[remember picture]
%     \node at (0,0) {\includegraphics[scale=0.2]{figures/heatmaps/H1L1MLPHead/only_bi_rate_out_dist_MC10Pos10_order1_L4H4d32_T50_heatmap_avg.pdf}};
%   \end{tikzpicture}
% \end{subfigure} &
% \rownote{No improvements.}
% \\[-20pt]

% ───────────────────────── Row 2: MLP ────────────────────────────────
\begin{subfigure}[t]{0.2\textwidth}\centering
  \rowstrut
  \begin{tikzpicture}[remember picture]
    \node at (0,0) {\includegraphics[scale=0.2]{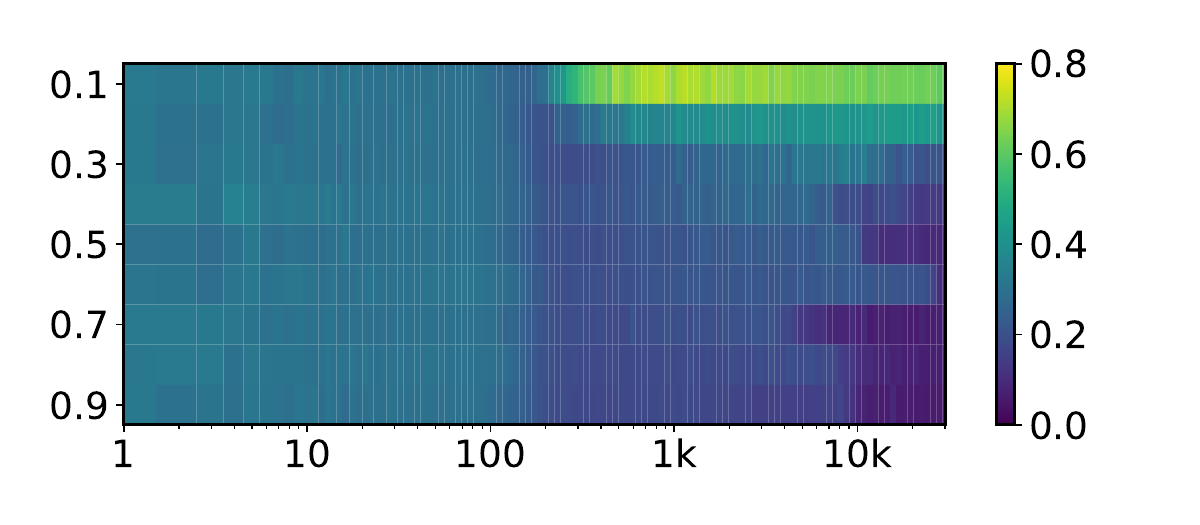}};
    \node[scale=0.75,rotate=90] at (-2.2,0.1) {\textbf{MLP}};
  \end{tikzpicture}
\end{subfigure} &
\begin{subfigure}[t]{0.2\textwidth}\centering
  \begin{tikzpicture}[remember picture]
    \node at (0,0) {\includegraphics[scale=0.2]{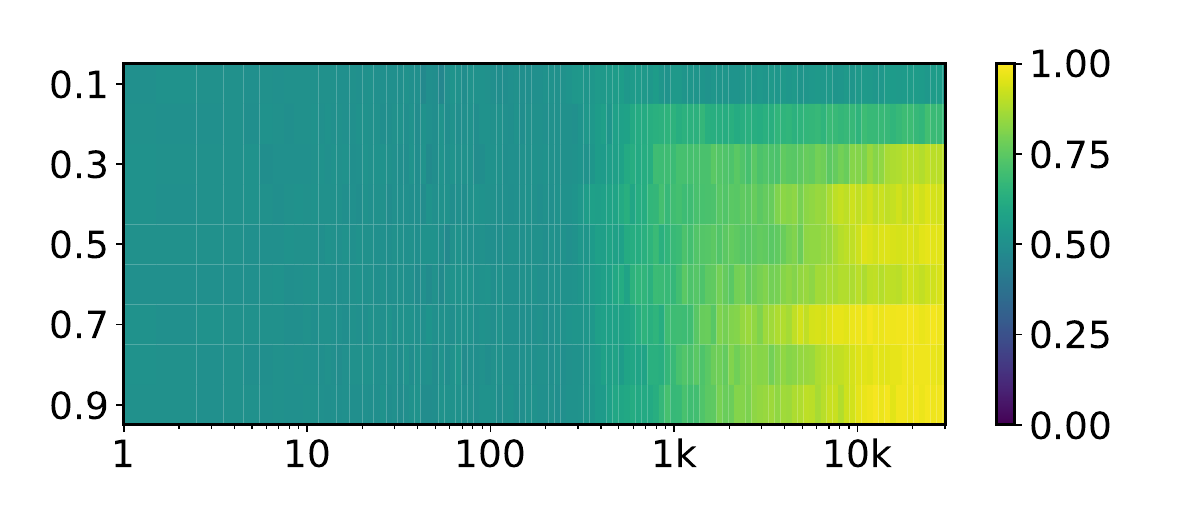}};
  \end{tikzpicture}
\end{subfigure} &
\begin{subfigure}[t]{0.2\textwidth}\centering
  \begin{tikzpicture}[remember picture]
    \node at (0,0) {\includegraphics[scale=0.2]{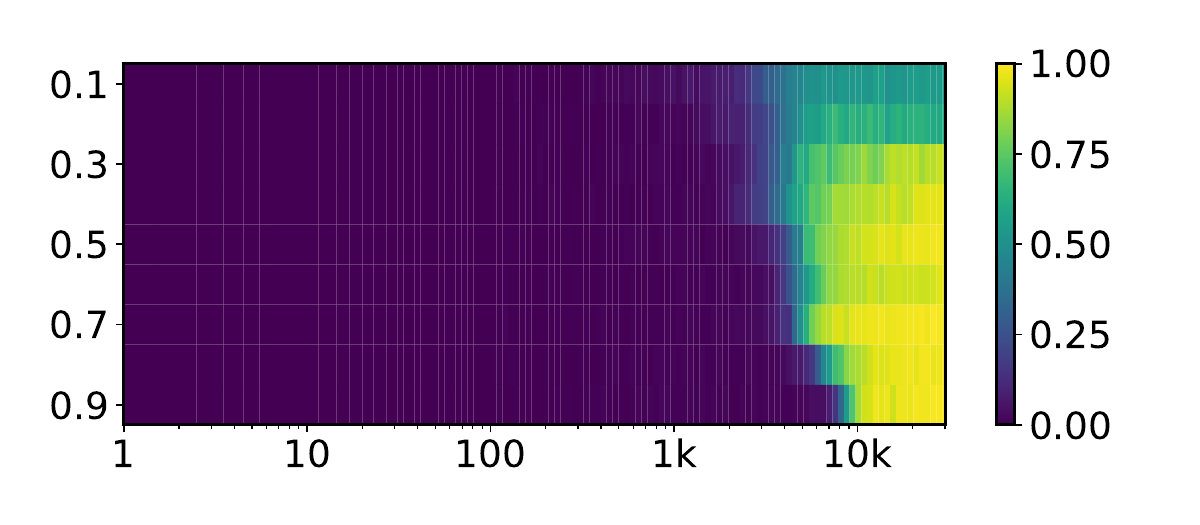}};
  \end{tikzpicture}
\end{subfigure} &
\rownote{No improvements.}
\\[-20pt]

% ───────────────────────── Row 9: V + O ──────────────────────────────
\begin{subfigure}[t]{0.2\textwidth}\centering
  \rowstrut
  \begin{tikzpicture}[remember picture]
    \node at (0,0) {\includegraphics[scale=0.2]{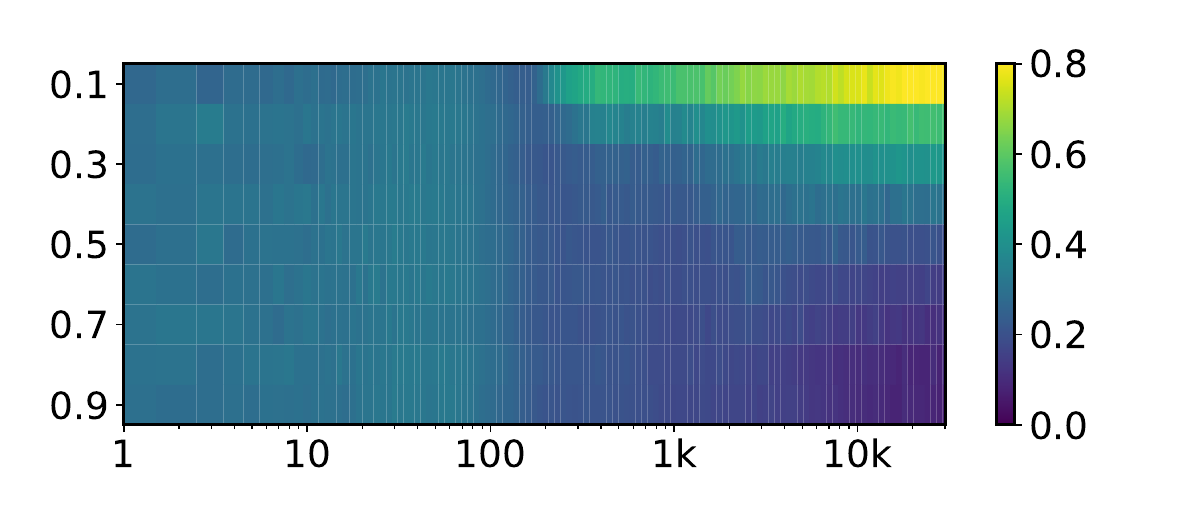}};
    \node[scale=0.75,rotate=90] at (-2.2,0.1) {$\mathbf{V},\mathbf{O}$};
  \end{tikzpicture}
\end{subfigure} &
\begin{subfigure}[t]{0.2\textwidth}\centering
  \begin{tikzpicture}[remember picture]
    \node at (0,0) {\includegraphics[scale=0.2]{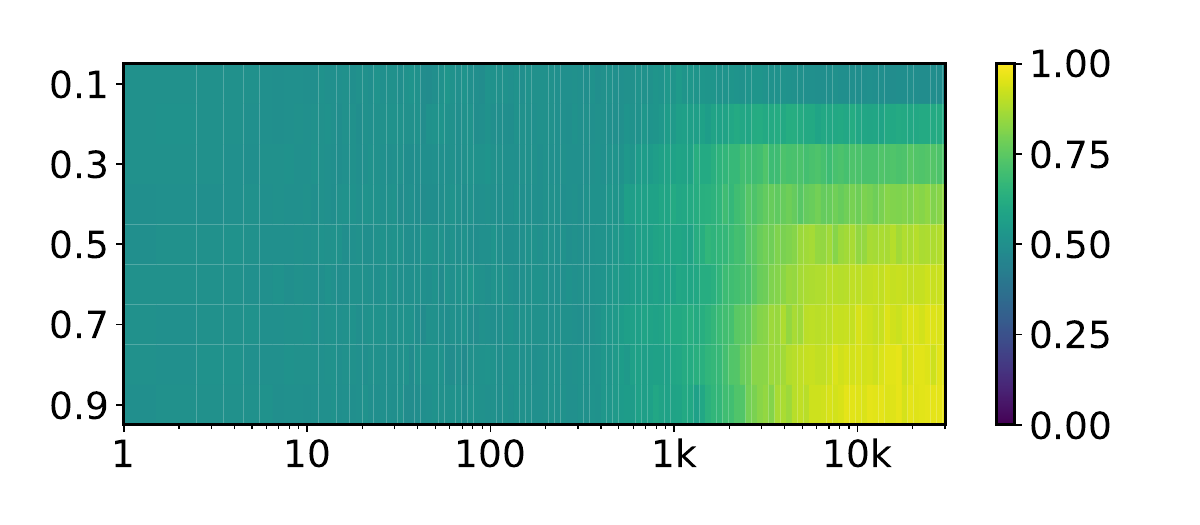}};
  \end{tikzpicture}
\end{subfigure} &
\begin{subfigure}[t]{0.2\textwidth}\centering
  \begin{tikzpicture}[remember picture]
    \node at (0,0) {\includegraphics[scale=0.2]{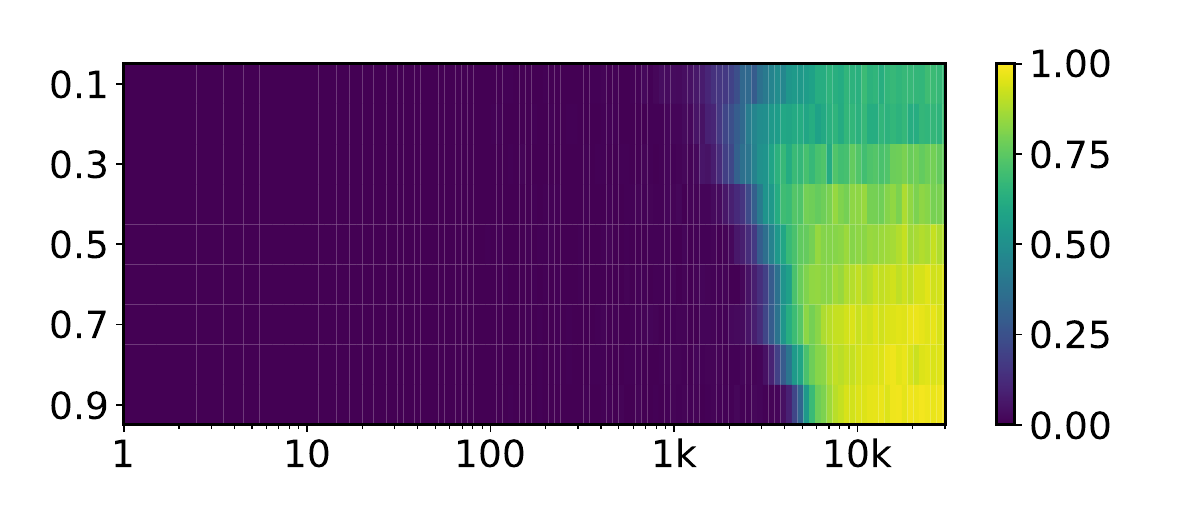}};
  \end{tikzpicture}
\end{subfigure} &
\rownote{}
\\[-20pt]

% ───────────────────────── Row 10: MLP + E ───────────────────────────
\begin{subfigure}[t]{0.2\textwidth}\centering
  \rowstrut
  \begin{tikzpicture}[remember picture]
    \node at (0,0) {\includegraphics[scale=0.2]{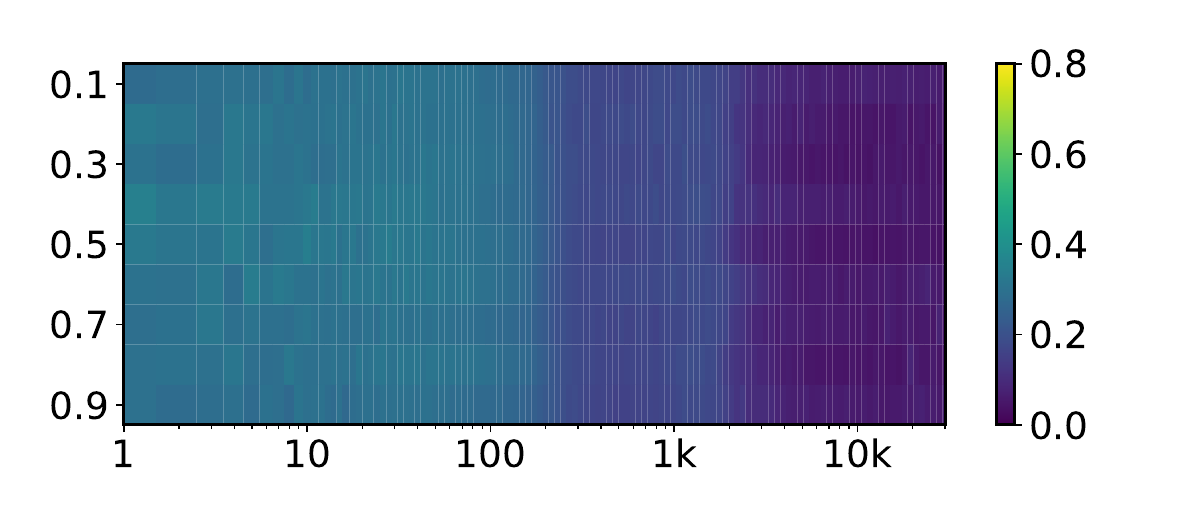}};
    \node[scale=0.75,rotate=90] at (-2.2,0.1) {$\textbf{MLP},\mathbf{E}$};
  \end{tikzpicture}
\end{subfigure} &
\begin{subfigure}[t]{0.2\textwidth}\centering
  \begin{tikzpicture}[remember picture]
    \node at (0,0) {\includegraphics[scale=0.2]{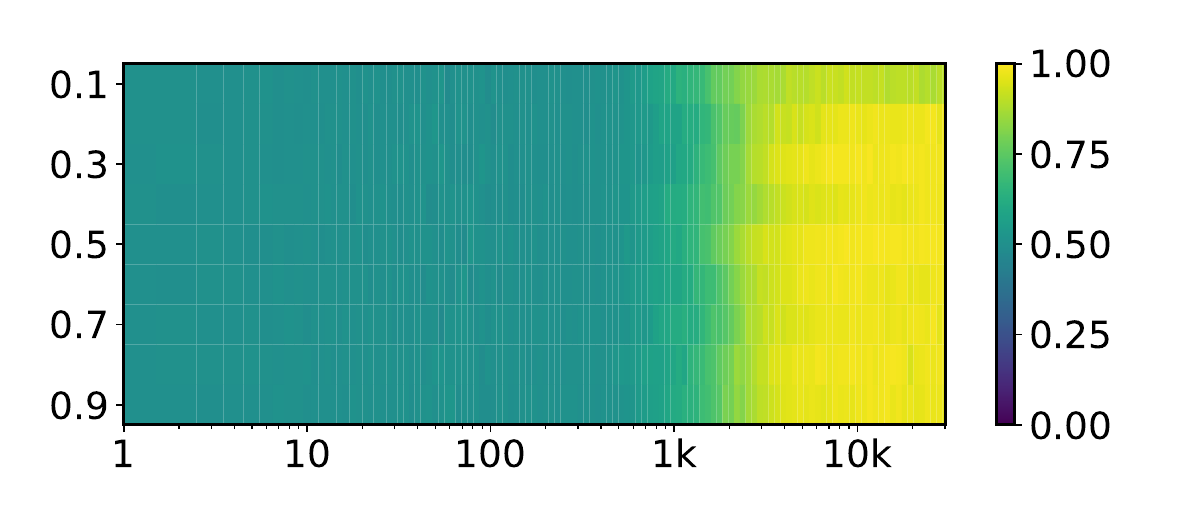}};
  \end{tikzpicture}
\end{subfigure} &
\begin{subfigure}[t]{0.2\textwidth}\centering
  \begin{tikzpicture}[remember picture]
    \node at (0,0) {\includegraphics[scale=0.2]{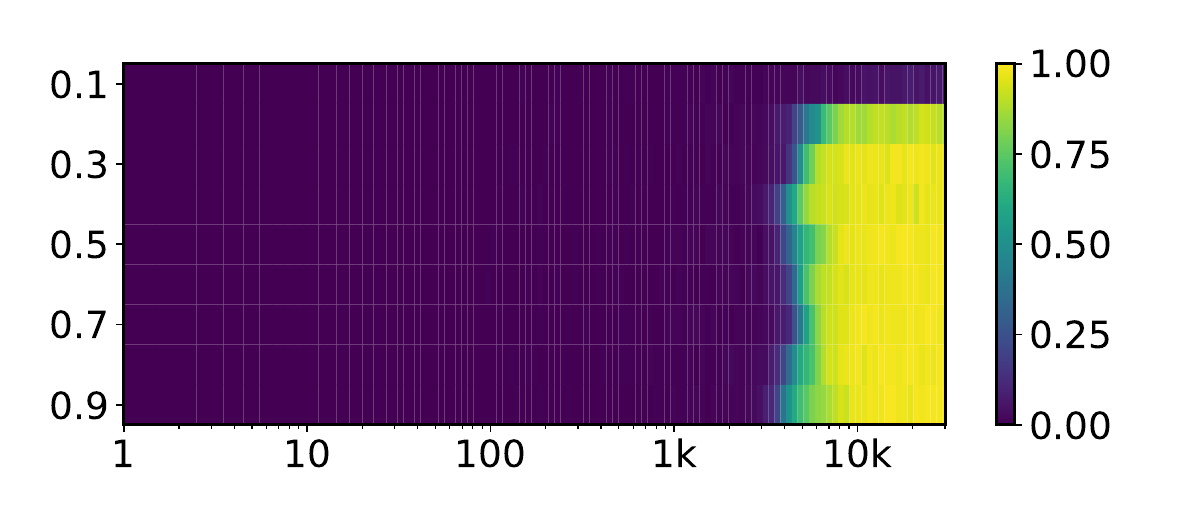}};
  \end{tikzpicture}
\end{subfigure} &
\rownote{}
\\[-20pt]

% ───────────────────────── Row 11: MLP + E + U ───────────────────────
\begin{subfigure}[t]{0.2\textwidth}\centering
  \rowstrut
  \begin{tikzpicture}[remember picture]
    \node at (0,0) {\includegraphics[scale=0.2]{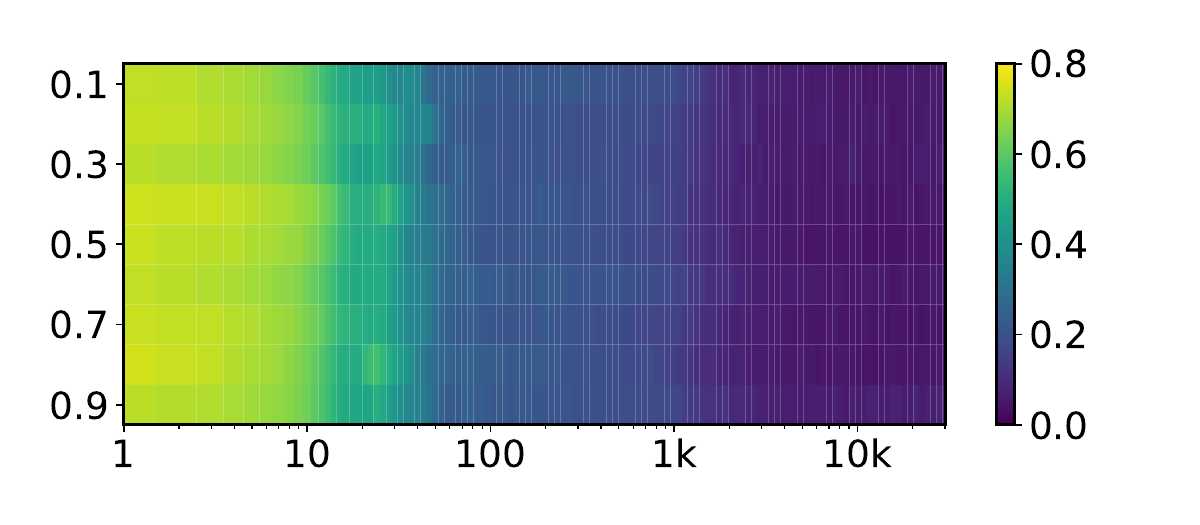}};
    \node[scale=0.75,rotate=90] at (-2.2,0.1) {$\textbf{MLP},\mathbf{E},\mathbf{U}$};
  \end{tikzpicture}
\end{subfigure} &
\begin{subfigure}[t]{0.2\textwidth}\centering
  \begin{tikzpicture}[remember picture]
    \node at (0,0) {\includegraphics[scale=0.2]{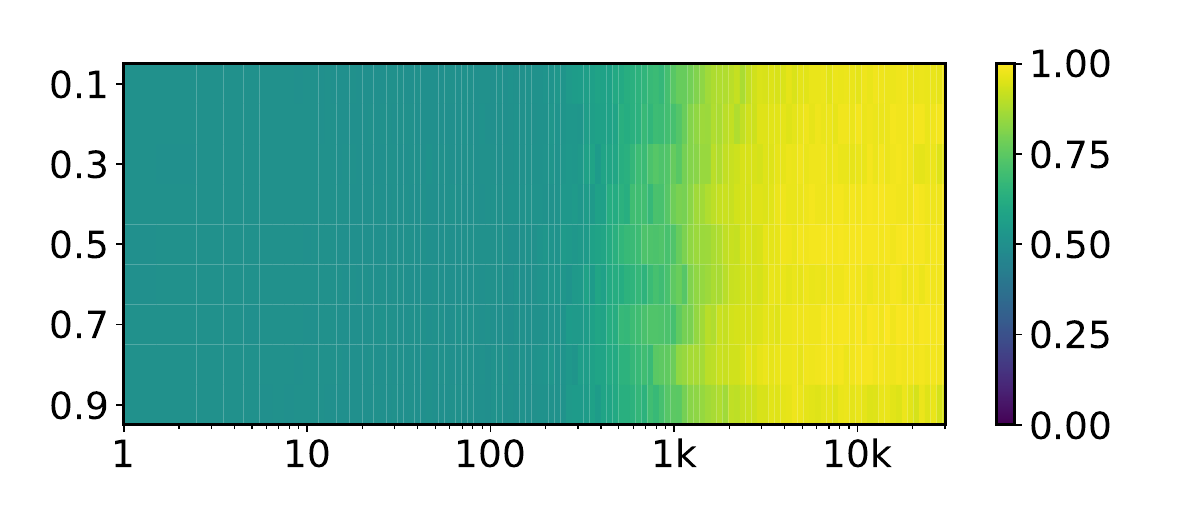}};
  \end{tikzpicture}
\end{subfigure} &
\begin{subfigure}[t]{0.2\textwidth}\centering
  \begin{tikzpicture}[remember picture]
    \node at (0,0) {\includegraphics[scale=0.2]{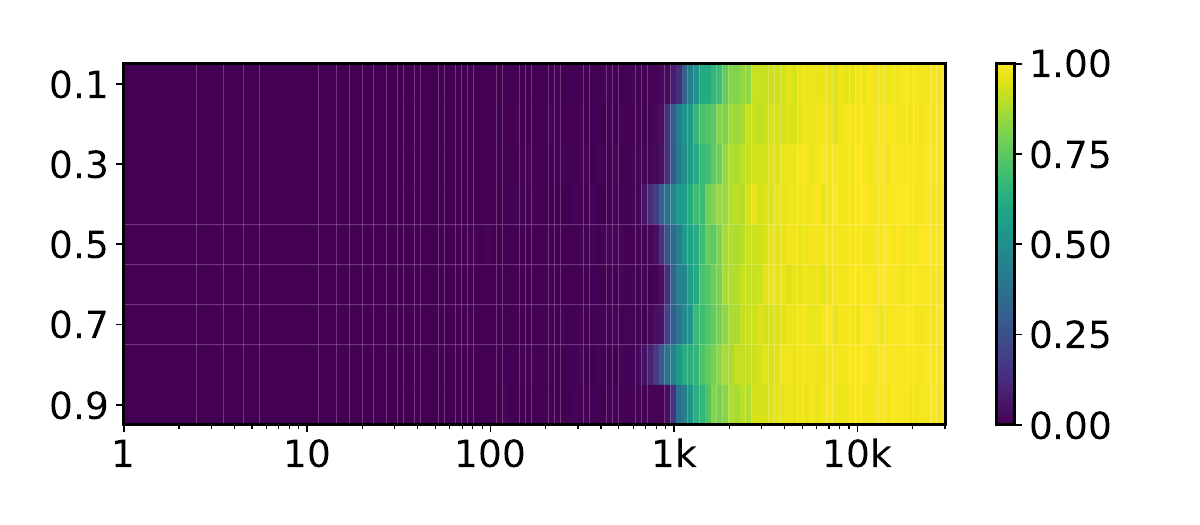}};
  \end{tikzpicture}
\end{subfigure} &
\rownote{}
\\[-20pt]

% ───────────────────────── Row 12: K + Q ─────────────────────────────
\begin{subfigure}[t]{0.2\textwidth}\centering
  \rowstrut
  \begin{tikzpicture}[remember picture]
    \node at (0,0) {\includegraphics[scale=0.2]{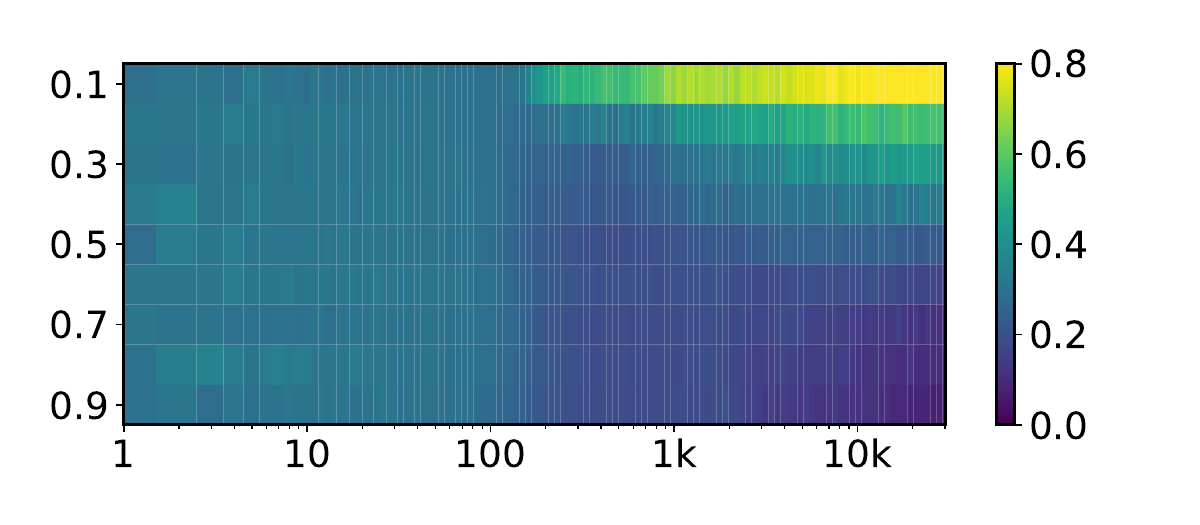}};
    \node[scale=0.75,rotate=90] at (-2.2,0.1) {$\mathbf{K},\mathbf{Q}$};
  \end{tikzpicture}
\end{subfigure} &
\begin{subfigure}[t]{0.2\textwidth}\centering
  \begin{tikzpicture}[remember picture]
    \node at (0,0) {\includegraphics[scale=0.2]{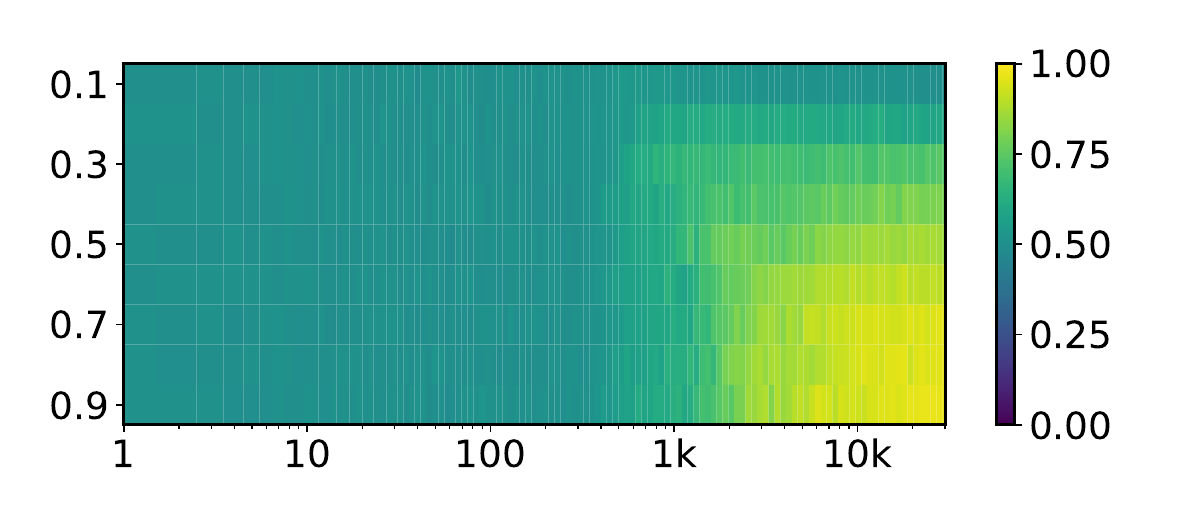}};
  \end{tikzpicture}
\end{subfigure} &
\begin{subfigure}[t]{0.2\textwidth}\centering
  \begin{tikzpicture}[remember picture]
    \node at (0,0) {\includegraphics[scale=0.2]{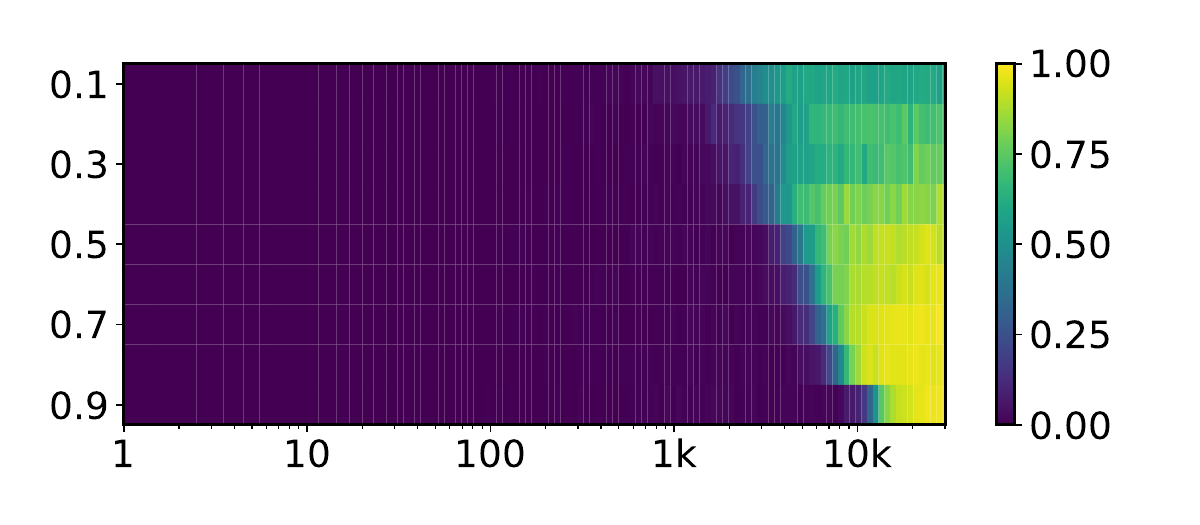}};
  \end{tikzpicture}
\end{subfigure} &
\rownote{}
\\[-20pt]

% ───────────────────────── Row 13: KQVOMLP ───────────────────────────
\begin{subfigure}[t]{0.2\textwidth}\centering
  \rowstrut
  \begin{tikzpicture}[remember picture]
    \node at (0,0) {\includegraphics[scale=0.2]{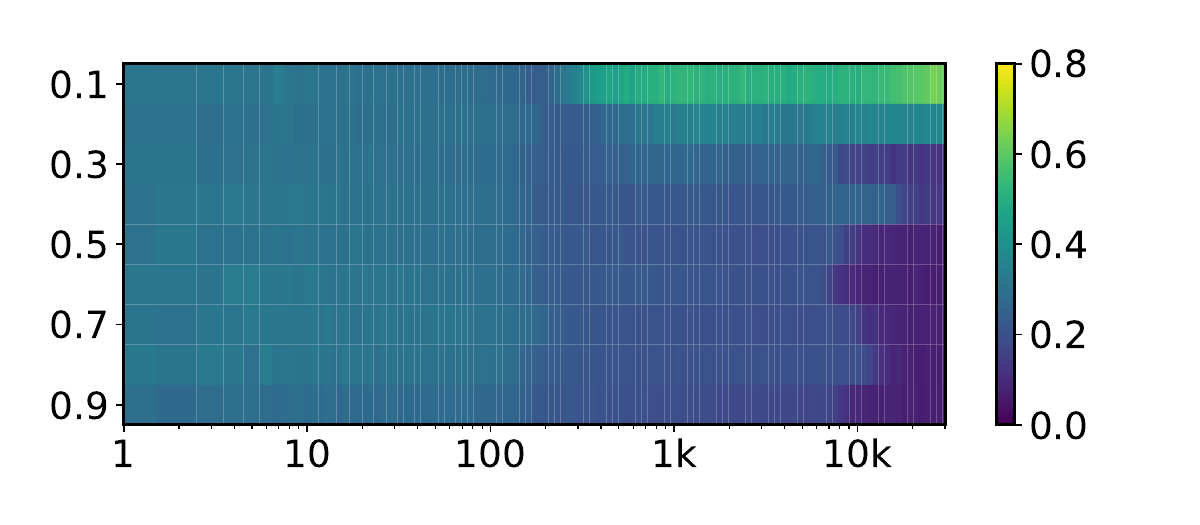}};
    \node[scale=0.65,rotate=90] at (-2.2,0.1) {$\mathbf{K,Q,V,O,MLP}$};
  \end{tikzpicture}
\end{subfigure} &
\begin{subfigure}[t]{0.2\textwidth}\centering
  \begin{tikzpicture}[remember picture]
    \node at (0,0) {\includegraphics[scale=0.2]{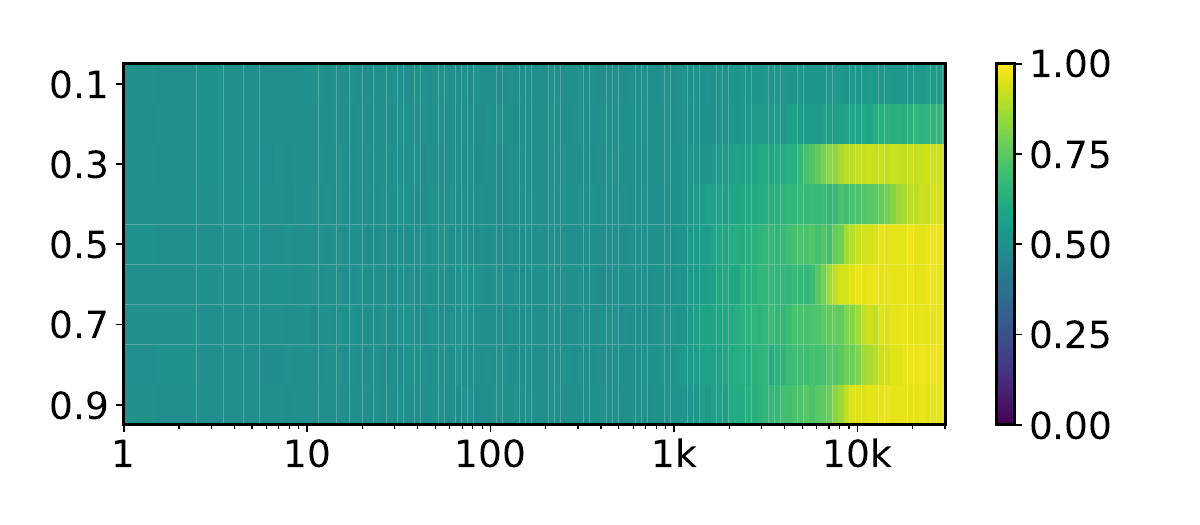}};
  \end{tikzpicture}
\end{subfigure} &
\begin{subfigure}[t]{0.2\textwidth}\centering
  \begin{tikzpicture}[remember picture]
    \node at (0,0) {\includegraphics[scale=0.2]{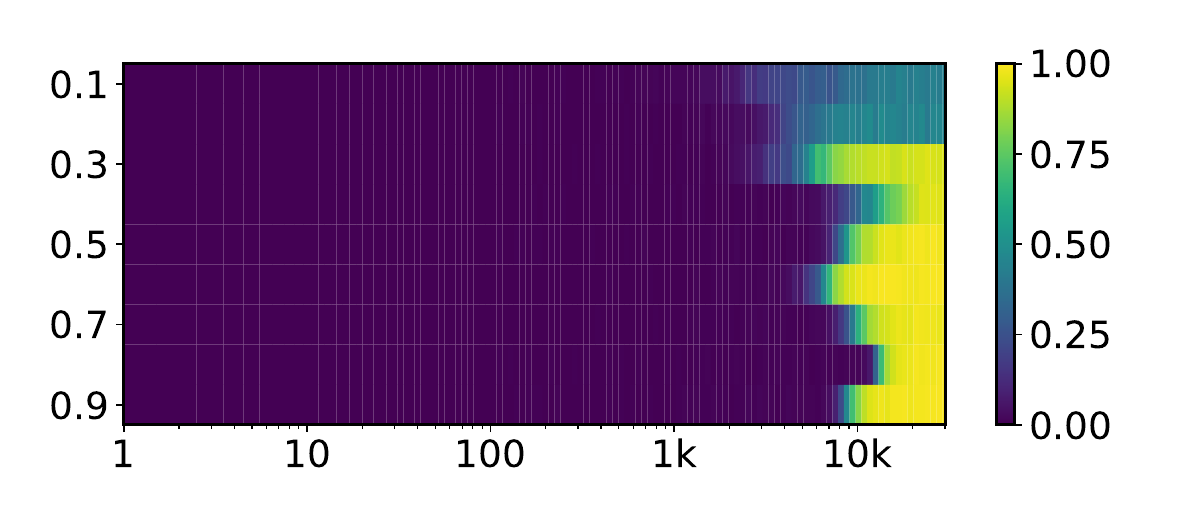}};
  \end{tikzpicture}
\end{subfigure} &
\rownote{}
\\[-0pt]

\end{tabular}
\caption{ Same as the intervention analysis from Fig.~\ref{fig:app_interventions} on a one-layer transformer. While the overall trends are similar to the four-layer model, the \textbf{Attn} intervention has a less significant impact here. This result aligns with the one-layer model's overall weaker performance (see also Fig.~\ref{fig:model_size}). }
\label{fig:app_interventions_L1}
\end{figure}

\begin{figure}[htbp]
\vspace{-10pt}
  \centering
  \begin{tabular}{@{\hspace{-30pt}}c@{\hspace{45pt}}c@{\hspace{45pt}}c@{}}

  % ─── header row: one cell for col 1, one spanning cols 2–3 ─────────
  % \multicolumn{1}{c@{\hspace{70pt}}}{\fontsize{9pt}{8pt}\selectfont\textbf{(a) ID mask ($\exposmatin$)}}
  % & \multicolumn{2}{c}{\fontsize{9pt}{8pt}\selectfont\textbf{(b) Factual accuracy}} \\[8pt]
  
  % ─── KL ───────────────────────────────
    \begin{subfigure}[t]{0.25\textwidth}
        \centering
        \begin{tikzpicture}[remember picture]
            \node at (0,0) {\includegraphics[scale=0.25]{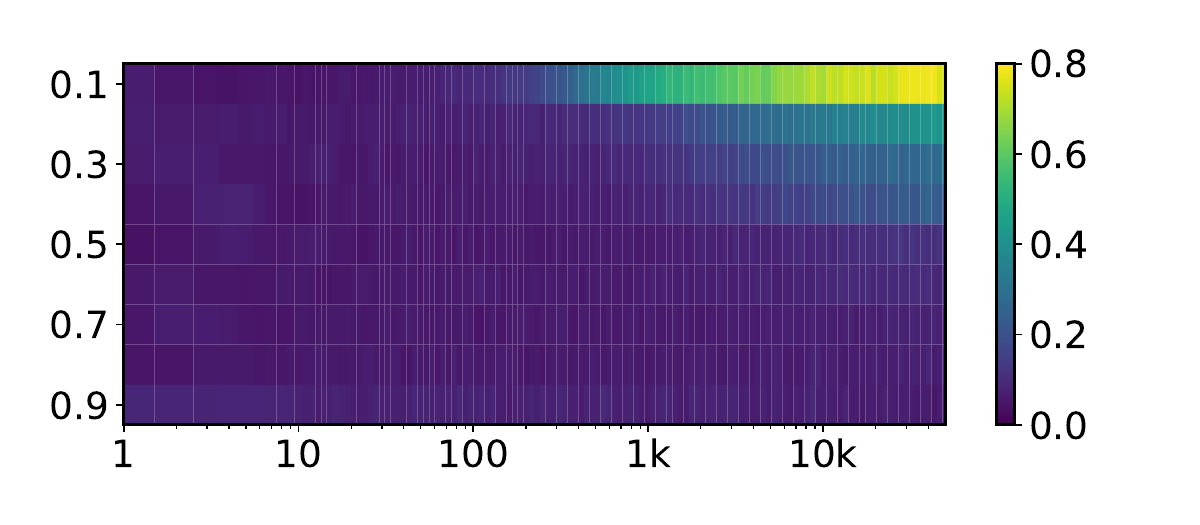}};
            % \node[scale=0.8] at (0.0, 1.3) { \textbf{ }};
            \node[scale=0.9,rotate=90] at (-2.6, 0.1) { $\dvr$};
            \node[scale=0.8] at (0.0, -1.2) {Iterations};
            \node[scale=0.8] at (0.0, 1.0) { \textbf{ $\KL$}};

        \end{tikzpicture}
    \end{subfigure} &
    % \hspace{70pt}%
    % \end{minipage}%
\colTwoThreeShift{
    \begin{subfigure}[t]{0.25\textwidth}
        \centering
        \begin{tikzpicture}[remember picture]
            \node at (0,0) {\includegraphics[scale=0.25]{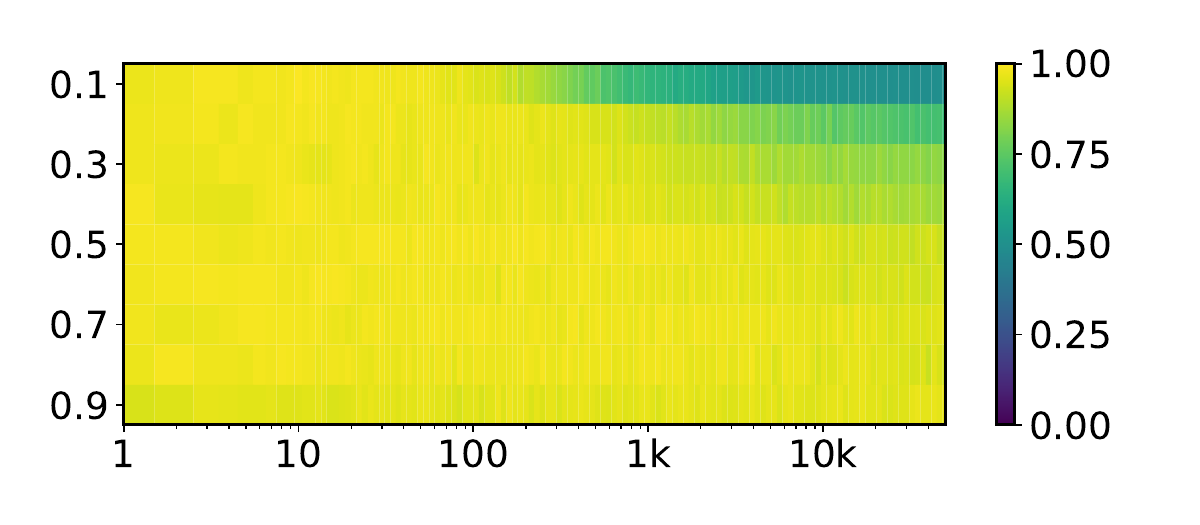}};
            % \node[scale=0.8,rotate=90] at (-2.7, 0.0) { \textbf{\posexp{10} }};            \node[scale=0.8] at (0.0, -1.2) { \textbf{ }};
            \node[scale=0.8] at (0.0, -1.2) {Iterations};
            \node[scale=0.8] at (0.0, 1.0) { \textbf{ $\posacc$}};
            % \node[scale=1.2] at (0.0, 1.5) { \textbf{(a) Statistical Loss}};

        \end{tikzpicture}
    \end{subfigure}
    }&%
    % \hspace{50pt}%
\colTwoThreeShift{    
    \begin{subfigure}[t]{0.25\textwidth}
        \centering
        \begin{tikzpicture}[remember picture]
            \node at (0,0) {\includegraphics[scale=0.25]{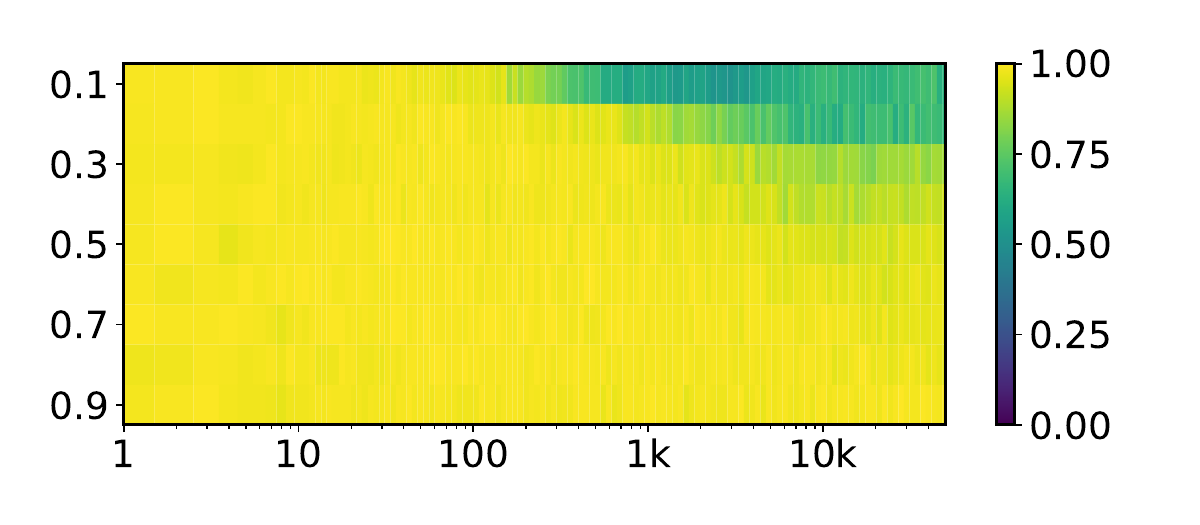}};            
            \node[scale=0.8] at (0.0, -1.2) {Iterations)};
            \node[scale=0.8] at (0.0, 1.0) { \textbf{ $\factacc$}};

        \end{tikzpicture}
    \end{subfigure}}
  \end{tabular}
  % \vspace{-0.1in}
  \caption{A one-layer transformer, initialized from a model trained on high-diversity data ($\dvr=0.9$), is then trained on a low-diversity dataset ($\dvr=0.1$). Although the model starts with perfect performance, its accuracy degrades across all metrics as training on the less diverse data progresses. This is similar to the degradation observed in Fig~\ref{fig:prob} (red).}  %\tina{fix alignments}
    
    % \vspace{-0.2in}
  \label{fig:high_div_init}
\end{figure}
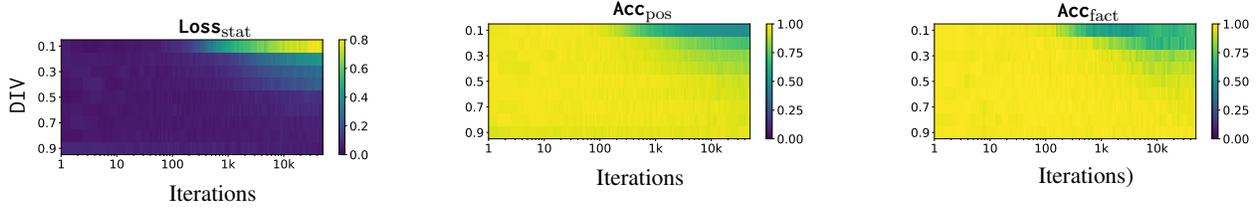

% \tina{table}

\begin{figure}[t]
  \hspace{-40pt}
  \centering
  \begin{subfigure}{0.6\linewidth}
    \centering
    \begin{tikzpicture}
      % Place the image in a node named 'image'
      % \node (image) {\includegraphics[width=\linewidth]{figures/otherSec5/L4H4_table_ood1_iter-1_.pdf}};
      \node (image) {\includegraphics[width=\linewidth]{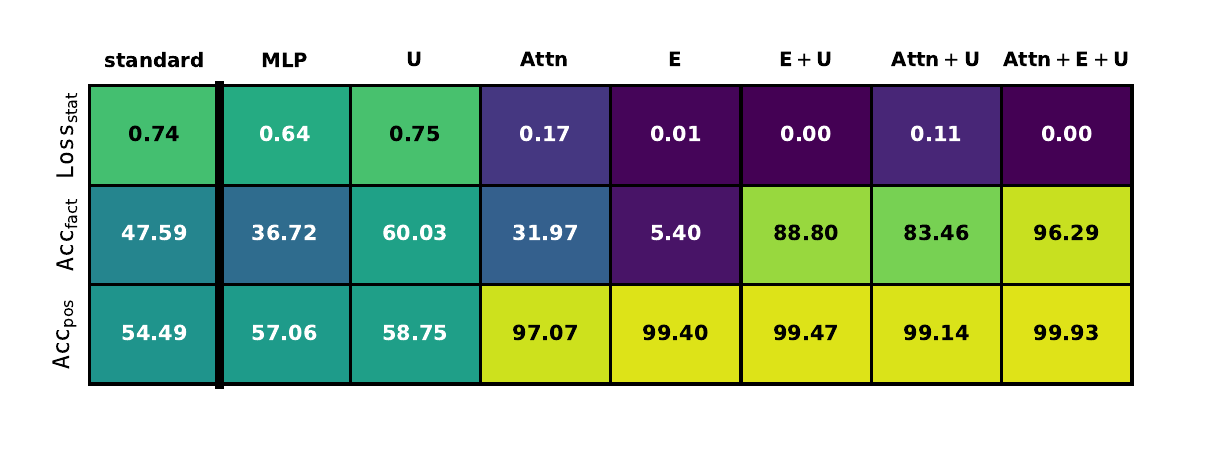}};
     \node[below=-9pt of image, font=\small] {};

      % Add the x-label below the image
      % \node[below=-7pt of image, font=\small] {Iterations};
      
      % Add the y-label to the left, rotated 90 degrees
      % \node[left=-1pt of image, rotate=90, anchor=center, font=\small] {Accuracy};
      % \node[right=-1pt of image, rotate=90, anchor=center, font=\small] {Loss};
    \end{tikzpicture}
    \vspace{-15pt}
    % \caption{}
    % \label{fig:left_large}
  \end{subfigure}%\hfill
  % \hspace{-10pt}

  \vspace{-5pt}
  \caption{Same metrics as Fig.~\ref{fig:intervention}-(a), but captured earlier in training (10k vs. 30k iterations). This mid-training snapshot more clearly reveals that the intervention on $\mathbf{U}$ accelerates learning facts, even on low-diversity data that yields sub-optimal performance.} 
  \label{fig:intervention_10k}
  \vspace{-0.1in}
\end{figure}

% \section{Additional discussion}\label{app:exp}
% \input{sections/app_extra}

\section{Minimal setting to understand the impact of low diversity}\label{app:minimal}
Focusing on factual recall, we consider a minimal toy setting with $\tmplnum$ templates and $\factsize=\tmplnum$ fact pairs with sequences of  length $\seqlen=2\times\tmplnum$. For further simplicity compared to our other settings, we let the generic tokens $\vmc$ to be drawn from uniform distribution over the $\vmsize=3$ tokens. For the source-target pair, we define each template $n\in[\tmplnum]$ by a position pair $\pospair_n=(n,n+\tmplnum)$. We compare the performance on two diversity levels: low diversity $\dvr=1/\tmplnum$ and high diversity $\dvr=(\tmplnum-1)/\tmplnum$.

Fig.~\ref{fig:minimal_setup} reports OOD factual recall in this setup for two diversity levels high (blue) and low (red) in three minimal settings:  (a) $\tmplnum=3$ with a 1-layer model, (b) $\tmplnum=3$ with a 4-layer model, (c) $\tmplnum=5$ with a 4-layer model. In all cases ID performance reaches $100\%$ by the end of training, so we only show the OOD results. 
% \tina{right? double check}
% 
As seen in panel (a), a 1-layer model is expressive enough to achieve perfect factual recall performance on the task, as it achieves perfect OOD (and ID) factual recall when trained with high diversity. However, with low diversity, the training algorithm fails to find this generalizing solution. Instead it converges to a solution that generalizes for the ID templates, but does not necessarily perform well on OOD templates. Increasing the model capacity roughly helps with the performance in the low-diversity case as shown in panel (b). However, increasing the task complexity by simply increasing $\tmplnum$, the same large model of panel (b), fails again at finding the generalizing solution. %\ct{I like it!}

We can formally think of this failure under low-diversity as follows. Following the notation in Sec.\ref{sec:setup}, the ultimate learning goal is to find model parameters  $\thetab^*$ that minimize the next-token prediction (NTP) loss over the complete distribution over the choice of the templates and facts, i.e., 
\begin{align*}
\thetab^*\in\arg\min_\thetab\,\,\Big\{\Lc_\text{tot}\left(\thetab\right) := 
\sum_{k\in[\factsize]}\,\sum_{n\in[\tmplnum]}\,\Ebb_{\xb\sim\{\Dc_n^k\}} \ellb_\text{NTP}\,\left(\xb;\,\thetab\right)\Big\}\,,
%\Ebb_{k\in[\factsize]}\,\Ebb_{n\in[\tmplnum]}\,\Ebb_{\xb\sim\{\Dc_n^k\}} \ellb_\text{NTP}\,\left(\xb;\,\thetab\right),
\end{align*}
where $\Dc_n^k$ is the distribution over sequences drawn from the $n$-th template with the fact placeholders filled with the $k$-th fact $(\source_k,\target_k)$, and $\ellb_\text{NTP}$ is the NTP loss on sequence $\xb$ parameterized by model parameters $\thetab$. {Note here that the total loss averages over \emph{all} $N$ templates.} {We assume henceforth that the model is sufficiently expressive such that $\Lc_\text{tot}(\thetab^*)$ attains the loss lower bound (over all possible parameterization). This is the case in all our settings.}

We can now decompose this loss into two components as $\Lc_\text{tot}\left(\thetab\right) = \Lc_\text{ID}\left(\thetab\right) + \Lc_\text{OOD}\left(\thetab\right)$, where $\Lc_\text{ID}(\cdot)$ aggregates the ID templates and the complement $\Lc_\text{OOD}(\cdot)$ term contains the OOD templates for each fact. Concretely, let %\tina{does it need some reweighting??}
\begin{align*}
\Lc_\text{ID}\left(\thetab\right)  &:= \sum_{k\in[\factsize]}\,\sum_{n\,:\,\exposmatin[n,k]=1}\,\Ebb_{\xb\sim\{\Dc_n^k\}} \ellb_\text{NTP}\,\left(\xb;\,\thetab\right),\\ 
    \Lc_\text{OOD}\left(\thetab\right) &:= \sum_{k\in[\factsize]}\,\sum_{n\,:\,\exposmatin[n,k]=0}\,\Ebb_{\xb\sim\{\Dc_n^k\}} \ellb_\text{NTP}\,\left(\xb;\,\thetab\right).
    % \Lc_\text{ID}\left(\thetab\right)  &:= \Ebb_{k\in[\factsize]}\,\Ebb_{n\in\{n';\exposmatin[n',k]=1\}}\,\Ebb_{\xb\sim\{\Dc_n^k\}} \ellb_\text{NTP}\,\left(\xb;\,\thetab\right),\\ 
    % \Lc_\text{OOD}\left(\thetab\right) &:= \Ebb_{k\in[\factsize]}\,\Ebb_{n\in\{n';\exposmatin[n',k]=0\}}\,\Ebb_{\xb\sim\{\Dc_n^k\}} \ellb_\text{NTP}\,\left(\xb;\,\thetab\right).
\end{align*}

During training, where we only get access to a subset of facts-template pairs $(k,n)$ for which $\exposmatin[n,k]=1$, we are essentially minimizing $\Lc_\text{ID}(\thetab)$. Intuitively this is the case because recall that we train the model such that at each iteration we see a fresh sequence $\x$ sampled from the ID templates and thus in the long run of many iterations, the training loss closely approximates the ID population loss $\Lc_\text{ID}(\thetab)$. This is also empirically verified, since with sufficiently long training we always reach 100\% ID accuracies. Thus, during training we find model parameters $\thetabtrain$ that  minimize the ID population risk, i.e.,
\[
\thetabtrain\in\arg\min_{\thetab} \Lc_\text{ID}(\thetab)\,.
\]
We remark that the set of minimizers can possibly contain multiple solutions (and we shortly argue that it does!). Also note that {in the assumed setting of $\Lc_\text{tot}(\thetab^*)$ attaining the total-loss lower bound,} a minimizer $\thetab^*$ of $\Lc_\text{tot}$ is certainly a minimizer of the ID risk. 

The interesting question is: \emph{Does training find model parameters 
$\thetabtrain$ that not only minimize the ID risk, but additionally minimize the 
OOD risk?} If that is the case, then $\thetabtrain=\thetab^*$, i.e., $\thetabtrain$ is a minimizer of the total loss $\Lc_\text{total}$.

Our experiments (both in the original setup of Fig.~\ref{fig:heatmaps_main} and even more evidently in the minimal setup of this section) reveal a compelling diversity-dependent dichotomy:
%the answer
%to the question whether training yields a minimizer $\thetab^*$ of $\Lc_\text{total}$ 
%depends on diversity. 
% 
On the one hand, under low-diversity, training converges to non-generalizing solutions that while they minimize $\Lc_\text{ID}(\thetab)$, they do not minimize the OOD risk $\Lc_\text{ID}(\thetab)$. Thus, $\thetabtrain$ is a minimizer of the ID risk $\Lc_\text{ID}(\thetab)$ but a different one than the total loss minimizer $\thetab^*$.
On the other hand, as diversity increases, training finds generalizing solutions, i.e. $\thetabtrain$ is now a minimizer of both the ID and the total loss. 

This dichotomy admits two possible explanations. With increasing diversity,  either (1) the non-generalizing solutions are removed from the set of global optimizers of the ID loss $\Lc_\text{ID}$, or (2) the landscape of the ID loss becomes more benign around the generalizing solutions (aka $\thetab^*)$, which in turn makes it easier for the model to find them. Fig.~\ref{fig:minimal_setup}-(b) also suggests that increased model capacity can partially help with making the ID landscape more benign.\looseness=-1

Precisely characterizing how the context diversity and model capacity reshape the ID loss landscape is an exciting direction for future work. We believe the minimal setup and intuitions introduced in this section can serve as a starting point for such analysis.

\begin{figure}[t]
  \centering
  \begin{tabular}{@{\hspace{-55pt}}c@{\hspace{50pt}}c@{\hspace{40pt}}c@{}}
  % ─── header row: one cell for col 1, one spanning cols 2–3 ─────────
  % \multicolumn{2}{c}{\fontsize{9pt}{8pt}\selectfont\textbf{(a) $\dvr=0.8$}}
  % & \multicolumn{2}{c}{\fontsize{9pt}{8pt}\selectfont\textbf{(b) $\dvr=0.1$}} \\[8pt]
  % ─── two rows of subfigures ───────────────────────────────
    % \hspace{-25pt}
    \begin{subfigure}{0.22\textwidth}
        \centering
        \begin{tikzpicture}[remember picture]
            \node at (0,0) {\includegraphics[scale=0.22]{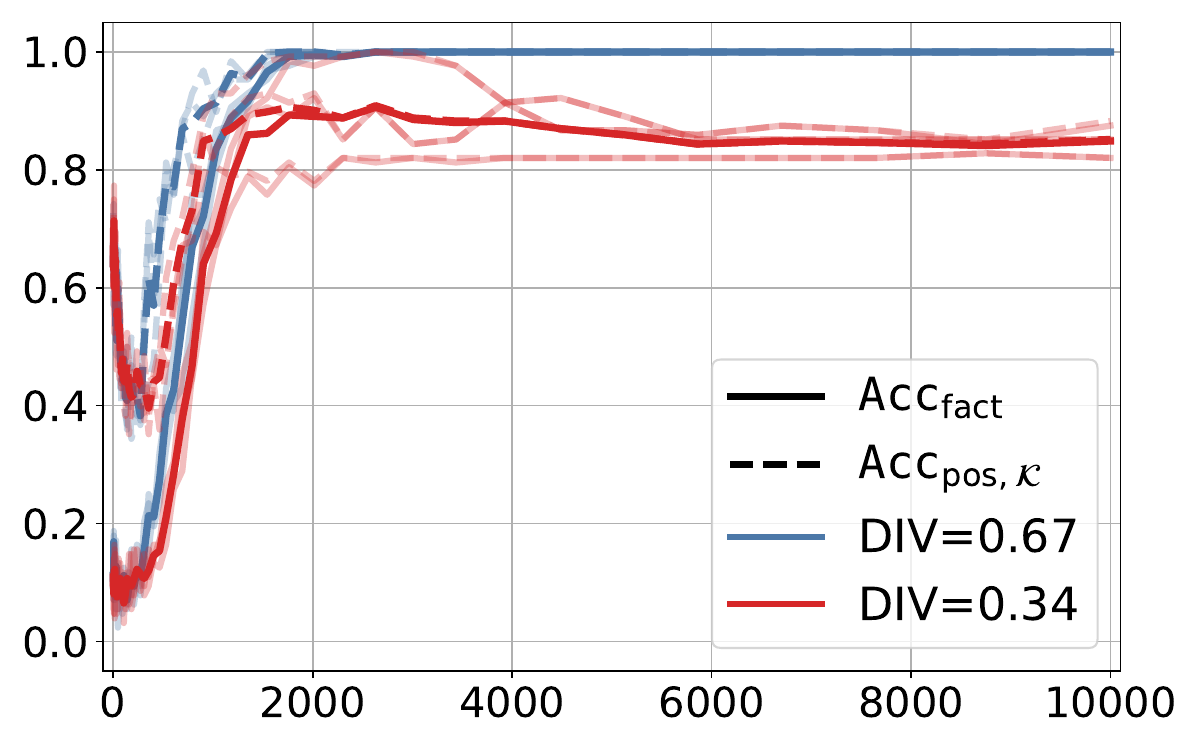}};
            \node[scale=0.9] at (0.0, 1.6) {$\tmplnum=3, \text{ 1-layer}$};
            \node[scale=0.9] at (0.0, -1.5) { iteration};
        \end{tikzpicture}
    \end{subfigure} &
    % \hspace{70pt}%
    \begin{subfigure}{0.22\textwidth}
        \centering
        \begin{tikzpicture}[remember picture]
            \node at (0,0) {\includegraphics[scale=0.22]{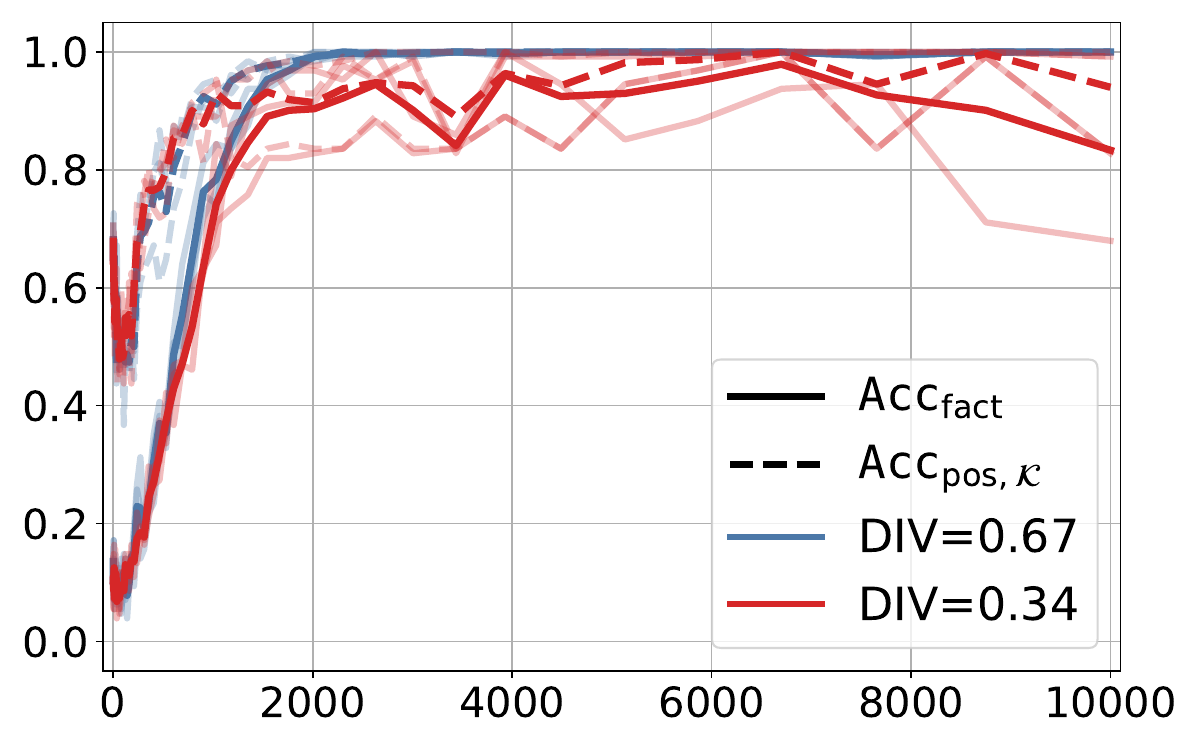}};
            \node[scale=0.9] at (0.0, 1.6) {$\tmplnum=3, \text{ 4-layer}$};
            \node[scale=0.9] at (0.0, -1.5) { iteration};
        \end{tikzpicture}
    \end{subfigure} &
    % \hspace{50pt}%
    \begin{subfigure}{0.22\textwidth}
        \centering
        \begin{tikzpicture}[remember picture]
            \node at (0,0) {\includegraphics[scale=0.22]{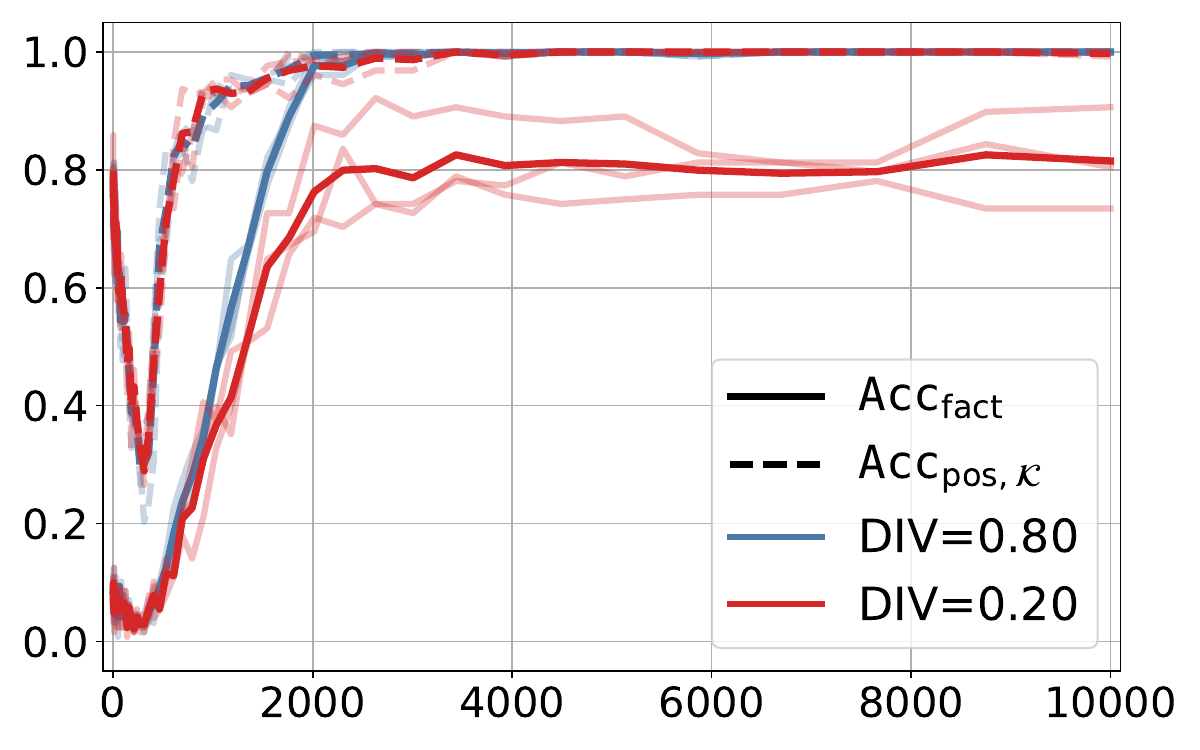}};
            \node[scale=0.9] at (0.0, 1.6) {$\tmplnum=5,\text{ 4-layer}$};
            \node[scale=0.9] at (0.0, -1.5) { iteration};
        \end{tikzpicture}
    \end{subfigure}
  \end{tabular}
  \vspace{-0.1in}
  \caption{\textbf{Minimal setup to replicate the impact of diversity.} Factual recall $\factacc$ (solid) and target-position accuracy $\posacck$ (dashed). See Sec.~ \ref{app:minimal} for discussion.   
  }
  \vspace{-0.1in}
    \label{fig:minimal_setup}
\end{figure}

\section{Representation Analysis}\label{sec:representation_analysis}
Building on our experiments probing the model's last-layer representations (Sec. \ref{sec:prob}), we now examine the clustering properties of hidden representations from all layers and the impact of training diversity.

To analyze the structure of these internal representations, for a given sequence drawn from the $n$-th template and carrying fact $(\source_k,\target_k)$, we probe hidden layer representations (for any layer $\ell$) at source fact position $\posSource_n$ and test whether the transformer encodes a \emph{template-invariant} representations of each fact $(\source_k, \target_k)$. 

For every (fact, template) pair (both ID and OOD), we first sample $M$ sequences. For each sequence, we collect hidden vectors $\hb_{\ell}^{(\posSource_n)}$ at source fact position $\posSource_n$ from the $\ell$-th transformer layer. For each layer $\ell$, we then stack the vectors into $\Hb^{(\ell)}\in\mathbb{R}^{P \times d}$ and keep the top $d' = \min(30,d,P)$ principal components, where $d$ denotes the dimensionality of the hidden layer representations, and $P = M \times N \times K$ denotes the total number of hidden vectors extracted per layer. This gives us a PCA-reduced matrix $\Hbtilde^{(\ell)}\!\in\!\mathbb{R}^{P\times d'}$ where $\hbtilde_i^{(\ell)}, \, \, i \in [P]$ denotes the PCA-reduced representation at fact position for the $i$-th sequence. Recall that $N$ denotes the number of templates, and $K$ denotes the number of atomic facts. We set $M$ to $250$ in our experiments. 

For every layer $\ell$, we take the PCA-reduced matrix $\Hbtilde^{(\ell)}\!\in\!\mathbb{R}^{P\times d'}$ and evaluate clustering
quality of the vector embeddings hen they are labeled in two different ways: 1) each vector tagged with the fact index $k$, and 2) each vector tagged with the template index $n$. We measure the clustering quality with \texttt{silhouette\_score} (\textsc{sklearn}).
% 
% along two label axes (we use the factual identity $a$ or template identity $n$ of the hidden representations $\Hbtilde^{(\ell)}$), as follows.  
% 
For a given hidden layer $\ell$, and every representation $\hbtilde_i^{(\ell)}, \, \, i \in [P]$, we compute 1) $e_i^{(\ell)}$, the average Euclidean distance to all other vectors that share its label and 2)  $f_i^{(\ell)}$, the smallest average distance to a group with a \emph{different} label. Formally, if $C_i$ is the set of indices with the same label,
then
\begin{align*}
    e_i^{(\ell)} \;=\; \frac{1}{|C_i|-1}\sum_{j\in C_i,\; j\neq i}\bigl\|\hbtilde_i^{(\ell)} -
                 \hbtilde_j^{(\ell)}\bigr\|_2,\quad 
    f_i^{(\ell)} \;=\;
\min_{C\neq C_i} \;
\frac{1}{|C|}\sum_{j\in C}
\bigl\|\hbtilde_i^{(\ell)} -
       \hbtilde_j^{(\ell)}\bigr\|_2 .
\end{align*}
% \[ 
% e_i^{(\ell)} \;=\; \frac{1}{|C_i|-1}\sum_{j\in C_i,\; j\neq i}\bigl\|\hbtilde_i^{(\ell)} -
%                  \hbtilde_j^{(\ell)}\bigr\|_2 .
% \]

% We then evaluate $f_i^{(\ell)}$, which is the smallest average distance to a group with a \emph{different} label. For each label $C\!\neq\!C_i$ we compute the mean distance of
% $\tilde{\mathbf h}^{(\ell)}_i$ to vectors in $C$ and find the smallest of those means,
% \[
% f_i^{(\ell)} \;=\;
% \min_{C\neq C_i} \;
% \frac{1}{|C|}\sum_{j\in C}
% \bigl\|\hbtilde_i^{(\ell)} -
%        \hbtilde_j^{(\ell)}\bigr\|_2 .
% \]

Using these two metrics, the score for each vector embedding $\hbtilde_i^{(\ell)}$ is defined as 
\[
  s_i^{(\ell)} \;=\;\frac{f_i^{(\ell)}-e_i^{(\ell)}}{\max\{f_i^{(\ell)},e_i^{(\ell)}\}}\in[-1,1], \quad i \in [P].
\]
The silhouette value attains $1$ when $\hbtilde_i^{(\ell)}$ lies well inside a compact cluster whose members share the same label, drops to $0$ when clusters of different labels overlap, and becomes negative if the vector is closer to another label’s cluster than to its own. The layer‑level score $s^{(\ell)}$ is the average of these values across all vectors in the layer, i.e., $s^{(\ell)}=\frac{1}{P}\sum_{i\in[P]}s_i^{(\ell)}$. To differentiate the two labeling schemes, we denote the score as $s^{(\ell)}_{\text{fact}}$ when clusters are labeled using factual indices $k$, and as $s^{(\ell)}_{\text{tmpl}}$ when template indices $n$ are used as labels. If $s^{(\ell)}_{\text{fact}}$ is high, it indicates that the representations are \emph{template-invariant}: the hidden representations learned for any given fact $\source$ only depends on the fact itself and not the context template it appears in during training.  In turn, if $s^{(\ell)}_{\text{tmpl}}$ is high, it suggests that the fact hidden representations from the same template cluster together even when the facts differ.

Figure \ref{fig:cluster_scores} reveals three consistent trends in the clustering structure of hidden representations as training diversity grows. First, \emph{factual identity is always the dominant organizing principle}: across layers the fact curves sit well above the template curves, indicating stronger clustering by fact than by template. Second, this \emph{fact-centric structure generalizes to unseen pairings}—the ID-only and ID+OOD fact curves roughly remain similar, showing that vectors from unseen fact–template combinations fall into the same clusters as their seen counterparts. Third, \emph{greater template diversity progressively weakens template-based structure while fact-based structure remains intact}, so the gap between the two widens.

\begin{figure}
    \centering
    \includegraphics[scale=0.5]{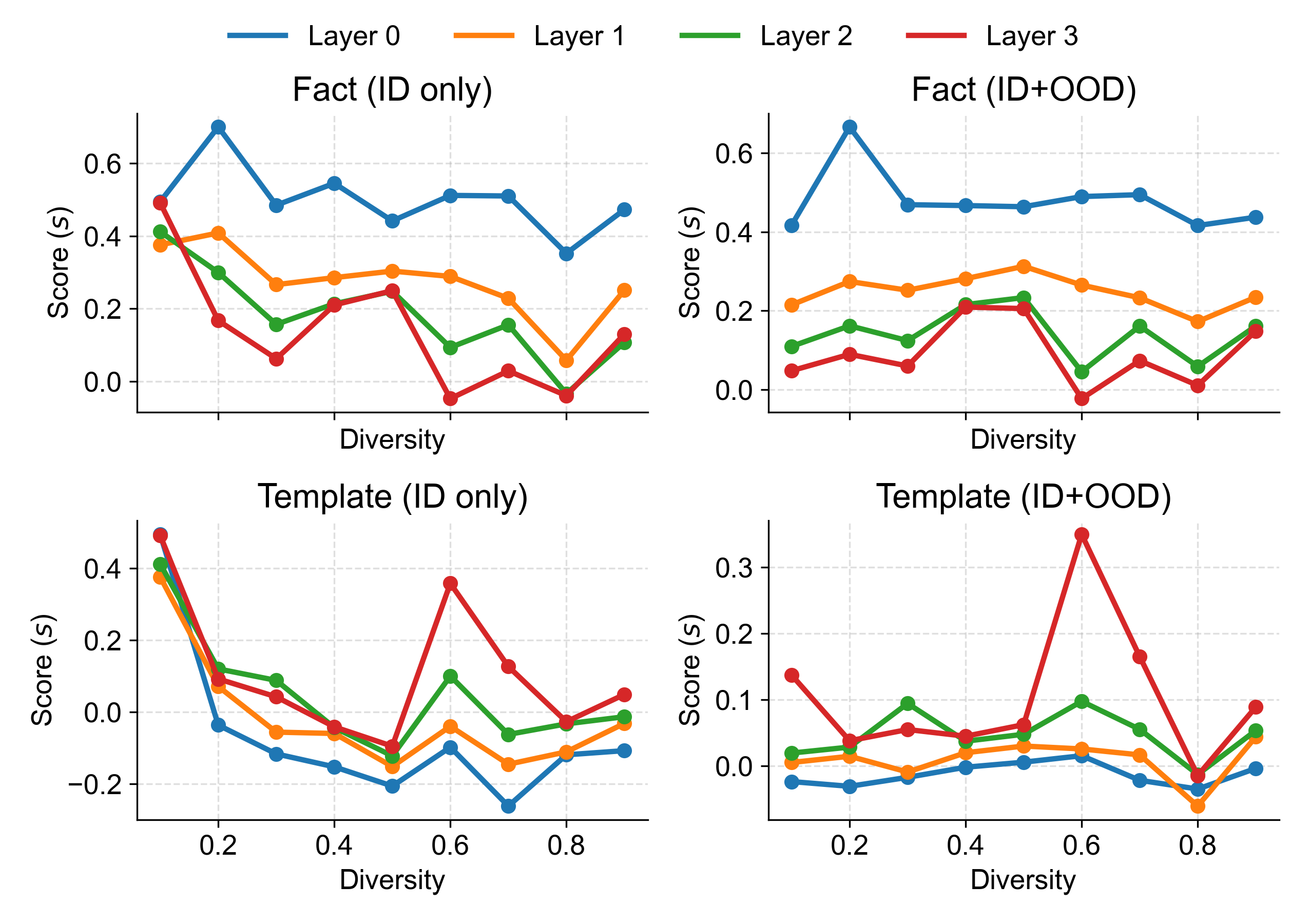}
    \caption{Clustering quality (see Sec.~\ref{sec:representation_analysis}) of hidden representations as a function of training diversity ($\dvr$) for the  $\posexp{10}$ template type. \textbf{Top}: Clustering scores when representations are labelled based on factual identity $k$. \textbf{Bottom}: Clustering scores when representations are labeled based on template identity $n$. \emph{Left} shows computation of score using representations of \emph{ID} (template, fact) pairs. \emph{Right} shows the clustering score using representations of both $\emph{ID}$ and $\emph{OOD}$ (template, fact) pairs.
    % $\emph{Left}$ and $\emph{right}$ similarly illustrate scores on only $\emph{ID}$, and $\emph{ID}$ and $\emph{OOD}$ (template, fact) pairs, respectively.
    \textbf{Fact clustering dominates:} In both “Fact” panels (top row) every layer’s curve sits well above the corresponding “Template” curves (bottom row). Hidden vectors therefore cluster primarily by the underlying fact rather than by the template. \textbf{Strong fact clustering persists on unseen templates:} The two fact curves—one computed on seen (ID-only) pairs, the other on the full ID + OOD set are roughly similar. Vectors for unseen fact–template combinations land in the same clusters as their seen counterparts, showing that the model abstracts the fact beyond the specific templates it saw during training. \textbf{Template invariance improves with diversity:} Moving from low to high diversity the template scores drift toward (or below) 0, while the fact scores remain roughly stable. This widening gap indicates that training on a broader mix of templates gradually removes template details from the representations while still keeping the facts separated.}\label{fig:cluster_scores}
    
     % Template-based clustering drops significantly with increasing diversity ($\dvr$), indicating that the influence of surface-level prompt structure diminishes when each fact is presented in many different forms. In contrast, fact-based clustering remains relatively stable across layers, with a noticeable improvement only at the final layer when comparing low vs. high diversity—indicating that robust fact representations begin to emerge primarily at the top of the network.} \label{fig:cluster_scores}
\end{figure}

\begin{figure}[h!]
    \centering
    \includegraphics[width=0.48\linewidth]{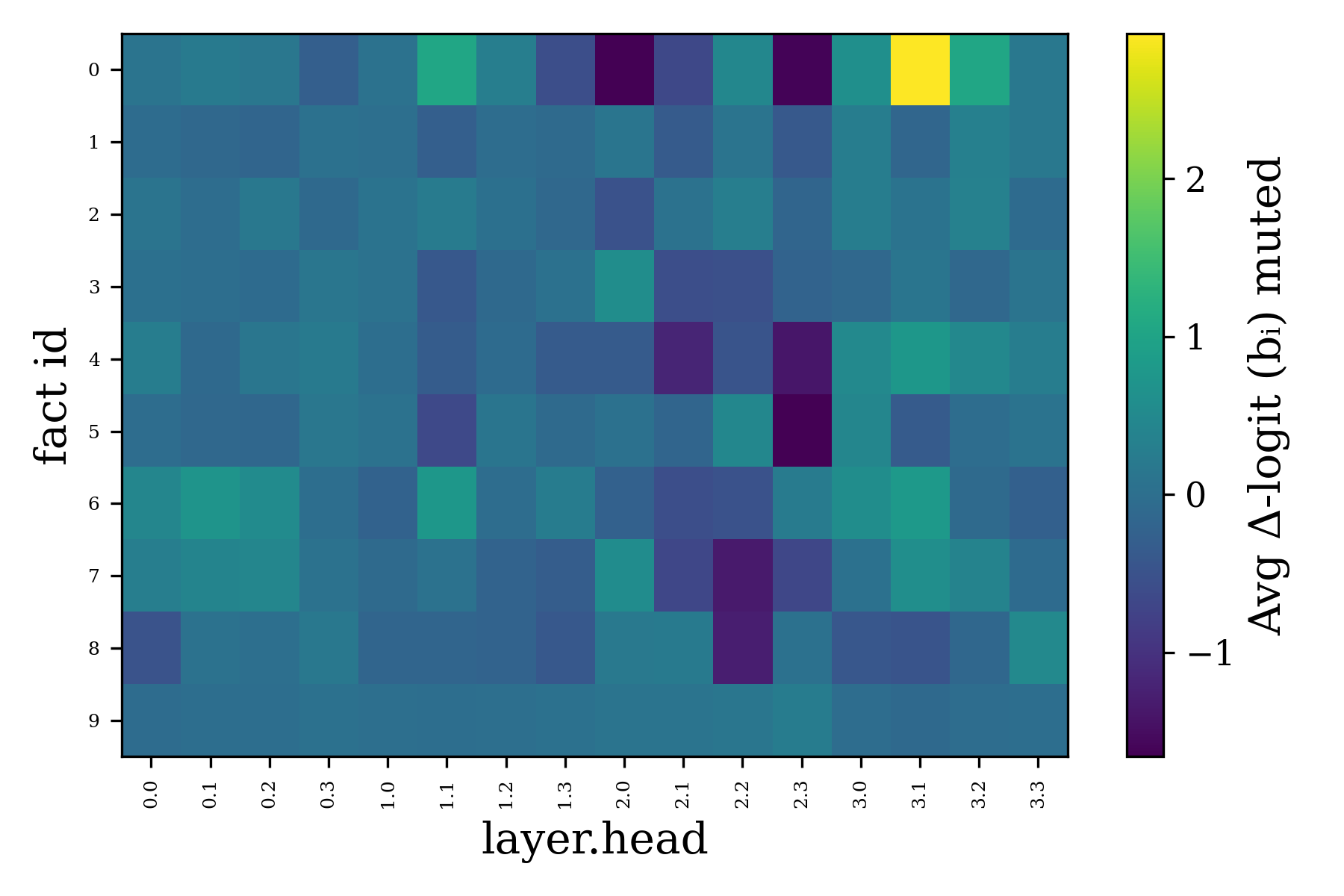}
    \includegraphics[width=0.48\linewidth]{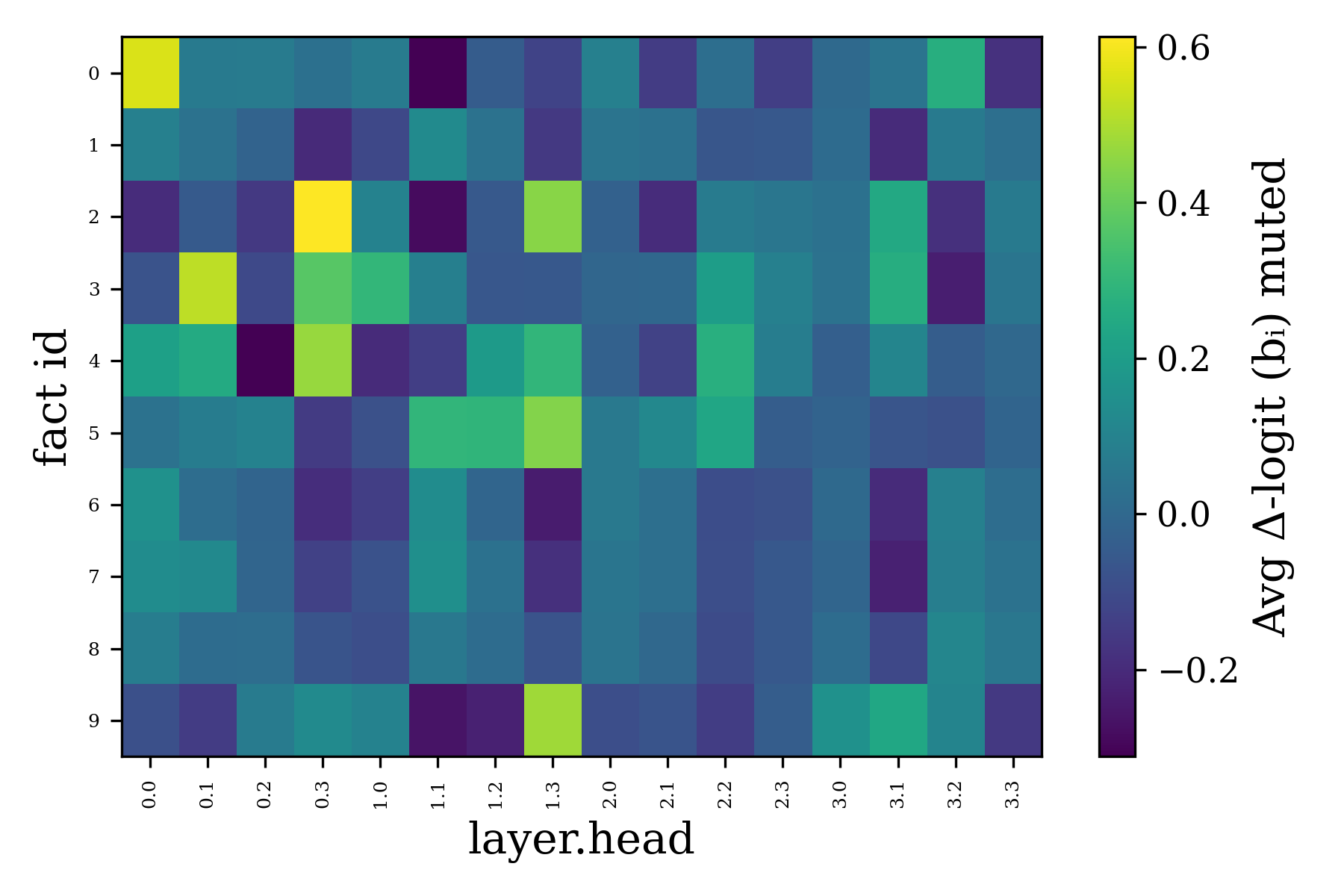}
    \caption{Per-fact head importance heatmaps in the factual recall task for the $\posexp{10}$ template type. Each row corresponds to a different fact, and each column to a specific attention head (indexed as layer.head). The color indicates the average change in the logit of the correct answer token $\target_i$ when the corresponding head is ablated, averaged over multiple in-distribution sequences for that fact. \textbf{Left}: model trained with low diversity ($\dvr = 0.1$). \textbf{Right}: model trained with high diversity ($\dvr = 0.9$). In the low diversity regime, head importance is diffuse and uniform across heads, suggesting no clear specialization. In contrast, at high diversity, certain heads become more consistently important for specific facts, indicating emergent specialization and more structured factual encoding.}
    \label{fig:fact_head}
\end{figure}

\noindent \textbf{Fact Heads.} To better understand how factual knowledge is stored across attention heads, we compute a per-fact head attribution heatmap. For each fact, we iteratively ablate individual heads (by zeroing their contribution) and measure the drop in the model's confidence for the correct token $\target$. This is averaged over multiple in-distribution contexts where the fact appears, yielding a (fact × head) matrix of logit drops. Figure \ref{fig:fact_head} shows these heatmaps for a model trained under low diversity (left) and high diversity (right). In the low diversity case, head importance is broadly distributed, with no head clearly emerging as critical for any fact. By contrast, in the high diversity regime, certain heads show strong and localized importance for specific facts, suggesting that the model has developed specialized storage heads. This points to a more structured encoding strategy that emerges only when the model sees the same fact across various different templates.

\end{document}